\documentclass[a4paper,11pt,fleqn]{book}

\pdfoutput=1

\usepackage[T1]{fontenc}
\usepackage[utf8]{inputenc}
\usepackage[square,sort,comma,numbers]{natbib}
\usepackage[french,german,english]{babel}

\usepackage{lmodern} 

\usepackage{setspace} 

\usepackage{graphicx,xcolor}
\graphicspath{{images/}}

\usepackage{booktabs}
\usepackage{lipsum}
\usepackage{microtype}
\usepackage{url}

\usepackage{fancyhdr}

\fancypagestyle{plain}{
	\fancyhf{}

	\fancyfoot[EL,OR]{\thepage}}
\fancypagestyle{addpagenumbersforpdfimports}{
	\fancyhead{}
	
	\fancyfoot{}
	\fancyfoot[RO,LE]{\thepage}
}

\usepackage{listings}
\usepackage{hyperref}
\hypersetup{pdfborder={0 0 0},
	colorlinks=true,
	linkcolor=black,
	citecolor=black,
	urlcolor=black}
\ifpdf
\usepackage[final]{pdfpages}
\else
\usepackage{calc}
\usepackage{breakurl}
\usepackage[nlwarning=false]{hypdvips}
\usepackage{backref}
\renewcommand*{\backref}[1]{}
\fi
\usepackage{bookmark}

\makeatletter
\renewcommand\@pnumwidth{20pt}
\makeatother

\makeatletter
\def\cleardoublepage{\clearpage\if@twoside \ifodd\c@page\else
    \hbox{}
    \thispagestyle{empty}
    \newpage
    \if@twocolumn\hbox{}\newpage\fi\fi\fi}
\makeatother \clearpage{\pagestyle{plain}\cleardoublepage}

\usepackage{color}
\usepackage{tikz}
\usepackage[explicit]{titlesec}
\newcommand*\chapterlabel{}
\titleformat{\chapter}[display]  
	{\normalfont\bfseries\Huge} 
	{\gdef\chapterlabel{\thechapter\ }}     
 	{0pt} 
 	  {\begin{tikzpicture}[remember picture,overlay]
    \node[yshift=-8cm] at (current page.north west)
      {\begin{tikzpicture}[remember picture, overlay]
        \draw[fill=black] (0,0) rectangle(35.5mm,15mm);
        \node[anchor=north east,yshift=-7.2cm,xshift=34mm,minimum height=30mm,inner sep=0mm] at (current page.north west)
        {\parbox[top][30mm][t]{15mm}{\raggedleft \rule{0cm}{0.6cm}\color{white}\chapterlabel}};  
        \node[anchor=north west,yshift=-7.2cm,xshift=37mm,text width=\textwidth,minimum height=30mm,inner sep=0mm] at (current page.north west)
              {\parbox[top][30mm][t]{\textwidth}{\rule{0cm}{0.6cm}\color{black}#1}};
       \end{tikzpicture}
      };
   \end{tikzpicture}
   \gdef\chapterlabel{}
  } 
\titlespacing*{name=\chapter,numberless}{-3.7cm}{83.2pt-\parskip}{-3.2pt+\parskip}
\titlespacing*{\chapter}{-3.7cm}{50pt-\parskip-\parskip}{30pt+\parskip+\parskip}
\titlespacing*{\section}{0pt}{13.2pt}{1em-\parskip}  
\titlespacing*{\subsection}{0pt}{13.2pt}{1em-\parskip}
\titlespacing*{\subsubsection}{0pt}{13.2pt}{1em-\parskip}
\titlespacing*{\paragraph}{0pt}{13.2pt}{1em-\parskip}

\newcounter{myparts}
\newcommand*\partlabel{}
\titleformat{\part}[display]  
	{\normalfont\bfseries\Huge} 
	{\gdef\partlabel{\thepart\ }}     
 	{0pt} 
 	  {\ifpdf\setlength{\unitlength}{20mm}\else\setlength{\unitlength}{0mm}\fi
	  \addtocounter{myparts}{1}
	  \begin{tikzpicture}[remember picture,overlay]
    \node[anchor=north west,xshift=-65mm,yshift=-6.9cm-\value{myparts}*20mm] at (current page.north east) 
      {\begin{tikzpicture}[remember picture, overlay]
        \draw[fill=black] (0,0) rectangle(62mm,20mm);   
        \node[anchor=north west,yshift=-6.1cm-\value{myparts}*\unitlength,xshift=-60.5mm,minimum height=30mm,inner sep=0mm] at (current page.north east)
        {\parbox[top][30mm][t]{55mm}{\raggedright \color{white}Part \partlabel \rule{0cm}{0.6cm}}};  
        \node[anchor=north east,yshift=-6.1cm-\value{myparts}*\unitlength,xshift=-63.5mm,text width=\textwidth,minimum height=30mm,inner sep=0mm] at (current page.north east)
              {\parbox[top][30mm][t]{\textwidth}{\raggedleft \rule{0cm}{0.6cm}\color{black}#1}};
       \end{tikzpicture}
      };
   \end{tikzpicture}
   \gdef\partlabel{}
  } 
\titlespacing*{\part}{11.06cm}{26.4pt-\parskip-\parskip}{0pt}

\usepackage{amsmath, amssymb}
\usepackage{amsfonts}
\usepackage{amsthm}

\usepackage[normalem]{ulem}
\usepackage{bm}
\usepackage{mathrsfs}
\usepackage[]{algorithm2e}
\usepackage{caption}
\usepackage{subcaption}
\usepackage{makecell}
\usepackage{multirow}
\usepackage{enumitem}
\usepackage{float}
\usepackage{xfrac}
\usepackage{tikz}
\usepackage{pgfplots}
\usepackage{algpseudocode}
\usepackage{balance}
\usepackage{wrapfig}
\usepackage{mathtools}

\usepackage[capitalize]{cleveref}
\crefname{section}{Sec.}{Secs.}
\Crefname{section}{Section}{Sections}
\crefname{appendix}{App.}{App.}
\Crefname{table}{Table}{Tables}
\crefname{table}{Tab.}{Tabs.}

\newtheorem*{theorem*}{Theorem}
\newtheorem{definition}{Definition}

\newtheorem*{proposition*}{Proposition}

\newtheorem*{conjecture*}{Conjecture}

\newcommand{\argmin}[1]{\underset{#1}{\text{argmin }}}
\newcommand{\argmax}[1]{\underset{#1}{\text{argmax }}}

\newcommand{\R}{\mathbb{R}}

\newcommand*{\MyToprule}{%
  \cmidrule[\heavyrulewidth]%
}
\newcommand*{\MyMidrule}{%
  \cmidrule%
}

\newcommand\blfootnote[1]{%
  \begingroup
  \renewcommand\thefootnote{}\footnote{#1}%
  \addtocounter{footnote}{-1}%
  \endgroup
}

\renewcommand{\epsilon}{\varepsilon}

\pgfplotsset{width=7cm,compat=1.16}
\usepgfplotslibrary{statistics}
\usetikzlibrary{pgfplots.groupplots}
\usetikzlibrary{plotmarks}
\usetikzlibrary{shapes}
\definecolor{ts1}{RGB}{ 0  0    0}    
\definecolor{ts2}{RGB}{127 127 127}  
\definecolor{ts3}{RGB}{229, 194, 36}  
\definecolor{ts4}{RGB}{238,113,27}  
\definecolor{ts5}{RGB}{19,219,30}  
\definecolor{ts6}{RGB}{0,58,236}  

\usepackage{xspace}
\def\etal{\emph{et al}.\xspace}
\def\eg{\emph{e.g}.\xspace}

\begin{document}
\setlength{\parindent}{0pt}
\setlength{\parskip}{0pt} 
\frontmatter
\includepdf{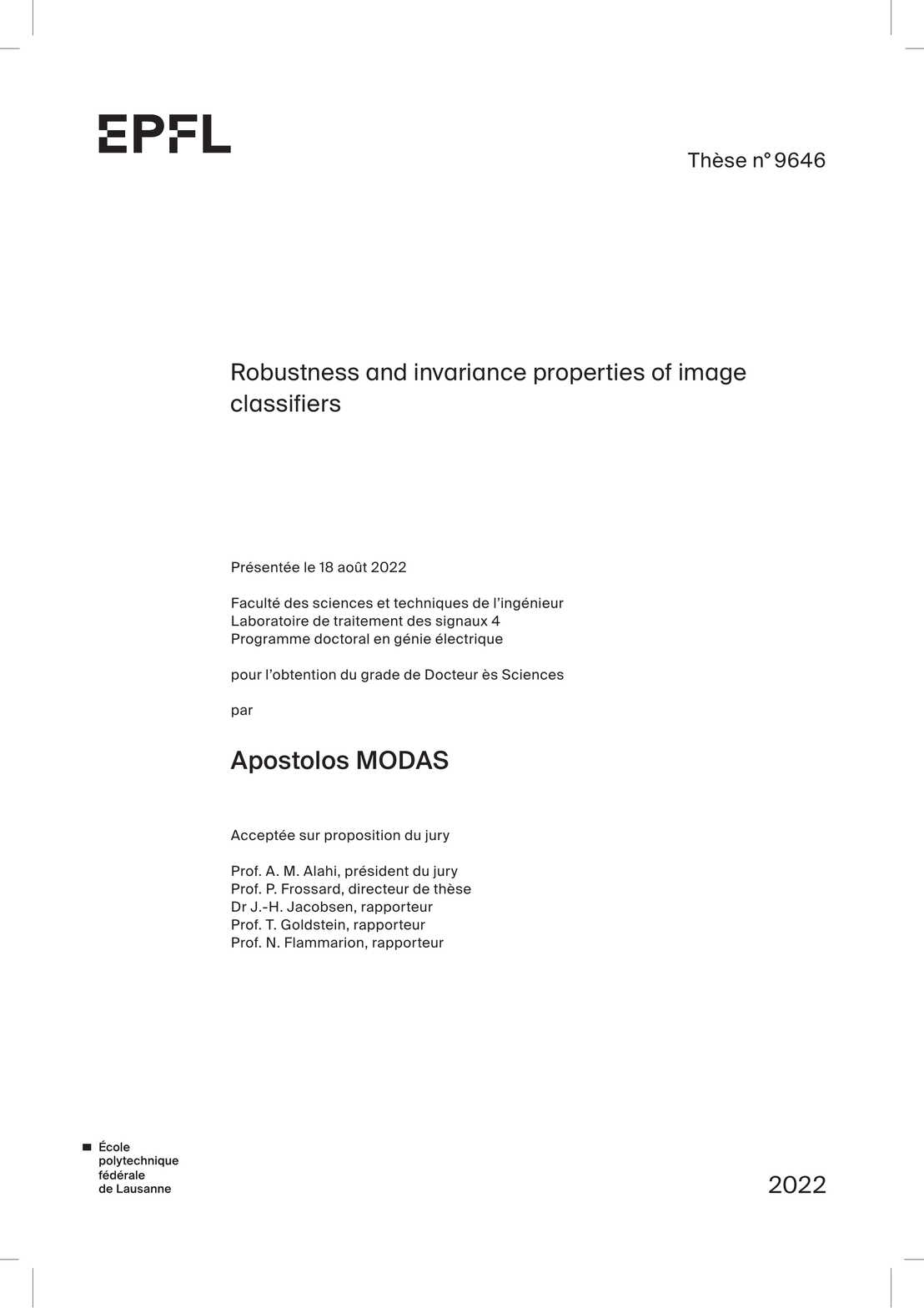}
\cleardoublepage
\thispagestyle{empty}

\vspace*{3cm}

\begin{raggedleft}
    	Philosophers have hitherto only interpreted the world in various ways. \\
    	The point is to change it. \\
     --- Karl Marx \\
\end{raggedleft}

\vspace{4cm}

\begin{center}
    To my beloved family\dots
\end{center}

\setcounter{page}{0}
\chapter*{Acknowledgements}
\markboth{Acknowledgements}{Acknowledgements}
\addcontentsline{toc}{chapter}{Acknowledgements}

First of all, I would like to express my sincere gratitude to my supervisor Prof. Pascal Frossard. Not only he trusted me and gave me the opportunity to pursue my PhD with him, but he also guided and supported me with patience even during the hardest moments of my life. I will always be grateful!

I would also like to thank the members of my thesis committee, Prof. Alexandre Alahi, Prof. Nicolas Flammarion, Prof. Tom Goldstein, and Dr. J\"orn-Henrik Jacobsen for their fruitful discussions and feedbacks.

I am extremely glad -- and lucky -- to have met and worked with Seyed and Guillermo. I definitely enjoyed every single minute of our collaboration. Their fascinating knowledge and intelligence helped me to evolve both as a researcher and as a person, and I am not sure how much of this thesis would exist without their continued support.

I would like to extend my sincere thanks to Mathieu Sinn and Beat Buesser for their hospitality during my internship at IBM Ireland. I also want to thank Prof. Andrea Cavallaro from QMUL for our 3-year collaboration. I really enjoyed our many discussions, and his constant motivation has always been a great support.

I also sincerely thank all the current and former LTS4 labmates \'Ad\'am, Ahmet, Alessandro, Arun, Bastien, Beril, Cl\'ement, Cl\'ementine, Dorina, Eda, Ersi, Guille, Harshitha, Hermina, Isabela, Javier, Jelena, Mattia, Mireille, Nikos, Ortal, Renata, Roberto, Sahar, Seyed, Stefano, William, Yamin. My special thanks go to Mireille for sharing the office with me during a big part of my PhD years. It was really a pleasure to have such a kind and understanding friend as an officemate. I also want to thank our administrative Anne for her genuine help and for her incredible efficiency.

A very big thank you to my good friends Christos, Dimitris, George, Haris, Irene, Kostas, Maksym, Manos, Marios, Nikos, Sotiris, Thanos, Vaggelis, Victoria and Vlasis for giving me strength during all these years away from home.

Finally, I am indebted to my family for their unconditional love and support, and for always being there for me since the beginning. And I am especially grateful to Meli for her support and sacrifices during all these years. To all of you, I know that I cannot give back all those things you have offered me.

\bigskip
 
\noindent\textit{Lausanne, June 1, 2022}
\hfill Apostolos Modas


\cleardoublepage
\chapter*{Abstract}
\markboth{Abstract}{Abstract}
\addcontentsline{toc}{chapter}{Abstract (English/Français)} 

Deep neural networks have achieved impressive results in many image classification tasks. However, since their performance is usually measured in controlled settings, it is important to ensure that their decisions remain correct when deployed in noisy environments. In fact, deep networks are not robust to a large variety of semantic-preserving image modifications, even to imperceptible image changes -- known as adversarial perturbations -- that can arbitrarily flip the prediction of a classifier. The poor robustness of image classifiers to small data distribution shifts raises serious concerns regarding their trustworthiness. To build reliable machine learning models, we must design principled methods to analyze and understand the mechanisms that shape robustness and invariance. This is exactly the focus of this thesis.

First, we study the problem of computing sparse adversarial perturbations, and exploit the geometry of the decision boundaries of image classifiers for computing sparse perturbations very fast. We evaluate the robustness of deep networks to sparse adversarial perturbations in high-dimensional datasets, and reveal a qualitative correlation between the location of the perturbed pixels and the semantic features of the images. Such correlation suggests a deep connection between adversarial examples and the data features that image classifiers learn.

To better understand this connection, we provide a geometric framework that connects the distance of data samples to the decision boundary, with the features existing in the data. We show that deep classifiers have a strong inductive bias towards invariance to non-discriminative features, and that adversarial training exploits this property to confer robustness. We demonstrate that the invariances of robust classifiers are useful in data-scarce domains, while the improved understanding of the data influence on the inductive bias of deep networks can be exploited to design more robust classifiers. 

Finally, we focus on the challenging problem of generalization to unforeseen corruptions of the data, and we propose a novel data augmentation scheme that relies on simple families of max-entropy image transformations to confer robustness to common corruptions. We analyze our method and demonstrate the importance of the mixing strategy on synthesizing corrupted images, and we reveal the robustness-accuracy trade-offs arising in the context of common corruptions. The controllable nature of our method permits to easily adapt it to other tasks and achieve robustness to distribution shifts in data-scarce applications.

Overall, our results contribute to the understanding of the fundamental mechanisms of deep image classifiers, and pave the way for building more reliable machine learning systems that can be deployed in real-world environments.

\paragraph{Keywords:} image classification, robustness, invariance, adversarial examples, distribution shifts, sparse perturbations, image transformations, data augmentation, decision boundary, deep learning, convolutional neural networks.

\begin{otherlanguage}{french}
\cleardoublepage
\chapter*{Résumé}
\markboth{Résumé}{Résumé}

Les réseaux de neurones profonds ont obtenu des résultats impressionnants dans de nombreuses tâches de classification d'images. Cependant, comme leurs performances sont généralement mesurées dans des environnements contrôlés, il est important de s'assurer que leurs décisions restent correctes lorsqu'ils sont déployés dans des environnements bruyants. En fait, les réseaux profonds ne sont pas robustes à une grande variété de modifications d'images préservant la sémantique, même à des changements d'images imperceptibles -- connus sous le nom de perturbations adverses -- qui peuvent arbitrairement faire basculer la prédiction d'un classificateur. La faible robustesse des classificateurs d'images aux petits changements de distribution des données soulève de sérieuses inquiétudes quant à leur fiabilité. Pour construire des modèles d'apprentissage automatique fiables, nous devons concevoir des méthodes fondées sur l'analyse et la compréhension des mécanismes qui façonnent la robustesse et l'invariance. C'est l'objet de cette thèse.

Tout d'abord, nous étudions le problème du calcul de perturbations adverses éparses et nous exploitons la géométrie des limites de décision des classificateurs d'images pour calculer très rapidement des perturbations éparses. Nous évaluons la robustesse des réseaux profonds aux perturbations adverses dispersées dans des ensembles de données à haute dimension, et nous révélons une corrélation qualitative entre l'emplacement des pixels perturbés et les caractéristiques sémantiques des images. Cette corrélation suggère une connexion profonde entre les exemples adverses et les caractéristiques des données que les classifieurs d'images apprennent.

Pour mieux comprendre cette connexion, nous fournissons un cadre géométrique qui fait le lien entre la distance des échantillons de données à la frontière de décision et les caractéristiques existant dans les données. Nous montrons que les classifieurs profonds ont un fort biais inductif vers l'invariance des caractéristiques non-discriminatives, et que l'entraînement contradictoire exploite cette propriété pour conférer de la robustesse. Nous démontrons que les invariances des classifieurs robustes sont utiles dans les domaines où les données sont rares, tandis que la meilleure compréhension de l'influence des données sur le biais inductif des réseaux profonds peut être exploitée pour concevoir des classifieurs plus robustes. 

Enfin, nous nous concentrons sur le problème difficile de la généralisation aux corruptions imprévues des données, et nous proposons un nouveau schéma d'augmentation des données qui s'appuie sur des familles simples de transformations d'images à entropie maximale pour conférer de la robustesse aux corruptions courantes. Nous analysons notre méthode et démontrons l'importance de la stratégie de mélange pour synthétiser les images corrompues, et nous révélons les compromis entre robustesse et précision dans le contexte des corruptions courantes. La nature contrôlable de notre méthode permet de l'adapter facilement à d'autres tâches et d'atteindre la robustesse aux changements de distribution dans les applications où les données sont rares.

Dans l'ensemble, nos résultats contribuent à la compréhension des mécanismes fondamentaux des classifieurs d'images profonds et ouvrent la voie à la construction de systèmes d'apprentissage automatique plus fiables pouvant être déployés dans des environnements réels.

\paragraph{Mots clés:} classification d’images, robustesse, invariance, exemples adverses, changements de distribution, perturbations éparses, transformations d’images, augmentation des données, limite de décision, apprentissage profond, réseaux de neurones convolutifs.

\end{otherlanguage}


\cleardoublepage
\pdfbookmark{\contentsname}{toc}
\tableofcontents

\setlength{\parskip}{1em}

\mainmatter
\chapter{Introduction}
\label{ch:introduction}

\begin{raggedleft}
    \textit{``It ain’t what you don’t know that gets you into trouble. \\
    It’s what you know for sure that just ain’t so.''} \\
    --- Mark Twain \\
    
\end{raggedleft}
\vspace*{3cm}

In recent years, deep neural networks have become the state-of-the-art in most machine learning benchmarks with the emergence of deep learning. Driven by the vast amounts of available data, deep learning systems have achieved outstanding performance in a wide range of applications, especially in the field of image classification. A standard way of assessing the ``outstanding performance'' of a classifier, is through its generalization. In practice, the generalization of a classifier is usually determined by its test accuracy, which is the performance on some held-out (test) data that have never been observed during training, but are typically assumed to come from the same distribution as the training data. 

When deploying deep learning models in the real world, though, we expect to face very different environments than those of the training data, which sometimes can be noisy or even hostile. Building classifiers that are able to generalize to such conditions is crucial, especially for safety-critical applications like autonomous driving or biomedical imaging. Therefore, it is of utmost importance that the decisions of the classifiers remain robust, even in the presence of worst-case scenarios.

However, deep classifiers are actually brittle and far from robust, and their generalization performance significantly degrades when evaluated in conditions that are slightly different from the training ones. In particular, the decisions of deep classifiers can easily change through small semantic-preserving modifications of their images, even if such changes would not affect the human perception in general. 

Building robust deep networks that are invariant to small modifications of their data is a challenging problem, and a large body of research has focused on developing a variety of techniques for improving the robustness of image classifiers. Nevertheless, the problem of creating classifiers that are able to robustly generalize under different settings is far from being solved, while the exact mechanisms behind the profound vulnerability of deep networks are still not well understood. In this thesis, we propose novel algorithms for evaluating and understanding the robustness and invariance properties of deep networks, and build upon our insights to design methodologies for improving the robustness of image classifiers to distribution shifts and nuisances of their data.

\section{Robustness of deep networks}
\label{sec:introduction-robustness-of-deep-networks}

In general, the low robustness of deep networks to different types of data distribution shifts can be observed in various scenarios and settings. For instance, deep classifiers struggle to generalize to images that have been slightly modified by common types of distortions that may occur during the acquisition or processing of the images (i.e., blur, color jitter etc)~\cite{Hendrycks2019Benchmarking}. In the spatial domain, carefully crafted shifts on the image pixels are enough to cause significant performance drops~\cite{Fawzi2015Manitest}. In the spectral domain, the predictions of image classifiers are more sensitive to small perturbations in the low-frequency part of the image spectrum rather than the high-frequency one~\cite{Yin2019FourierPerspective}. Furthermore, image classifiers tend to rely on spurious features of the image background, and hence when image objects appear on backgrounds that are typically presented in different object categories (i.e., a fish on a grass field), is enough to force the network to an erroneous decision (i.e., rabbit instead of fish)~\cite{Xiao2021NoiseOrSignal}. Finally, image classifiers do not exhibit low robustness only in cases where the images are somehow manipulated or distorted, but they can also struggle with completely new images, even if they are collected in the exact same way and from a similar distribution as the one used for training~\cite{Recht2019DoImagenet,Hendrycks2021NaturalAdversarial}. 

One way to improve the robustness of the classifiers to distribution shifts or distortions, is by increasing the variability of the training data. The hope is that, the more samples the classifier observes, the better knowledge it obtains about the data distribution; hence, it is more likely to generalize to new samples. In this regard, an obvious approach for improving the robustness of deep classifiers is to increase the amount of training samples~\cite{Taori2020MeasuringRobustnessToNatural}. However, this technique might be impractical, since training on huge datasets requires a lot of computational power, and for many tasks the available training data can be quite scarce. In practice, one of the most common techniques for artificially increasing the amount and variability of training data is the so-called data augmentation, where one expects that the classifier becomes invariant -- to some extent -- to the transformations used to generate the augmented data during training~\cite{Shorten2019SurveyDataAugm}. That is, the decision of the classifier does not vary when test images are manipulated with the transformations used during training. Note here that the choice of transformations used during data augmentation might be crucial for the overall robustness of the model. For instance, if we want the classifier to be robust to image rotations, then applying a random rotation in $(0^{\circ},10^{\circ}]$ on the training images might not be sufficient to achieve robustness to rotations that are larger than $10^{\circ}$. 

In general, although data augmentation or the use of additional data might work in practice for specific tasks, these methods are not definitive solutions for building robust classifiers. Ideally, given access to a finite amount of training data, we would still want  a robust classifier  to maintain its predictions when new data come from a slightly shifted distribution, or in the presence of distortions that do not change the semantics of the image. This implies that, for achieving robust and invariant image classification, the classifiers should ideally learn the underlying “concepts” that characterize the class of an image, rather than learning spurious image features that provide generalization only to specific test sets.

\section{Adversarial robustness}
\label{sec:introduction-adversarial-robustness}

Building classifiers with the desired invariance properties is however far from trivial. In fact, the robustness of deep neural networks is heavily challenged by the existence of the so-called \emph{adversarial examples}~\cite{Szegedy2014Intriguing}. These are data samples that have been modified by carefully-crafted, semantic-preserving perturbations that are often imperceptible to the human eye, but can change the prediction of the classifier to any arbitrary class. 
Adversarial examples represent one of the most intriguing phenomenon in the robustness literature, since their existence is not only constrained to image classifiers~\cite{Goodfellow2015Explaining,Moosavi2016DeepFool,Moosavi2017Universal}, but extends to virtually any application of deep learning such as autonomous driving~\cite{Modas2020TowardsRobustSensing}, malware detection~\cite{Grosse2017AdversarialMalware}, natural language processing~\cite{Azantot2018LanguageAdversarial}, speech recognition~\cite{Carlini2018AudioAdversarial} and reinforcement learning~\cite{Huang2017AdversarialAttackPolicies}.

\begin{figure}[t]
    \centering
    \includegraphics[width=0.8\columnwidth]{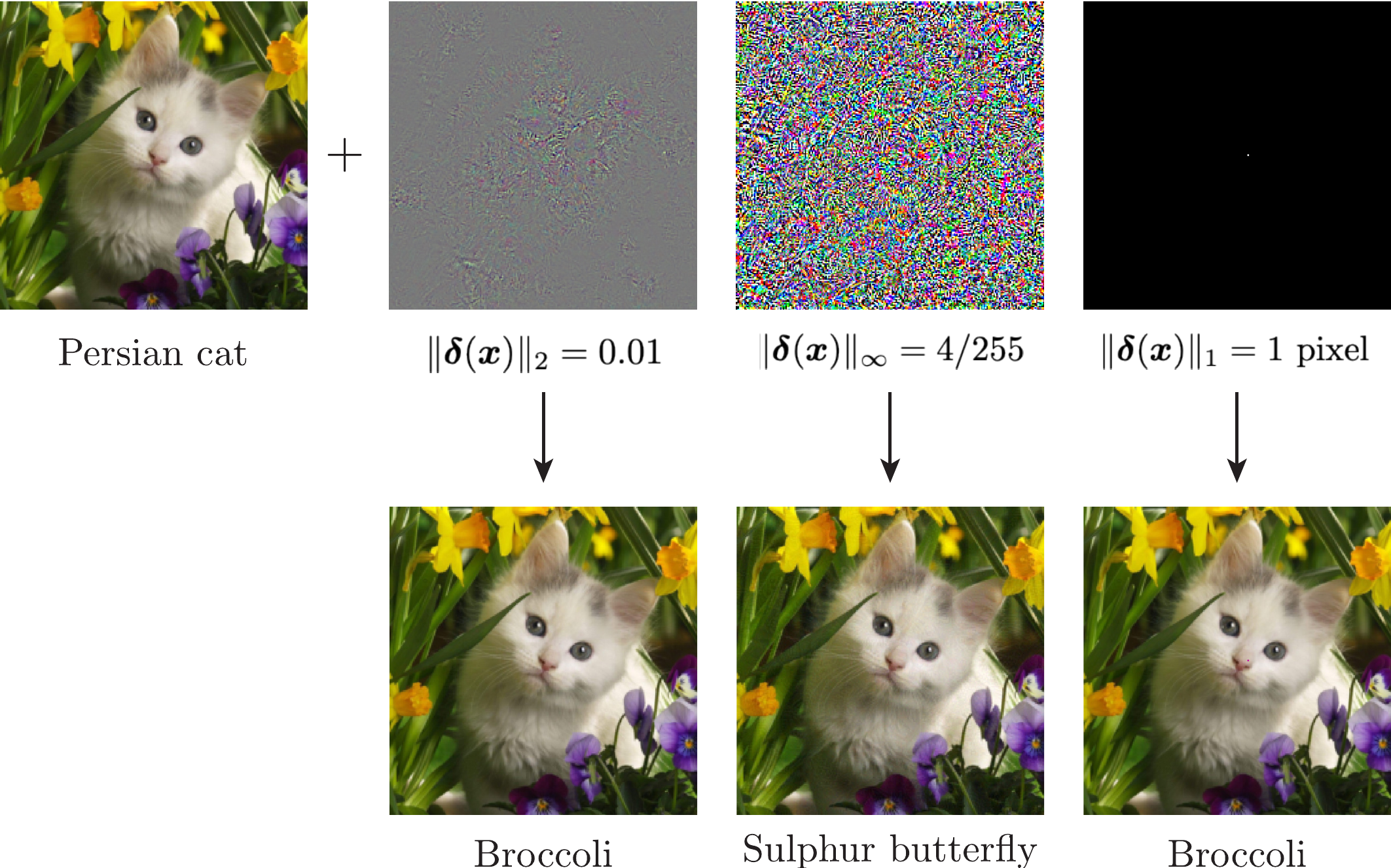}
    \caption{Illustration of different additive adversarial perturbations ($\ell_1, \ell_2$ and $\ell_\infty$) (top) and the corresponding adversarial examples (bottom) that fool a deep neural network. The norm of each perturbation is indicated below the corresponding image, except for the $\ell_1$ perturbation (sparse) where the number of perturbed pixels is provided. The resulting misclassified labels are shown below each adversarial example. In all cases the adversarial example is hardly distinguishable by a human observer. The original image is taken from the web.}
    \label{fig:adv_examples}
\end{figure}

Formalizing adversarial examples is an ill-posed problem, since there can exist multiple ways for defining and computing them depending on the data they are applied to, while the notion of perceptibility is subjective and can be arbitrary. Nevertheless, it is common in the literature to compute the $\ell_p$-norm of the perturbations as a perceptibility proxy. The goal then is to find adversarial perturbations of a small $\ell_p$-norm. An illustration of adversarial examples generated with adversarial perturbations of different $\ell_p$ norms are shown in \cref{fig:adv_examples}.

The simplest and most common adversarial perturbations are the additive ones. 
Let $\bm{x}\in\R^D$ be a $D$-dimensional input, and $f:\R^D\to \R^K$ be the final layer of a neural network (i.e., logits), such that, for any input $\bm{x}$, $F(\bm{x})=\text{argmax}_k f_k(\bm{x})$ represents the decision of that network, where $f_k(\bm{x})$ denotes the component of $f(\bm{x})$ that corresponds to the $k$th class.
Formally, for a given data sample $\bm x$ along with its associated label $y$, an additive $\ell_p$-norm adversarial perturbation $\bm\delta\in\R^D$ is defined as the one that maximizes the classifier's loss $\mathcal{L}_\theta$ within an $\ell_p$-ball of radius $\epsilon$ around $\bm x$
\begin{equation}
\begin{split}
    \argmax{\bm\delta} & \mathcal{L}_\theta(\bm x+\bm\delta, y) \\
    \text{s.t. } & \|\bm\delta\|_p \leq \epsilon \\
    & \bm\delta\in\mathcal{C},
\end{split}
\label{eq:eps-adv-pert}
\end{equation}
where $\theta$ are the parameters of the classifier and $\mathcal{C}$ denotes a general set of constraints, e.g., the perturbed image is within a valid pixel value range $\mathcal{C}=\{\bm\delta: \bm x + \bm\delta \in [0,1]^D\}$. Alternatively, one can also define the \emph{minimal} additive $\ell_p$-norm adversarial perturbation as the smallest additive perturbation that changes the decision of the classifier
\begin{equation}
\begin{split}
    \argmin{\bm\delta} & \|\bm\delta\|_p \\
    \text{s.t. } & F(\bm x) \neq F(\bm x + \bm\delta) \\
    & \bm\delta\in\mathcal{C}.
\end{split}
\label{eq:minimal-adv-pert}
\end{equation}
In fact, this definition of minimal adversarial perturbations encapsulates the notion of distance between a data sample and the decision boundary of a classifier, and has been used to study the local geometry of the decision boundaries~\cite{Fawzi2017GeometricPerspective,OrtizJimenez2020optimism}. Finally, we note that adversarial perturbations are not restricted to the additive $\ell_p$-norm definitions of \cref{eq:eps-adv-pert} and \cref{eq:minimal-adv-pert}, but can rather extend for instance to geometric~\cite{Xiao2018Spatially,Kanbak2018GeometricRobustness}, functional~\cite{Laidlaw2019Functional}, or even physical perturbations~\cite{Eykholt2018RobustPhysical,Athalye2018SynthesizingRobust}.

Building classifiers that are robust against adversarial perturbations is still an open problem. Surprisingly, performing standard data augmentation does not improve adversarial robustness, even when it boosts robustness to specific types of distortions and transformations. Instead, one might need to adapt data augmentation in the worst-case settings and perform \emph{adversarial training}~\cite{Moosavi2016DeepFool,Madry2018TowardsDeepLearning}. Adversarial training is a data augmentation method that replaces the clean images with their adversarial examples during training, and has been empirically shown to consistently result into more robust classifiers. Nevertheless, performing adversarial training can be computationally expensive, and it also comes at the cost of building classifiers that may perform worse on their standard test sets~\cite{Fawzi2018AnalysisOfClassifiers,Tsipras2019RobustnessOddsAccuracy}. Finally, adversarial training typically results into classifiers that are robust mostly on the perturbations they were trained on (e.g., perturbations of a specific $\ell_p$ norm)~\cite{Maini2020RobustnessUnion}, while its contribution to the robustness to non-adversarial distortions might not be significant~\cite{kireev2021}.


\section{Towards better understanding of deep networks}
\label{subsec:introduction-beyond-adversarial}

The profound gap between human and machine perception, along with the low robustness of deep neural networks, raises serious concerns regarding the implications on the security, safety, and fairness of deep learning systems. In order to integrate deep learning in sensitive tasks, such robustness vulnerabilities should be addressed. Therefore, it is imperative that we focus on understanding the mechanisms that govern the robustness properties of deep networks, and eventually design principled methods for building accurate and robust classifiers. 

The main theme of this thesis is to exploit the connection between adversarial robustness and the geometry of deep networks~\cite{Moosavi2019PhDThesis,OrtizJimenez2020optimism} as a proxy for understanding some fundamental properties of deep neural networks, and build upon our intuitions to design novel methodologies that confer robustness to distribution shifts of the data. Our main contribution is a better understanding of the role of data in the inductive biases of (robust) deep neural networks, which enables us to design new methods for improving the invariance properties of image classifiers and achieve better robustness to different types of distribution shifts. In what follows, we provide more details regarding the individual contributions of this thesis.

First, we focus on the computation of sparse adversarial perturbations. In general, evaluating the robustness of deep networks to perturbations in non-standard $\ell_p$ regimes, for instance sparse perturbations with $p\in\{0,1\}$, can be useful for identifying salient image features, due to the localized structure of the perturbations. However, generating sparse adversarial perturbations can be computationally expensive, and existing methods~\cite{Papernot2016JSMA,CarliniWagner2017,Su2019OnePixel} are impractical in large datasets. As a first contribution of this thesis, we provide a very fast, geometry-inspired sparse attack that exploits the low mean curvature of the decision boundaries to generate adversarial perturbations. This efficient algorithm enables us to thoroughly analyze the transferability and spatial properties of sparse perturbations, and provides empirical insights on the correlation between the data features and adversarial perturbations.

As a second step, we study the relationship between adversarial perturbations and the features of the data. The features of real datasets are not known a priori, and hence current literature~\cite{Jetley2018WithFriends,Ilyas2019AreNotBugs} relies either on experimental evaluations or on synthetic examples in order to investigate the connections between adversarial perturbations and data features. It is still not established which mechanisms are responsible for creating these connections, and it is unclear if this phenomenon depends on the data, the network architecture, or the learning algorithm. Propelled again by the geometric properties of adversarial examples, we propose a new geometric framework that connects the norm of minimal adversarial perturbations (i.e., distance to the classifier decision boundary) with the data features. By carefully manipulating the input data, we demonstrate that deep classifiers have a strong inductive bias towards invariance to non-discriminative features. In fact, we show that adversarial training actually exploits this property to build more robust classifiers. Furthermore, our insights explain why some methods for crafting adversarial examples are more efficient if they constrain the perturbations to the low-frequency subspaces of the Fourier basis: the reason is that the discriminative features of standard image datasets are aligned with low-frequency directions. Finally, we also demonstrate that the invariance properties of robust classifiers are beneficial in tasks where the available training data are scarce.

Last, we extend our focus to the more general case of robustness to non-adversarial perturbations of the data. In particular, we deal with the problem of robustness to common corruptions, a term that generally refers to typical image distortions (i.e., blur, color jitter, brightness variations, random noise etc.) that can occur during acquisition, storage, or processing of the images. In these settings, achieving robustness is more complex than merely inducing invariance through simple data feature manipulations. At the same time, defining such ``common corruptions'' is an ill-posed problem. For achieving robustness to common corruptions, prior works~\cite{Hendrycks2020AugMix,Hendrycks2021TheManyFaces,Calian2022DefendingImageCorruptions} have built complex data augmentation strategies, combining multiple methods to enrich the training data. These works typically introduce intricate design choices or heuristics, and it is hard to understand which elements of the methods are indeed crucial for improving robustness. We formulate a set of primitive image transformations in the spatial, color and spectral domains, and propose a systematic data augmentation scheme for improving the robustness of deep classifiers to common corruptions of their data. Our method achieves state-of-the-art robustness in multiple benchmarks, while its simplicity permits to perform an in-depth analysis of robustness in the context of common corruptions. In particular, we highlight the importance of deploying a mixing strategy during the generation of the augmentation instances, and analyze the potential robustness-accuracy trade-offs and the benefits of generating the data augmentations during training (on-line). Finally, we demonstrate that our method can serve as an off-the-shelf solution for achieving robustness to distribution shifts that extend beyond the concept of common corruptions.

It should be noted that, although our focus in this thesis is on images, our geometric framework for connecting the data features to the decision boundaries, as well as the basic principles of our novel augmentation strategy can be easily extended to other modalities of data.

\section{Thesis outline}
\label{sec:introduction-thesis_outline}


The rest of thesis is organized as follows:

In Chapter \ref{ch:related-work}, we review some of the prior works related to the problem of (i) evaluating the robustness of image classifiers, (ii) improving the robustness of classifiers to adversarial examples and distribution shifts, and (iii) studying the connections between adversarial robustness and the features learned by deep classifiers.

In Chapter \ref{ch:sparse-adversarial-perturbations}, we study the problem of efficiently generating sparse adversarial perturbations. In particular, we design a geometry-inspired algorithm that is by orders of magnitude faster than existing methods and efficiently scales to high-dimensional datasets. Our empirical analysis sheds new light on the transferability of sparse adversarial perturbations, and on the connections between the image semantics and the features that deep networks learn.

In Chapter \ref{ch:hold-me}, we develop a novel methodology to characterize the relationship between the distance of a set of samples to the decision boundary, and the discriminative features of the dataset that are used by a classifier. We demonstrate that convolutional neural networks are invariant to non-discriminative features of a dataset. We further show that the decision boundary is very sensitive to the position of the training samples, and that adversarial training exploits this sensitivity and invariance bias to build robust classifiers. Finally, we demonstrate that the invariance properties of robust classifiers prevent overfitting in the non-classical task of estimating the amount of content within a container in scarce data regimes.

In Chapter \ref{ch:prime}, we study the general problem of robustness to common corruptions of the images. We formulate a new model for semantically-preserving image corruptions, and build on basic concepts to characterize the notions of transformation strength and diversity using a few transformation primitives. We propose a general data augmentation scheme that relies on simple yet rich families of max-entropy image transformations. Our method tops the current baselines on different common corruption datasets, while its simplicity makes it an effective tool for understanding common corruption robustness, and build classifiers with improved out-of-distribution generalization properties. In particular, we demonstrate that our method can easily be tuned for the context of classifying the filling level within a container, and generate augmentations with properties that resemble those of test-time distribution shifts.

Finally, in Chapter \ref{ch:conclusion} we summarize the main outcomes of this thesis, and outline some of the potential future research directions.
\chapter{Related work}
\label{ch:related-work}

In this chapter, we review some of the relevant works from the literature that are linked to the problems studied in this thesis. In particular, in Section \ref{sec:related-robustness-distribution-shifts} we summarize the methods for evaluating and improving the robustness of deep neural networks to distribution shifts, while in Section \ref{sec:related-robustness-adversrial-perturbations} we focus on robustness under adversarial settings. Finally, in Section \ref{sec:related-understanding-deep-learning} we review related works that connect the robustness of deep networks to different other properties of deep learning.

\section{Robustness to distribution shifts}
\label{sec:related-robustness-distribution-shifts}

\subsection{Evaluating robustness to distribution shifts}
\label{susbec:related-shifts-eval}

\textbf{Natural distribution shifts}\quad
Reliable classification under distribution shifts has received a growing amount of interest in the field of machine learning~\cite{Candela2009DatasetShiftInML}. Especially in the context of image classification~\cite{MorenoTorres2012UnifyingViewOnDatasetShift}, measuring robustness to natural distribution shifts (i.e., not synthetically induced through perturbations or transformations) that can arise in the real world is very important for a broad deployment of machine learning models. In fact, the authors in~\cite{Recht2018DoCIFAR10,Recht2019DoImagenet} showed that classifiers that achieve state-of-the-art performance on standard benchmarks practically overfit to the given test sets of such benchmarks. In particular, they exhibit a significant accuracy drop on newly collected unseen images (new test set), despite being sampled from a distribution that is very similar to the one of the original test set. This phenomenon can be even more pronounced, as shown in~\cite{Hendrycks2021NaturalAdversarial}, if the new test set is explicitly selected (adversarially) to cause extreme accuracy drops (up to $90\%$). Having a unified approach for evaluating the robustness of classifiers to natural distribution shifts is very important. To this end, the authors in~\cite{Taori2020MeasuringRobustnessToNatural} recently defined a very large testbed, and demonstrated that classifiers trained on more diverse data typically achieve better robustness. At the same time, they also demonstrated that robustness to synthetic shifts (i.e., through artificial perturbations or transformations) does not necessarily imply natural robustness.

\textbf{Common corruptions and nuisances}\quad
One of the most broadly studied distribution shifts are those resulting from the different types of visual distortions, or nuisances, that can synthetically/artificially happen on the images (i.e., during acquisition or processing of the images, or through artificial occlusions). Note, that, it is important to focus on cases where the distortions do not alter the semantic information of the images. For instance, although deep networks are relatively robust to small random noise~\cite{Fawzi2016RobustnessRandomNoise,Franceschi2018RobustnesUniform}, they can still change their decision for larger random noise regimes, despite the fact that the image semantics are preserved. In particular, as shown in~\cite{Dodge2017StudyAndComparisonOfHuman,Geirhos2018GeneralisationInHumans}, although the human visual system can be quite robust to different visual distortions (e.g., strong Gaussian noise or blurs), deep networks are not. And even if the classifiers are trained (i.e., through data augmentation) to be robust to some specific distribution shift (i.e., distortions introduced with Gaussian noise), they tend to overfit to that specific shift and perform poorly on other type of distortion types~\cite{Geirhos2018GeneralisationInHumans}.

Apart from the different types of additive noise or blur~\cite{Dodge2016UnderstandingImageQuality,Vasiljevic2016ExaminingImpactBlur}, the generalization of deep classifiers can also be affected by the existence of different types of nuisances in the data distribution. For instance, deep classifiers can exhibit a very low robustness to slight combinations of translations and rotations of the images~\cite{Fawzi2015Manitest,Engstrom2019Exploring}. 
In general, measuring the robustness of a classifier on every possible corruption or nuisance is an ill-posed problem that lacks a formal description. For this reason, the research community has developed multiple standardized benchmarks to measure the robustness of classifiers to different distribution shifts, such as (i) common corruptions (e.g., noises, blurs, weather effects, digital transforms, spectral transforms)~\cite{Hendrycks2019Benchmarking,Mintun2021Cbar}, (ii) naturally captured blurry images~\cite{Hendrycks2021TheManyFaces}, or (iii) visual artistic renditions~\cite{Hendrycks2021TheManyFaces}.

\textbf{Inductive and distribution biases}\quad
Beyond standard corruptions or nuisances, deep networks might have problems generalizing to other types of distribution shifts, where one should not consider what the networks ``have not seen'', but rather what they ``have already seen'' or how they learn to generalize. Hence, we have to focus on the different types of (inductive) biases that deep classifiers have, or inherit from the data. For instance, the large accuracy drop on newly collected unseen images (new test set) that was observed in~\cite{Recht2019DoImagenet} has been later found to be mainly caused by some statistical bias introduced during the collection/replication of the new dataset~\cite{Engstrom2020StatisticalBias}. Another example of bias that can cause a generalization drop is the image resolution discrepancy between train and test time~\cite{Touvron2019FixingTrainTest}: if random resize and crop applied during training generate images where the objects are larger, then the classifier might not perfectly generalize to test images of smaller objects.

Apart from data- or user-induced biases, the inductive bias of the architecture and/or the learning algorithm can also cause problems in generalizing to different distribution shifts. The authors in~\cite{Geirhos2018TextureBias} showed that CNNs are more biased towards the texture rather than the shape of visual objects, and that they exhibit a significant accuracy drop when the texture changes or is totally absent (i.e., edge-based images or silhouettes). Furthermore, the authors in~\cite{Xiao2021NoiseOrSignal} showed that deep networks learn spurious background features rather than focusing on the actual object, hence irrelevant background changes result in quite low accuracy. Additionally, the authors in~\cite{Yin2019FourierPerspective} showed that Gaussian data augmentation and adversarial training, bias the model towards low-frequency information, which increases its robustness to high-frequency changes but makes it more vulnerable to low-frequency distortions. 
Furthermore, the authors in~\cite{OrtizJimenez2020NADs} and~\cite{OrtizJimenez2021Underspecification} showed that deep classifiers have a very strong inductive bias on fitting the data along specific directions, and that they cannot generalize to data where the information is not aligned with these directions. Finally, a recent study in~\cite{Geirhos2020Shortcut} has assigned the low robustness properties of deep networks to the problem of shortcut learning, where the classifier learns some ``easy'' features (i.e., background information) for specific objects, but those features are not representative and generalizable for the actual object class.

\subsection{Improving robustness to distribution shifts}
\label{susbec:related-shifts-improving}

\textbf{Additional training data}\quad
One straightforward approach to improve the robustness of deep networks to natural distribution shifts, consists in increasing the size of the training set by collecting new samples. However, this technique might be impractical, since training on huge datasets requires a lot of computational power, while for many tasks the amount of available data can be quite scarce. In particular, the authors in~\cite{Xie2020SelfTraining} showed that exploiting unlabelled data in combination with knowledge distillation can improve the generalization of a classifier to natural distribution shifts and corruptions. Furthermore, the authors in~\cite{Taori2020MeasuringRobustnessToNatural} demonstrated through a big testbed that using more data was the only technique that consistently led to better robustness on multiple distribution shifts. Interestingly, though, they observed that some models did not have any robustness benefits, since the are limited in exploiting further information from additional data (i.e., due to limited capacity).

\textbf{Data augmentation}\quad
The most common technique for improving the robustness to image distortions/transformations and nuisances, is to artificially increase the size and variability of the training set through data augmentation. The most standard augmentation is to apply a random horizontal flip and crop on the training image, in order to train classifiers so that they are invariant to the horizontal orientation and location of the object, which can also be beneficial for handling occlusions. For the latter, typical methods randomly mask part of the input~\cite{DeVries2017CutOut,Zhong2020RandomErasing} in order to force the network to focus on different image features and become robust to occlusions. Other methods that prevent the network from overfitting to specific image features replace parts of the image with crops of another image~\cite{Takahasi2018RICAP,Yun2019CutMix} or synthesize a new image as an interpolation of two images~\cite{Zhang2018Mixup,Guo2019MixUpAsLocally}. Note here that such techniques necessarily require the use of soft labels during classification. Also, the texture-shape bias introduced in~\cref{susbec:related-shifts-eval} can be avoided by training the classifier on a ``'stylized'' version of the dataset, such that the classifier becomes invariant to the actual texture of the object~\cite{Geirhos2018TextureBias}.

Another line of research has managed to increase the generalization performance of image classifiers by exploiting during training a set of different spatial (e.g., translations, rotations), color (e.g., brightness), and spectral (e.g., sharpness) image transformations~\cite{Cubuk2019Autoaugment,Cubuk2020RandAugment}. When properly combined with a mixing strategy, such transformations during training have been later shown to be quite beneficial in improving the robustness of classifiers to different corruption benchmarks~\cite{Hendrycks2020AugMix}; especially if the mixing is performed in an adversarial (worst-case) way~\cite{Wang2021AugMax}. Nevertheless, such transformations work well on small benchmarks (i.e., CIFAR-10) but do not perform equally well on higher-dimensional datasets. For this reason, the research community has proposed to address these limitations (i) either with more complex methods that use large autoencoders for generating more diverse augmentations~\cite{Hendrycks2021TheManyFaces,Calian2022DefendingImageCorruptions} or ``denoise'' the data~\cite{Kim2021QualityAgnostic}, or (ii) with the creation of multi-view networks in the spectral domain for learning invariant representations of the data~\cite{Chen2021AmplitutePhase}. Nevertheless, such methods can be conceptually -- and computationally -- complex, which prevents from pinpointing their elements that actually contribute to the overall robust.

Adversarial training has also been used as a data augmentation method for improving the robustness of image classifiers to different non-adversarial corruptions. The authors in~\cite{Rusak2020SimpleWay} proposed an adversarial training scheme that uses noise generated from uncorrelated distributions that maximize the classification loss. The authors in~\cite{Yi2021ImprovedOOD} studied theoretically and empirically the settings in which adversarial training improves robustness to out-of-distribution samples, while the authors in~\cite{kireev2021} analyzed the effectiveness of standard $\ell_p$ adversarial training against common image corruptions, and proposed a relaxation of adversarial training in the embedding space~\cite{Laidlaw2020perceptual}.

\textbf{Architecture choice}\quad
Finally, it is important to mention that the deep network architecture plays an important role in the robustness to distribution shifts. For instance, compressing through pruning a standard CNN can significantly improve its robustness to common corruptions~\cite{Diffenderfer2021WinningHand}. In addition, Vision Transformers~\cite{Dosovitskiy2021AnImageIsWorth} have been shown to be more robust to common corruptions of their data compared to standard CNNs~\cite{Bhojanapalli2021Understanding,Morrison2021Exploring}, when trained on very large data regimes, i.e., ImageNet-21K~\cite{KolesnikovBiT}. This is due, in part, to their different inductive bias~\cite{Morrison2021Exploring}. Besides, performing properly tailored variants of adversarial training on ViTs can further boost their robustness to common corruptions~\cite{Herrmann2021Pyramid,Mao2021Towards}.

\section{Robustness to adversarial perturbations}
\label{sec:related-robustness-adversrial-perturbations}

\subsection{Evaluating robustness to adversarial perturbations}
\label{susbec:related-adv-eval}

\textbf{Additive adversarial perturbations}\quad
The most widely studied way of measuring the adversarial robustness of deep networks is through the use of $\ell_p$-norm additive adversarial perturbations. These perturbations were first introduced in~\cite{Szegedy2014Intriguing} and were computed using a box-constrained L-BFGS algorithm, which is a simple method but not scalable to high dimensional image classification tasks. Nevertheless, computing adversarial examples is as fast and easy as slightly moving the image along the direction of the gradient of the loss function. That was initially shown with the FGSM algorithm~\cite{Goodfellow2015Explaining}, which inspired the creation of multiple gradient-based adversarial attacks that either solve \cref{eq:eps-adv-pert} for a given perturbation budget $\epsilon$~\cite{Kurakin2017AdvScale,Madry2018TowardsDeepLearning,Tramer2020AdaptiveAttacks}, or compute the minimal adversarial perturbation of \cref{eq:minimal-adv-pert} that changes the decision of the classifier~\cite{Moosavi2016DeepFool,CarliniWagner2017,Brendel2019AccReliable,Croce2020MinimallyDistorted}. Note here that one quite particular case is the so-called \emph{universal perturbation}, which is a single perturbation that can be applied to every image and still change the classifier's decision with high probability~\cite{Moosavi2017Universal}. Another interesting case is the construction of $\epsilon$-constrained adversarial perturbations (input space) that minimize the distance between specific internal representations of the network~\cite{Sabour2016AdvManipulation}. 

In some applications, evaluating the robustness of classifiers to perturbations with certain properties, such as sparsity, may be required. Sparse perturbations represent a very special case, since they are constrained to non-standard $\ell_p$ norms, such as $p=0$. Interestingly, the vulnerability of deep classifiers is extreme, since they can rather easily ``break'' by perturbing just a single or a few pixels of the image~\cite{Su2019OnePixel,CarliniWagner2017,Papernot2016JSMA,Narodytska2017,Croce2019SparseImperceivable}. However, finding such perturbations is an NP-hard problem and computationally expensive, and some of the existing methods cannot scale in very high-dimensional datasets. 

Another interesting type of structured perturbations is that of the subspace-constrained perturbations, which were first studied in~\cite{Fawzi2016RobustnessRandomNoise}. Such perturbations can be constrained to any desired subspace, i.e., different frequency bands defined by the Fourier basis, and can shed light onto multiple spectral properties of deep networks. In particular, it has been shown that deep classifiers are more sensitive to low-frequency perturbations, compared to their high-frequency counterparts~\cite{Zhou2018Transferable}. Furthermore, low-frequency perturbations can be more effective even against adversarially trained models~\cite{Sharma2019EffectivenessLowFrequency,Tsuzuku2019StructuralSens}, while they have also been exploited to design more efficient black-box attacks~\cite{Guo2019SimBA,Liu2019QFool,Rahmati2020GeoDA}. That is, adversarial attacks where the adversary has access only to the output of the classifier.

\textbf{Non-additive adversarial perturbations}\quad
Beyond additive perturbations, one can think of more sophisticated ways to construct adversarial examples. For instance, the robustness of deep classifiers to adversarial geometric transformations is studied in~\cite{Fawzi2015Manitest,Kanbak2018GeometricRobustness,Engstrom2019Exploring}, where image classifiers are shown to be very vulnerable to small rotations, translations and affine transformations. Furthermore, some works have evaluated the robustness of deep classifiers to other adversarial perturbation regimes such as color transformations~\cite{Laidlaw2019Functional,Hosseini2018Semantic,Shamsabadi2020ColorFool}, occlusions~\cite{Sharif2016Accessorize,Fawzi2016MeasuringEffectNuisance,Eykholt2018RobustPhysical}, and deformations~\cite{Fawzi2016MeasuringEffectNuisance,Xiao2018Spatially}.

\textbf{Black-box settings}\quad
Finally, for the sake of completeness, note that the robustness of deep networks can be evaluated in black-box settings, where the only available information during the construction of the adversarial examples is represented by the predictions or the class probabilities of the network. This scenario is quite realistic and many methods have been proposed for computing black-box adversarial perturbations~\cite{Chen2017ZOO,Brendel2018BoundaryAttack,Uesato2018AdvRisk,Chen2019HSJA,Maksym2020square}. However, this is mainly a security concern, which is not the main scope of this thesis. 

\subsection{Improving robustness to adversarial perturbations}
\label{susbec:related-adv-improving}

\textbf{Adversarial training}\quad
The most standard way for building classifiers that are robust to adversarial examples is the so-called \emph{adversarial training}, which can be seen as a type of data augmentation that replaces the clean images with their adversarial examples during training. 
Adversarial training was early introduced together with the first methods for computing adversarial examples~\cite{Szegedy2014Intriguing,Goodfellow2015Explaining,Moosavi2016DeepFool}. One of the risks of using adversarial training is that it can cause the classifier to overfit to specific types of adversarial examples (i.e., generated with FGSM). One can avoid this issue if the algorithm that generates adversarial perturbations is selected properly. In this sense, the scheme that seems to consistently result into more robust classifiers against a variety of adversarial attacks is the one proposed in~\cite{Madry2018TowardsDeepLearning}, which generates adversarial examples using the PGD algorithm~\cite{Kurakin2016PhysicalWorld}. This algorithm maximizes the classifier's loss within a ball of specific radius around the data samples. An alternative, as shown recently~\cite{Laidlaw2020perceptual}, would be to perform adversarial training with adversarial examples that maximize the classifier's loss, but at the same time constrain the adversarial representations of the classifier to be close to the original representations. This technique has been shown to build classifiers that are more robust to a variety of adversarial attacks that also extend beyond $\ell_p$-norm settings. Nevertheless, the main problem with adversarial training is that it is computationally expensive, and the main focus of current adversarial training schemes is to achieve similar or higher robustness, but at a lower cost~\cite{Shafahi2019ForFree,WongFastAT}.

\textbf{Regularization}\quad
In order to avoid the computational cost of adversarial training, many works have focused on increasing the stability of the classifiers through different types of regularization. The authors in~\cite{Gu2014Towards} proposed to smooth the norm of the gradient at each layer of the classifier, while other works have attempted to improve the robustness by regularizing the input gradient or the full Jacobian~\cite{Lyu2015Unified,Jakubovitz2018Improving,Ross2018Improving}. Furthermore, it has been shown that, second-order regularization techniques that penalize the curvature of the input loss function~\cite{Moosavi2019CURE,Qin2019LocalLinearization,Singla2020SecondOrder} result into classifiers that exhibit robustness similar to adversarial training. Lately, some works have focused on penalizing the curvature of the loss landscape in the weight space to improve robustness~\cite{Wu2020WeightPert}, and others that improve the stability of adversarial training by imposing constraints between the clean and noisy input gradients~\cite{Andriushchenko2020UnderstandingFast}. 

Note here, that, reducing the curvature of the input loss landscape to create more robust models can also be exploited to achieve robustness with guarantees, i.e., theoretical certificates that demonstrate a specific level of robustness for a given neural network. Indeed, certifiable adversarial defenses, like randomized smoothing~\cite{Cohen2019RandomizedSmoothing,Yang2020RandomizedSmoothingAllShapes,salman_provably_2020}, also implicitly regularize curvature by averaging the decision of a classifier on randomly perturbed samples. This way, one effectively convolves the loss landscape of a classifier with the probability density function of the perturbation distribution, hence, reducing the mean curvature of the loss landscape and smoothing the input geometry of the classifier. 

\section{Understanding deep learning through robustness}
\label{sec:related-understanding-deep-learning}

\textbf{Geometric insights}\quad
Although the lack of robustness in deep networks raises serious concerns regarding their security and trustworthiness, the whole process of evaluating and improving their robustness has revealed many important properties of deep learning. The first benefit of adversarial perturbations is that they enable us to study the local geometry of the decision boundaries~\cite{Fawzi2017GeometricPerspective} and obtain multiple insights regarding the topology and the geometry of the decision regions~\cite{Fawzi2018EmpiricalStudyTopology}. Furthermore, it has been shown that adversarial perturbations span a low-dimensional subspace of the input space~\cite{Tramer2017SpaceTransferable,Moosavi2018UniversalGeometric}, and that such subspaces of different networks are also aligned, which can justify the transferability properties of adversarial examples across neural networks~\cite{Tramer2017SpaceTransferable}. In addition, it has been observed that adversarial perturbations can exist in multiple directions~\cite{Fawzi2016RobustnessRandomNoise,Tramer2017SpaceTransferable,Fawzi2018EmpiricalStudyTopology} and by studying these different perturbations one can reveal geometric properties such as flatness and curvature of the decision boundaries. In fact, adversarial directions are mostly assigned to curved decision boundaries, and universal adversarial perturbations correspond to those shared curved directions~\cite{Moosavi2018UniversalGeometric}. Moreover, it has been observed that adversarial training creates decision boundaries that lie further away from most data samples~\cite{Moosavi2016DeepFool} and with lower curvature compared to standard models~\cite{Moosavi2019CURE}, which justifies why regularization methods improve the robustness of deep networks. Nevertheless, the fact that the decision boundaries exhibit a low mean curvature might in some cases increase the vulnerability of deep networks, since this ``flatness'' property can exploited as a prior for designing better adversarial attacks~\cite{Liu2019QFool, Rahmati2020GeoDA,Rahmati2021DoubleEdged}.

\textbf{Connection to data features}\quad
In deep learning, the networks are supposed to find ``good'' features of the training data, for a given learning task. It thus means that the key to the success of deep learning is the choice of features exploited by a neural network, and adversarial examples are actually correlated with such features. In particular, the authors in~\cite{Jetley2018WithFriends} showed that adversarial perturbations span a low-dimension but highly discriminative subspace of the input, and that deep networks exploit simple and brittle features of the dataset, i.e., non-robust features that are aligned with adversarial perturbations. Furthermore, this was also supported by the authors in~\cite{Ilyas2019AreNotBugs}. They showed that training on adversarial examples and with the corresponding adversarial labels, results into classifiers that achieve non-trivial accuracy on the original unmodified dataset. This means that, the only way that the network trained on the adversarial samples can generalize to the unmodified test set, is by exploiting the non-human-aligned features introduced by the adversarial perturbations themselves, i.e., the non-robust features. On the other hand, it has been shown that adversarially trained (robust) classifiers learn features that correlate better with semantically meaningful features of the input images~\cite{Tsipras2019RobustnessOddsAccuracy,Santukar2019ImageSynthesis,Engstrom2019AsPrior,AllenZhu2020Purification}. In this sense, one can actually use the (robust) representations of these models as effective primitives for semantic image manipulations, in order to perform complex tasks such as image generation or inpainting~\cite{Santukar2019ImageSynthesis}. 

\textbf{Generalization}\quad
Adversarial training filters out the non-robust spurious features which, however, would typically be used by the network to achieve good accuracy on the test set. The effect of this is reflected on the so-called robustness/accuracy tradeoff~\cite{Fawzi2018AnalysisOfClassifiers,Fawzi2018VulnerAny,Tsipras2019RobustnessOddsAccuracy}, while it also justifies the empirical observation that one needs more data to generalize when adversarial training is used~\cite{Schmidt2018AdversariallyRobustGeneralization}. In practice, it has also been shown that adding more data into the adversarial training process can improve robustness and decrease the generalization gap~\cite{AlayracAreLabelsRequired,CarmonUnlabeledAT,RaghunathanUnderstanding}. On the contrary, humans are not susceptible to adversarial perturbations, suggesting that they do not exploit non-robust features~\cite{Serre2019Deep}. Hence, it is argued that deep networks should also be able to achieve good generalization by using only robust features~\cite{Xie2020Improve,Tang2020OnlineAugment}. Some theoretical results indicate that there exist at least some synthetic distributions in which adversarial robustness and accuracy are positively correlated~\cite{Dohmatob2019NoFreeLunch,Gilmer2018AdversarialSpheres}. In fact, it has been recently shown that adversarial robustness and generalization are tightly close, and that the gap between the data samples and the decision boundary (minimal adversarial perturbations) can be used to predict the test accuracy~\cite{jiangPredictingGeneralizationGap2019,WerpachowskiDetectingOverfitting}. 

\textbf{Dynamics of learning and inductive bias}\quad
Finally, by tracking the evolution of adversarial perturbations during training, one can reveal different inductive biases and geometric properties of deep networks. 
For instance, the inductive bias of deep networks towards invariance has recently been argued to be prejudicial for classification as it can decrease the alignment between our human perception and the network's decisions~\cite{JacobsenExcessiveInvariance,TramerFundamentalTradeoffs}. This happens because adversarial training forces the networks to latch onto overly-robust features of the training set that are not human-aligned. Furthermore, in the deep learning community it is a common belief that neural networks are relatively immune to overfitting~\cite{zhangUnderstandingDeepLearning2016}. Indeed, the train and the validation losses of a neural network during training are clearly decreasing. Nevertheless, with adversarial training, it has recently been shown that the best robustness in the validation set is consistently found at the middle stages of training~\cite{RiceRobustOverfitting}. This confirms that adversarially trained neural networks have a tendency to overfit to the adversarial examples observed during training. This phenomenon is known as robust overfitting, and it has recently been shown to be one reason for some of the reported differences between different adversarial defenses. In fact, the state-of-the-art adversarial training technique uses early-stopping to obtain the best robustness results~\cite{Gowal2020Uncovering}. 

\section{Summary}
\label{sec:related-summary}

We summarize the main points of this chapter, in the light of the contributions of this thesis and upcoming challenges:
\begin{itemize}
    \item Deep image classifiers are extremely vulnerable to adversarial manipulations of their input samples. Different methods have been developed to assess their robustness properties, even in very challenging settings where the perturbations are governed by sparsity constraints. However, such methods are computationally expensive, thus rendering them impractical for high-dimensional datasets. One of the goals of this thesis is to provide a fast and scalable method for evaluating the robustness of deep networks to sparse adversarial perturbations. This will allow to analyze potential correlations between the spatial location of the perturbed sparse pixels and the semantic features of the images.
    
    \item The role of data in the generalization and robustness properties of deep classifiers, and the mechanisms that drive the networks to learn or ignore specific image features are not fully understood yet. We here provide a framework that connects the local geometry of the decision boundaries with the features of the dataset, and demonstrate that adversarial training exploits the invariance bias of deep networks and their sensitivity to the position of the training samples for building robust classifiers.
    
    \item Deep networks are not only vulnerable to adversarial manipulations, but they also exhibit poor robustness to common corruptions of their data. For improving this vulnerability, prior works have mostly focused on increasing the complexity of their training pipelines in the name of diversity, making it hard to pinpoint which elements of these methods meaningfully contribute to the overall robustness. In this thesis, we formulate a model for characterizing semantically-preserving image corruptions, and propose a principled and simple approach that is based on a mixture of few transformation primitives to confer robustness to common corruptions.
    
\end{itemize}
\cleardoublepage
\chapter{Sparse adversarial perturbations and image features}
\label{ch:sparse-adversarial-perturbations}

\begin{raggedleft}
    \textit{``It is the little bits of things that fret and worry us. \\
    We can dodge an elephant, but we can’t dodge a fly.''} \\
    --- Josh Billings \\
\end{raggedleft}
\vspace*{2cm}

\section{Introduction}
\label{sec:sparse-introduction}

Most of the existing methods in the adversarial robustness literature compute $\ell_p$-norm adversarial perturbations for $p\in\{2, \infty\}$. However, understanding the vulnerabilities of deep neural networks in non-standard $\ell_p$ regimes is also important. In particular, sparse perturbations for $p\in\{0,1\}$ are quite interesting, since their localized nature can reveal important parts of the image that such perturbations exploit~\cite{Xu2018StrAttack}. Prior works on sparse perturbations change the pixels either based on their saliency score~\cite{Papernot2016JSMA}, or using evolutionary algorithms~\cite{Su2019OnePixel}, or with greedy local search algorithms~\cite{Narodytska2017}. In general though, computing sparse adversarial perturbations with minimal $\ell_0$ norm is an NP-hard problem, and current algorithms are all characterized by high complexity and can hardly scale to high-dimensional datasets. Hence, a fast and accurate method for computing sparse perturbations is still needed to easily analyze different robustness properties of image classifiers.
\blfootnote{Part of this chapter has been published in}
\blfootnote{``SparseFool: A few pixels make a big difference''. In \textit{IEEE Conference on Computer Vision and Pattern Recognition (CVPR)}, 2019~\cite{ModasSparseFool}.}

In this chapter, we propose SparseFool, a geometry inspired algorithm that exploits the low mean curvature of the decision boundaries to linearize the sparsity constraints, and thus compute adversarial perturbations efficiently. We show through extensive evaluations that (i) our method computes sparse perturbations much faster than the existing methods, and (ii) it can scale efficiently to high dimensional datasets. We further propose a method to control the magnitude of the perturbation applied on every pixel -- and hence the perceptibility of the resulting perturbation --, while retaining the levels of sparsity and complexity. We analyze visually the image features affected by our attack, and show the existence of some shared semantic information across different images and networks, which suggests a strong correlation between adversarial examples and the semantic features of the images. Finally, we show that classifiers that are adversarially trained with $\ell_\infty$ perturbations are not robust to sparse perturbations, which indicates that the image features that are related to $\ell_1$ perturbations are different from those of $\ell_\infty$ perturbations.

The rest of the chapter is organized as follows: in \cref{sec:sparse-sparse-adversarial-perturbations}, we describe the challenges for computing minimal sparse adversarial perturbations, and provide an efficient method that linearizes the initial optimization problem to obtain an approximate solution. In \cref{sec:sparse-experimental-evaluation} we evaluate our algorithm on multiple datasets and networks, and perform comparisons with other state-of-the-art methods. Finally, in \cref{sec:sparse-analysis} we analyze empirically the resulting perturbations and demonstrate visual correlations between the perturbed pixels and the semantic information of the images.

\section{Minimal sparse adversarial perturbations}
\label{sec:sparse-sparse-adversarial-perturbations}

\subsection{Sparsity constraints}
\label{subsec:sparse-sparsity-constraints}

Recall from \cref{sec:introduction-adversarial-robustness} that minimal adversarial perturbations are defined as
\begin{equation}
\begin{split}
    \argmin{\bm\delta} & \|\bm\delta\|_p \\
    \text{s.t. } & F(\bm{x}) \neq F(\bm{x} + \bm\delta) \\
    & \bm\delta\in\mathcal{C},
\end{split}
\label{eq:sparse-minimal-adv-pert}
\end{equation}
where $f:\R^D\to \R^K$ is the final layer of a neural network (i.e., logits), such that, for any input $\bm{x}\in\R^D$, $F(\bm{x})=\text{argmax}_k f_k(\bm{x})$ represents the decision of that network, with $f_k(\bm{x})$ denoting the $k$th component of $f(\bm{x})$ that corresponds to the $k$th class.

Most of the existing adversarial algorithms solve \cref{eq:sparse-minimal-adv-pert} for $p=2$ or $\infty$, resulting in dense but imperceptible perturbations. For the case of sparse perturbations, the goal is to minimize the number of perturbed pixels required to fool the network, which corresponds to minimizing $\|\bm\delta\|_0$. Unfortunately, this is an NP-hard problem, for which reaching a global minimum cannot be guaranteed in general~\cite{Blumensath2008IterativeThresholding, Nikolova2013Description, Patrascu2015RandoomCoord}. There exist different methods~\cite{Nagahara2014SparsePacketized, Patrascu2015RandoomCoord} to avoid the computational burden of this problem, with the $\ell_1$ relaxation being the most common: the minimization of $\|\bm\delta\|_0$ under linear constraints can be approximated by solving the corresponding convex $\ell_1$ problem~\cite{Candes2005Decoding, Donoho2006Compressed, Natarajan1995SparseApprox}\footnote{Under some conditions, the solution of such approximation is indeed optimal~\cite{Candes2005ErrorCorr, Donoho2003Optimally, Gribonval2003SparseRepr}.}.

DeepFool~\cite{Moosavi2016DeepFool} is an algorithm that exploits such a relaxation, by adopting an iterative procedure that includes a linearization of the classifier at each iteration, in order to estimate the minimal adversarial perturbation $\bm\delta$. Specifically, at each iteration $i$, the classifier $f$ is linearized around the current point $\bm{x}^{(i)}$, the minimal perturbation $\bm\delta^{(i)}$ (in an $\ell_2$ sense) is computed as the projection of $\bm{x}^{(i)}$ onto the linearized hyperplane, and the next iterate $\bm{x}^{(i+1)}$ is updated. Such a linearization procedure could actually be used to solve \cref{eq:sparse-minimal-adv-pert}
for $p=1$, so as to obtain an approximation to the $\ell_0$ solution. In fact, by generalizing the projection to any $\ell_p$ norm, $\ell_1$-DeepFool provides an efficient way for computing sparse adversarial perturbations using the $\ell_1$ projection.

Although the $\ell_1$-DeepFool efficiently computes sparse perturbations, it does not explicitly respect the validity of the adversarial image values; that is $\mathcal{C}=\{\bm{\delta}:\bm{x}+\bm{\delta}\in[0,255]^D\}$ in \cref{eq:sparse-minimal-adv-pert}. For $\ell_2$ and $\ell_\infty$ perturbations, almost every pixel of the image is typically distorted with noise of small magnitude. Hence, one can practically ``ignore'' such constraints~\cite{Goodfellow2015Explaining,Moosavi2016DeepFool} since it is unlikely that many pixels will be out of their valid range; and even then, clipping the invalid values after the computation of such adversarial images could have a minor impact. This, however, is not the case for sparse perturbations, which typically result in a few distorted pixels but of high-magnitude perturbation, and clipping the values after computing the adversarial image can have a significant impact on the success of the attack.

We investigate the effect of such clipping operation on the quality of adversarial perturbations generated by $\ell_1$-DeepFool. For example, with perturbations computed for a VGG-16~\cite{Simonyan2015VeryDeep} trained on ImageNet~\cite{Deng2009ImageNet}, we observed that $\ell_1$-DeepFool achieves almost $100\%$ of fooling rate by perturbing only $0.037\%$ of the pixels on average. However, clipping the pixel values of adversarial images to $[0,255]$ results in a fooling rate of merely $13\%$. Furthermore, incorporating the clipping operator inside the iterative procedure of the algorithm does not improve the results. In other words, $\ell_1$-DeepFool fails to properly compute sparse perturbations. This underlies the need for an improved attack algorithm that natively takes into account the validity of generated adversarial images, as proposed in the next sections.

\subsection{Linearization and boundary approximation}
\label{subsec:sparse-problem-linearization}

Based on the above discussion, minimal sparse perturbations should be obtained as
\begin{equation}
\begin{split}
    \argmin{\bm\delta} & \|\bm\delta\|_0 \\
    \text{s.t. } & F(\bm{x}) \neq F(\bm{x} + \bm\delta) \\
    & \bm{l}\preccurlyeq \bm{x}+\bm{\delta}\preccurlyeq \bm{u},
\end{split}
\label{eq:sparse-l0_optim}
\end{equation}
where $\bm{l}, \bm{u}\in\mathbb{R}^D$ denote the lower and upper bounds of the values of $\bm{x} + \bm\delta$, such that $l_i \leq x_i + \delta_i \leq u_i, \enskip i=1\dots D$.

\begin{figure}[t]
\begin{center}
\includegraphics[width=0.45\linewidth, page=1]{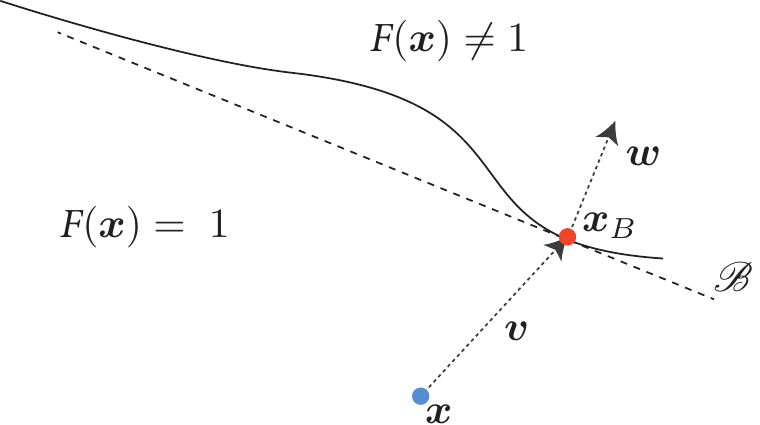}
\end{center}
   \caption{The approximated decision boundary $\mathscr{B}$ in the vicinity of the datapoint $\bm x$ that belongs to class $F(\bm x)=1$. $\mathscr{B}$ can be seen as a one-vs-all linear classifier for class $1$.}
\label{fig:sparse-flat}
\end{figure}

To find an efficient relaxation to \cref{eq:sparse-l0_optim}, we focus on the geometric characteristics of the decision boundary, and specifically on its curvature. It has been shown~\cite{Fawzi2017GeometricPerspective, Fawzi2018EmpiricalStudyTopology, Jetley2018WithFriends} that the decision boundaries of state-of-the-art deep networks have a quite low mean curvature in the neighborhood of data samples. In other words, for a datapoint $\bm{x}$ and its corresponding minimal $\ell_2$ adversarial perturbation $\bm v$, the decision boundary at the vicinity of $\bm{x}$ can be locally approximated by a hyperplane passing through the datapoint $\bm{x}_B=\bm{x}+\bm v$, and a normal vector $\bm w$~(see \cref{fig:sparse-flat} for an illustration). Hence, by exploiting this property we can relax \cref{eq:sparse-l0_optim} so that sparse adversarial perturbations are computed by solving the following $\ell_1$ box-constrained optimization problem
\begin{equation}
\begin{split}
    \argmin{\bm\delta} & \|\bm\delta\|_1 \\
    \text{s.t. } & \bm{w}^T\Big((\bm{x} + \bm\delta)- \bm{x}_{B}\Big)=0 \\
    & \bm{l}\preccurlyeq \bm{x}+\bm{\delta}\preccurlyeq \bm{u}.
\end{split}
\label{eq:sparse-l1_linear}
\end{equation}

For solving \cref{eq:sparse-l1_linear}, simply computing the $\ell_1$ projection of $\bm x$ onto the approximated hyperplane does not guarantee a solution. For a perturbed image, consider the case where some of its values exceed the bounds defined by $\bm{l}$ and $\bm{u}$. Thus, by readjusting the invalid values to match the constraints, the resulted adversarial image may eventually not lie onto the approximated hyperplane. For this reason, we propose an iterative procedure, where at each iteration we project only towards one single coordinate of the normal vector $\bm w$ at a time. If projecting $\bm x$ towards a specific direction does not provide a solution, then the perturbed image at this coordinate has reached its extrema value. Therefore, at the next iteration this direction should be ignored, since it cannot contribute any further to finding a better solution.

Formally, let $S$ be a set containing all the directions of $\bm w$ that cannot contribute to the minimal perturbation anymore. Then, the perturbation $\bm\delta$ is updated through the $\ell_1$ projection of the current perturbed iterate $\bm{x}^{(i)}$ onto the estimated hyperplane as
\begin{equation}
    \bm{\delta}_d\leftarrow\dfrac{|\bm{w}^T(\bm{x}^{(i)}-\bm{x}_B)|}{|w_d|}\cdot\text{sign} (w_d),
\label{eq:sparse-delta_d}
\end{equation}
where $d$ is the index of the maximum absolute value of $\bm{w}$ that has not already been used
\begin{equation}
    d\leftarrow\argmax{j\notin S}{|w_j|}.
\label{eq:sparse-argmax}
\end{equation}

Before proceeding to the next iteration, we must ensure the validity of the values of the next iterate $\bm{x}^{(i+1)}$. For this reason, we use a projection operator $Q(\cdot)$ that readjusts the values of the updated point that are out of bounds, by projecting $\bm{x}^{(i)} + \bm\delta$ onto the box-constraints defined by $\bm{l}$ and $\bm{u}$. Hence, the new iterate $\bm{x}^{(i+1)}$ is updated as
$\bm{x}^{(i+1)}\leftarrow Q(\bm{x}^{(i)} + \bm\delta)$.
Note here that the bounds $\bm{l}$, $\bm{u}$ are not limited to only represent the dynamic range of an image, but can be generalized to satisfy any similar restriction. For example, as we will describe later in \cref{sec:sparse-experimental-evaluation}, they can be used to control the perceptibility of the computed adversarial images.

\AlgoDontDisplayBlockMarkers
\RestyleAlgo{ruled}
\SetAlgoNoLine
\LinesNumbered
\begin{algorithm}[t]
    \small
 	\KwIn{image $\bm{x}$, normal $\bm{w}$, boundary point $\bm{x}_B$, projection operator $Q$.}
 	\KwOut{perturbed point $\bm{x}^{(i)}$}
 	\BlankLine
	Initialize: $\bm{x}^{(0)}\leftarrow\bm{x}$,\enskip$i\leftarrow 0$,\enskip$S=\{\}$\\
	
	\smallskip
	\While{$\bm{w}^T(\bm{x}^{(i)}-\bm{x}_B)\neq 0$}
	{
	    $\bm\delta\leftarrow\bm{0}$
	    
	    \smallskip
	    $d\leftarrow\argmax{j\notin S}{|w_j|}$
	    
	    \smallskip
	    $\delta_d\leftarrow\dfrac{|\bm{w}^T(\bm{x}^{(i)}-\bm{x}_B)|}{|w_d|}\cdot\text{sign} (w_d)$
	    
	    \smallskip
	    $\bm{x}^{(i+1)}\leftarrow Q(\bm{x}^{(i)}+\bm\delta)$
	    
		\smallskip
	    $S \leftarrow S\cup \{d\}$
	    
		\smallskip
		$i\leftarrow i+1$
	}	
 	\KwRet{$\bm{x}^{(i)}$}
\caption{LinearSolver}
\label{alg:sparse-linear_solver}
\end{algorithm}

The next step is to check if the new iterate $\bm{x}^{(i+1)}$ has reached the approximated hyperplane. Otherwise, it means that the perturbed image at the coordinate $d$ has reached its extrema value, and thus we cannot change it any further; perturbing towards the corresponding direction will have no effect. Thus, we reduce the search space by adding the direction $d$ to the forbidden set $S$, and repeat the procedure until we reach the approximated hyperplane. The algorithm for solving the linearized problem is summarized in \cref{alg:sparse-linear_solver}.

Finally, in order to complete our solution we focus on the linear approximation of the decision boundary. Recall that we need to find a boundary point $\bm{x}_B$, along with the corresponding normal vector $\bm w$. Finding $\bm{x}_{B}$ is analogous to computing (in a $\ell_2$ sense) the minimal adversarial example of $\bm{x}$. Recall that DeepFool iteratively moves $\bm{x}$ towards the decision boundary, and stops as soon as the perturbed data point reaches the other side of the boundary. Therefore, the final point usually lies very close to the decision boundary, and thus, $\bm{x}_{B}$ can be very well approximated by $\bm{x}+\bm\delta_{\text{DF}}$, with $\bm\delta_{\text{DF}}$ being the $\ell_2$-DeepFool perturbation. Let us describe the decision boundary between the class assigned to $\bm{x}_B$, $F(\bm{x}_B)$, and any other class $F(\bm{x})$, by considering the zero level set of $f$ 
\begin{equation*}
     \mathscr{B} = \Big\{ \bm{x} : f_{F(\bm{x}_B)}(\bm{x}) - f_{F(\bm{x})}(\bm{x}) = 0 \Big\}.
\end{equation*}
Using a first-order Taylor expansion at $\bm{x}_B$, the decision boundary can be expressed as
\begin{equation*}
    f_{F(\bm{x}_B)}(\bm{x}_B) + \nabla f_{F(\bm{x}_B)}(\bm{x}_B)^T(\bm{x}-\bm{x}_B) - f_{F(\bm{x})}(\bm{x}_B) - \nabla f_{F(\bm{x})}(\bm{x}_B)^T(\bm{x}-\bm{x}_B) = 0.
\end{equation*}
Since $\bm{x}_B$ lies very close to the decision boundary, then $f_{F(\bm{x}_B)}(\bm{x}_B) \approx f_{F(\bm{x})}(\bm{x}_B)$ and
\begin{equation*}
    \Big(\nabla f_{F(\bm{x}_B)}(\bm{x}_B)^T - \nabla f_{F(\bm{x})}(\bm{x}_B)^T\Big)(\bm{x}-\bm{x}_B) = 0.
\end{equation*}
We can then define the estimated normal vector $\bm w$ to the decision boundary  as
\begin{equation}
    \bm w:=\nabla f_{F(\bm{x}_B)}(\bm{x}_B) - \nabla f_{F(\bm{x})}(\bm{x}_B),
\label{eq:sparse-normal_vec}
\end{equation}
Hence, the decision boundary can now be approximated through the affine hyperplane $\mathscr{B}\triangleq\big\{\bm{x}:\bm{w}^T(\bm{x}- \bm{x}_B)=0\big\}$, and sparse adversarial perturbations are computed by applying \cref{alg:sparse-linear_solver}.

\subsection{SparseFool}
\label{subsec:sparse-sparsefool}
However, although we expected a single-step solution, in many cases the algorithm did not fool the classifier. This is due to the fact that the decision boundaries of the networks are only \emph{locally flat}. Thus, if the $\ell_1$ perturbation moves the datapoint $\bm{x}$ away from the flat area, then the perturbed point will not reach the other side of the decision boundary.

\begin{figure}[t]
\begin{center}
\includegraphics[width=0.45\linewidth, page=2]{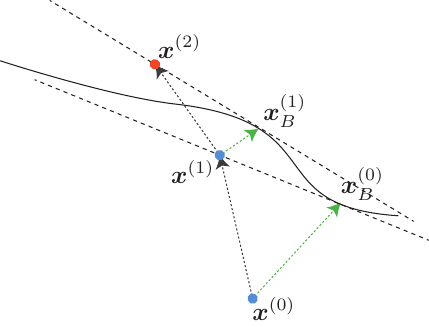}
\end{center}
   \caption{Illustration of SparseFool algorithm. With green we denote the $\ell_2$-DeepFool adversarial perturbations computed at each iteration. In this example, the algorithm converges after $2$ iterations, and the total perturbation is $\bm\delta=\bm{x}^{(2)}-\bm{x}^{(0)}$.}
\label{fig:sparse-curv_convg}
\end{figure}

\AlgoDontDisplayBlockMarkers
\RestyleAlgo{ruled}
\SetAlgoNoLine
\LinesNumbered
\begin{algorithm}[t]
    \small
	\SetKwFunction{Union}{Union}\SetKwFunction{deepfool}{DeepFool}\SetKwFunction{linear}{LinearSolver}
 	\KwIn{image $\bm{x}$, projection operator $Q$, classifier $f$.}
 	\KwOut{perturbation $\bm\delta$}
 	\BlankLine
	Initialize: $\bm{x}^{(0)}\leftarrow\bm{x},\enskip i\leftarrow 0$\\
	
	\While{$F(\bm{x}^{(i)})=F(\bm{x}^{(0)})$}
	{
	    \smallskip
		$\bm\delta_\text{DF}=$ \deepfool{$\bm{x}^{(i)}$}
		
		\smallskip
		$\bm{x}^{(i)}_B=\bm{x}^{(i)}+\bm\delta_\text{DF}$
		
		\smallskip
		$\bm{w}^{(i)}=\nabla f_{F(\bm{x}^{(i)}_B)}(\bm{x}^{(i)}_B) - \nabla f_{F(\bm{x}^{(i)})}(\bm{x}^{(i)}_B)$
		
		\smallskip
		$\bm{x}^{(i+1)}=$ \linear{$\bm{x}^{(i)}$, $\bm{w}^{(i)}$, $\bm{x}^{(i)}_B$, $Q$}
		
		\smallskip
		$i\leftarrow i+1$
	}	
 	\KwRet{$\bm\delta=\bm{x}^{(i)}-\bm{x}^{(0)}$}
\caption{SparseFool}
\label{alg:sparse-SparseFool}
\end{algorithm}

We mitigate this convergence issue with an iterative method, namely {\em SparseFool}, where each iteration includes the linear approximation of the decision boundary. Specifically, at iteration $i$, the boundary point $\bm{x}_B^{(i)}$ and the normal vector $\bm{w}^{(i)}$ are estimated using $\ell_2$-DeepFool based on the current iterate $\bm{x}^{(i)}$. Then, the next iterate $\bm{x}^{(i+1)}$ is updated through the solution of \cref{alg:sparse-linear_solver}, having though $\bm{x}^{(i)}$ as the initial point. The algorithm terminates when $\bm{x}^{(i)}$ changes the label of the network. An illustration of SparseFool is given in \cref{fig:sparse-curv_convg}, and the algorithm is summarized in \cref{alg:sparse-SparseFool}.

\begin{figure*}[t]
\centering
\begin{subfigure}{0.32\textwidth}
    \centering
    \includegraphics[width=\linewidth]{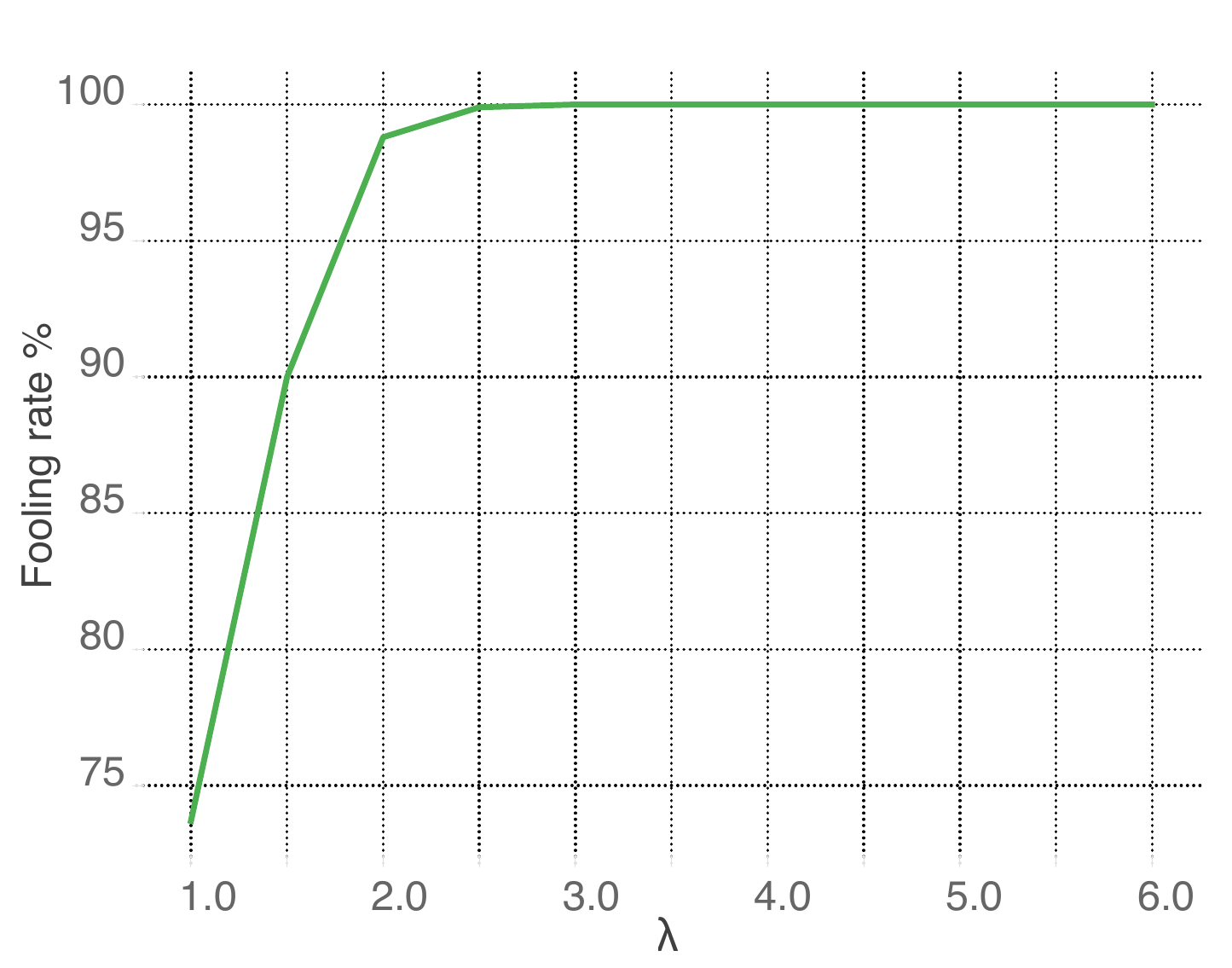}
\end{subfigure}\hfill
\begin{subfigure}{0.32\textwidth}
    \centering
    \includegraphics[width=\linewidth]{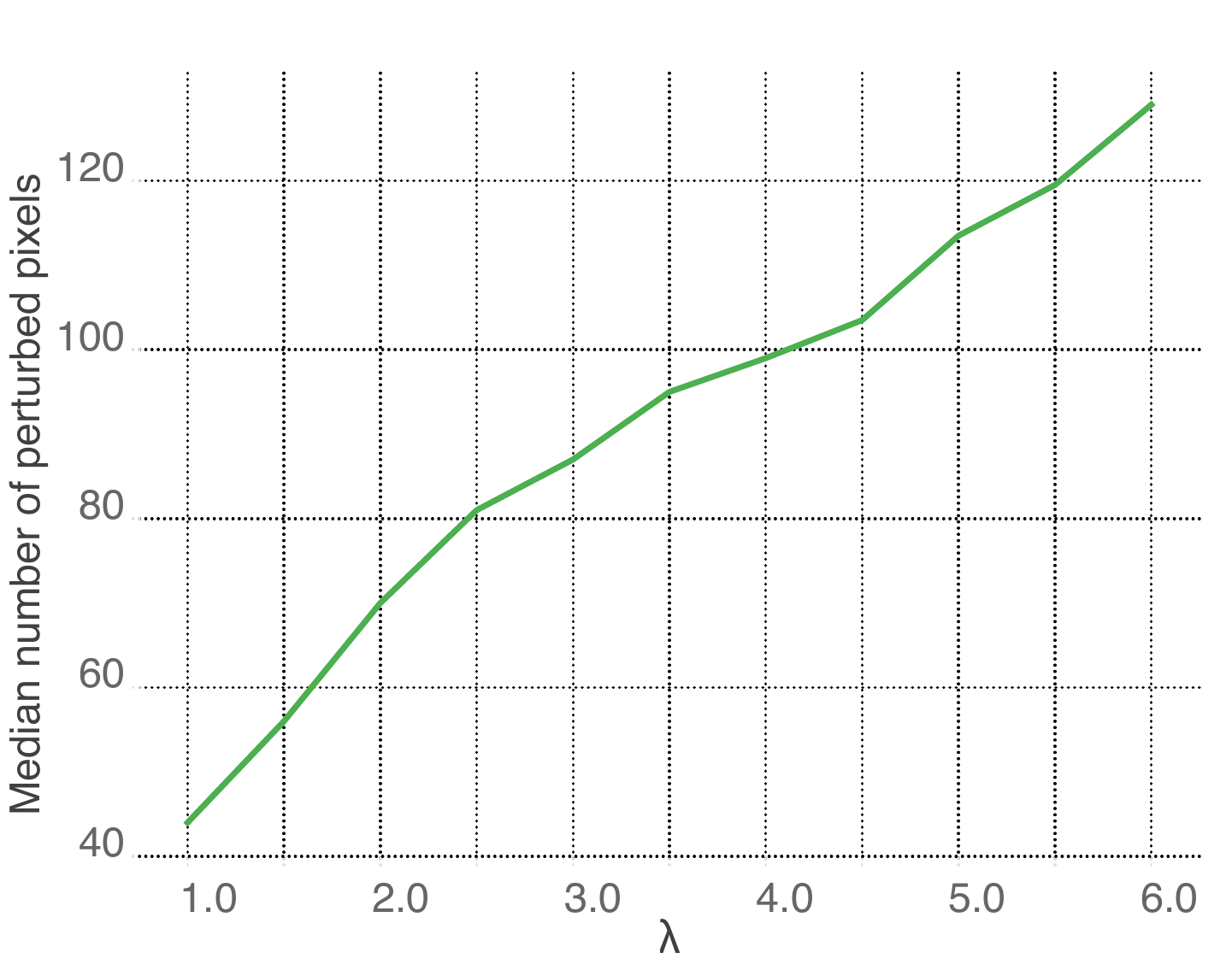}
\end{subfigure}\hfill
\begin{subfigure}{0.32\textwidth}
    \centering
    \includegraphics[width=\linewidth]{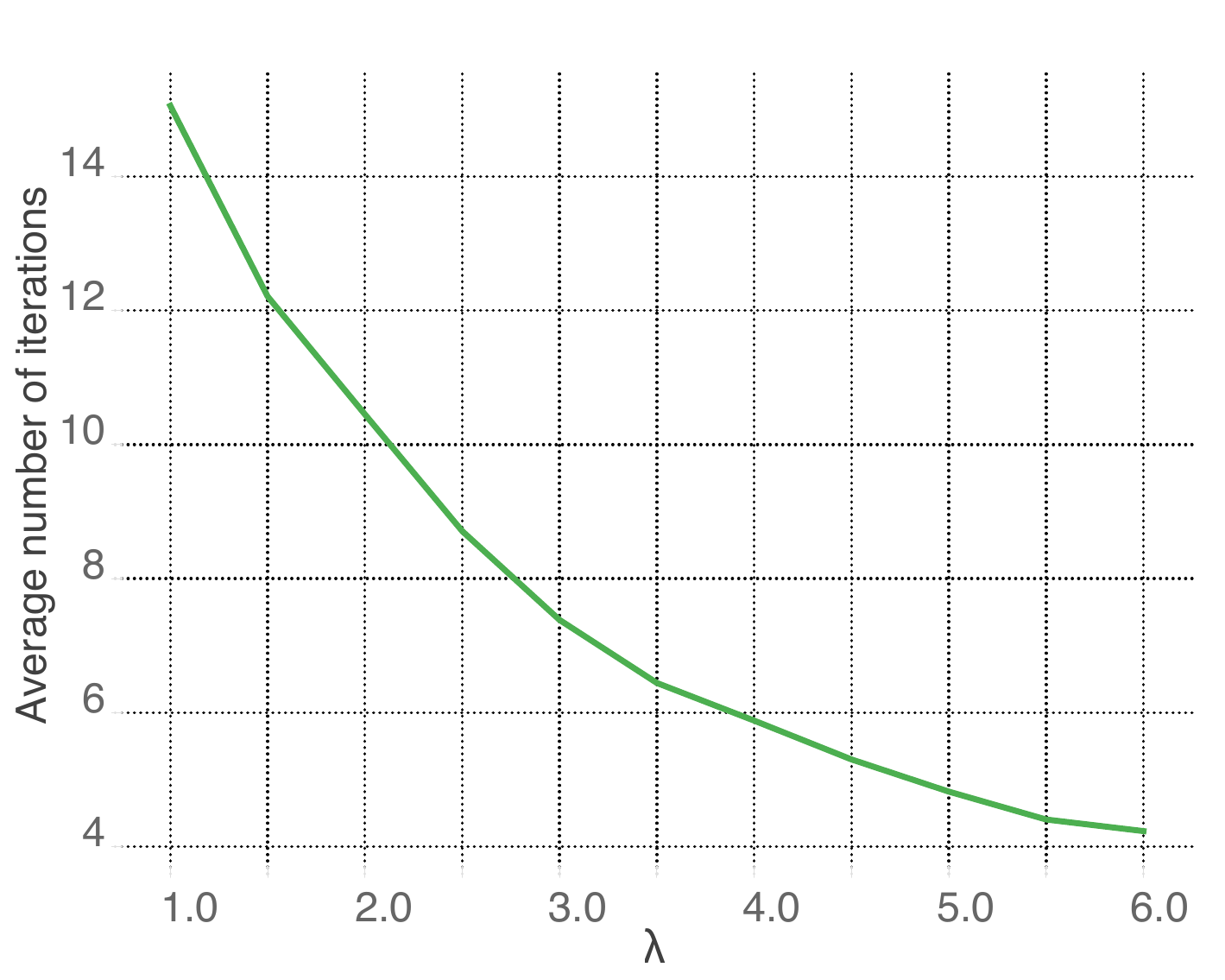}
\end{subfigure}\hfill
\caption{Fooling rate, sparsity of the perturbations, and average iterations of SparseFool for different values of $\lambda$, on $4000$ images from ImageNet using an Inception-v3~\protect\cite{Szegedy2016Inception}.}
\label{fig:sparse-lambda_effect}
\end{figure*}

In our experiments, we observed that instead of using the boundary point $\bm{x}^{(i)}_B$ at the step $6$ of SparseFool, better convergence can be achieved by going further into the other side of the boundary, and find a solution for the hyperplane passing through the datapoint $\bm{x}^{(i)}+\lambda (\bm{x}^{(i)}_B-\bm{x}^{(i)})$, where $\lambda\geq1$. Specifically, as shown in \cref{fig:sparse-lambda_effect}, this over-shooting parameter is used to control the trade-off between the fooling rate, the sparsity, and the complexity. Values close to $1$, lead to sparser perturbations, but also to lower fooling rate and increased complexity. On the contrary, higher values of $\lambda$ lead to fast convergence -- even one step solutions --, but the resulted perturbations are less sparse. Since $\lambda$ is the \emph{only parameter} of the algorithm, it can be easily adjusted to meet the corresponding needs in terms of fooling rate, sparsity, and complexity.

Finally, note that our approach is different from DeepFool's. DeepFool approximates the decision boundary by linearizing the classifier, while we rather linearize the approximated decision boundary $\mathscr{B}$. Furthermore, $\mathscr{B}$ is the boundary between the adversarial and the estimated true class, and thus it can be seen as an affine binary classifier. Since at each iteration the adversarial class is computed as the closest (in an $\ell_2$ sense) to the true one, we can say that SparseFool operates as an untargeted attack. Nevertheless, it can be easily transformed to a targeted one, by simply computing at each iteration the adversarial example -- and thus approximating the decision boundary -- of a target class.

\section{Experimental evaluation}
\label{sec:sparse-experimental-evaluation}

We evaluate SparseFool on deep convolutional neural network architectures with $10000$ images of the MNIST~\cite{LeCun2010MNIST} test set, $10000$ images of the CIFAR-10~\cite{krizhevskyLearningMultipleLayers2009} test set, and $4000$ randomly selected images from the ImageNet ILSVRC2012 validation set. In order to evaluate our algorithm and compare with related works, we compute the fooling rate, the median perturbation percentage, and the average execution time. Given a dataset $\mathscr{D}$, the fooling rate measures the efficiency of the algorithm based on the formula
\begin{equation}
    \dfrac{\big| \bm{x}\in\mathscr{D}:F(\bm{x}+\bm\delta_{\bm{x}})\neq F(\bm{x})\big|}{\big| \mathscr{D} \big|},
\end{equation}
where $\bm\delta_{\bm{x}}$ is the perturbation of the image $\bm{x}$, while the median perturbation percentage is the median (across all images) percentage of pixels that are perturbed.

We compare SparseFool with JSMA~\cite{Papernot2016JSMA}. Since JSMA is a targeted attack, we use its ``untargeted'' version, where the target is chosen at random. We also make a modification at the success condition; instead of checking if the predicted class is equal to the randomly selected target, we simply check if it is different from the initial class. Note that JSMA is not evaluated on ImageNet, due to its huge computational cost for searching over all pairs of candidates, as also mentioned in~\cite{CarliniWagner2017}. We also compare SparseFool with ``One-pixel attack'' (1-PA)~\cite{Su2019OnePixel}. Since 1-PA perturbs exactly $k$ pixels, for every image we start with $k=1$ and increase it till 1-PA finds an adversarial example. Again, we do not evaluate 1-PA on ImageNet, due to its computational cost in high dimensional images.

\subsection{Performance analysis}
\label{subsec:sparse-performance-analysis}

\begin{table*}[tb]
\resizebox{\textwidth}{!}
{%
	\begin{tabular}{| c | c | c | c | c | c| c | c | c | c | c | c |}
	\hline
	\multirow{2}{*}{Dataset}
	& \multirow{2}{*}{Network}
	& \multirow{2}{*}{Acc. (\%)}
	& \multicolumn{3}{c|}{Fooling rate (\%)}
	& \multicolumn{3}{c|}{Perturbation (\%)}
	& \multicolumn{3}{c|}{Time (sec)} \\
	\cline{4-12}
	\multicolumn{1}{|c|}{}
	& \multicolumn{1}{c|}{}
	& \multicolumn{1}{c|}{}
	& \multicolumn{1}{c|}{SF}
	& \multicolumn{1}{c|}{JSMA}
	& \multicolumn{1}{c|}{$1$-PA}
	& \multicolumn{1}{c|}{SF}
	& \multicolumn{1}{c|}{JSMA}
	& \multicolumn{1}{c|}{$1$-PA}
	& \multicolumn{1}{c|}{SF}
	& \multicolumn{1}{c|}{JSMA}
	& \multicolumn{1}{c|}{$1$-PA} \\
	\hline
	\multicolumn{1}{|c|}{MNIST}
	& \multicolumn{1}{c|}{LeNet~\protect\cite{LeNet}}
	& \multicolumn{1}{c|}{$99.14$}
	& \multicolumn{1}{c|}{$99.93$}
	& \multicolumn{1}{c|}{$95.73$} 
	& \multicolumn{1}{c|}{$100$} 
	& \multicolumn{1}{c|}{$1.66$}
	& \multicolumn{1}{c|}{$4.85$} 
	& \multicolumn{1}{c|}{$9.43$} 
	& \multicolumn{1}{c|}{$0.14$}
	& \multicolumn{1}{c|}{$0.66$}
	& \multicolumn{1}{c|}{$310.2$}  \\
	\hline
	\hline
	\multirow{2}{*}{CIFAR-10}
	& \multicolumn{1}{c|}{VGG-19}
	& \multicolumn{1}{c|}{$92.71$}
	& \multicolumn{1}{c|}{$100$}
	& \multicolumn{1}{c|}{$98.12$} 
	& \multicolumn{1}{c|}{$100$} 
	& \multicolumn{1}{c|}{$1.07$}
	& \multicolumn{1}{c|}{$2.25$} 
	& \multicolumn{1}{c|}{$0.15$} 
	& \multicolumn{1}{c|}{$0.34$}
	& \multicolumn{1}{c|}{$6.28$}
	& \multicolumn{1}{c|}{$102.7$}  \\
	\cline{2-12}
	& \multicolumn{1}{c|}{ResNet18~\protect\cite{He2016ResNet}}
	& \multicolumn{1}{c|}{$92.74$}
	& \multicolumn{1}{c|}{$100$}
	& \multicolumn{1}{c|}{$100$}
	& \multicolumn{1}{c|}{$100$}  
	& \multicolumn{1}{c|}{$1.27$}
	& \multicolumn{1}{c|}{$3.91$}
	& \multicolumn{1}{c|}{$0.2$}  
	& \multicolumn{1}{c|}{$0.69$}
	& \multicolumn{1}{c|}{$8.73$}
	& \multicolumn{1}{c|}{$167.4$}  \\
	\hline
\end{tabular}
}
\caption{The performance of SparseFool (SF), JSMA~\cite{Papernot2016JSMA}, and ``One-pixel attack'' ($1$-PA)~\cite{Su2019OnePixel} on MNIST and CIFAR-10. Due to its high complexity, $1$-PA is evaluated on only $100$ samples. All the experiments conducted on a GTX TITAN X.}
\label{tab:sparse-mnist_cifar}
\end{table*}

\begin{table}[t]
\begin{center}
{
    \small
	\begin{tabular}{| c | c | c | c | c |}
	\hline
	\multicolumn{1}{|c|}{Network}
	& \multicolumn{1}{c|}{\makecell{Acc. (\%)}}
	& \multicolumn{1}{c|}{\makecell{Fooling\\rate (\%)}}
	& \multicolumn{1}{c|}{\makecell{Pert.\\(\%)}}
	& \multicolumn{1}{c|}{\makecell{Time \\ (sec)}}\\
	\hline
	\multicolumn{1}{|c|}{VGG-16}
	& \multicolumn{1}{c|}{$71.59$}
	& \multicolumn{1}{c|}{$100$} 
	& \multicolumn{1}{c|}{$0.18$}
	& \multicolumn{1}{c|}{$5.09$} \\
	\hline
	\multicolumn{1}{|c|}{ResNet-101}
	& \multicolumn{1}{c|}{$77.37$}
	& \multicolumn{1}{c|}{$100$} 
	& \multicolumn{1}{c|}{$0.23$}
	& \multicolumn{1}{c|}{$8.07$}\\
	\hline
	\multicolumn{1}{|c|}{DenseNet-161}
	& \multicolumn{1}{c|}{$77.65$}
	& \multicolumn{1}{c|}{$100$}
	& \multicolumn{1}{c|}{$0.29$} 
	& \multicolumn{1}{c|}{$10.07$} \\
	\hline
	\multicolumn{1}{|c|}{Inception-v3}
	& \multicolumn{1}{c|}{$77.45$}
	& \multicolumn{1}{c|}{$100$}
	& \multicolumn{1}{c|}{$0.14$} 
	& \multicolumn{1}{c|}{$4.94$} \\
	\hline
\end{tabular}
}
\end{center}
\caption{The performance of SparseFool on the ImageNet dataset, using the pre-trained models provided by PyTorch~\protect\cite{Paszke2019PyTorch}. All experiments were conducted on a GTX TITAN X.}
\label{tab:sparse-imagenet}
\end{table}

We first evaluate the performance of SparseFool, JSMA, and 1-PA on MNIST and CIFAR-10. The control parameter $\lambda$ in SparseFool was set to $1$ and $3$ for the MNIST and CIFAR-10 datasets respectively. We observe in \cref{tab:sparse-mnist_cifar} that SparseFool computes $2.9$x sparser perturbations, and is $4.7$x faster compared to JSMA for the MNIST dataset. This behavior remains similar for the CIFAR-10 dataset, where SparseFool computes on average perturbations of $2.4$x higher sparsity, and is $15.5$x faster. Notice here the difference in the execution time: JSMA becomes much slower as the dimension of the input data increases, while SparseFool's time complexity remains at very low levels.

Compared to 1-PA, SparseFool computes $5.5$x sparser perturbations on MNIST, and is more than $3$ orders of magnitude faster. On CIFAR-10, SparseFool still finds very sparse perturbations, but less so than the 1-PA in this case. The reason is that our method does not solve the $\ell_0$ optimization problem, but it rather computes sparse perturbations through the $\ell_1$ relaxation. The solution is often sub-optimal, and may be optimal when the image is very close to the boundary, where the linear approximation is more accurate. However, solving the linearized problem is fast and enables our method to efficiently scale to high dimensional data, which is not the case for 1-PA. Considering the tradeoff between the sparsity and complexity, we choose to sacrifice the former. In fact, our method is able to compute sparse perturbations $270$x faster than 1-PA.

Finally, due to the computational cost of JSMA and 1-PA, we do not evaluate them on ImageNet. Instead, we compare SparseFool with an algorithm that randomly selects a subset of elements from each color channel, and replaces their intensity with a random value from $V=\{0,255\}$. The cardinality of each channel subset is constrained to match SparseFool's per-channel median number of perturbed elements: for each channel, we select as many elements as the median, across all images, of SparseFool's perturbed elements for this channel. The performance of SparseFool on ImageNet is reported in \cref{tab:sparse-imagenet}, while the corresponding fooling rates of the random algorithm are $18.2\%$, $13.2\%$, $14.5\%$, and $9.6\%$ respectively. The fooling rates obtained by the random algorithm are far from comparable to SparseFool's, indicating that our algorithm cleverly finds sparse solutions. Our method is consistent among different architectures, perturbing on average $0.21\%$ of the pixels, with an average execution time of $7$ seconds per sample.

To the best of our knowledge, we are the first to provide an adequate sparse attack that efficiently achieves such fooling rates and sparsity, and at the same time scales to high dimensional data. 1-PA does not necessarily find good solutions for all the studied datasets, however, SparseFool -- as it relies on the high dimensional geometry of the classifiers --  successfully computes sparse enough perturbations for all three datasets.

\subsection{Perceptibility}
\label{subsec:sparse-perceptibility}

\begin{figure}[b]
    \centering
    \begin{subfigure}[b]{0.45\columnwidth}
        \centering
        \captionsetup{font=scriptsize}
        \includegraphics[width=0.7\linewidth, page=1]{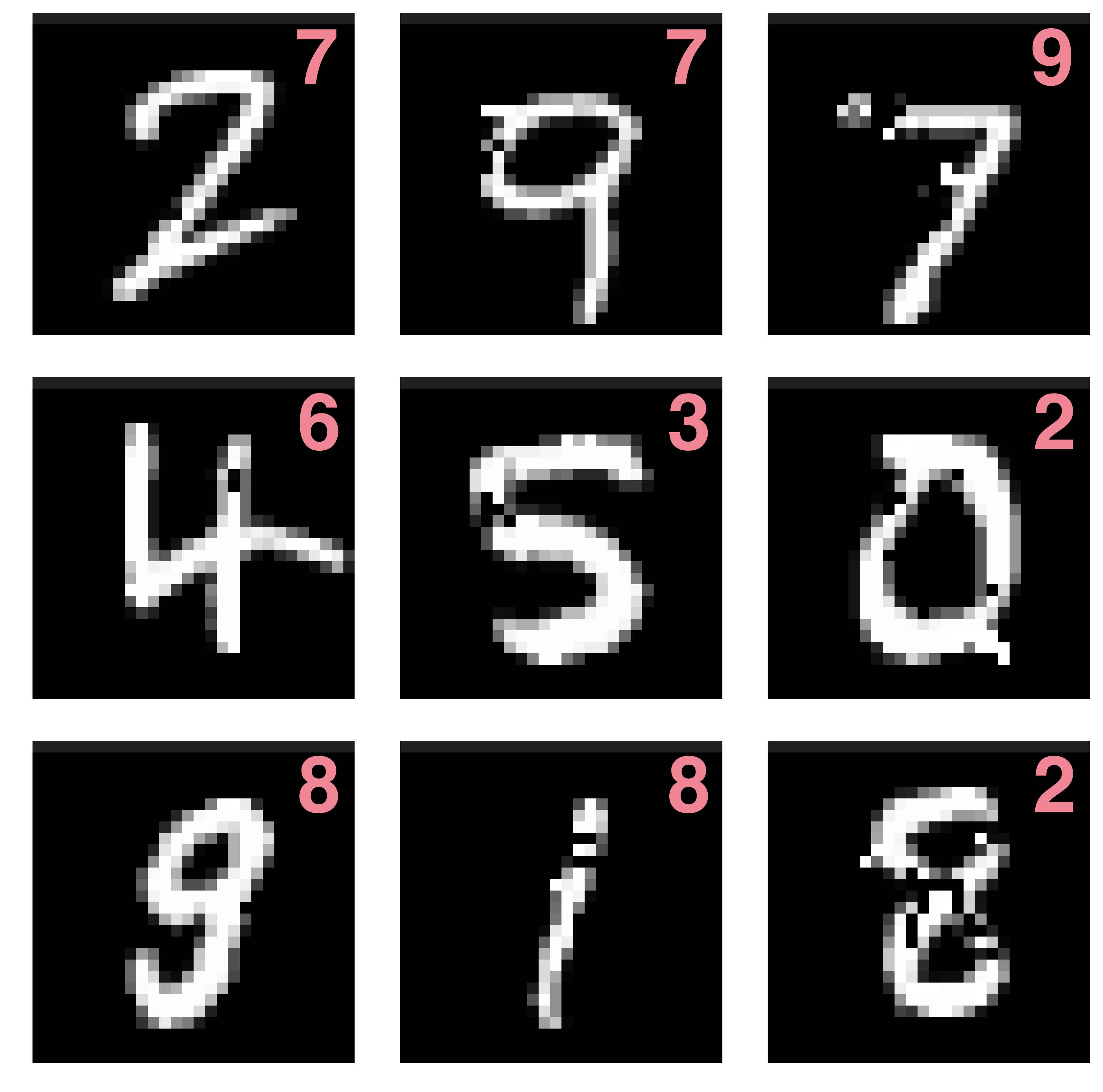}
    \captionsetup{font=small}
    \caption{MNIST}
    \label{fig:sparse-pert_mnist}
    \end{subfigure}\hspace{-0.15em}%
    \begin{subfigure}[b]{0.45\columnwidth}
        \centering
        \captionsetup{font=scriptsize}
        \includegraphics[width=0.7\linewidth, page=2]{images/ch3_sparse/mnist_cifar.pdf}
    \captionsetup{font=small}
    \caption{CIFAR-10}
    \label{fig:sparse-pert_cifar}
    \end{subfigure}
    \caption{SparseFool adversarial examples for (a) MNIST and (b) CIFAR-10. Each column corresponds to different level of perturbed pixels.}
    \label{fig:sparse-pert_mnist_cifar}
\end{figure}
\begin{figure}[!ht]
\centering
    \begin{subfigure}[b]{0.25\columnwidth}
        \includegraphics[width=\linewidth]{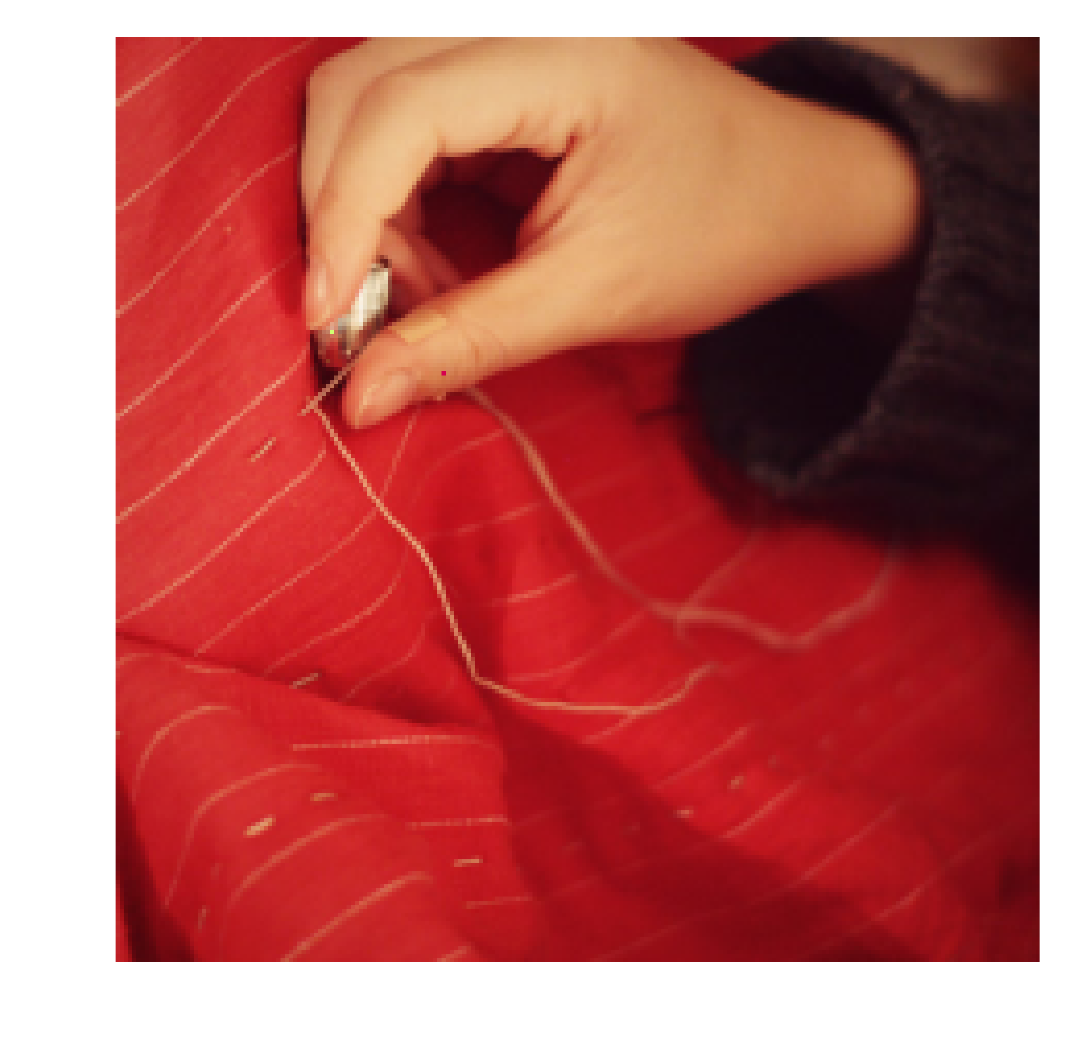}
        \vspace*{-7mm}
        \caption*{cockroach}
    \end{subfigure}\!
    \begin{subfigure}[b]{0.25\columnwidth}
        \includegraphics[width=\linewidth]{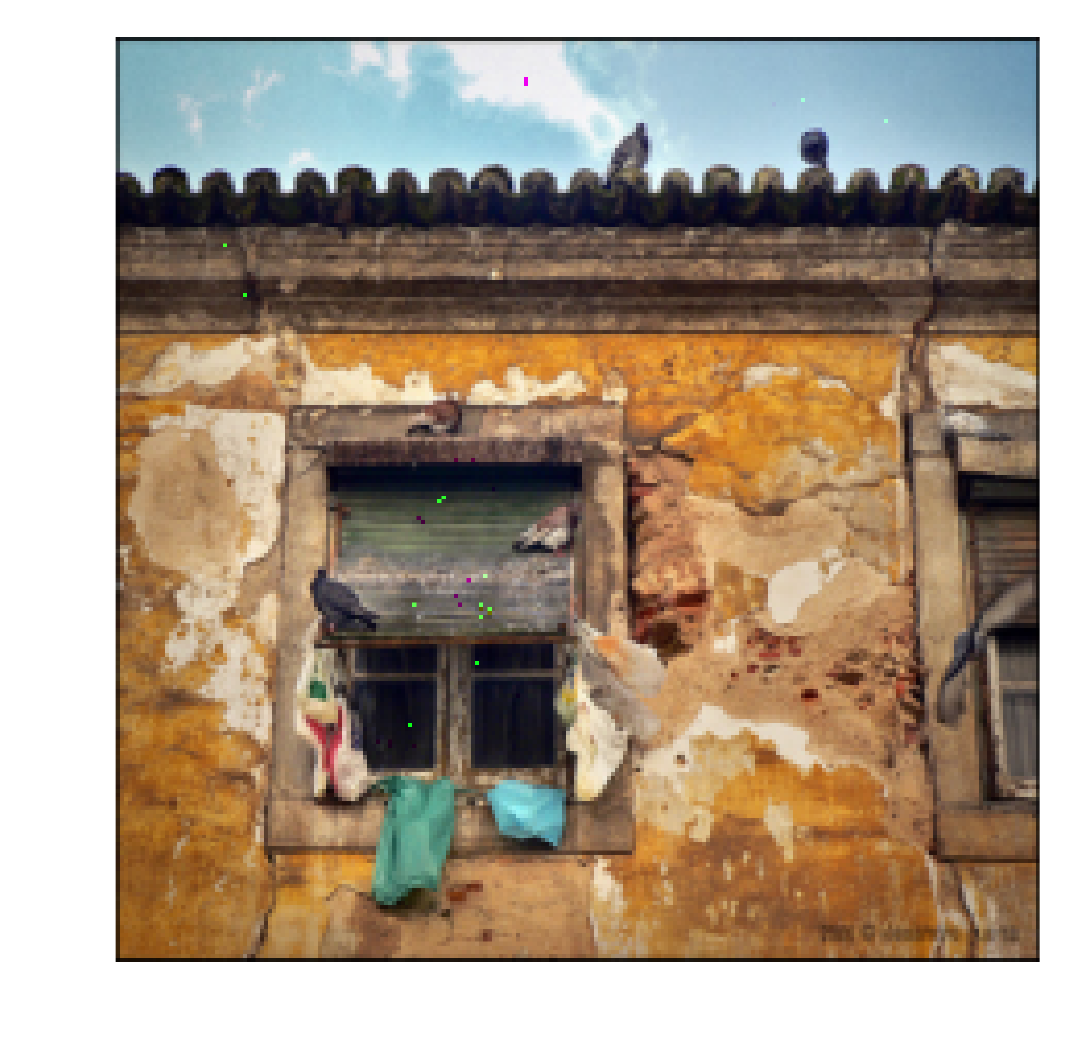}
        \vspace*{-7mm}
        \caption*{palace}
    \end{subfigure}\!
    \begin{subfigure}[b]{0.25\columnwidth}
        \includegraphics[width=\linewidth]{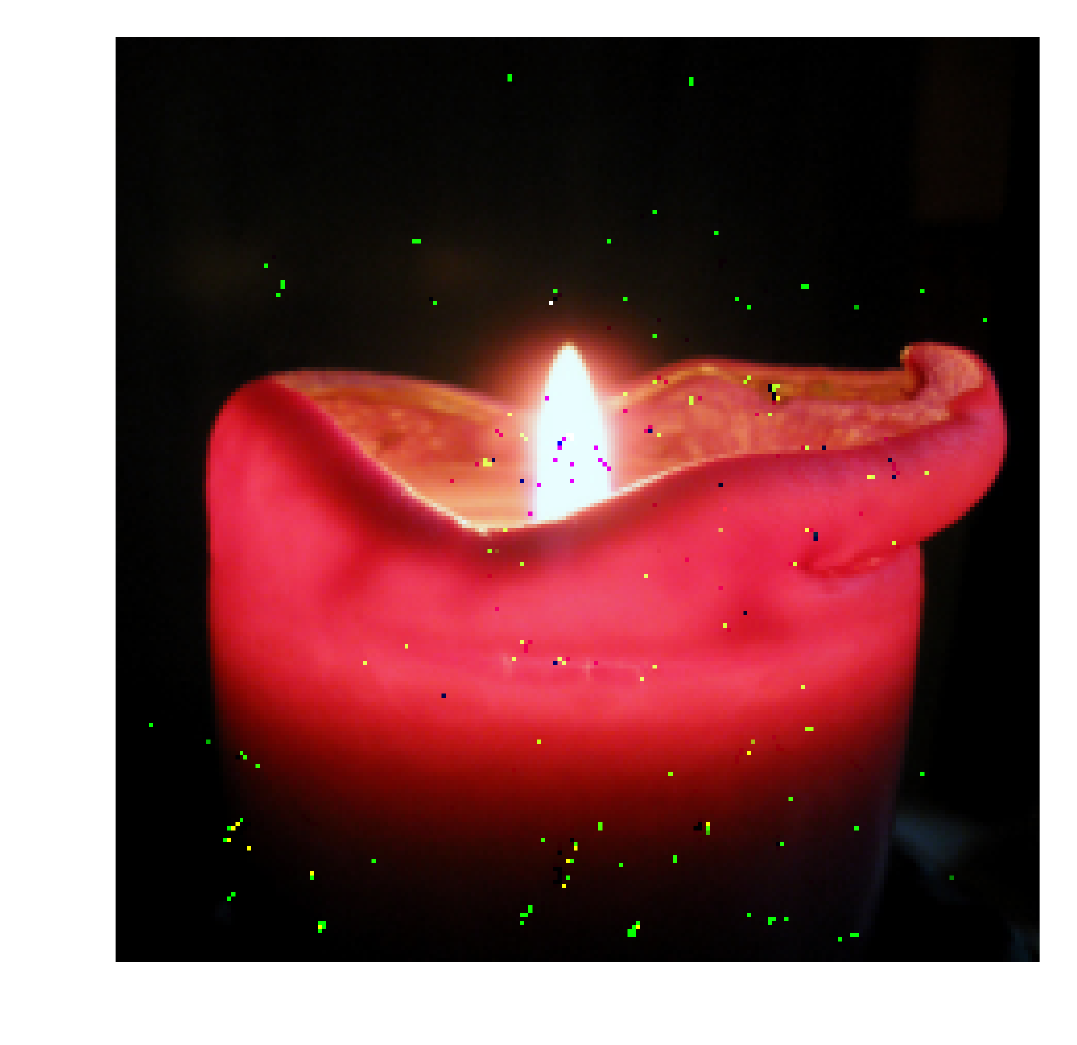}
        \vspace*{-7mm}
        \caption*{bathtub}
    \end{subfigure}
    
    \begin{subfigure}[b]{0.25\columnwidth}
        \includegraphics[width=\linewidth]{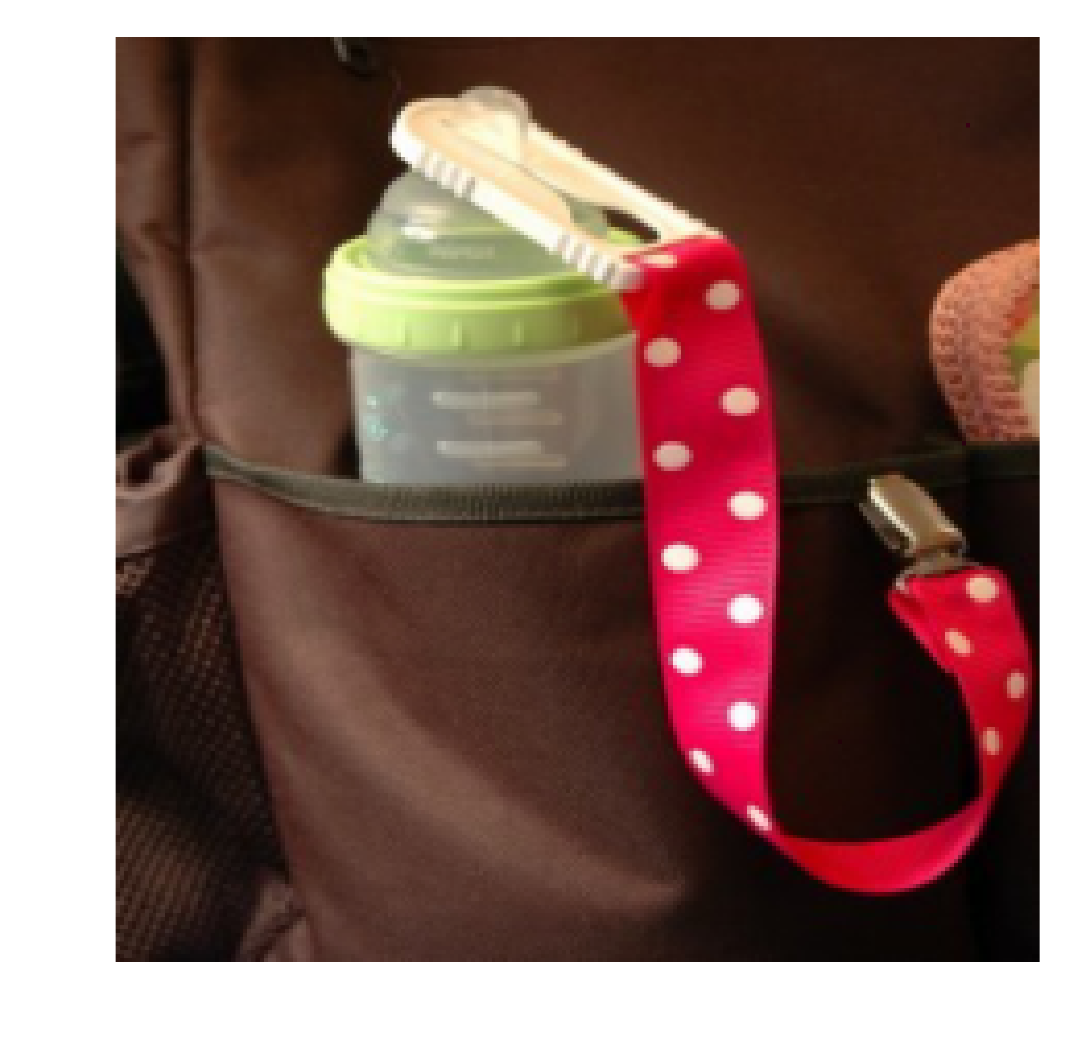}
        \vspace*{-7mm}
        \caption*{sandal}
    \end{subfigure}\!
    \begin{subfigure}[b]{0.25\columnwidth}
        \includegraphics[width=\linewidth]{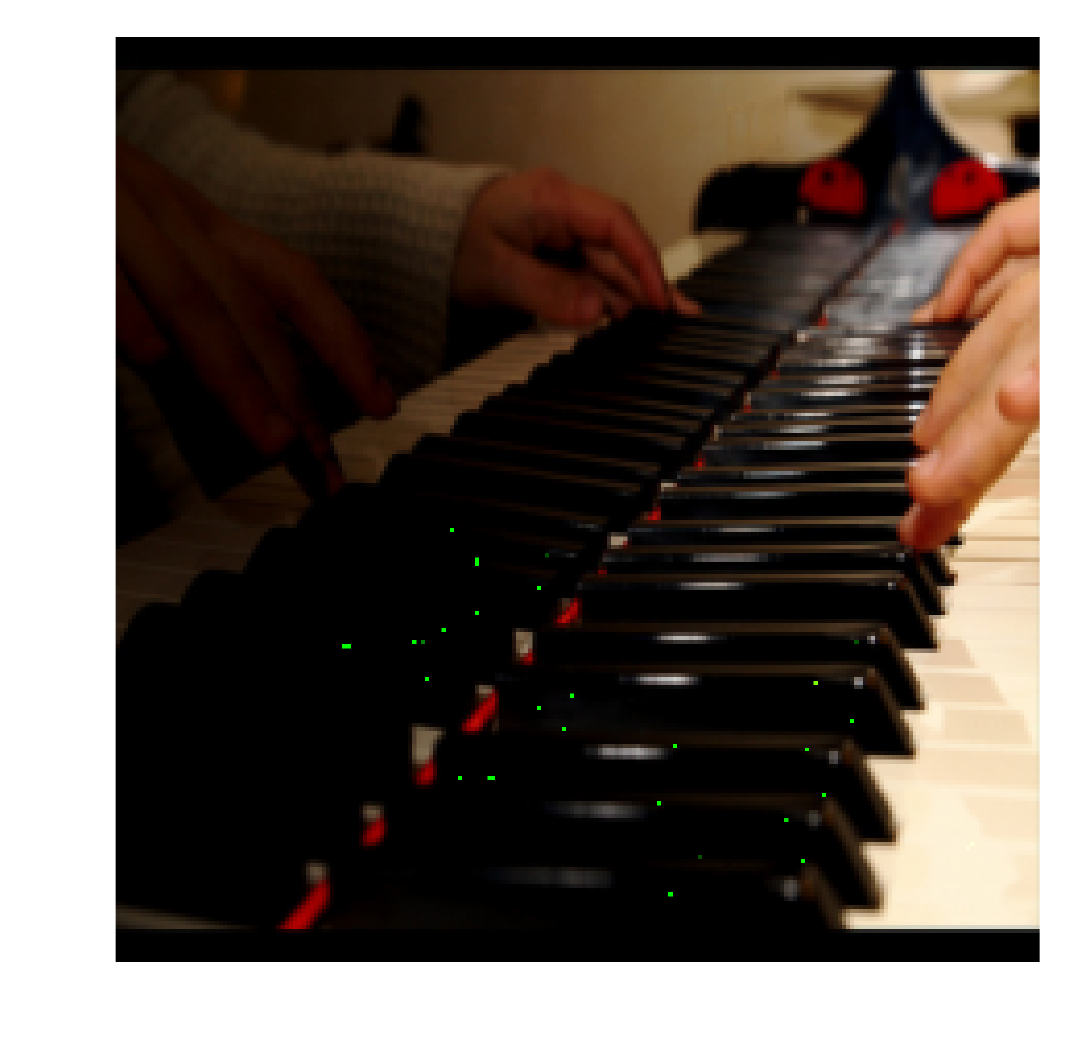}
        \vspace*{-7mm}
        \caption*{wine bottle}
    \end{subfigure}\!
    \begin{subfigure}[b]{0.25\columnwidth}
        \includegraphics[width=\linewidth]{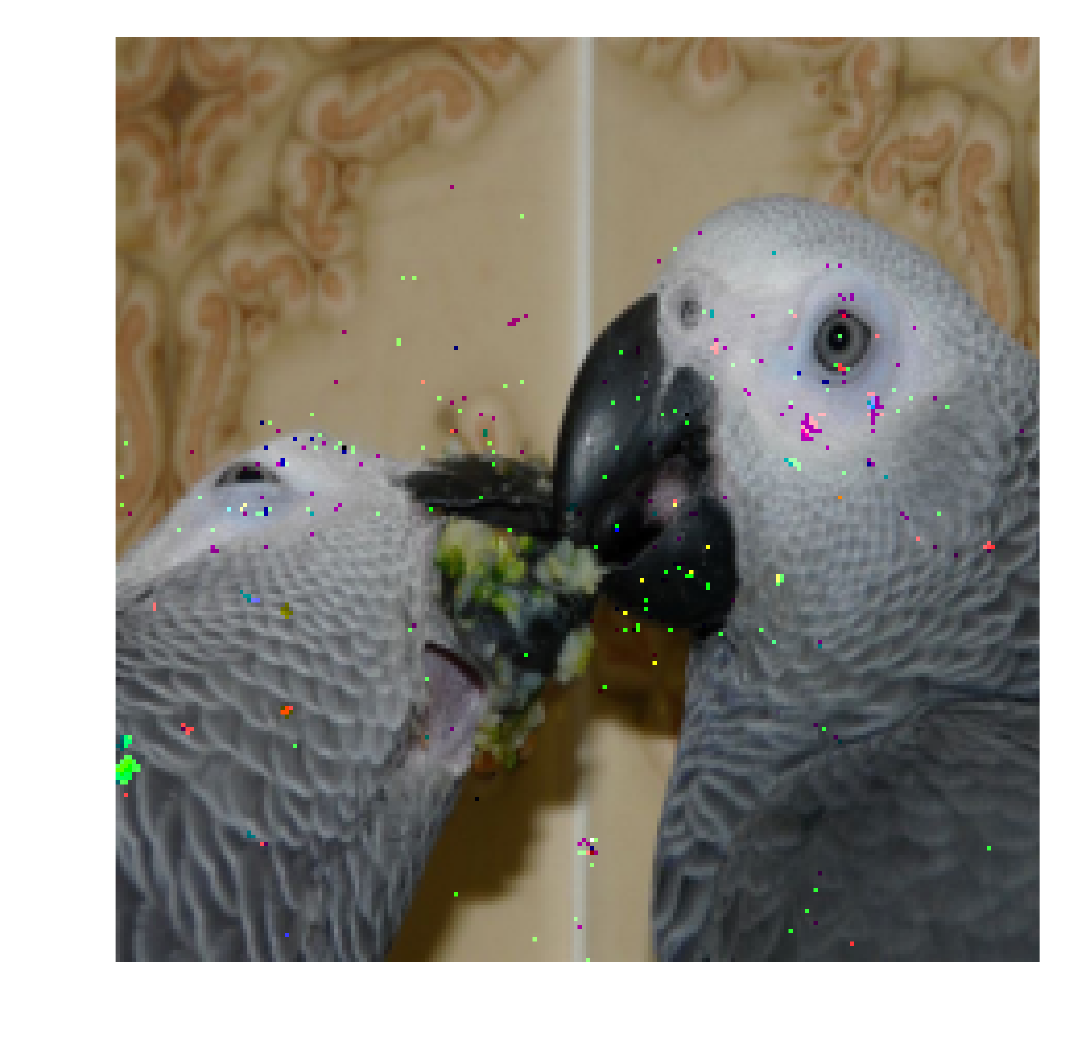}
        \vspace*{-7mm}
        \caption*{bubble}
    \end{subfigure}
    \caption{SparseFool adversarial examples for ImageNet. Each column corresponds to different level of perturbed pixels. The fooling labels are shown below the images.}
    \label{fig:sparse-pert_imagenet}
\end{figure}

We now illustrate some adversarial examples generated by SparseFool for three different levels of sparsity: highly sparse perturbations, sparse perturbations, and somewhere in the middle. For MNIST and CIFAR-10 (\cref{fig:sparse-pert_mnist} and~\cref{fig:sparse-pert_cifar} respectively), we observe that for highly sparse cases, the perturbation is either imperceptible or can be easily ignored. However, as the number of perturbed pixels increases, the distortion becomes even more perceptible, and in some cases the noise is detectable and far from imperceptible. A similar behavior is also observed for the ImageNet dataset (\cref{fig:sparse-pert_imagenet}).

To eliminate this perceptibility effect, we focus on the lower/upper bounds of the values of the adversarial image $\hat{\bm{x}}=\bm{x}+\bm\delta$. Recall from \cref{subsec:sparse-problem-linearization} that the bounds $\bm{l}$, $\bm{u}$ are defined such that $l_i \leq \hat{x}_i \leq u_i, \enskip i=1\dots D$. If these bounds represent the dynamic range of the image, then $\hat{x}_i$ can take every possible value from this range, and the magnitude of the noise at the element $i$ can reach visible levels. However, if the perturbed values lie close to the original values $x_i$, then we might prevent the magnitude from reaching very high levels. Hence, assuming a dynamic range of $[0,255]$, we explicitly constrain the values of $\hat{x}_i$ to lie in a small interval $\pm\alpha$ around $x_i$, such that $0\leq x_i-\alpha\leq\hat{x}_i\leq x_i+\alpha \leq 255$.

\begin{figure}[b]
\centering
\includegraphics[width=0.43\columnwidth]{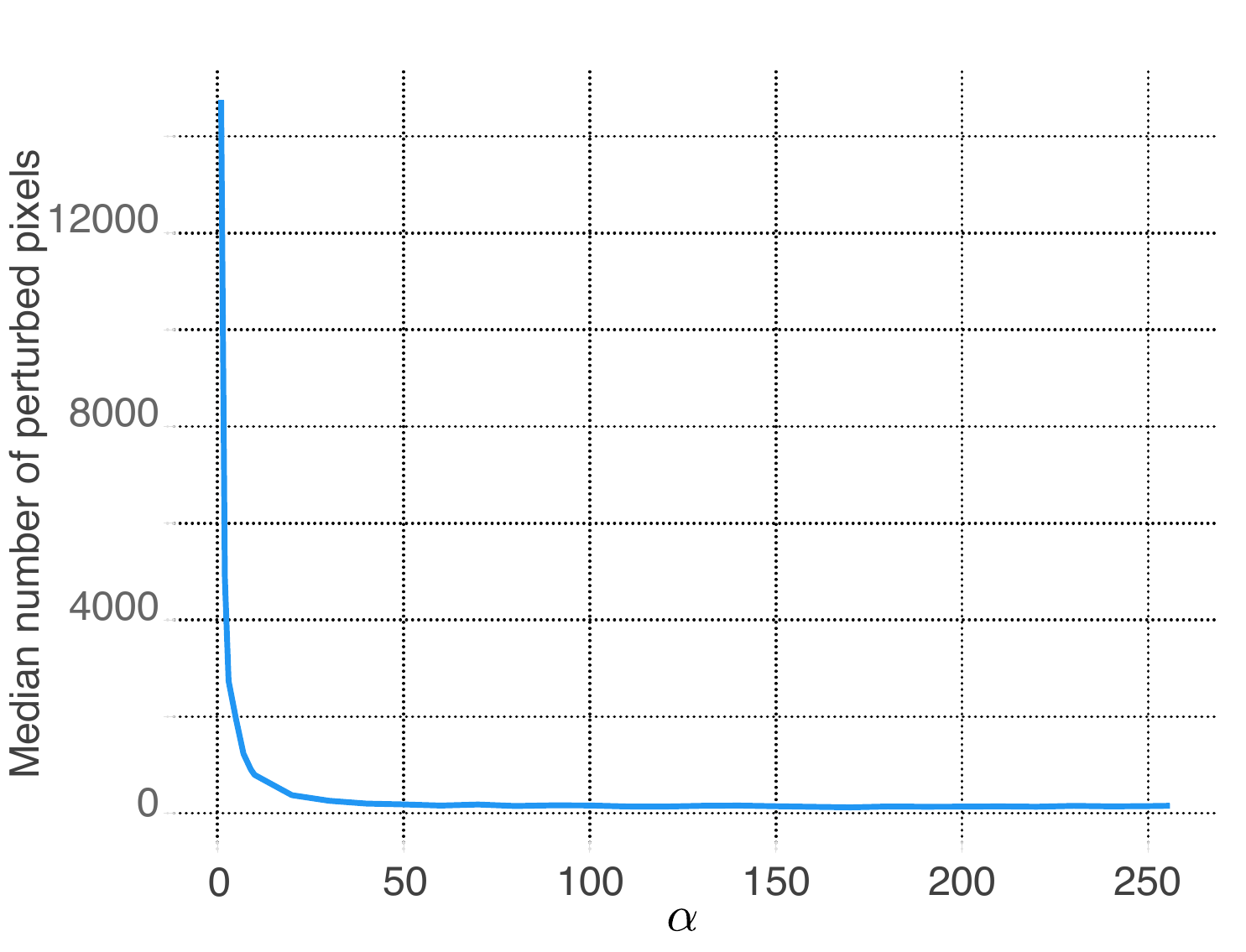}
\caption{The resulted sparsity of SparseFool perturbations for $\pm \alpha$ around the values of $\bm{x}$, for $100$ samples from ImageNet on a ResNet-101 architecture.}
\label{fig:sparse-perc_sparse_const}
\end{figure}

\begin{figure}[!ht]
    \captionsetup[subfigure]{justification=centering}
    \centering
    \begin{subfigure}[b]{0.3\columnwidth}
        \caption*{$x_i \pm 255$}
        \includegraphics[width=0.8\linewidth]{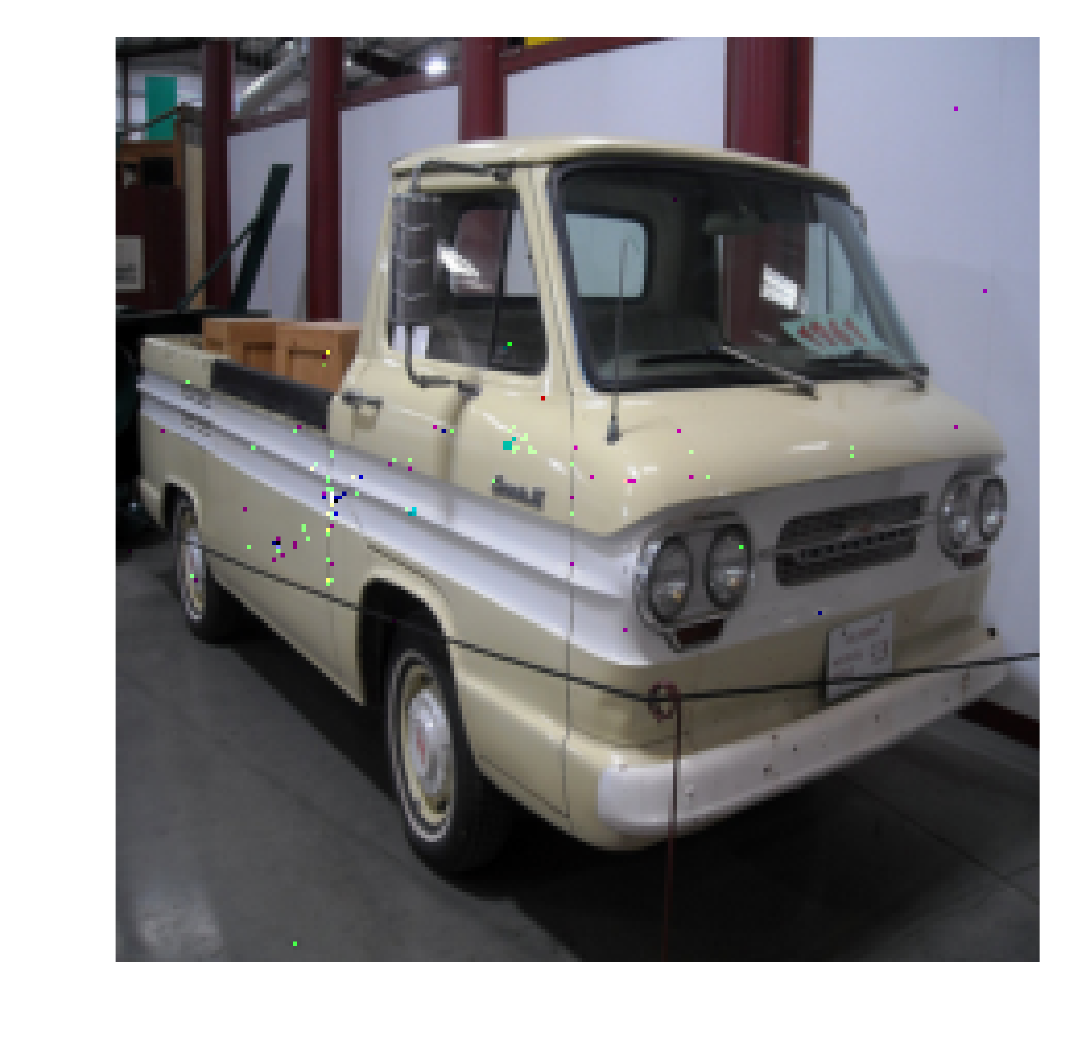}
        \caption*{amphibian\\($0.227\%$)}
    \end{subfigure}
    \begin{subfigure}[b]{0.3\columnwidth}
        \caption*{$x_i \pm 30$}
        \includegraphics[width=0.8\linewidth]{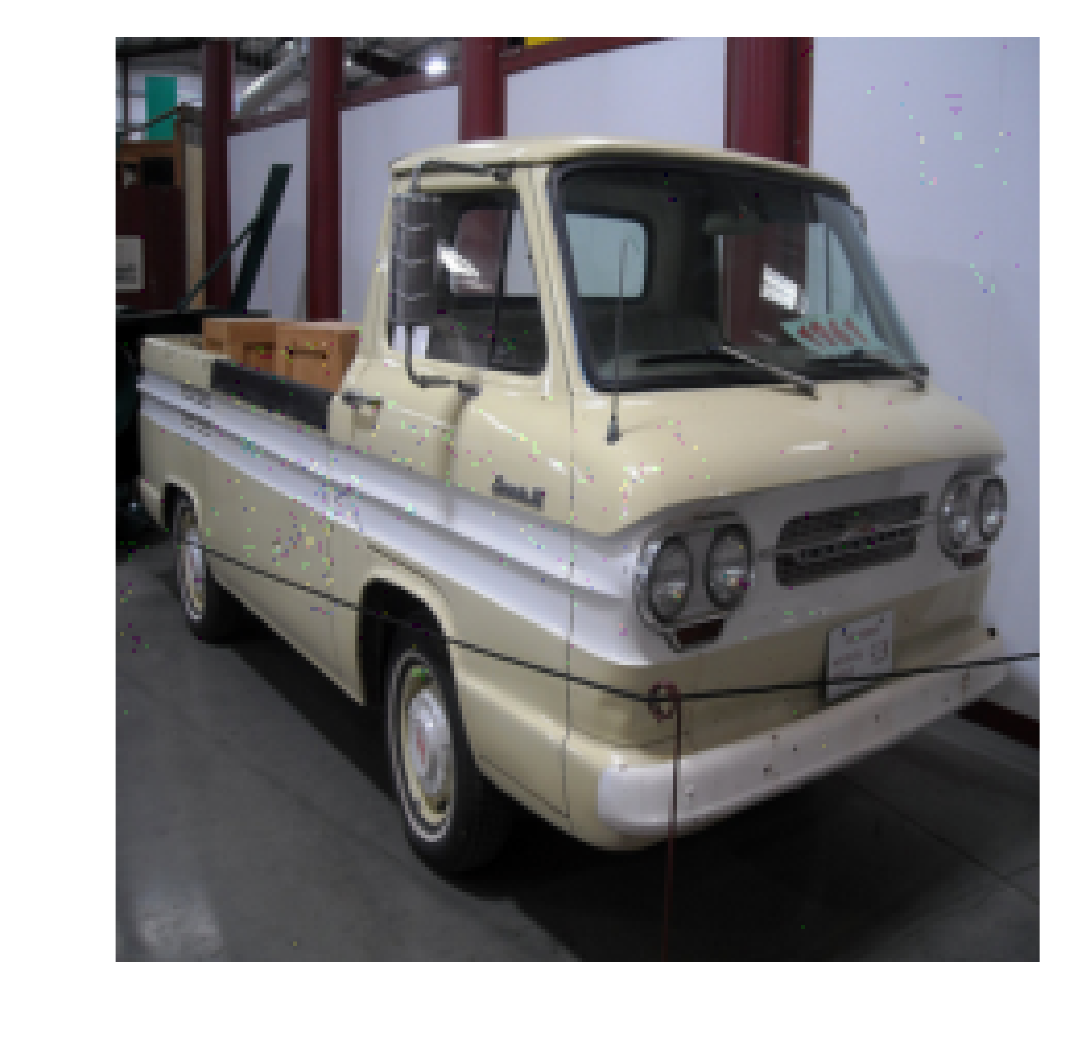}
        \caption*{amphibian\\($1.058\%$)}
    \end{subfigure}
    \begin{subfigure}[b]{0.3\columnwidth}
        \caption*{$x_i \pm 10$}
        \includegraphics[width=0.8\linewidth]{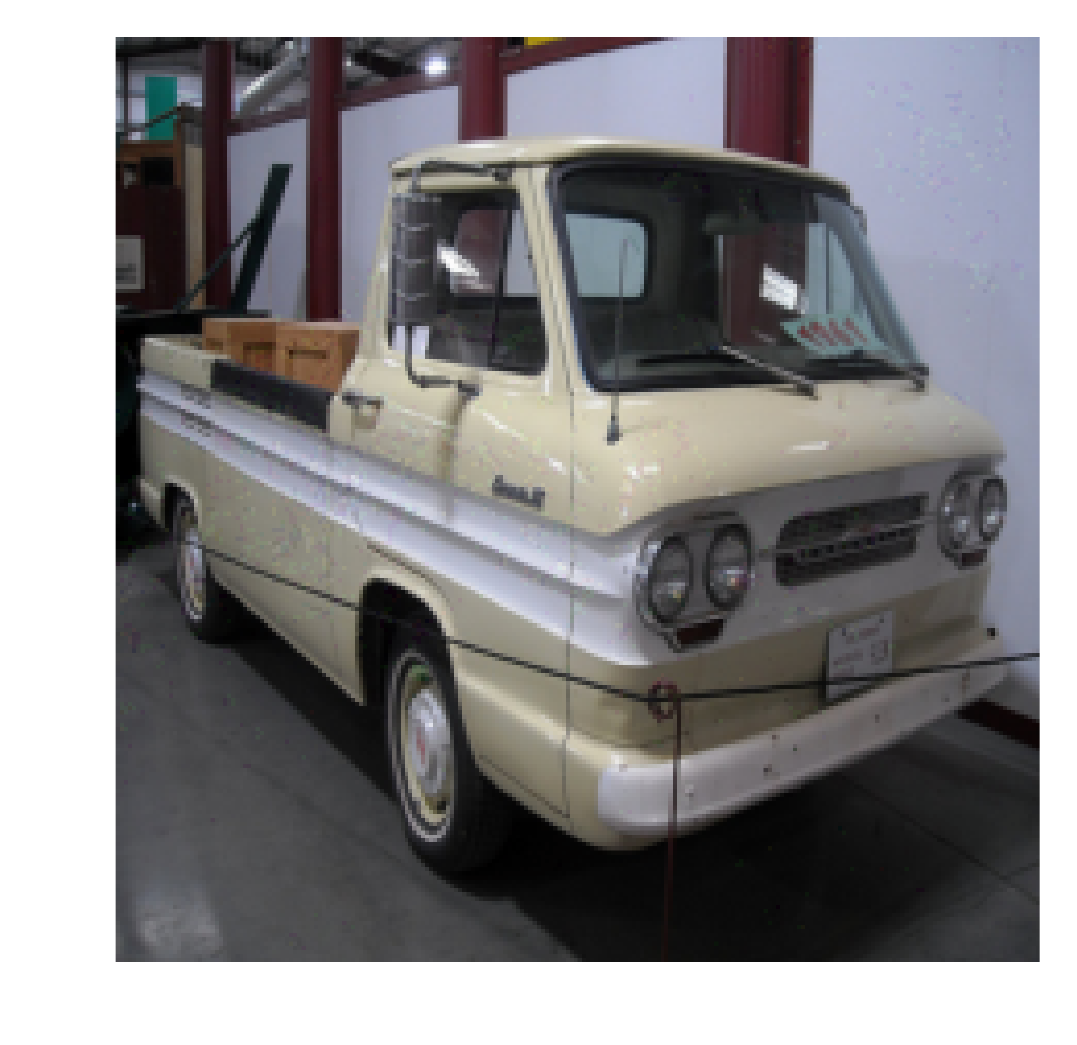}
        \caption*{amphibian\\($4.296\%$)}
    \end{subfigure}
    
    \begin{subfigure}[b]{0.3\columnwidth}
        \includegraphics[width=0.8\linewidth]{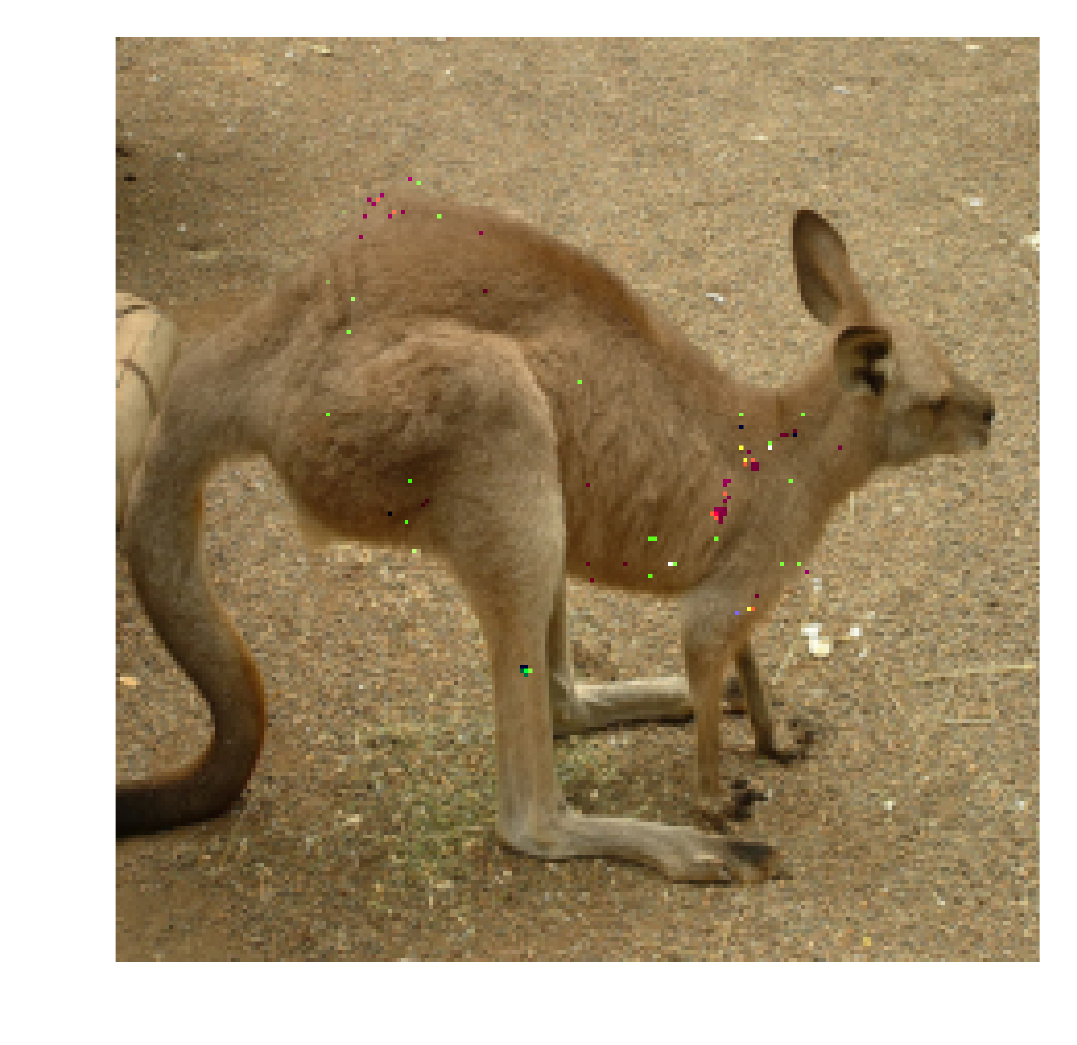}
        \caption*{Arabian camel\\($0.169\%$)}
    \end{subfigure}
    \begin{subfigure}[b]{0.3\columnwidth}
        \includegraphics[width=0.8\linewidth]{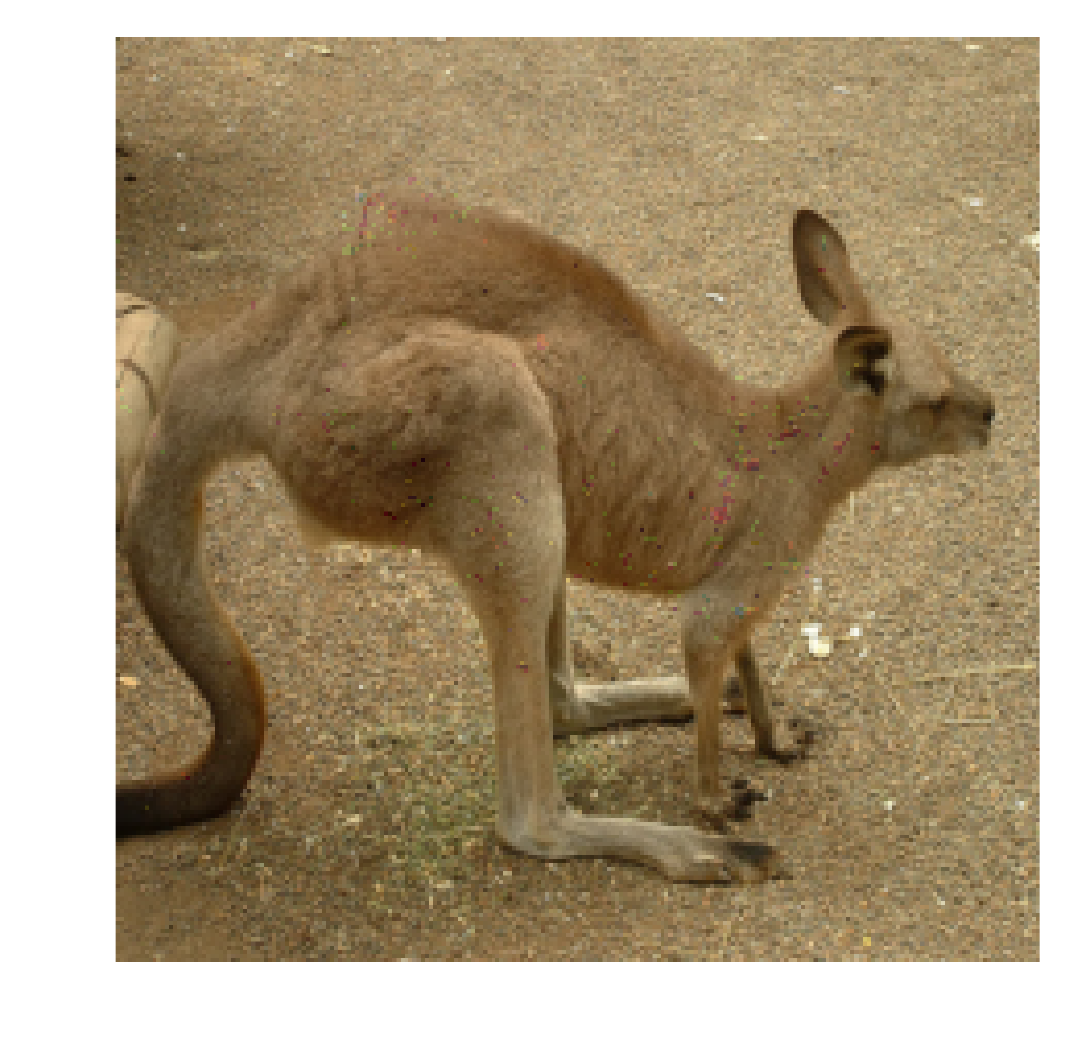}
        \caption*{Arabian camel\\($0.839\%$)}
    \end{subfigure}
    \begin{subfigure}[b]{0.3\columnwidth}
        \includegraphics[width=0.8\linewidth]{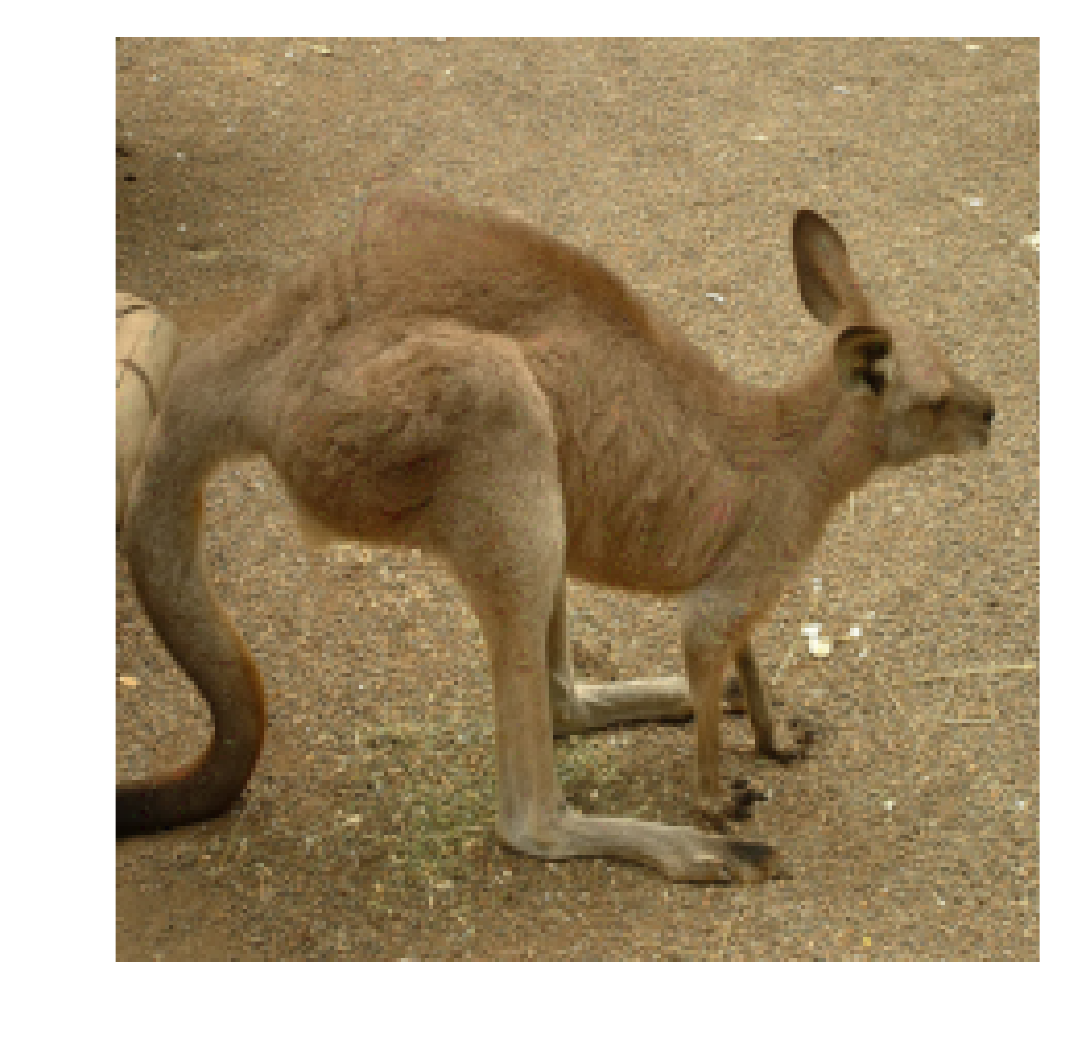}
        \caption*{Arabian camel\\($3.202\%$)}
    \end{subfigure}
    \caption{The effect of $\alpha$ on the perceptibility and the sparsity of SparseFool perturbations. The values of $\alpha$ are shown on top of each column, while the fooling label and the percentage of perturbed pixels are written below each image.}
    \label{fig:sparse-percept_delta}
\end{figure}

The sparsity for different values of $\alpha$ is shown in \cref{fig:sparse-perc_sparse_const}. Higher values give  more freedom to the perturbations, but after $\alpha\approx25$ the sparsity levels remain almost constant, suggesting that we do not need the whole dynamic range. Furthermore, we observed that the average execution time per sample of SparseFool from this value onward remains constant as well, while the fooling rate is $100\%$ regardless $\alpha$. Thus, by properly selecting $\alpha$, we can control the perceptibility of the perturbations and retain sparsity at a sufficient level. The influence of $\alpha$ on the perceptibility and sparsity is demonstrated in ~\cref{fig:sparse-percept_delta}.

\section{Analysis of sparse perturbations}
\label{sec:sparse-analysis}

\subsection{Shared semantic features}
\label{subsec:sparse-shared-semantic-features}

We now analyze different properties of the perturbations generated with SparseFool. We first investigate if the perturbations transfer across different architectures. For the VGG-16, ResNet-101, and DenseNet-161 architectures, we report in Table~\ref{tab:sparse-transfer} the fooling rate of each model when fed with adversarial examples generated for another one. We observe that sparse perturbations can generalize only to some extent, and that they are more transferable from larger to smaller architectures. This indicates that there should be some shared semantic information between different architectures that SparseFool exploits, but the exact structure of the perturbations is mostly network dependent.

\begin{table}[t]
\begin{center}
{
    \small
	\begin{tabular}{| m{1.8cm} | m{1.7cm} | m{1.7cm} | m{1.6cm} |}
	\cline{2-4}
	\multicolumn{1}{c|}{}
	& \multicolumn{1}{c|}{\makecell{VGG16}}
	& \multicolumn{1}{c|}{\makecell{ResNet101}}
	& \multicolumn{1}{c|}{\makecell{DenseNet161}}\\
	\hline
	VGG16
	& \multicolumn{1}{c|}{$100\%$}
	& \multicolumn{1}{c|}{$10.8\%$} 
	& \multicolumn{1}{c|}{$8.2\%$} \\
	\hline
	ResNet101
	& \multicolumn{1}{c|}{$25.3\%$}
	& \multicolumn{1}{c|}{$100\%$} 
	& \multicolumn{1}{c|}{$12.1\%$} \\
	\hline
	\multicolumn{1}{|c|}{DenseNet161}
	& \multicolumn{1}{c|}{$28.2\%$}
	& \multicolumn{1}{c|}{$17.5\%$} 
	& \multicolumn{1}{c|}{$100\%$} \\
	\hline
\end{tabular}
}
\end{center}
\caption{Fooling rates of SparseFool perturbations between pairs of models for $4000$ samples from ImageNet. Row/column denote the source/target model respectively.}
\label{tab:sparse-transfer}
\end{table}

\begin{figure}[t]
\centering
    \begin{subfigure}[b]{0.25\columnwidth}
        \caption{VGG-16}
        \includegraphics[width=\linewidth]{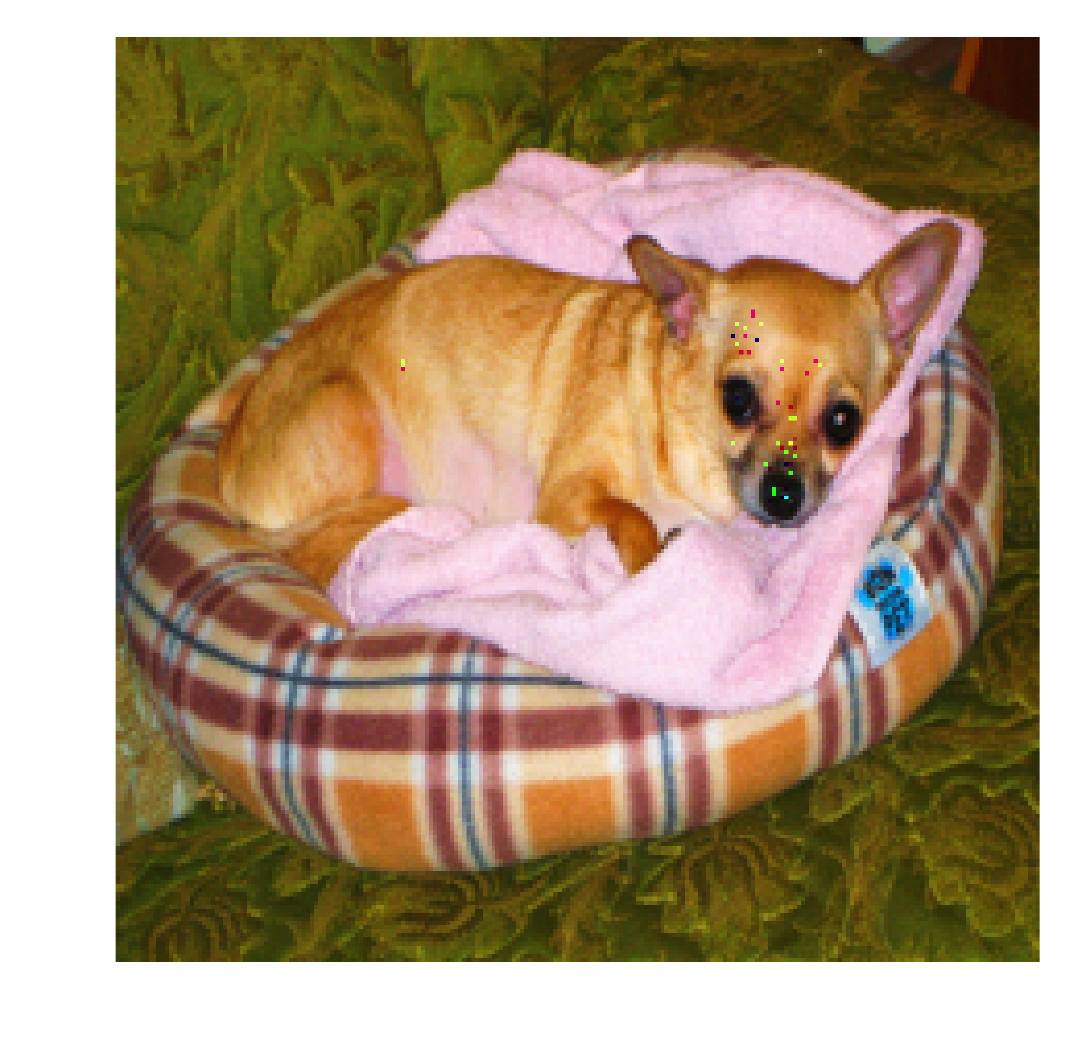}
    \end{subfigure}\!
    \begin{subfigure}[b]{0.25\columnwidth}
        \caption{ResNet-101}
        \includegraphics[width=\linewidth]{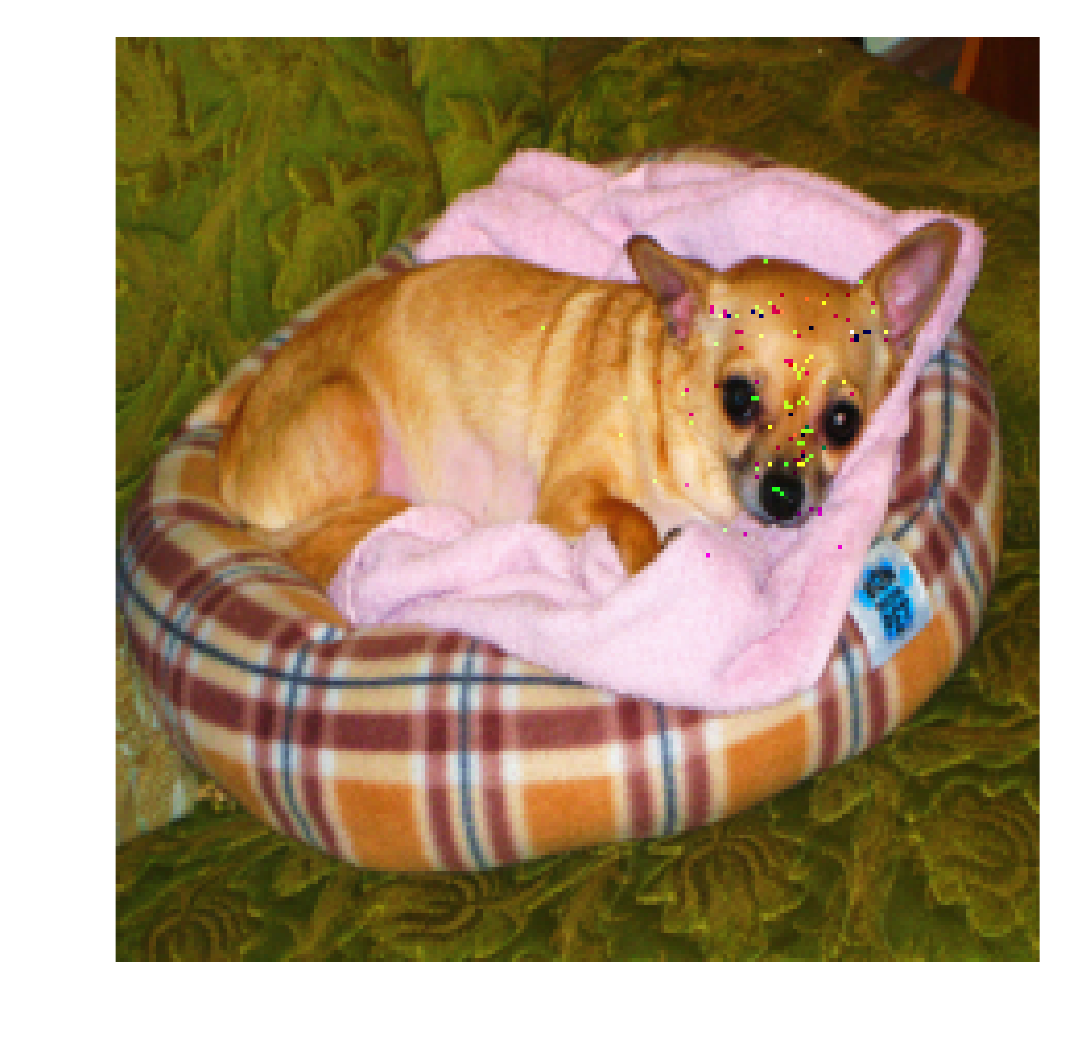}
    \end{subfigure}\!
    \begin{subfigure}[b]{0.25\columnwidth}
        \caption{DenseNet-161}
        \includegraphics[width=\linewidth]{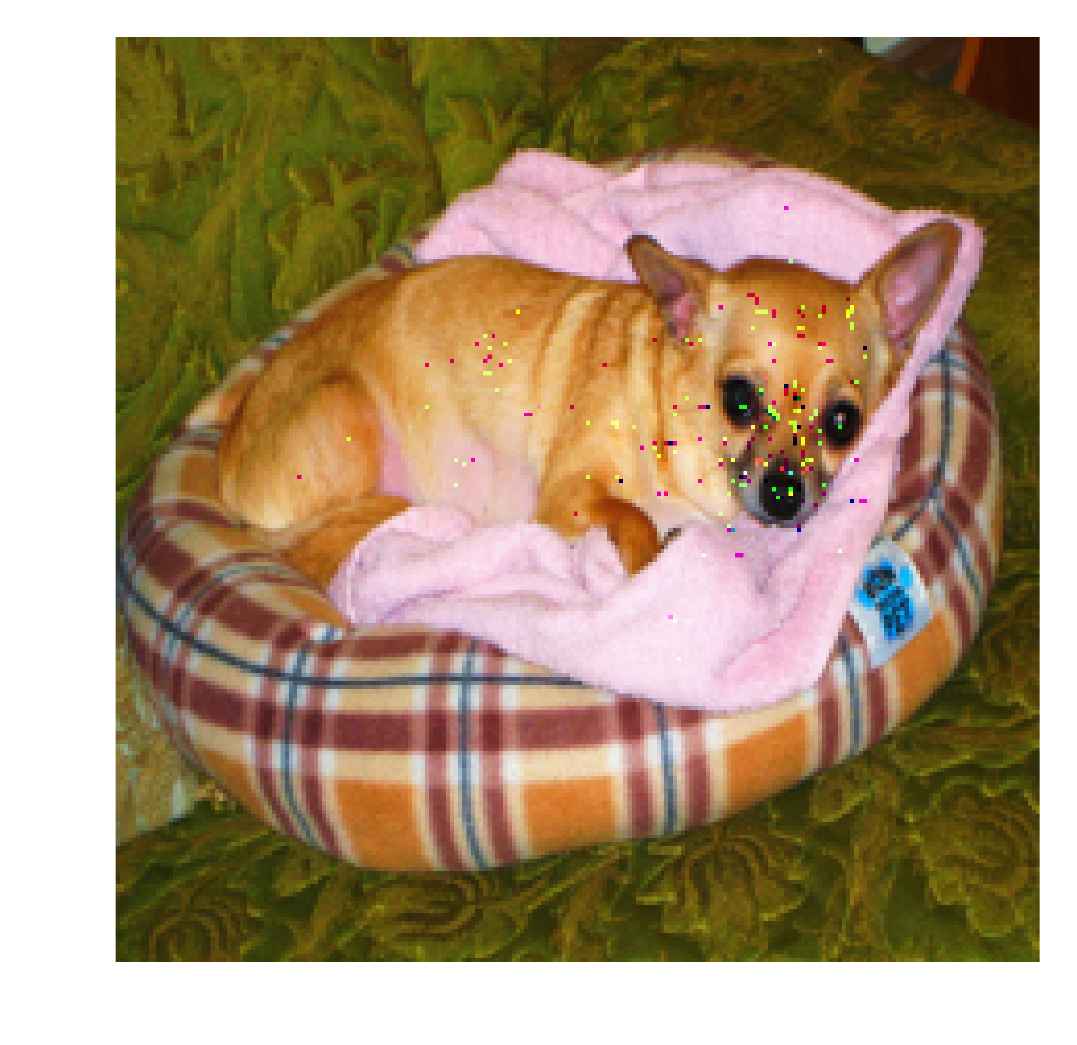}
    \end{subfigure}
    
    \begin{subfigure}[b]{0.25\columnwidth}
        \includegraphics[width=\linewidth]{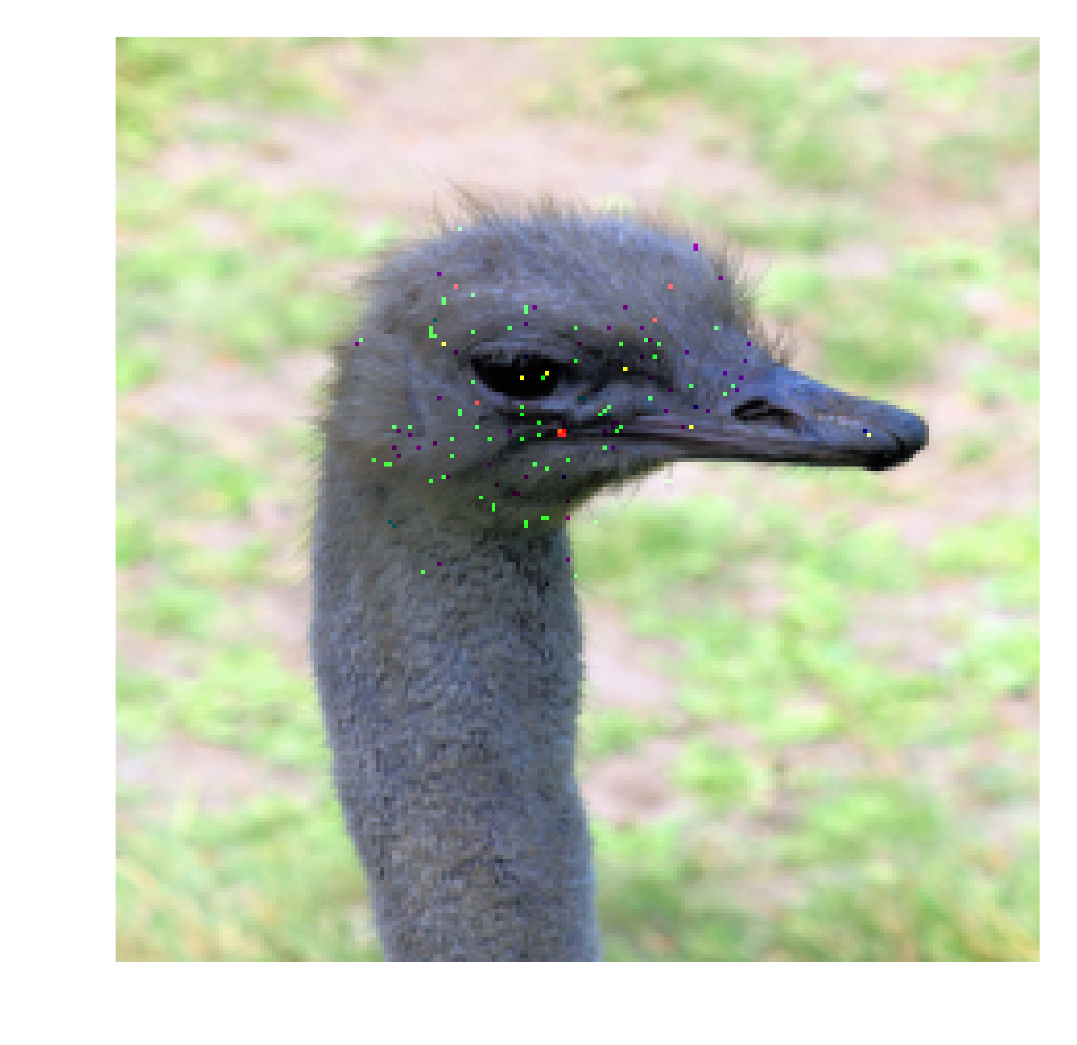}
    \end{subfigure}\!
    \begin{subfigure}[b]{0.25\columnwidth}
        \includegraphics[width=\linewidth]{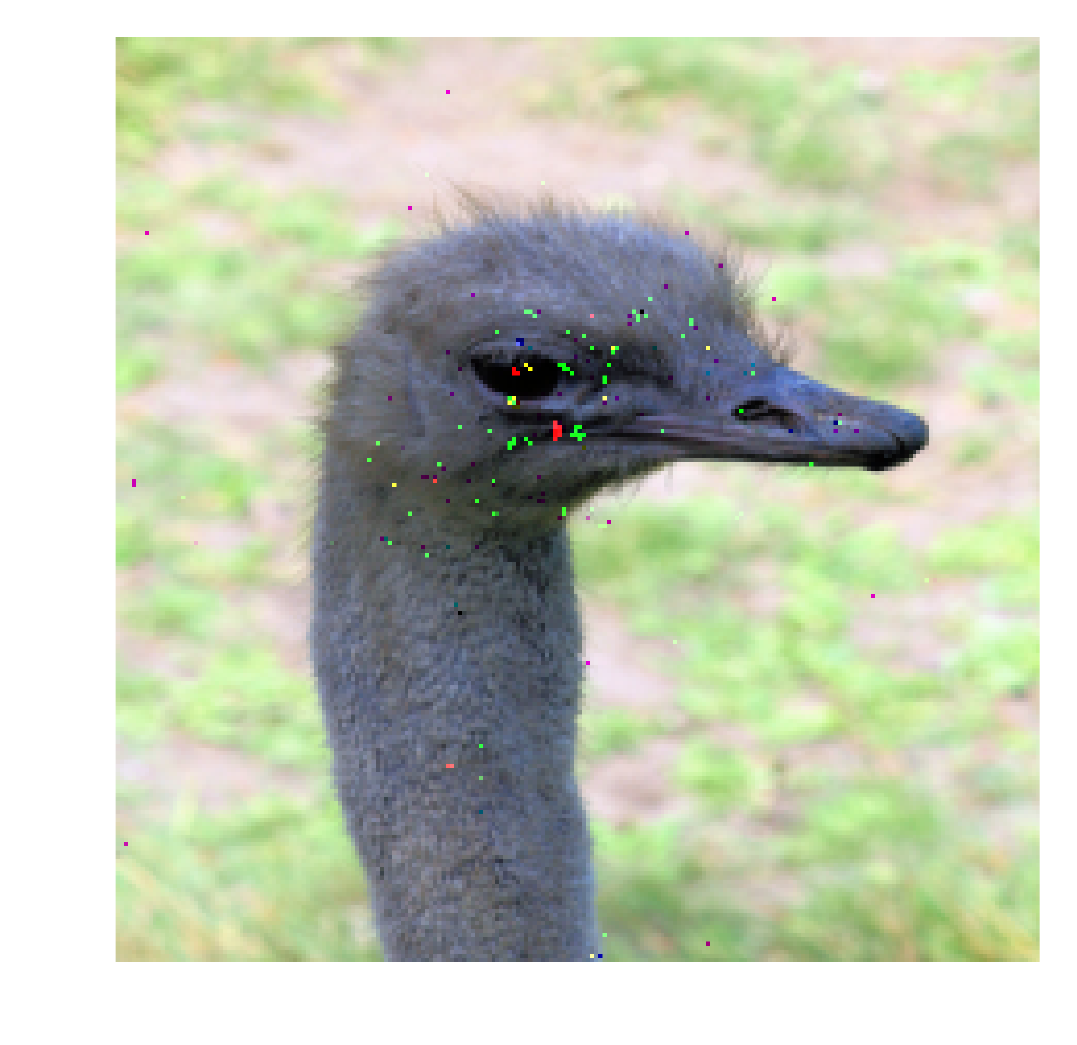}
    \end{subfigure}\!
    \begin{subfigure}[b]{0.25\columnwidth}
        \includegraphics[width=\linewidth]{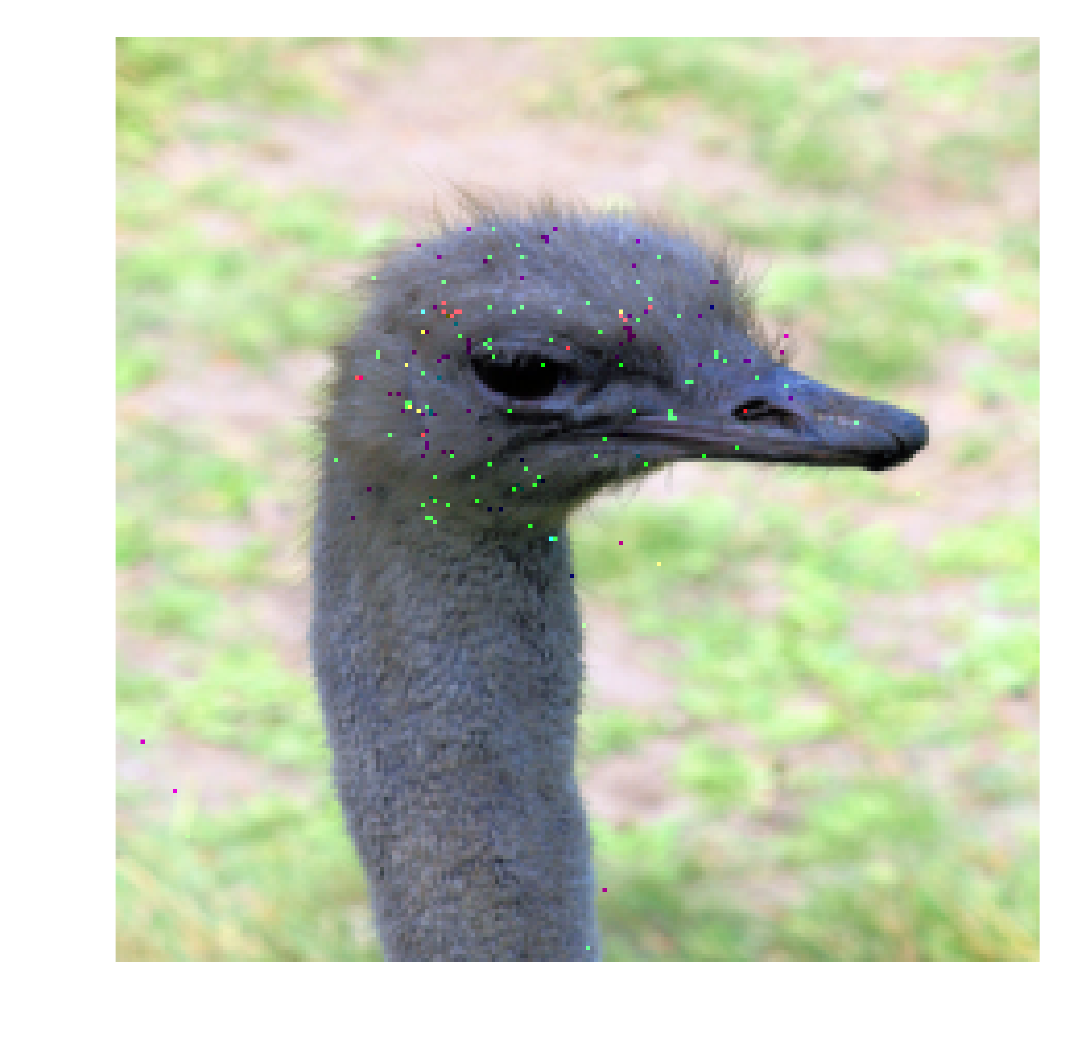}
    \end{subfigure}
    \caption{Shared information of SparseFool perturbations across three different networks. For all networks, the first row image was classified as ``Chihuahua'' and misclassified as ``French Bulldog'', and second row image as ``Ostrich'' and ``Crane'' respectively.}
    \label{fig:sparse-univ_imagenet}
\end{figure}

We inspect some animal categories of the ImageNet dataset
and examine if semantic information is shared across different architectures (\cref{fig:sparse-univ_imagenet}). For all networks, the perturbation consistently lies around the important -- from a human perspective -- areas (i.e., head) of the image, but the way it concentrates or spreads differs for each network. One could say that this is different from dense ($p\in\{2,\infty\}$) perturbations, which exploit more shared input directions and hence have higher transferability. In this sense, we can say that sparse perturbations are identifying directions that align well with human intuition, but at the same time are more unique to each architecture.

\begin{figure}[t]
    \centering
    \begin{subfigure}[b]{0.4\columnwidth}
        \centering
        \includegraphics[width=\linewidth, page=1]{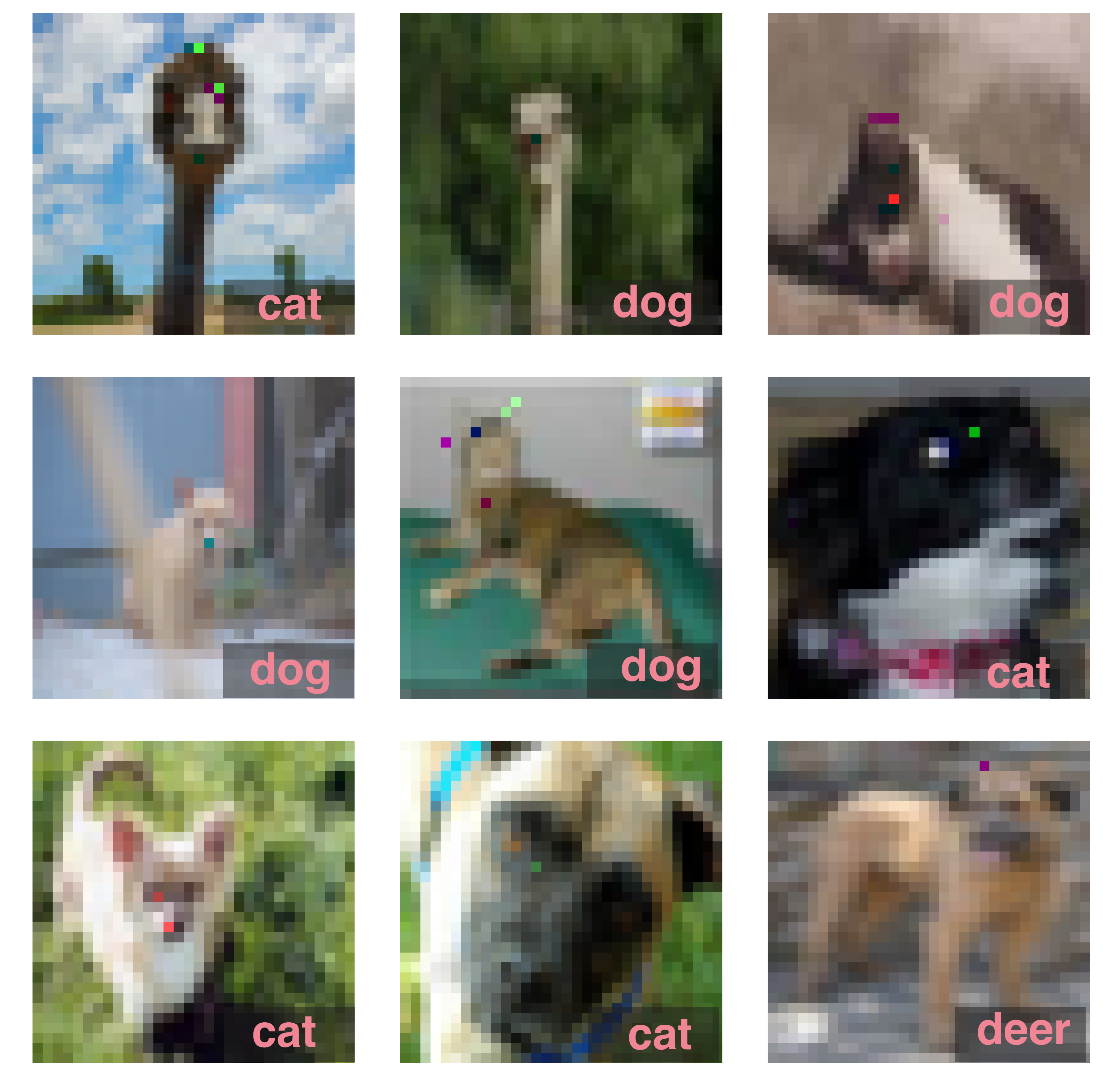}
    \caption{Common features}
    \label{fig:sparse-univ_features}
    \end{subfigure}\hspace{2em}%
    \begin{subfigure}[b]{0.4\columnwidth}
        \centering
        \includegraphics[width=\linewidth, page=2]{images/ch3_sparse/univ_feat.pdf}
    \caption{Fooling class features}
    \label{fig:sparse-univ_target}
    \end{subfigure}
    \caption{Semantic information of SparseFool perturbations for the CIFAR-10 dataset, on a ResNet-18. Observe that the perturbation is concentrated (a) on some features around the area of the face, or (b) on areas that are important for the fooling class.}
    \label{fig:sparse-univ_cifar}
\end{figure}

For the CIFAR-10 dataset, we observe that in many cases of animal classes, SparseFool tends to perturb some common features around the area of the head (i.e., eyes, ears, nose, mouth etc.), as shown in \cref{fig:sparse-univ_features}. Furthermore, we tried to understand if there is a correlation between the perturbed pixels and the fooling class. Interestingly, as shown in \cref{fig:sparse-univ_target}, we observe that in many cases the algorithm perturbs those regions of the image that correspond to important features of the fooling class, i.e., when changing a ``bird'' label to a ``plane'', where the perturbation seems to resemble some parts of the plane (i.e., wings, tail, nose, turbine). This behavior becomes even more evident when the fooling label is a ``deer'', where the noise lies mostly around the area of the head in a way that resembles the antlers.

\subsection{Exclusiveness of adversarial robustness}
\label{subsec:sparse-exclusiveness-adversarial-robustness}

Finally, we want to understand to what extent sparse perturbations can be ``covered'' by dense perturbations of larger $\ell_p$ regimes, when building robust classifiers. To this end, we perform adversarial training on a ResNet-18 on the CIFAR-10 dataset, using $\ell_\infty$  perturbations of $\epsilon=8/255$ crafted with the PGD attack, as described in~\cite{Madry2018TowardsDeepLearning}. Note that these are the most commonly used settings for building more robust classifiers. The accuracy of this more robust model on the CIFAR-10 test set is $82.17\%$. Then, we compute the adversarial examples for this model using SparseFool and we measure the fooling rate, median perturbation percentage, and average execution time.

Compared to the results of \cref{tab:sparse-mnist_cifar}, the fooling rate is still $100\%$. Regarding the average execution time, it dropped from $0.69$ to $0.3$ sec, which means that it is ``easier'' to fool the more robust model compared to the original one. This behaviour can be explained by the fact that adversarially trained classifiers have decision boundaries of very small curvature~\cite{Moosavi2019CURE}, hence the linear approximation of SparseFool is better. 

On the other hand, although the perturbation percentage increased from $1.27\%$ to $2.44\%$, SparseFool is still able to compute very sparse perturbations compared to $\ell_\infty$ ones, which perturb almost $100\%$ of the pixels. This result indicates that the effect of $\ell_1$ and $\ell_\infty$ perturbations on the robustness of the networks is somehow ``mutually exclusinve''. This also suggests that the image features exploited by $\ell_1$ perturbations are different from those of $\ell_\infty$ ones, and that building classifiers that are robust to a specific $\ell_p$ regime does not guarantee robustness to other regimes.

\section{Conclusions}
\label{sec:sparse-conclusions}

In this chapter, we proposed a novel geometry-inspired algorithm to compute sparse adversarial perturbations. In order to avoid the NP-hardness of minimizing the $\ell_0$ norm to compute sparse perturbations, we focused on finding an efficient relaxation. To this end, we exploited the low mean curvature of the decision boundaries in the vicinity of the data samples and designed an iterative method that we coin SparseFool. At each iteration SparseFool performs a linear approximation of the decision boundary and solves the simpler $\ell_1$ box-constrained problem to compute sparse adversarial perturbations.

We experimentally demonstrated that SparseFool computes very sparse perturbations, is by orders of magnitude faster than existing methods, and can easily scale to high-dimensional datasets. Furthermore, it incorporates a simple technique to improve the perceptibility of the perturbations, without sacrificing either sparsity or complexity.

By visually inspecting the generated adversarial examples, we observed that SparseFool alters features that are shared among different images, and that in many cases the perturbations resemble image features that are correlated with the fooling class. Finally, we demonstrated that adversarial training with $\ell_\infty$ perturbations does not build classifiers that are invariant to sparse perturbations, suggesting that the image features exploited by $\ell_1$ perturbations are different from those of $\ell_\infty$ ones.

One intriguing observation comes from the visual inspection of the adversarial examples. It actually hints that adversarial perturbations might not necessarily be a ``hole'' in the system, but they might actually reflect some strong connection/correlation between the features of the dataset and the features that the networks use for taking their decisions. In fact, the authors in~\cite{Jetley2018WithFriends} showed that adversarial perturbations span a low-dimension but highly discriminative subspace of the input, and that deep networks exploit simple and brittle features of the dataset, i.e., non-robust features that are aligned with adversarial perturbations. However, the potential explanations behind these connections is still an open question. Hence, in the next chapter we investigate this question and provide a novel framework for connecting the image features with the geometry of the decision boundaries and the inductive biases of deep learning.
\chapter{Analysis of learned features using adversarial proxies}
\label{ch:hold-me}

\begin{raggedleft}
    \textit{``An idea is always a generalization, and generalization is a property of thinking. \\
    To generalize means to think''} \\
    --- Georg Wilhelm Friedrich Hegel \\
\end{raggedleft}
\vspace*{2cm}

\section{Introduction}
\label{sec:hmt-introduction}



The existence of adversarial perturbations implies that the decision boundaries of deep classifiers lie very close to any input sample, and in the previous chapter we demonstrated that the local geometry of the decision boundaries can be used to design a novel sparse attack. This unintuitive behaviour contradicts the common belief that classifiers should be invariant to non-discriminative information of the data. However, using our sparse perturbations we also revealed interesting correlations between the adversarial perturbations and the semantic features of the images. This suggest that adversarial examples might actually be something more than just superficial.
\blfootnote{Part of this chapter has been published in}
\blfootnote{``Hold me tight! Influence of discriminative features on deep network boundaries''. In \textit{Neural Information Processing Systems (NeurIPS)}, 2020~\cite{OrtizJimenez2020HoldMeTight}.}
\blfootnote{``Redundant features can hurt robustness to distribution shift''. In \textit{Uncertainty \& Robustness in Deep Learning Workshop (ICML)}, 2020~\cite{OrtizJimenezRedundantFeatures}.}
\blfootnote{``Improving filling level classification with adversarial training''. In \textit{IEEE International Conference on Image Processing (ICIP)}, 2021~\cite{Modas2021Improving}.}

In fact, recent works have established that such perturbations are indeed not irrelevant signals, but rather discriminative features of the training set~\cite{Jetley2018WithFriends,Ilyas2019AreNotBugs}. This has led to the conjecture that, in most datasets, there exist both robust and non-robust features that neural networks exploit to construct their decision boundaries. Besides, it has also been argued that the excessive invariance in the decision boundaries introduced by adversarial training can be harmful for standard accuracy, since this invariance causes the classifier to rely on overly-robust features~\cite{TramerFundamentalTradeoffs}. But, what exactly are these features, and how do networks construct these boundaries? This is still unclear. In this chapter, we propose a novel geometric framework that connects the discriminative features of the training dataset to the norm of the adversarial perturbations. We shed light on these phenomena by describing (i) the strong inductive bias of the networks towards invariance to non-discriminative features, and (ii) the sensitivity of training to small perturbations.

\begin{figure}[t]
\centering
\includegraphics[width=\textwidth]{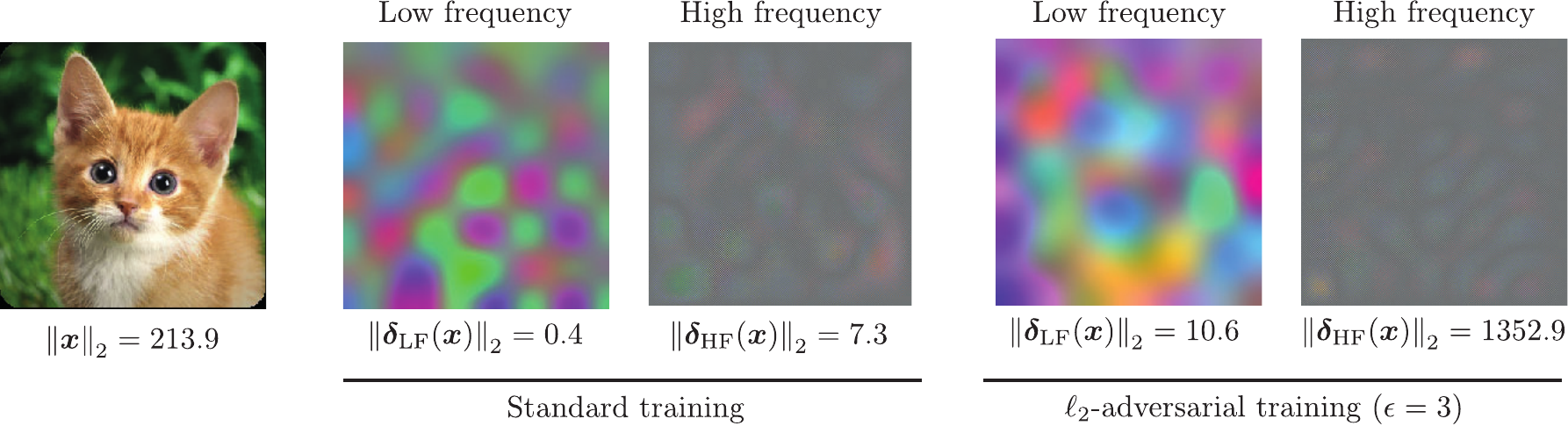}
\caption{Minimal adversarial perturbations constrained in different DCT frequency bands ($8\times 8$ subspaces taken from the top left and bottom right of the $224\times224$ DCT matrix) for a ResNet-50 trained (\textbf{left}), and adversarially trained (\textbf{right}) on ImageNet.}
\label{fig:hmt-introduction}
\end{figure}

Explaining the mechanisms that construct the decision boundaries is key to understand the dynamics of adversarial training~\cite{Madry2018TowardsDeepLearning}, which only differs from standard training in that it slightly perturbs the training samples during optimization. However, these small perturbations can utterly change the geometric properties of these classifiers~\cite{Moosavi2019CURE}. An example of such change can be seen in \cref{fig:hmt-introduction}, which shows the minimal perturbations -- constrained to lie on a low and a high frequency subspace -- required to flip the decision of a network. The norm of the perturbations measures the distance (margin) to the decision boundary in these subspaces. Clearly, reaching the boundary using high frequency perturbations requires much more energy (larger $\ell_2$ norm) than using low frequency ones~\cite{Sharma2019EffectivenessLowFrequency,Yin2019FourierPerspective,Maiya2021Frequency}. But, surprisingly enough, when the network is adversarially trained~\cite{Madry2018TowardsDeepLearning}, the largest increase in margin happens in the high frequency subspace. Note that, on the standard network, this distance is already much greater than the size of the training perturbations. Based on this observation, we pose the following questions:
\begin{enumerate}[leftmargin=*]
\item \emph{How is the margin in different directions related to the features in the training data?}
\item \emph{How can very small perturbations significantly change the geometry of deep networks?}
\end{enumerate}

In this chapter, we propose a novel approach to answer these questions, and provide a new perspective on the relationship between the distance of a set of samples to the boundary, and the discriminative features used by a network. In particular, we develop a new methodology to construct a local summary of the decision boundary from margin observations along a sequence of orthogonal directions. This framework permits to carefully tweak the properties of the training samples and measure the induced changes on the boundaries of deep classifiers trained on synthetic and large-scale vision datasets. Via a series of carefully designed experiments, we rigorously confirm the ``common belief'' that CNNs tend to behave as ideal classifiers and are approximately invariant to non-discriminative features of a dataset. Furthermore, we show that the construction of the decision boundary is very sensitive to the position of the training samples, such that very small perturbations in certain directions can utterly change the decision boundaries in these directions. In fact, we show that adversarial training exploits this training sensitivity and invariance bias to build robust classifiers. Finally, we demonstrate that the invariance properties of robust classifiers can be very beneficial for downstream applications where the available training data are scarce.

The reset of the chapter is organized as follows: In \cref{sec:hmt-decision-boundary-and-discriminative-featuers} we define our framework and demonstrate on synthetic data how it can be used to associate small margin directions with discriminative features. Then, in \cref{sec:hmt-discriminative-features-of-real} we deploy our framework on real datasets, and demonstrate the strong inductive bias of deep networks towards invariance to non-discriminative features. In \cref{sec:hmt-sensitivity-to-position} we use our framework to explain how adversarial training exploits the sensitivity of the network to the position of the training samples, in order to build robust models. Finally, in \cref{sec:hmt-icip} we demonstrate that the invariances of robust classifiers improve the performance on the off-the-shelf task of estimating the filling level within containers, where the available training data are scarce.

\section{Decision boundary and discriminative features}
\label{sec:hmt-decision-boundary-and-discriminative-featuers}

\subsection{Proposed framework}
\label{sec:hmt-subspace-adversarial-perturbations}

Recall from \cref{sec:introduction-adversarial-robustness} that we denote as $f:\R^D\to \R^K$ the final layer of a neural network (i.e., logits), such that, for any input $\bm{x}\in\R^D$, $F(\bm{x})=\text{argmax}_{k} f_k(\bm{x})$ represents the decision function of that network, where $f_k(\bm{x})$ denotes the $k$th component of $f(\bm{x})$ that corresponds to the $k$th class. The decision boundary between classes $k$ and $\ell$ is the set $\mathcal{B}_{k,\ell}(f)=\{\bm{x}\in\R^D: f_k(\bm{x})-f_\ell(\bm{x})=0\}$ (in general, we will omit the dependency with $k,\ell$ for simplicity). Unless stated otherwise, we assume that all networks are trained using a cross-entropy loss function and some variant of (stochastic) gradient descent. We also assume that training has been conducted for many epochs, and that it has approximately converged to a local minimum of the loss, achieving $100\%$ accuracy on the training data~\cite{zhangUnderstandingDeepLearning2016}. In general, all our experimental details are listed in \cref{ch:appendix-ch4}.

In this work, we study the role that the training set $\mathcal{T}=\{(\bm{x}^{(i)}, y^{(i)})\}_{i=0}^{N-1}$ has on the boundary $\mathcal{B}(f)$. Specifically, we propose to use adversarial proxies to measure the distribution of distances to the decision boundary along a sequence of well defined subspaces. The main quantities of interest are:
\begin{definition}[Subspace-constrained minimal adversarial perturbations]
    Based on the definition of \cref{eq:minimal-adv-pert}, for a decision function $F$, a sample $\bm{x}\in\R^D$, and a sub-region of the input space $\mathcal{S}\subseteq\R^D$, we define the ($\ell_2$) minimal adversarial perturbation of $\bm{x}$ in the subspace $\mathcal{S}$ as
    \begin{equation*}
        \begin{split}
        \bm{\delta}_\mathcal{S}(\bm{x}) &= \argmin{\bm\delta} \|\bm\delta\|_2 \\
        & \text{s.t. } F(\bm x) \neq F(\bm x + \bm\delta) \\
        & \bm\delta\in\mathcal{S}.
    \end{split}
    \label{eq:hmt-subspace-minimal-adv-pert}
    \end{equation*}
    In general, we will use $\bm{\delta}(\bm{x})$ to refer to $\bm{\delta}_{\R^D}(\bm{x})$.
\end{definition}

\begin{definition}[Margin]
    The magnitude $\|\bm{\delta}_\mathcal{S}(\bm{x})\|_2$ is the margin of $\bm{x}$ in $\mathcal{S}$.
\end{definition}

Our main objective is to obtain a local summary of $\mathcal{B}(f)$ around a set of samples, by measuring their margin in a sequence of distinct subspaces $\{\mathcal{S}_j\}_{j=0}^{R-1}$. In practice, we use a subspace-constrained version of DeepFool~\cite{Moosavi2016DeepFool}\footnote{We do not enforce the $[0,1]^D$ box constraints on the adversarial images, as we are not interested in finding ``plausible'' adversarial perturbations, but in measuring the distance to $\mathcal{B}(f)$.} to approximate the margins in each $\mathcal{S}_j$.

DeepFool is regarded as one of the most efficient methods to identify minimal adversarial perturbations. Since we want to measure margin, norm-constrained attacks like PGD~\cite{Madry2018TowardsDeepLearning} are not suitable for our study. Besides, more complex attacks like C\&W~\cite{CarliniWagner2017}, or, even, using unconstrained gradient descent in the input space, are computationally more demanding and harder to tune than DeepFool. Since they in general find very similar adversarial perturbations as DeepFool, we decided to opt for DeepFool in our work.s

\subsection{Evidence on synthetic examples}
\label{subsec:hmt-synthetic-example-discriminative-features}

In general, the distance from a sample to the boundary of a neural network can greatly vary depending on the search direction~\cite{Fawzi2017GeometricPerspective}. This behaviour is typically translated into classifiers with small margins along some directions, and large margins along others. We now investigate if neural networks only construct boundaries along discriminative directions, and remain invariant in every other direction\footnote{This is indeed a desired property for any classification method, but note that for neural networks the existence of adversarial examples contests the idea of it being a reasonable assumption.}. 

To this end, we generate a balanced training set $\mathcal{T}_1(\epsilon, \sigma)$  by independently sampling $N$ points $\bm{x}^{(i)}=\bm{U}(\bm{x}_1^{(i)}\oplus\bm{x}_2^{(i)})$, with $\bm{x}_1^{(i)}=\epsilon y^{(i)}$ and $\bm{x}_2^{(i)}\sim\mathcal{N}(0, \sigma^2 \bm{I}_{D-1})$, where $\oplus$ is the concatenation operator, $\epsilon>0$ the feature size, and $D=100$. The labels $y^{(i)}$ are uniformly sampled from $\{-1, +1\}$. The multiplication by a random orthonormal matrix $\bm{U}\in \operatorname{SO}(D)$ is performed to avoid possible biases of the classifier towards the canonical basis. $\mathcal{T}_1$ is a linearly separable dataset with a single discriminative feature parallel to $\bm{u}_1$ (i.e., first row of $\bm{U}$), and all other dimensions filled with non-discriminative noise.

\begin{table}
\begin{center}
\small
\begin{tabular}{lcccc}
\toprule
& $\bm{u}_1$ & $\operatorname{span}\{\bm{u}_1\}^\perp$ & $\mathcal{S}_{\text{orth}}$ & $\mathcal{S}_{\text{rand}}$ \\
\midrule
5-perc. & $1.74$ & $4.85$ & $30.68$  & $17.21$ \\
Median & $2.50$ & $12.36$ & $102.0$  & $27.90$ \\
95-perc. & $3.22$ & $31.60$ & $229.5$  & $80.61$\\
\bottomrule
\end{tabular}
\end{center}
\caption{Margin statistics of an MLP trained on $\mathcal{T}_1(\epsilon=5, \sigma=1)$ along different directions ($N=10,000$, $M=1,000$, $S=3$).}
\label{tab:hmt-invariance_synthetic}
\end{table}

To evaluate our hypothesis, we train an overparameterized multi-layer perceptron (MLP) with 10 hidden layers of 500 neurons using SGD (test accuracy: $100\%$). Table~\ref{tab:hmt-invariance_synthetic} shows the margin statistics on the linearly separable direction $\bm{u}_1$; its orthogonal complement $\operatorname{span}\{\bm{u}_1\}^\perp$; a fixed random subspace of dimension S, $\mathcal{S}_\text{rand}\subset\R^D$; and a fixed random subspace of the same dimensionality, but orthogonal to $\bm{u}_1$, $\mathcal{S}_{\text{orth}}\subset\operatorname{span}\{\bm{u}_1\}^\perp$. From these values we can see that along the direction where the discriminative feature lies, the margin is much smaller than in any other direction. Therefore, we can see that the classification function of this network is only creating a boundary in $\bm{u}_1$ with median margin $\epsilon/2$, and that it is approximately invariant in $\operatorname{span}\{\bm{u}_1\}^\perp$.

Comparing the margin values for $\mathcal{S}_{\text{orth}}$ and $\mathcal{S}_{\text{rand}}$ we see that, if the observation basis is not aligned with the features exploited by the network, the margin measurements might not be able to separate the small and large margin directions. Indeed, since $\mathcal{S}_{\text{orth}}$ is orthogonal to the only discriminative direction $\bm{u}_1$, we see that the margin values reported in this region are much higher than those reported in $\mathcal{S}_{\text{rand}}$. The reason for this is that the margin required to flip the label of a classifier in a randomly selected subspace is of the order of $\sqrt{S/D}$ with high probability~\cite{Fawzi2016RobustnessRandomNoise}, and hence the non-trivial correlation of a random subspace with the discriminative features will always hide the differences between small and large margin directions.

Finally, the margin fluctuations and the fact that the classifier is not completely invariant to $\operatorname{span}\{\bm{u}_1\}^\perp$ might indicate that the network has built a complex boundary. However, in \cref{sec:appendix-ch4-theoretical-margin} we show that similar fluctuations and finite values in $\operatorname{span}\{\bm{u}_1\}^\perp$ are observed even if the model is linear by construction and separates the training data perfectly.


\section{Discriminative features of real datasets}
\label{sec:hmt-discriminative-features-of-real}

In contrast to the synthetic data, where the discriminative features are known by construction, the exact description of the features presented in \emph{real} datasets is usually not known. In order to identify these features and understand their connection to the local construction of the decision boundaries, we apply the proposed framework on standard computer vision datasets, and investigate if deep networks trained on real data also present high invariance along the non-discriminative directions of the dataset.

\begin{figure}[!ht]
\begin{center}
\includegraphics[width=0.40\textwidth]{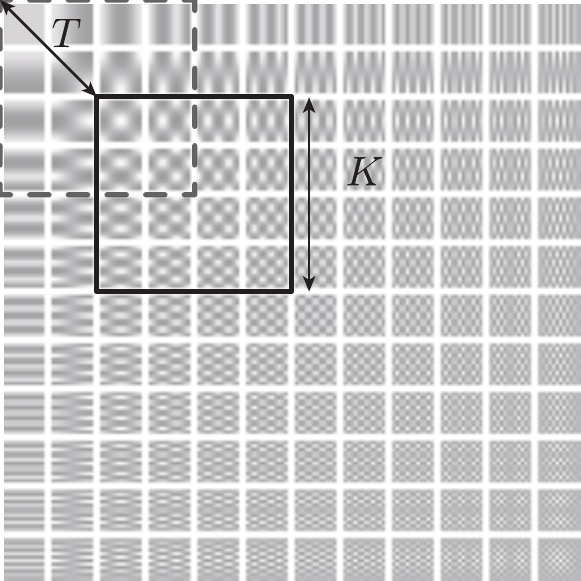}
\end{center}
\caption{Illustration of sampling from the diagonal of the DCT basis.}
\label{fig:hmt-dct12x12}
\end{figure}

In our study, we train multiple networks on MNIST~\cite{LeCun2010MNIST} and CIFAR-10~\cite{krizhevskyLearningMultipleLayers2009}, and for ImageNet~\cite{Deng2009ImageNet} we use several of the pretrained networks provided by PyTorch~\cite{Paszke2019PyTorch}\footnote{Experiments on more CNNs (with similar findings) are presented in Appendix~I of~\cite{OrtizJimenez2020HoldMeTight}.}. Let $W,H,C$ denote the width, height, and number of channels of the images in those datasets, respectively. In our experiments we use the 2-dimensional discrete cosine transform (2D-DCT)~\cite{Ahmed1974DCT} basis of size $H\times W$ to generate the observation subspaces. In particular, let $\bm{\mathcal{D}}\in\R^{H\times W\times H\times W}$ denote the 2D-DCT generating tensor, such that $\operatorname{vec}(\bm{\mathcal{D}}(i,j,:,:)\otimes \bm{I}_C)$ represents one basis element of the image space. We generate the subspaces by sampling (see \cref{fig:hmt-dct12x12}) $K\times K$ blocks from the diagonal of the DCT tensor using a sliding window with step-size $T$:
$\mathcal{S}_j=\operatorname{span}\{\operatorname{vec}\left(\bm{\mathcal{D}}\left(j\cdot T+k, j\cdot T+k, :,:\right)\otimes \bm{I}_C\right)\;k=0,\dots,K-1\}$.

The sliding window on the diagonal of the DCT gives a good trade-off between visualization abilities in simple one-dimensional plots, and a diverse sampling of the spatial spectrum of natural images, with a well-defined gradient flowing from low to high frequencies\footnote{A similar analysis including off-diagonal subspaces is presented in Appendix~I of~\cite{OrtizJimenez2020HoldMeTight}.}. The DCT has a long application tradition in image processing due to its good approximation of the decorrelation transform (KLT)~\cite{Gonzalez2017DigitalImageProcessing}. Furthermore, in previous studies on the robustness of deep networks to different frequencies, the DCT was also the basis of choice~\cite{Sharma2019EffectivenessLowFrequency} because it avoids dealing with complex subspaces. 

\begin{figure}[t]
\centering
\begin{subfigure}[b]{0.33\linewidth}
    \includegraphics[width=\linewidth]{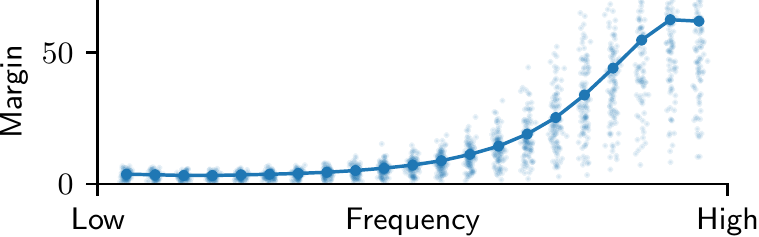}
    \caption{MNIST ($99.4\%$)}
    \label{subfig:hmt-invariance_mnist}
\end{subfigure}\hfill
\begin{subfigure}[b]{0.33\linewidth}
    \includegraphics[width=\linewidth]{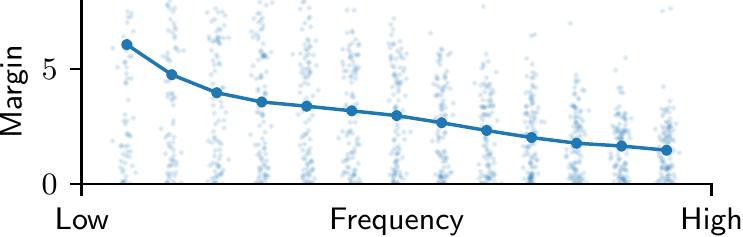}
    \caption{CIFAR-10 ($93.0\%$)}
    \label{subfig:hmt-invariance_cifar}
\end{subfigure}\hfill
\begin{subfigure}[b]{0.33\linewidth}
    \includegraphics[width=\linewidth]{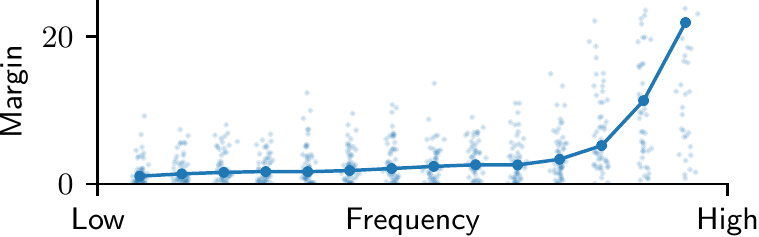}
    \caption{ImageNet ($76.2\%$)}
    \label{subfig:hmt-invariance_imagenet}
\end{subfigure}\hfill
\par\bigskip
\begin{subfigure}[b]{0.33\linewidth}
    \includegraphics[width=\linewidth]{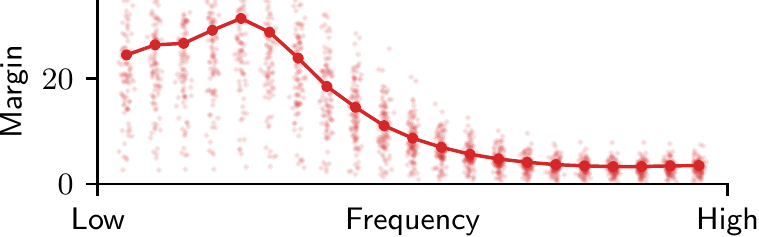}
    \caption{MNIST flipped ($99.3\%$)}
    \label{subfig:hmt-invariance_mnist_flip}
\end{subfigure}\hfill
\begin{subfigure}[b]{0.33\linewidth}
    \includegraphics[width=\linewidth]{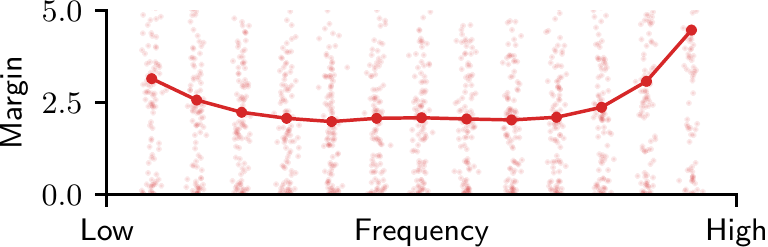}
    \caption{CIFAR-10 flipped ($91.2\%$)}
    \label{subfig:hmt-invariance_cifar_flip}
\end{subfigure}\hfill
\begin{subfigure}[b]{0.33\linewidth}
    \includegraphics[width=\linewidth]{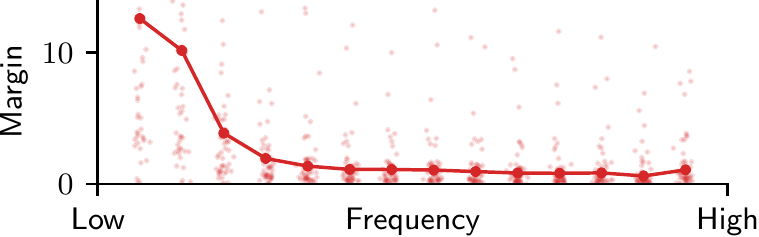}
    \caption{ImageNet flipped ($68.1\%$)}
    \label{subfig:hmt-invariance_imagenet_flip}
\end{subfigure}\hfill
\caption{Margin distribution of test samples in subspaces taken from the diagonal of the DCT (low to high frequencies). Thick lines indicate the median values of the margin, and shaded points represent its distribution. Parentheses contain the test accuracy. \textbf{Top}: (a) MNIST (LeNet), (b) CIFAR-10 (DenseNet-121) and (c) ImageNet (ResNet-50) \textbf{Bottom}: (d) MNIST, (e) CIFAR-10 and (f) ImageNet trained on frequency-``flipped'' versions of the standard datasets.}
\label{fig:hmt-invariance_real}
\end{figure}

The margin distribution of the evaluated test samples is presented in the top of \cref{fig:hmt-invariance_real}. For MNIST and ImageNet, the networks present a strong invariance along high frequency directions and small margin along low frequency ones. We will later show that this is related to the fact that these networks mainly exploit discriminative features in the low frequencies of these datasets. Notice, however, that for CIFAR-10 the margin values are more uniformly distributed; an indication that the network exploits discriminative features across the full spectrum as opposed to the human vision system~\cite{Campbell1968ApplicationFourier}.

\begin{figure}[!hb]
\vspace{0.5em}
\begin{center}
\begin{subfigure}[b]{0.38\textwidth}
\includegraphics[width=\textwidth]{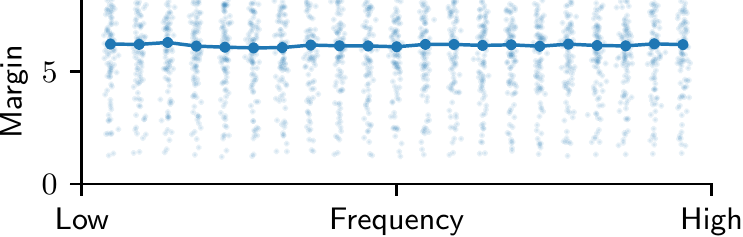}
\caption{MNIST (Test: $99.35\%$)}
\end{subfigure}
\begin{subfigure}[b]{0.38\textwidth}
\includegraphics[width=\textwidth]{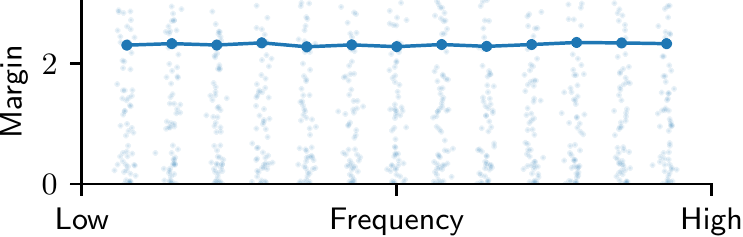}
\caption{CIFAR-10 (Test: $93.03\%$)}
\end{subfigure}\hfill
\begin{subfigure}[b]{0.38\textwidth}
\includegraphics[width=\textwidth]{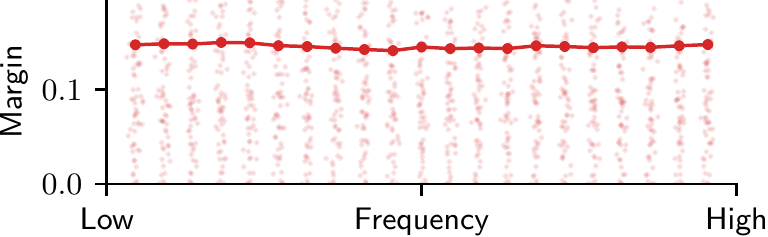}
\caption{MNIST flipped (Test: $99.34\%$)}
\end{subfigure}
\begin{subfigure}[b]{0.38\textwidth}
\includegraphics[width=\textwidth]{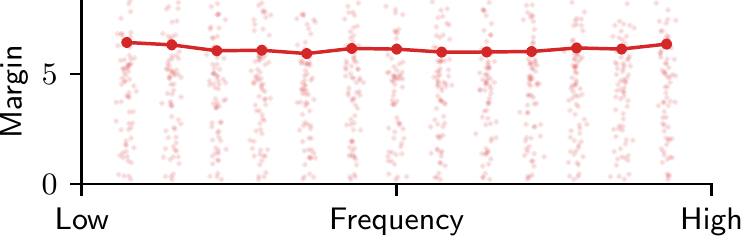}
\caption{CIFAR-10 flipped (Test: $91.19\%$)}
\end{subfigure}\hfill
\caption{Margin distribution of test samples in subspaces taken from a random orthonormal matrix arranged as a tensor of the same dimensionality as the DCT tensor. Subspaces are taken from the diagonal with the same parameters as in DCT. \textbf{Top}: (a) MNIST (LeNet), (b) CIFAR-10 (DenseNet-121) \textbf{Bottom}: (d) MNIST (LeNet) and (e) CIFAR-10 (DenseNet-121) trained on frequency ``flipped'' versions of the datasets.}
\label{fig:hmt-invariance_real_random}
\end{center}
\end{figure}

We observe in practice that the DCT basis is also quite aligned to the features of these datasets, and hence it can give precise information about the discriminative features exploited by the networks. A more aligned basis with respect to the discriminative features would probably show a sharper transition between low and high margins. However, finding such network-agnostic bases is a challenging task without knowing the features \emph{a priori}. The DCT is not perfectly feature-aligned, but it seems to be a good choice for comparing different architectures, especially if we compare its results to those obtained using a random orthonormal basis where differences in margin cannot be identified. Indeed, as shown in \cref{fig:hmt-invariance_real_random} it is clear that a random basis is not valid for this task as the margin in any random subspace is of the same order with high probability~\cite{Fawzi2016RobustnessRandomNoise}.

\subsection{Boundary adaptation to data representation}
\label{subsec:hmt-boundary-adaptation}

Towards verifying that the proposed framework can capture the relation between the data features and the local construction of the decision boundaries, we must first ensure that the direction of the observed invariance (large margin) is related to the features presented in the dataset, rather than being just an effect of the network itself.

\begin{figure}[!hb]
\centering
\includegraphics[width=0.65\textwidth]{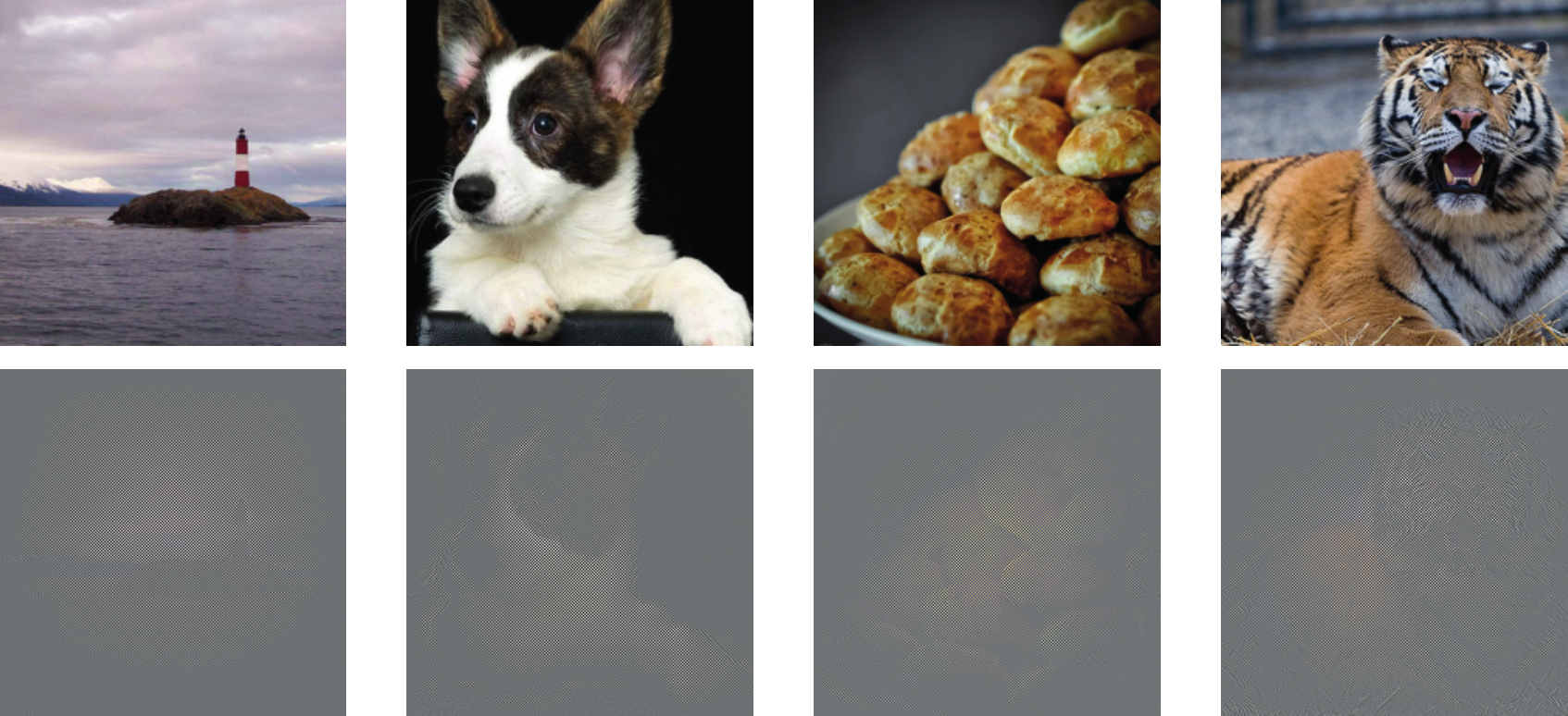}
\caption{``Flipped'' image examples from ImageNet. \textbf{Top}: original. \textbf{Bottom}: ``flipped''.}
\label{fig:hmt-flipped_examples}
\end{figure}

Based on our observation that the margin tends to be small in low frequency directions and large in high frequency ones, we carefully tweak the representation of the data such that the low frequencies are swapped with the high frequencies. In practice, if $\mathfrak{D}$ denotes the forward DCT transform operator, the new image representation $\bm{x}'$ is expressed as $\bm{x}'=\mathfrak{D}^{-1}(\operatorname{flip}(\mathfrak{D}(\bm{x})))$, where $\operatorname{flip}$ corresponds to one horizontal and one vertical flip of the DCT transformed image. Some ``flipped'' examples are shown in \cref{fig:hmt-flipped_examples}.

Thus, if the direction of the resulting margin is strongly related to the data features, the constructed decision boundaries should also adapt to this new data representation, and the margin along the invariant directions (high frequencies) should swap with the margin of the discriminative ones (low frequencies). Informally speaking, the margin distribution should ``flip''.

We apply our framework on multiple networks trained on the ``flipped'' datasets, and the margin distribution is depicted at the bottom of \cref{fig:hmt-invariance_real}. For both MNIST and ImageNet, the directions of the decision boundaries indeed \emph{follow} the new data representation -- although they are not an exact mirroring of the original representation. This indicates that the margin strongly depends on the data distribution, and it is not solely an effect of the network architecture. Note again that for CIFAR-10 the effect is not as obvious, due to the quite uniform distribution of the margin.

\subsection{Invariance and elasticity of decision boundary}
\label{subsec:hmt-invariance-and-elasticity}

The second property we need to verify is that the small margins reported in \cref{fig:hmt-invariance_real} do indeed correspond to directions containing discriminative features in the training set. For doing so, we use the insights of \cref{subfig:hmt-invariance_cifar} on CIFAR-10 -- where, opposed to the other datasets, we assume that there are exploited discriminative features in the whole spectrum -- and show that, by explicitly modifying its features, we can induce a high margin response in the measured curve in a set of selected directions.

\begin{figure}[ht]
\begin{center}
\includegraphics[width=0.45\textwidth]{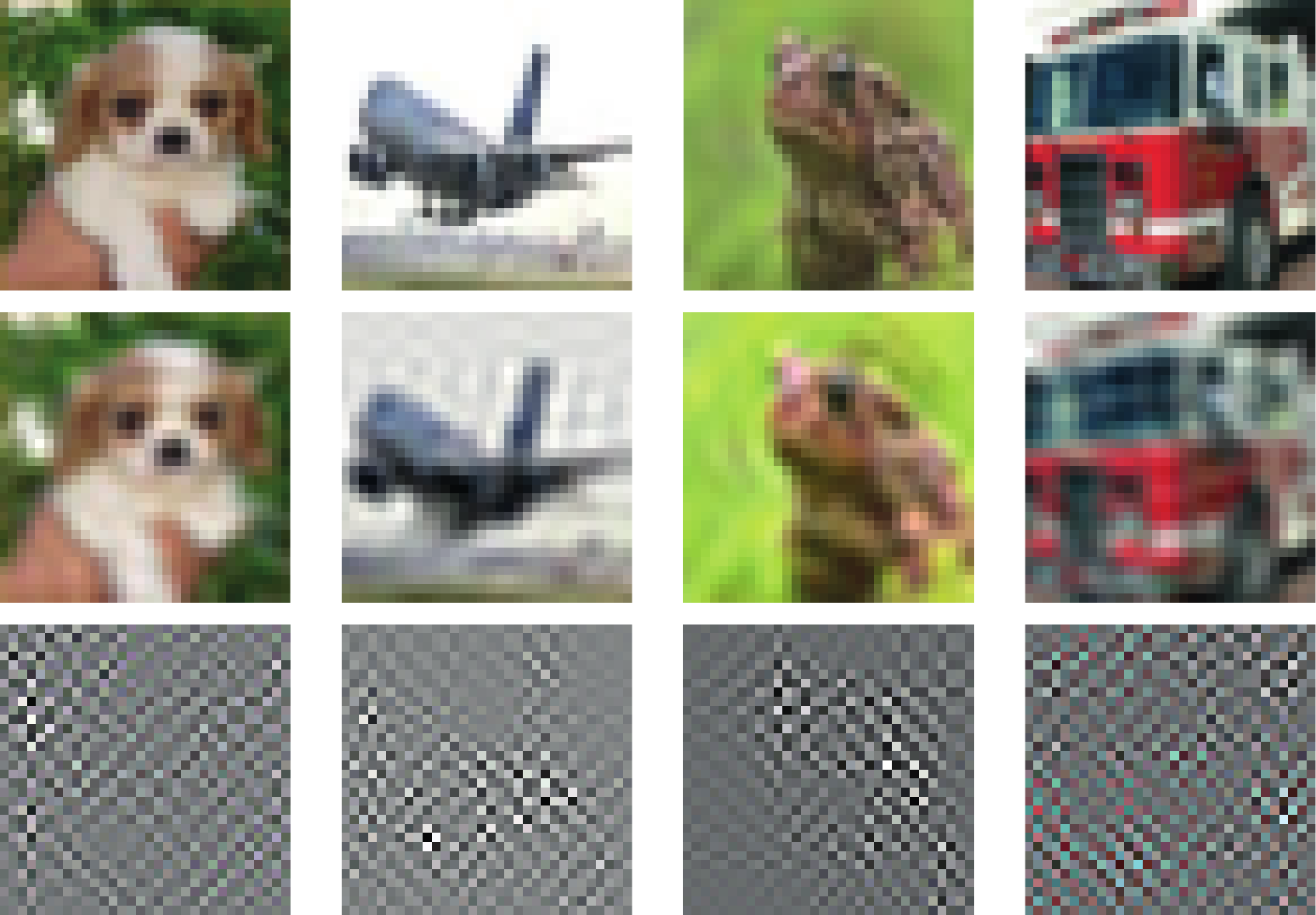}
\caption{Examples of filtered CIFAR-10 images. \textbf{Top} original images, \textbf{middle} low-pass and \textbf{bottom} high-pass versions.}
\label{fig:hmt-samples_filtered}
\end{center}
\end{figure}

In particular, we create a low-pass filtered version of CIFAR-10 ($\mathcal{T}_\text{LP}$), where we retain only the frequency components in a $16\times 16$ square at the top left of the diagonal of the DCT-transformed images. This way we ensure that no training image has any energy/information outside of this frequency subspace. Examples of $\mathcal{T}_\text{LP}$ images are shown in the second row of \cref{fig:hmt-samples_filtered}.

\begin{figure}[!hb]
\centering
\vspace{0.5em}
\includegraphics[width=0.65\textwidth]{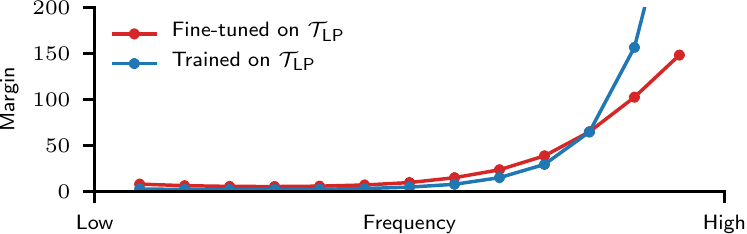}
\caption{Median margin of test samples from CIFAR-10 for a DenseNet-121 (i) trained on CIFAR-10 and fine-tuned on $\mathcal{T}_{\text{LP}}$ ($90.79\%$), and (ii) trained on $\mathcal{T}_{\text{LP}}$ from scratch ($89.67\%$). Parentheses contain the test accuracy.}
\label{fig:hmt-invariance_LP_cifar}
\end{figure}

The median margin\footnote{We do not plot the full distribution to avoid clutter. The $5$-percentile of the margin in the last subspace is $5.05$.} of CIFAR-10 test samples for a network trained on $\mathcal{T}_{\text{LP}}$ is illustrated in \cref{fig:hmt-invariance_LP_cifar}. Indeed, by eliminating the high frequency content, we have forced the network to become invariant along these directions. This clearly demonstrates that there existed discriminative features in the high frequency spectrum of CIFAR-10, and that by removing these from all the samples, the inductive bias of training pushes the network to become invariant to them.

Moreover, this effect can \emph{also} be triggered during training. To show this, we start with the CIFAR-10 trained network studied in \cref{subfig:hmt-invariance_cifar} and continue training it for a few more epochs with a small learning rate using only $\mathcal{T}_{\text{LP}}$. \Cref{fig:hmt-invariance_LP_cifar} shows the new median margins of this network. The fine-tuned network is again invariant to the high frequencies.

Finally, note that by training with only low frequency data, the test accuracy of the network on the original CIFAR-10 only drops around $3\%$\footnote{Similar effect was shown on ImageNet~\cite{Yin2019FourierPerspective}, although the network was only tested on filtered data. For MNIST, training on low-pass data yields no accuracy drop, since MNIST trained networks exploit mostly low frequencies and already have large margins in the high frequencies. There might exist discriminative information in the high frequencies, but the network does not exploit it.}. Because $\mathcal{T}_\text{LP}$ has no high frequency energy, a network trained on it will uniformly extend its boundaries in this part of the spectrum and no high frequency perturbation will be able to flip the network's output. In contrast, testing $\mathcal{T}_\text{LP}$ data on a CIFAR-10 trained network only achieves $27.45\%$ test accuracy. This is because networks trained on CIFAR-10 do have boundaries in the high frequencies, and hence showing them original samples perturbed in this frequency range (i.e., $\mathcal{T}_\text{LP}$) can greatly change their decisions.

\subsection{Connections to catastrophic forgetting}
\label{subsec:hmt-catastrophic-forgetting}

The elasticity to the modification of features during training gives a new perspective to the theory of catastrophic forgetting~\cite{McCloskey1989Catastrophic}, as it confirms that the decision boundaries of a neural network can only exist for as long as the classifier is trained with the samples (features) that hold them together. In particular, we demonstrate this by adding and removing points from a dataset such that its discriminative features are modified during training, and hence artificially causing an elastic response on the network.

\begin{figure}[!t]
\captionsetup[subfigure]{font=small, justification=centering}
\centering
\vspace{0.5em}
\begin{subfigure}[b]{0.5\linewidth}
    \includegraphics[width=\linewidth]{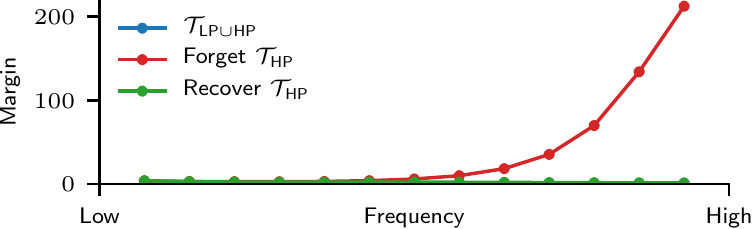}
    \caption*{Zoom-out for observing the general invariance.}
\end{subfigure}\hfill
\begin{subfigure}[b]{0.5\linewidth}
    \includegraphics[width=\linewidth]{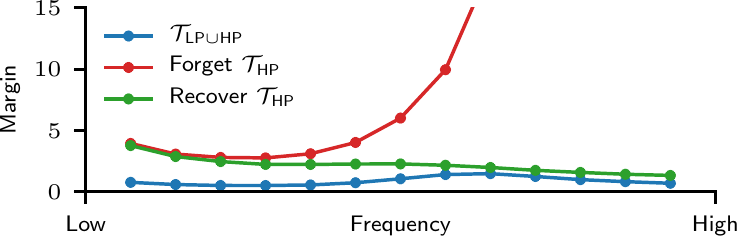}
    \caption*{Zoom-in for a more detailed observation.}
\end{subfigure}\hfill
\caption{Median margin of $\mathcal{T}_{\text{LP}}$ test samples for a DenseNet-121. \textbf{Blue:} trained on $\mathcal{T}_{\text{LP}\cup\text{HP}}$; \textbf{Red:} after forgetting $\mathcal{T}_{\text{HP}}$; \textbf{Green:} after recovering $\mathcal{T}_{\text{HP}}$.}
\label{fig:hmt-forgetting_recovering}
\end{figure}

To this end, we train a DenseNet-121 on a new dataset $\mathcal{T}_{\text{LP}\cup\text{HP}}=\mathcal{T}_{\text{LP}}\cup\mathcal{T}_\text{HP}$ formed by the union of two filtered variants of CIFAR-10: $\mathcal{T}_{\text{LP}}$ is constructed by retaining only the frequency components in a $16\times 16$ square at the top-left of of the DCT-transformed CIFAR-10 images (low-pass), while for $\mathcal{T}_{\text{HP}}$ only the frequency components in a $16\times 16$ square at the bottom-right of the DCT (high-pass; cf. third row of \cref{fig:hmt-samples_filtered}). This classifier has a test accuracy of $86.59\%$ and $57.29\%$ on $\mathcal{T}_{\text{LP}}$ and $\mathcal{T}_{\text{HP}}$, respectively. The median margin of $1,000$ $\mathcal{T}_{\text{LP}}$ test samples along different frequencies for this classifier is shown in blue in \cref{fig:hmt-forgetting_recovering}. As expected, the classifier has picked features across the whole spectrum with the low frequency ones probably belonging to boundaries separating samples in $\mathcal{T}_{\text{LP}}$, and the high frequency ones separating samples from $\mathcal{T}_{\text{LP}}$ and $\mathcal{T}_{\text{HP}}$\footnote{$\mathcal{T}_{\text{LP}}$ and $\mathcal{T}_{\text{HP}}$ have discriminative features only in the low and high frequency part, respectively.}.

After this, we continue training the network with a linearly decaying learning rate (max. $\alpha=0.05$) for another 30 epochs, but using only $\mathcal{T}_{\text{LP}}$, achieving a final test accuracy of $87.81\%$ and $10.01\%$ on $\mathcal{T}_{\text{LP}}$ and $\mathcal{T}_{\text{HP}}$, respectively. Again, \cref{fig:hmt-forgetting_recovering} shows in red the median margin along different frequencies on test samples from $\mathcal{T}_{\text{LP}}$. The new median margin is clearly invariant on the high frequencies -- where $\mathcal{T}_{\text{LP}}$ has no discriminative features  -- and the classifier has completely \emph{erased} the boundaries that it previously had in these regions, regardless of the fact that those boundaries did not harm the classification accuracy on $\mathcal{T}_{\text{LP}}$. 

Finally, we investigate if the network is able to recover the forgotten decision boundaries that were used to classify $\mathcal{T}_\text{HP}$. We continue training the network (``forgotten'' $\mathcal{T}_\text{HP}$) for another 30 epochs, but this time by using the whole $\mathcal{T}_{\text{LP}\cup\text{HP}}$. Now this classifier achieves a final test accuracy of $86.1\%$ and $59.11\%$ on $\mathcal{T}_{\text{LP}}$ and $\mathcal{T}_{\text{HP}}$ respectively, which are very close to the corresponding accuracies of the initial network trained from scratch on $\mathcal{T}_{\text{LP}\cup\text{HP}}$ (recall: $86.59\%$ and $57.29\%$). The new median margin for this classifier is shown in green in \cref{fig:hmt-forgetting_recovering}. As we can see by comparing the green to the blue curve, the decision boundaries along the high-frequency directions can be recovered quite successfully. 

\subsection{Discussion}
\label{subsec:hmt-discussion-discriminative-featuers}

The main claim in \cref{sec:hmt-decision-boundary-and-discriminative-featuers} and \cref{sec:hmt-discriminative-features-of-real} is that deep neural networks \emph{only} create decision boundaries in regions where they identify discriminative features in the training data. As a result, there is a big relative difference in the large margin along the invariant directions, and the smaller margin in the discriminative directions.

The main difficulty for establishing causation in this idea is the fact that the discriminative features of real datasets are not known. Hence, determining their role on the geometry of a trained neural network can only be done by artificially manipulating the data. In particular, there are two main confounding factors that might alternatively explain our results: the network architecture or the training algorithm. However, the experiments in \cref{sec:hmt-discriminative-features-of-real} are precisely designed to rule out their influence in this phenomenon.

Specifically, in the flipping experiments, flipping the data -- \emph{ceteris paribus} -- also flips the margin distribution, thus demonstrating that the margins are necessarily caused by the information present in the data. The other interventions we do on the samples (e.g., low-pass experiments) confirm that, in the absence of information in a certain discriminative subspace, the network becomes invariant along this discriminative subspace. Therefore, we believe that there is indeed a causal connection between the features of the data and the measured margins in these neural networks. In fact, parallel theoretical studies have demonstrated that the ability of neural networks to distinguish between discriminative and non-discriminative noise subspaces in a dataset is one of the main advantages of deep learning over kernel methods~\cite{Ghorbani2020NeuralNetworksOutperform}.

\section{Sensitivity to position of training samples}
\label{sec:hmt-sensitivity-to-position}

Our novel framework to relate boundary geometry and data features can help track the dynamics of learning. In this section, we use it to explain how training with a slightly perturbed version of the training samples can greatly alter the network geometry. We further analyze how adversarial training can be so successful in removing features with small margin to increase the network's robustness.

\subsection{Evidence on synthetic examples}
\label{subsec:hmt-synthetic-example-sensitivity}

\begin{figure}[ht]
\vspace{0.5em}
\centering
\includegraphics[width=0.6\textwidth]{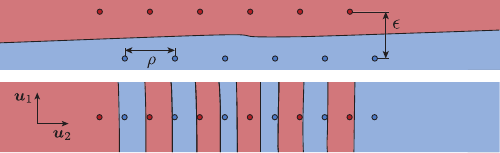}
\caption{Cross-section of an MLP trained on $\mathcal{T}_2(\rho=20, \epsilon, \sigma=1, K=3)$ with $\epsilon=1$ (\textbf{top}) and $\epsilon=0$ (\textbf{bottom}). Axes scaled differently.}
\label{fig:hmt-cross_sections}
\end{figure}

We train multiple times an MLP with the same setup as \cref{subsec:hmt-synthetic-example-discriminative-features}, but this time using slightly perturbed versions of the same synthetic dataset. In particular, we use a family of training sets $\mathcal{T}_2(\rho, \epsilon, \sigma, K)$ consisting in $N=10,000$ independent $D=100$-dimensional samples $\bm{x}^{(i)}=\bm{U}(\bm{x}_1^{(i)}\oplus\bm{x}_2^{(i)}\oplus\bm{x}_3^{(i)})$ such that $\bm{x}_1^{(i)}=\epsilon y^{(i)}$; $\bm{x}_2^{(i)}= \rho \cdot k$ when $y^{(i)}=+1$, and $\bm{x}_2^{(i)}=\rho \cdot \left(k + \frac{1}{2}\right)$ when $y^{(i)}=-1$, where $k$ is sampled from a discrete uniform distribution with values $\{-K,\dots,K-1\}$; and $\bm{x}_3^{(i)}\sim\mathcal{N}(0,\sigma^2\bm{I}_{D-2})$ (see \cref{fig:hmt-cross_sections}). Here, $\epsilon,\rho\geq0$ denote the feature sizes. Again, the multiplication by a random orthonormal matrix $\bm{U}\in\operatorname{SO}(D)$ avoids any possible bias of the network towards the canonical basis. Note that for $\epsilon>0$ this training set will always be linearly separable using $\bm{u}_1$, but without necessarily yielding a maximum margin classifier. Especially when $\rho\gg\epsilon$.

\begin{figure}[ht]
\centering
\includegraphics[width=0.6\textwidth]{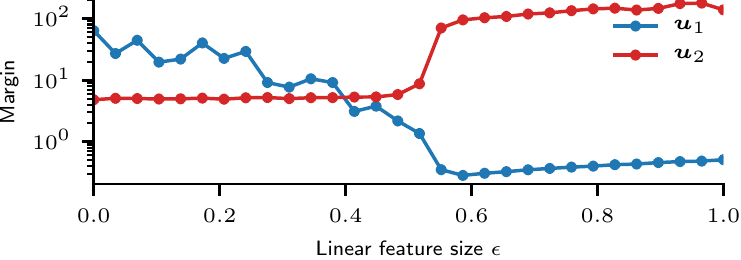}
\caption{Median margin values along $\bm{u}_1$ and $\bm{u}_2$ for MLPs (test: $100\%$ always) trained on $\mathcal{T}_2$ for different values of $\epsilon$ and $\rho=20$.}
\label{fig:hmt-unstable_boundary}
\end{figure}

\Cref{fig:hmt-unstable_boundary} shows the median margin of $M=1,000$ observation samples for an MLP trained on different versions of $\mathcal{T}_2(\rho, \epsilon, \sigma, K)$ with a fixed $\rho=20$, but a varying small $\epsilon$. Based on this plot, it is clear that for very small $\epsilon$ the neural network predominantly uses the information contained in $\bm{u}_2$ to separate the different classes. Indeed, for $\epsilon<0.2$, the network is almost invariant in $\bm{u}_1$, and it uses a non-linear alternating pattern in $\bm{u}_2$ to separate the data\footnote{This particular pattern can in principle classify any dataset with $\rho=20$, no matter the value of $\epsilon$.} (see bottom row of \cref{fig:hmt-cross_sections}). On the contrary, at $\epsilon>0.5$ we notice a sharp transition in which we see that the neural network suddenly changes its behaviour and starts to linearly separate the different points using only $\bm{u}_1$ (see top row of \cref{fig:hmt-cross_sections}).

We conjecture that this phenomenon is rooted on the strong inductive bias of the learning algorithm to build connected decision regions whenever geometrically and topologically possible, as empirically validated in~\cite{Fawzi2018EmpiricalStudyTopology}. Here, we go one step further and hypothesize that the inductive bias of the learning algorithm has a tendency to build classifiers in which every pair of training samples with the same label belongs to the same decision region. If possible, connected by a straight path.

We see \cref{fig:hmt-unstable_boundary} as a validation of this hypothesis. For small values of $\epsilon$, it is hard for the algorithm to find solutions that connect points from the same class with a straight path, as this is very aligned with $\bm{u}_2$. However, there is a precise moment (i.e., $\epsilon=0.5$) in which finding such a solution becomes much easier, and then the algorithm suddenly starts to converge to the linearly separating solution.

At this stage it is important to highlight that repeating the same experiment with a different random seed, or for a fixed initialization, does not affect the results. Furthermore, overfitting cannot be the cause of these results, as the MLP always achieves $100\%$ test accuracy for $\epsilon < 0.5$, as well. Finally, adding a small weight decay (i.e., $10^{-3}$) does not help the network find the linearly separable solution for $\epsilon < 0.5$; it rather hinders its convergence (i.e., final train accuracy is $50\%$).

It remains unclear whether this inductive bias is the only mechanism that can trigger a sharp transition in the type of learned decision boundaries, or if there are other types of biases that can cause the same effect. In any case, we believe that the significant difference in the type of function that the algorithm learns when trained with very similar training samples (see  \cref{fig:hmt-cross_sections}), is an unambiguous confirmation of the sensitivity of deep learning to the exact position of its training input.

Concurrent work~\cite{Shah2020Pitfalls} has also used a similarly constructed dataset to $\mathcal{T}_2(\rho, \epsilon, \sigma, K)$ to argue that the simplicity bias of a neural network when trained using standard procedures might be responsible for the selection of non-robust features in the dataset~\cite{Ilyas2019AreNotBugs}.

\subsection{Connections to adversarial training}
\label{subsec:hmt-connections-to-adversarial-training}

\begin{figure}[t]
\centering
\begin{subfigure}{0.33\textwidth}
\includegraphics[width=\textwidth]{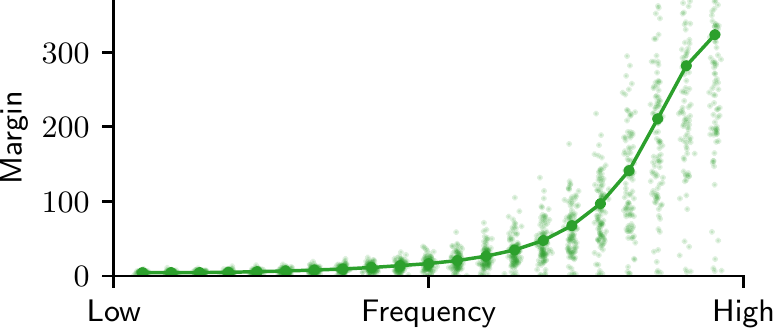}
\caption{MNIST ($\text{Test: }98\%$)}\label{subfig:hmt-invariance_robust_mnist}
\end{subfigure}\hfill
\begin{subfigure}{0.33\textwidth}
\includegraphics[width=\textwidth]{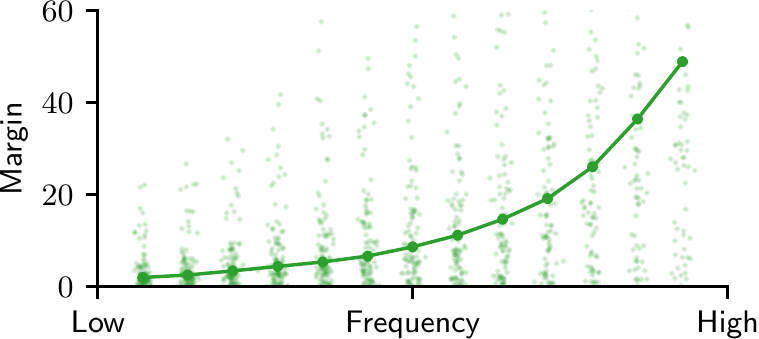}
\caption{CIFAR-10 ($\text{Test: }83\%$)}\label{subfig:hmt-invariance_robust_cifar}
\end{subfigure}\hfill
\begin{subfigure}{0.33\textwidth}
\includegraphics[width=\textwidth]{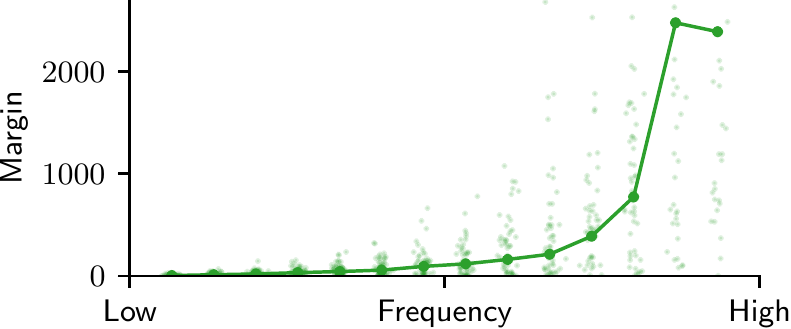}
\caption{ImageNet ($\text{Test: }76\%$)}\label{subfig:hmt-invariance_robust_imagenet}
\end{subfigure}\hfill
\caption{Margin distribution of test samples in subspaces taken from the diagonal of the DCT (low to high frequencies). Adversarially trained networks using $\ell_2$ PGD (a) LeNet ($\text{Adv: }76\%$), (b) DenseNet-121 ($\text{Adv: }55\%$) and (c) ResNet-50 ($\text{Adv: }35\%$).}
\label{fig:hmt-invariance_real_robust}
\end{figure}

Finally, we show that adversarial training exploits the type of phenomena described in \cref{subsec:hmt-synthetic-example-sensitivity} to reshape the boundaries of a neural network. In this regard, \cref{fig:hmt-invariance_real_robust} shows the margin distribution across the DCT spectrum of a few adversarially trained networks\footnote{The analogous effect for the frequency-``flipped'' datasets is detailed in Appendix~M of~\cite{OrtizJimenez2020HoldMeTight}}. As expected, the margins of the adversarially trained networks are significantly higher than those in \cref{fig:hmt-invariance_real}. 

Surprisingly, though, the largest increase can be noticed in the high frequencies for all datasets. Considering that adversarial training only differs from standard training in that it slightly moves the training samples, it is imperative that deep networks converge to very different solutions under such small modifications. The next experiments on CIFAR-10 shed light on the dynamics of this process.

\begin{figure}[ht]
\centering
\vspace{0.5em}
\includegraphics[width=0.7\textwidth]{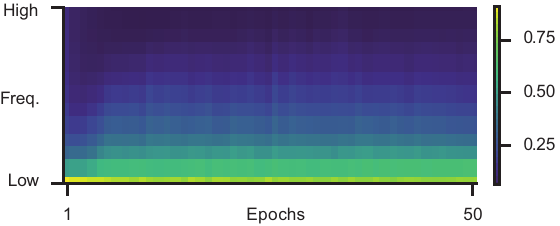}
\caption{Energy of adversarial perturbations on subspaces of the DCT during adversarial training of CIFAR-10 (DenseNet-121). Plot shows 95-percentile.}
\label{fig:hmt-spectral_cifar_densenet}
\end{figure}

\textbf{Very small adversarial perturbations can trigger large invariance}\quad
Slightly perturbing the training samples can remove features in an unpredictable manner. \Cref{fig:hmt-spectral_cifar_densenet} shows the spectral decomposition of the adversarial perturbations crafted during adversarial training of CIFAR-10. The energy of the perturbations during training is always concentrated in the low frequencies, and has hardly any high frequency content. However, the greatest effect on margin is seen on the high frequency directions (see \cref{fig:hmt-invariance_real_robust}). This is similar to what is seen in \cref{fig:hmt-cross_sections}, where slightly perturbing the training samples along $\bm{u}_2$ drastically affects the margin along $\bm{u}_2$. 

Overall, we see that adversarial training exploits the sensitivity of the network to small changes in the training samples to hide some discriminative features from the model. This is especially clear when we compare the CIFAR-10 values in \cref{subfig:hmt-invariance_robust_cifar} and \cref{subfig:hmt-invariance_cifar}, where it becomes evident that some previously used discriminative features in the high frequencies are completely overlooked by the adversarially trained network. In the following example, we show that, in practice, it is not actually necessary to change the position of all training points to induce a large invariance reaction.

\textbf{Invariance can be triggered by just a few samples}\quad
Modifying the position of just a \emph{minimal} number of training samples is enough to locally introduce excessive invariance on a classifier. To demonstrate this, we take a ResNet-18 (test: $90\%$) trained on CIFAR-10, and randomly select a set of $100$ training samples $\mathcal{P}\subset\mathcal{T}$. We fine-tune this classifier replacing those $100$ samples with $(\bm{x} + \bm{\delta}^\mathrm{o}(\bm{x}), y)$ in $\mathcal{P}$ (test: $90\%$), where $\bm{\delta}^\mathrm{o}$ and $\bm{\delta}^\mathrm{f}$ represent the adversarial perturbations for the original and fine-tuned network, respectively.

\begin{figure}[ht]
\centering
\vspace{0.5em}
\includegraphics[width=0.8\textwidth]{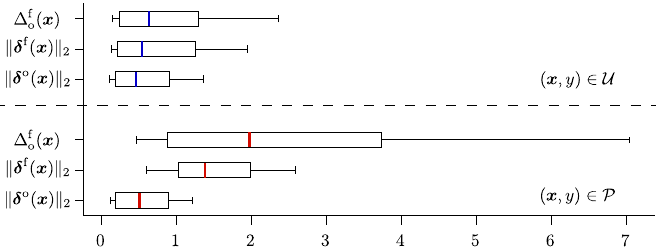}
\caption{Margin distribution in different directions of a ResNet-18 trained on CIFAR-10 and fine-tuned on 100 DeepFool examples.}
\label{fig:hmt-fine_tuning_cifar}
\end{figure}

\Cref{fig:hmt-fine_tuning_cifar} shows the magnitude of these perturbations both for the $100$ adversarially perturbed points $\mathcal{P}\subset \mathcal{T}$ and for a subset of $1,000$ unmodified samples $\mathcal{U}\subset \mathcal{T}$. Here, we can clearly see that, after fine-tuning, the boundaries around $\mathcal{P}$ have been completely modified, showing a large increase in the distance to the boundary in the direction of the original adversarial perturbation $\Delta_\mathrm{o}^\mathrm{f}(\bm{x})$ for $(\bm{x}, y)\in\mathcal{P}$. Meanwhile, the boundaries around $\mathcal{U}$ have not seen such a dramatic change. 

This means that modifying the position of \emph{only} a small fraction of the training samples can induce a large change in the shape of the boundary. Note that this dependency on a few samples resembles the one of support vector machines~\cite{Cortes1995SVM}, whose decision boundaries are defined by the position of a few supporting vectors. However, in contrast to SVMs, deep neural networks are not guaranteed to maximize margin in the input space (see \cref{fig:hmt-unstable_boundary}), and the points that support their boundaries need not be the ones closest to them, hence rendering their identification much harder.

\section{Implications in data-scarce applications}
\label{sec:hmt-icip}

Interestingly, the invariances obtained by robust classifiers that we analyzed in the previous chapters, have been recently shown to be quite beneficial when performing transfer learning~\cite{salman2020adversarially,utrera2020adversariallytrained}. In this section, we will demonstrate that the robust properties of deep networks can be exploited to improve the generalization performance in the off-the-shelf task of estimating the filling level within a container, where the available training data are typically very scarce.

Consider a real-world example of collaborative interactions between humans and robots. In such scenario, estimating through vision the physical properties of objects manipulated by humans is important for performing \emph{accurate} and \emph{safe} grasps of objects handed over by humans~\cite{SanchezMatilla2020BenchmarkHumanRobot}. For achieving successful grasps, one important property that should be estimated is the weight of the object based on the shape of the container~\cite{Xompero2020ICASSP_LoDE}, the type of content inside the container, and the amount of that content.

In particular, estimating the amount of content (filling level) within an \emph{unknown} container is a quite challenging problem due to distribution shifts occurred at test time related to, e.g., (i) occlusions caused by the hand holding the container, (ii) the transparencies of both the container and the filling (e.g.,~depth estimation may be highly inaccurate for transparent objects~\cite{Sajjan2020ICRA_ClearGrasp}), and (iii) by the differences in the shape of the containers. Typically, the few approaches designed to tackle this problem use RGB~\cite{Mottaghi2017ICCV}, thermal~\cite{Schenck2017ICRA}, or a combination of RGB and depth data~\cite{Do2016,Do2018}, and usually observe the action of pouring content in a container over multiple frames~\cite{Schenck2017ICRA,Schenck2017RSS,Do2016,Do2018}.

By approaching the problem as a classification task, the authors in~\cite{Mottaghi2017ICCV} showed that transfer learning~\cite{TanTransferLearning} was the best-performing strategy: self-collected data were used as task-specific dataset, the {\em target domain}, to fine-tune the parameters of a CNN pre-trained on the much larger ImageNet dataset~\cite{Deng2009ImageNet}, the {\em  source domain}. In the context of filling level estimation, the available training data are usually rather scarce, hence transfer learning introduces useful knowledge when fine-tuning on the target domain. 

\subsection{An off-the-shelf task: filling level classification}
\label{sec:icip-filling-level-classification}

\subsubsection{Task description and training strategies}
\label{subsec:icip-definitions-and-training-strategies}
\begin{figure}[t!]
    \centering
    \includegraphics[width=.6\columnwidth]{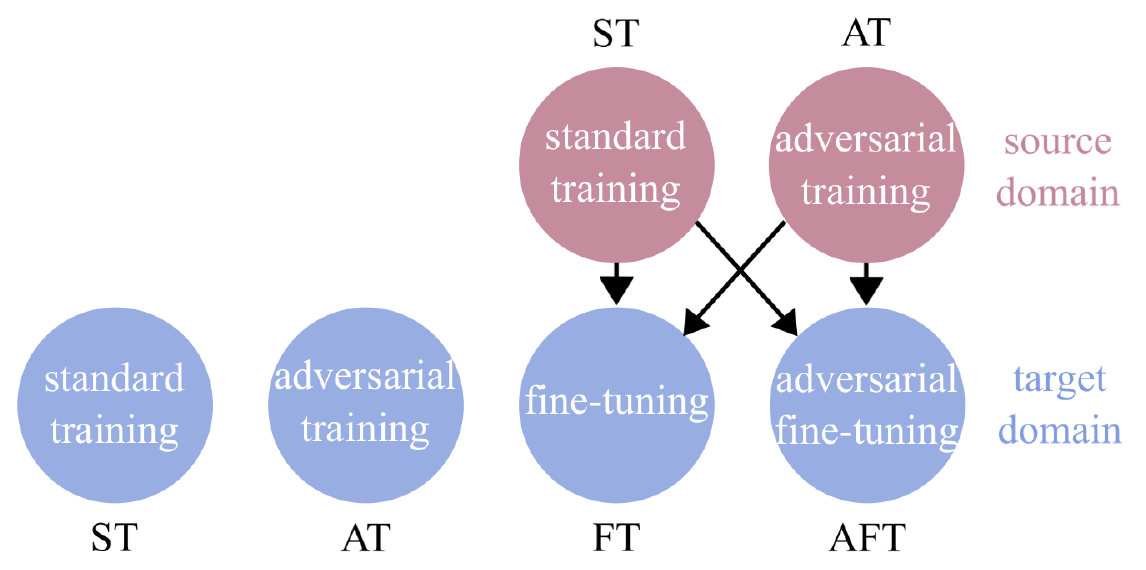}
    \caption{The six training strategies analyzed in our experiments: independent standard training (ST) and adversarial training (AT) on the target domain, and four transfer learning strategies from source to target domain via fine-tuning (FT).}
    \label{fig:icip-statediagram}
\end{figure}

We approach the problem of estimating the filling level, $y$, of a container captured in an image $\bm{x}$, as a classification task. We express the filling level as a percentage of the container's capacity:~$y \in \{0\%,50\%,90\%,\text{\emph{unknown}}\}$, where the \emph{unknown} class helps handling cases with opaque or translucent containers for which the filling level cannot be estimated through direct vision. Given a train set of image-label pairs $\mathcal{T}=\{(\bm{x}^i,y^i)\}_{i=1}^N$, the goal is to find a classifier that minimizes a suitable loss function $\mathcal{L}_\theta(\bm{x},y)$ such that $F$ correctly predicts $y$ for $\bm{x}\sim \mathcal{D}$ but $\bm{x}\notin \mathcal{T}$ (generalization).

We refer to the common strategy for training a classifier on a train set, $\mathcal{T}$, as Standard Training (ST). A good generalization may be achieved if the number of image-label pairs in $\mathcal{T}$ is very large, e.g.,~$N\approx1.2$ millions in ImageNet. However, for the target task of classifying the filling level  such amount of data is not available. Transfer learning helps to overcome this limitation by using an additional training set $\mathcal{S}$, with $|\mathcal{S}|=M \gg N$, that may not be related to the target task. Transfer learning pre-trains the parameters of $f$ on $\mathcal{S}$ (source domain) and then refines them on $\mathcal{T}$ (target domain) via fine-tuning (FT). We refer to this strategy as ST$\rightarrow$FT. With ST$\rightarrow$FT, the parameters of some layers in the pre-trained model are fixed and FT only refines those of the remaining layers. We will denote with $L$ the number of layers whose parameters are fixed. 

Recall that with Adversarial Training (AT), the resulting models learn features that correlate better with features of the classes of interest~\cite{Tsipras2019RobustnessOddsAccuracy,AllenZhu2020Purification,Engstrom2019AsPrior,Santukar2019ImageSynthesis}. Hence, $f$ is expected to learn more task-relevant features with  AT. We aim to evaluate AT on the filling-level classification task, and to compare it against five other strategies. As training strategies we consider ST$\rightarrow$FT~\cite{Mottaghi2017ICCV}; ST on the target domain; AT on the target domain; and three  combinations of AT with transfer learning, namely AT on the \emph{source} domain (AT$\rightarrow$FT), AT on the \emph{target} domain (ST$\rightarrow$AFT), and AT on \emph{both} domains (AT$\rightarrow$AFT).

Similarly to what was observed in~\cite{salman2020adversarially,utrera2020adversariallytrained}, we expect that the performance of fine-tuning on $\mathcal{T}$ will further improve if we use a model trained on $\mathcal{S}$ with AT instead of a model trained with ST, even if the classification performance of the robust model on $\mathcal{S}$ is worse than the performance of the model trained with ST. The exact reason behind this improvement is still an open question, but it is related to the differences in the learned features between standard and robust models. Also, this improvement depends on the value of $\epsilon$ used during AT, and the value that leads to better accuracy may differ across tasks and domains.  Smaller values for $\epsilon$  generally lead to better performance~\cite{utrera2020adversariallytrained}, but its value will be selected empirically.

Finally, the last two training  combinations apply AT either on the \emph{target} domain (ST$\rightarrow$AFT) via FT or on \emph{both} domains (AT$\rightarrow$AFT). Considering the effect of AT on the features learned by a classifier, we will investigate how $f$ is affected when the transferred learned features from $\mathcal{S}$ are further filtered by AT on $\mathcal{T}$. \cref{fig:icip-statediagram} summarizes the training strategies under analysis, which will be compared in the next section.

\subsubsection{A novel dataset}
\label{subsec:icip-dataset}

Since the task is to classify the filling level from a single RGB image, the CCM dataset~\cite{Xompero_CCM} is a suitable choice due to its large variability in terms of capturing conditions. CCM comprises of four views capturing under different backgrounds and illumination conditions cups and drinking glasses. The containers are transparent, translucent or opaque. The content is transparent (water) or opaque (pasta, rice). Each container stands upright on a surface or is being manipulated by a person. We only consider data of the public CCM repository, namely 4 cups and 4 drinking glasses. 
 
From the CCM video data, we automatically sampled and then visually verified 10,269 frames of containers for which a pouring action was completed. To increase the variability in the sampled data, we selected frames considering that the container is completely visible or occluded by the person's hand, and under different backgrounds. 
\begin{figure}[t!]
    \centering
    \includegraphics[width=.5\textwidth]{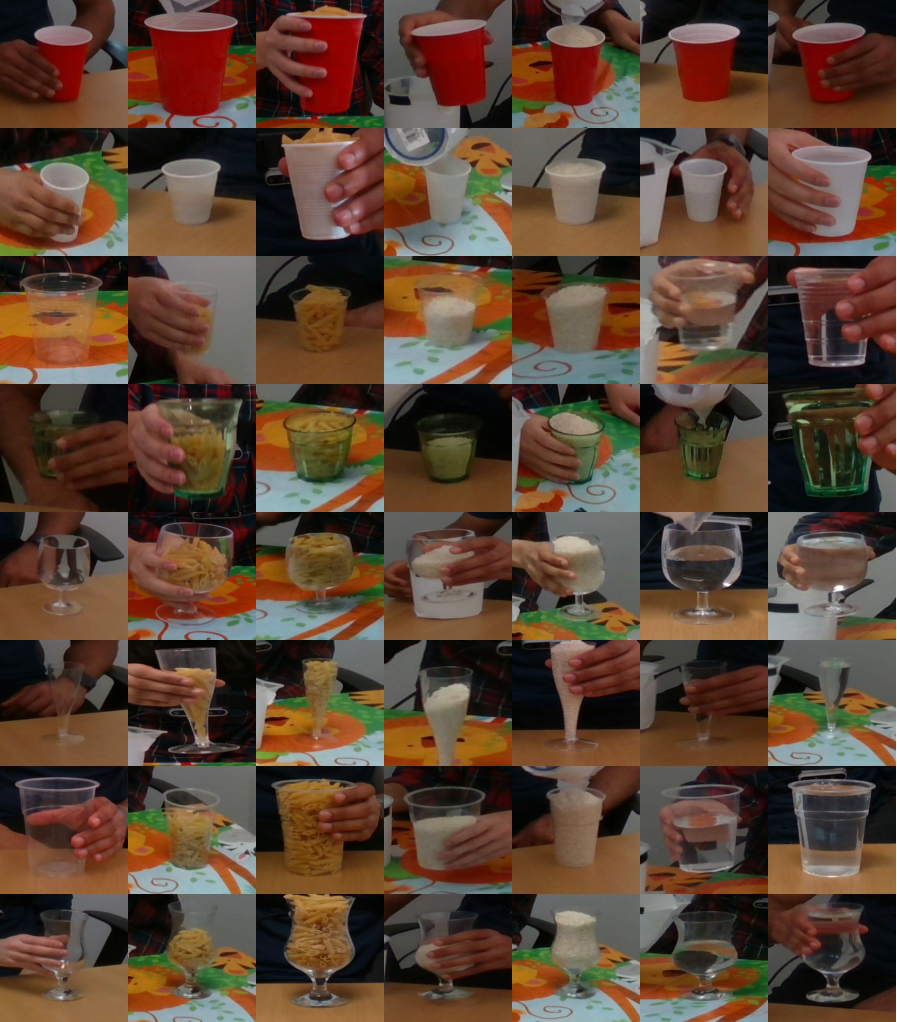}
    \caption{Sample images (resized crops) from the CORSMAL Containers Manipulation dataset~\protect\cite{Xompero_CCM}. Each column shows different filling types and levels, and each row shows different backgrounds and hand occlusions.
    }
    \label{fig:icip-dataset}
\end{figure}

For each frame, the final image is extracted by cropping only the region with the container using Mask R-CNN~\cite{He2017ICCV_MaskRCNN}, followed by visual verification. 
Each crop is associated to an annotation of filling type and filling level (empty or filled at 50\% or 90\% of the  capacity of the container), hand occlusion, and transparency of the container. We call this image dataset Crop-CCM or \emph{C-CCM}. Sample C-CCM images\footnote{Sampled images can be found at \protect\url{https://corsmal.eecs.qmul.ac.uk/filling.html}} are shown in \cref{fig:icip-dataset}.

\begin{figure}[t!]
    \centering
    \includegraphics[width=0.75\textwidth]{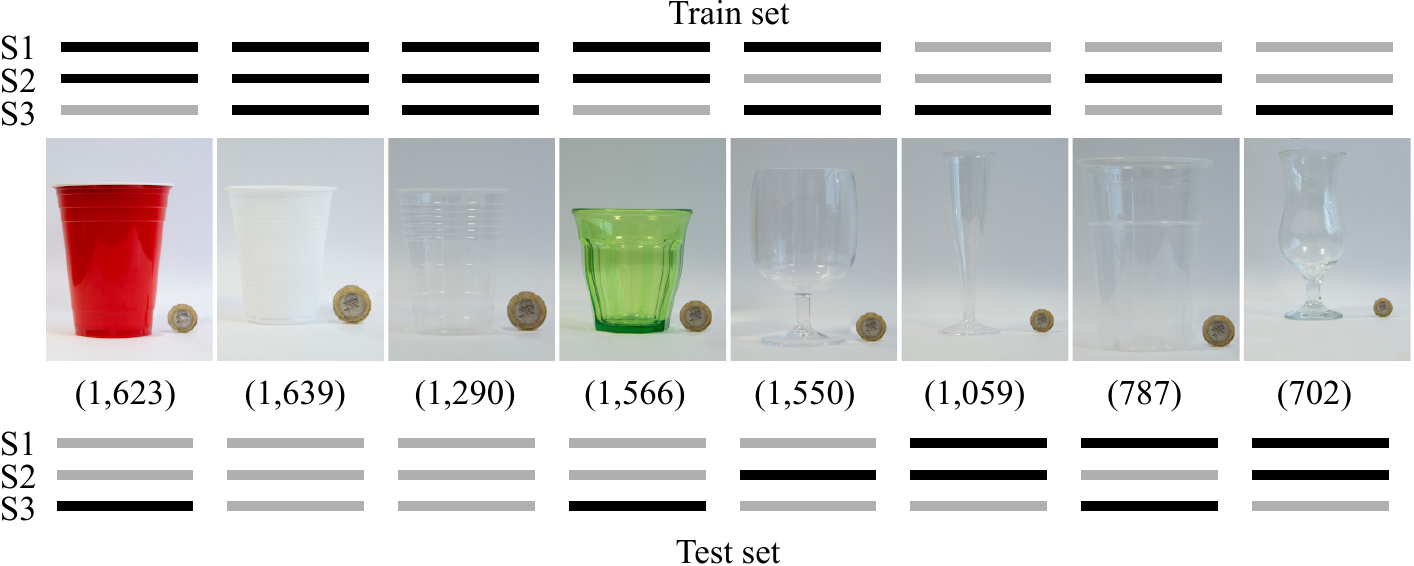}
    \caption{Comparison of three train and test splits (S1, S2, S3) of the public containers from CCM for the shape analysis in the experiments. Black lines mean that the set of images belonging to that container are part of the train (test) set in the data split. The number of images for each container are shown in parentheses. Note the diversity in shape, color, texture, transparency, and size.}
    \label{fig:icip-containers}
\end{figure}

Finally, in order to evaluate the robustness of the classifier to distribution shifts of the test data, we focus mainly on the shape of the containers, but also in their color, texture, transparency, and size. To investigate these aspects, we split C-CCM into train and test sets under three configurations. The first configuration ($\text{S}_1$) considers a {champagne flute} in the test set to further increase the shape variability of containers not previously seen in the train set. The second configuration ($\text{S}_2$) swaps a {beer cup} with a {wine glass} to analyze the influence of the stem of the wine glass. The last configuration ($\text{S}_3$) places all the containers with a stem in the train set, and the test set contains only cups without stem, as well as a red cup and a green transparent cup, which are characterized by differences in color, texture and transparency with respect to the training data. \cref{fig:icip-containers} shows the three configurations and the number of samples for each container type.

\subsection{Adversarial invariance prevents overfitting}
\label{sec:icip-robust-invariance-prevents}
In this section we analyze the effect of the transfer learning parameters, and then evaluate the generalization performance of the six different training strategies on the C-CCM dataset. We use as classifier a ResNet-18~\cite{He2016ResNet}. Note that we also conducted experiments using a ResNet-50 and a WideResNet-50~\cite{wide_resnet}, and the findings are similar to the ones of ResNet-18. We will focus on ResNet-18 as it is the least complex network among the three. With ST we train the classifier on C-CCM, whereas with AT we train the classifier on images modified with $\ell_2$ adversarial perturbations crafted with the $10$-iteration PGD~\cite{Madry2018TowardsDeepLearning}. With the transfer learning  strategies we fine-tune the available pre-trained models on C-CCM: for ST$\rightarrow$FT and ST$\rightarrow$AFT we use the pre-trained model provided by PyTorch~\cite{Paszke2019PyTorch}, whereas for AT$\rightarrow$FT and AT$\rightarrow$AFT we use the robust models provided by~\cite{salman2020adversarially}.

For each strategy, we train or fine-tune the classifier for $30$ epochs, using a cross-entropy loss and stochastic gradient descent. The learning rate for updating the weights is set to $0.1$ when training directly on C-CCM, and $0.005$ when performing transfer learning. The learning rate decays linearly during training. Note that the models we evaluate are the ones obtained at the end of the training epochs (no early-stopping), while for dealing with class imbalances, the training images in a batch are randomly sampled with probabilities that are inversely proportional to the number of images of each class. 

\subsubsection{Sensitivity analysis}
\label{subsec:icip-sensitivity-analysis}

We perform a sensitivity analysis on the number of fixed layers ($L$)  in  fine-tuning with ST$\rightarrow$FT,  ST$\rightarrow$AFT, AT$\rightarrow$FT and AT$\rightarrow$AFT; and to select the size of the bound for crafting the adversarial perturbation for  AT, ST$\rightarrow$AFT, AT$\rightarrow$FT and AT$\rightarrow$AFT.
Note that we differentiate $\epsilon$ for the  source, $\epsilon^s$, and target, $\epsilon^t$, domain. 
Specifically, we perform the sensitivity analysis only for $\epsilon^s$ with AT$\rightarrow$FT, and for each data split configuration we select the $\epsilon^s$ for which AT$\rightarrow$FT achieves the highest accuracy.
Then, based on these values of $\epsilon^s$, for each data configuration we set $\epsilon^t=\epsilon^s$: since we use $10$-iteration $\ell_2$-PGD, performing a sensitivity analysis or a grid search on $\epsilon^t$ is computationally inefficient, as it is analogous to increasing almost $10\times$ the training epochs.

\pgfplotstableread{images/ch5_corsmal_icip/data_splits/layers_scenarios.txt}\sensitivitylayers
\pgfplotstableread{images/ch5_corsmal_icip/data_splits/epsilon_scenarios.txt}\sensitivityepsilon
\begin{figure}[t!]
    \centering
    \setlength\tabcolsep{2pt}
    \begin{tabular}{cc}
    \begin{tikzpicture}
    \begin{axis}[
        width=.35\columnwidth,
        ymin=45,ymax=85,
        ylabel={Accuracy (\%)},
        tick label style={font=\footnotesize},
        xmin=0.5, xmax=5.5,
        xtick={1, 2, 3, 4, 5},
        xticklabels={0, 1, 2, 3, 4},
        xlabel={$L$},
        label style={font=\footnotesize},
    ]
    \addplot+[solid, color=black, mark=square*, mark options={mark size=2pt,fill=black}] table[x=Layers, y=Scenario1]{\sensitivitylayers};
    \addplot+[solid, color=black, mark=*, mark options={mark size=2pt,fill=black}] table[x=Layers, y=Scenario2]{\sensitivitylayers};
    \addplot[solid, color=black, mark=triangle*, mark options={mark size=2pt,fill=black}] table[x=Layers, y=Scenario3]{\sensitivitylayers};
    \addplot+[only marks, color=red, mark=square*, mark options={mark size=2pt,fill=red}] coordinates {(2, 78.34)
	};
	\addplot+[only marks, color=red, mark=*, mark options={mark size=2pt,fill=red}] coordinates {(2,65.63)
	};
	\addplot[only marks, color=red, mark=triangle*, mark options={mark size=2pt,fill=red}] coordinates {(2,82.32)
	};
    \end{axis}
    \end{tikzpicture} &
    \begin{tikzpicture}
    \begin{semilogxaxis}[
        width=.35\columnwidth,
        ymin=60,ymax=92,
        tick label style={font=\footnotesize},
        xmin=0.005, xmax=1.5,
        xtick={0.01,0.05,0.1,0.5,1},
        xticklabels={.01,.05,.1,.5,1},
        xlabel={$\epsilon^s$},
        label style={font=\footnotesize},
    ]
    \addplot+[solid, color=black, mark=square*, mark options={mark size=2pt,fill=black}] table[x=Epsilon, y=Scenario1]{\sensitivityepsilon};
    \addplot+[solid, color=black, mark=*, mark options={mark size=2pt,fill=black}] table[x=Epsilon, y=Scenario2]{\sensitivityepsilon};
    \addplot[solid, color=black, mark=triangle*, mark options={mark size=2pt,fill=black}] table[x=Epsilon, y=Scenario3]{\sensitivityepsilon};
	\addplot+[only marks, color=red, mark=square*, mark options={mark size=2pt,fill=red}] coordinates {(0.05, 80.97)
	};
	\addplot+[only marks, color=red, mark=*, mark options={mark size=2pt,fill=red}] coordinates {(1,73.27)
	};
	\addplot[only marks, color=red, mark=triangle*, mark options={mark size=2pt,fill=red}] coordinates {(0.5,88.23)
	};
    \end{semilogxaxis}
    \end{tikzpicture}
    \end{tabular}
    \caption{
    Sensitivity analysis for the number of fixed layers $L$ with ST$\rightarrow$FT (left) and for the maximum amount of perturbation bound, $\epsilon^s$, with AT$\rightarrow$FT on test set of the three  dataset splits: first split $\text{S}_1$ (\protect\raisebox{1pt}{\protect\tikz \protect\draw[black,fill=black] (0,0) rectangle (1.ex,1.ex);}), second split $\text{S}_2$ (\protect\raisebox{1pt}{\protect\tikz \protect\draw[black,fill=black] (1,1) circle (0.5ex);}), third split $\text{S}_3$ (\protect\raisebox{1pt}{\protect\tikz \protect\node[fill=black,regular polygon,regular polygon sides=3,inner sep=1.5pt] at (-0.5,0) {};}). Red indicates the highest achieved accuracy.
    Note the different scale of the y-axis, and the logarithmic scale for the x-axis (right).
    }
    \label{fig:icip-analysislayers}
\end{figure}

We first analyze the classification accuracy on the test sets of the three dataset splits when varying the number of fixed layers for ST$\rightarrow$FT as $L=\{0,1,2,3,4\}$. Here, $L=0$ denotes that no layer remains fixed during fine-tuning (the full network is updated). Note that for a ResNet-18 classifier, a layer is a ResNet block of convolutions and batch normalization (see the original ResNet paper~\cite{He2016ResNet}). Since the target dataset is small, it is reasonable to fix the first layer ($L=1$) in order to prevent the classifier from a possible overfitting~\cite{Yosinski2014}.
Indeed, \cref{fig:icip-analysislayers} (left) shows that the accuracy on the test set of all configurations (S1, S2, S3) is consistently higher for $L=1$ (78.34\%, 65.63\%, 82.32\%), while it gradually decays as $L$ grows. This is also expected~\cite{Yosinski2014}, since we allow fewer layers to be fine-tuned on the target datasets, and the classifiers then mostly use fixed features from ImageNet. Therefore, we set $L=1$ for ST$\rightarrow$FT as well as for ST$\rightarrow$AFT, AT$\rightarrow$FT, and AT$\rightarrow$AFT.

Then, we set$L=1$ and analyze the classification accuracy of AT$\rightarrow$FT when varying the perturbation size in the source domain, $\epsilon^s$. \cref{fig:icip-analysislayers} (right) shows that the highest achieved accuracy is different for each dataset configuration: 80.97\% for $\text{S}_1$ with $\epsilon^s=0.05$, 73.27\% for $\text{S}_2$ with $\epsilon^s=1$, and 88.23\% for $\text{S}_3$ with $\epsilon^s=0.5$. As mentioned previously, we use these values of $\epsilon^s$ also for $\epsilon^t$ when performing AT, ST$\rightarrow$AFT, and AT$\rightarrow$AFT. However, we observed that the model trained with ST$\rightarrow$AFT is unable to converge (train accuracy around 45\%) on $\text{S}_2$ for $\epsilon^t=1$ and on $\text{S}_3$ for $\epsilon^t=0.5$, while it successfully converges on $\text{S}_1$ for the smaller $\epsilon^t=0.05$. We believe that this might be caused by the fact that AT with larger $\epsilon^t$ values eliminates many non-robust, yet useful, features transferred from ImageNet, and prevents the model from fitting the remaining features. Hence, we  set $\epsilon^t=0.05$ for ST$\rightarrow$AFT across all dataset configurations for the rest of the experiments, since with this value the network converges for all dataset configurations.

\subsubsection{Generalization to unseen data}
\label{subsec:icip-generalization-to-unseen}

\pgfplotstableread{images/ch5_corsmal_icip/data_splits/accuracy_new.txt}\accscenarios
\begin{figure*}[t!]
    \centering
    \begin{tikzpicture}
    \begin{axis}[
    axis x line*=bottom,
    axis y line*=left,
    enlarge x limits=false,
    ybar,
    width=\linewidth,
	bar width=3pt,
    xmin=0,xmax=15,
    xtick=data,
    height=0.33\columnwidth,
    ymin=0,  ymax=100,
    ytick={0,20,40,60,80,100},
    ylabel={Accuracy (\%)},
    label style={font=\footnotesize},
    tick label style={font=\footnotesize},
    ymajorgrids=true,
    ]
    \addplot+[ybar, black, fill=ts1, draw opacity=0.5] table[x=CoID,y=S1]{\accscenarios};
    \addplot+[ybar, black, fill=ts2, draw opacity=0.5] table[x=CoID,y=S2]{\accscenarios};
    \addplot+[ybar, black, fill=ts3, draw opacity=0.5] table[x=CoID,y=S3]{\accscenarios};
    \addplot+[ybar, black, fill=ts4, draw opacity=0.5] table[x=CoID,y=S4]{\accscenarios};
    \addplot+[ybar, black, fill=ts5, draw opacity=0.5] table[x=CoID,y=S5]{\accscenarios};
    \addplot+[ybar, black, fill=ts6, draw opacity=0.5] table[x=CoID,y=S6]{\accscenarios};
    \end{axis}
    \begin{axis}[
        axis x line*=top,
        axis y line*=right,
        width=\linewidth,
        height=.33\columnwidth,
        xmin=0,xmax=15,
        tick label style={font=\footnotesize, align=center,text width=3cm},
        xtick={2.5,7.5,12.5},
        xticklabels={$\text{S}_1$,$\text{S}_2$,$\text{S}_3$},
        typeset ticklabels with strut,
        label style={font=\footnotesize},
        ymin=0,ymax=100,
        yticklabels={},
    ]
    \end{axis}
    \node[inner sep=0pt] (flute1) at (0.86,-1)
    {\includegraphics[width=.08\columnwidth]{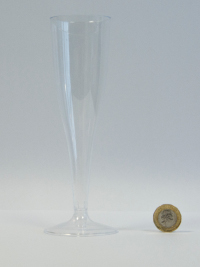}};
    \node[inner sep=0pt] (flute1) at (2.19,-1)
    {\includegraphics[width=.08\columnwidth]{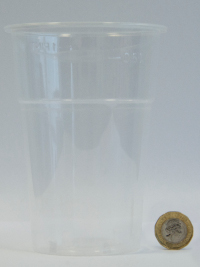}};
    \node[inner sep=0pt] (flute1) at (3.53,-1)
    {\includegraphics[width=.08\columnwidth]{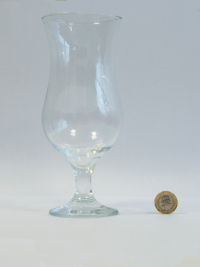}};
    \node[inner sep=0pt] (flute1) at (5.26,-1)
    {\includegraphics[width=.08\columnwidth]{images/ch5_corsmal_icip/containers/champagne_flute_glass.jpg}};
    \node[inner sep=0pt] (flute1) at (6.59,-1)
    {\includegraphics[width=.08\columnwidth]{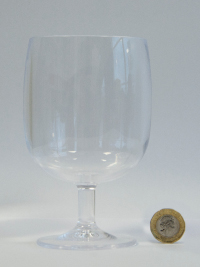}};
    \node[inner sep=0pt] (flute1) at (7.93,-1)
    {\includegraphics[width=.08\columnwidth]{images/ch5_corsmal_icip/containers/cocktail_glass.jpg}};
    \node[inner sep=0pt] (flute1) at (9.66,-1)
    {\includegraphics[width=.08\columnwidth]{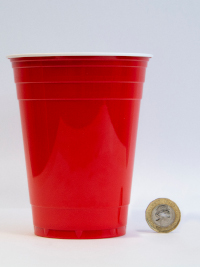}};
    \node[inner sep=0pt] (flute1) at (10.99,-1)
    {\includegraphics[width=.08\columnwidth]{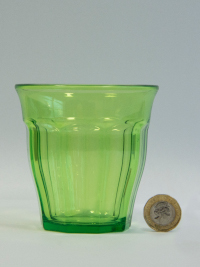}};
    \node[inner sep=0pt] (flute1) at (12.33,-1)
    {\includegraphics[width=.08\columnwidth]{images/ch5_corsmal_icip/containers/beer_cup.jpg}};
    \end{tikzpicture}
   \caption{
   Comparison of the per-container filling level classification accuracy (\%) for the six training strategies. Note the different containers in the test set for each dataset split (see \protect\cref{fig:icip-containers} for the train set of each split).
   Legend:
    \protect\raisebox{2pt}{\protect\tikz \protect\draw[ts1,line width=2] (0,0) -- (0.3,0);}~ST,
    \protect\raisebox{2pt}{\protect\tikz \protect\draw[ts2,line width=2] (0,0) -- (0.3,0);}~AT,
    \protect\raisebox{2pt}{\protect\tikz \protect\draw[ts3,line width=2] (0,0) -- (0.3,0);}~ST$\rightarrow$FT,
    \protect\raisebox{2pt}{\protect\tikz \protect\draw[ts4,line width=2] (0,0) -- (0.3,0);}~ST$\rightarrow$AFT,
    \protect\raisebox{2pt}{\protect\tikz \protect\draw[ts5,line width=2] (0,0) -- (0.3,0);}~AT$\rightarrow$FT,
    \protect\raisebox{2pt}{\protect\tikz \protect\draw[ts6,line width=2] (0,0) -- (0.3,0);}~AT$\rightarrow$AFT.
   }
    \label{fig:icip-shapeanalysis}
\end{figure*}

Since all the parameters have been decided, we can now proceed with evaluating the generalization performance of every training strategy on each container in the test set individually. Note that, in this way, we can practically measure the robustness of the classifier to the test-time distribution shifts. 

\Cref{fig:icip-shapeanalysis} shows the filling level classification performance. Constrained by the amount, and hence by the diversity, of training images, the differently trained classifiers could potentially develop biases or overfit to some features, such as the shape of a container. AT$\rightarrow$FT achieved superior performance most of the times. With transfer learning, the features introduced from ImageNet appear to decrease such biases, and enable the classifiers to identify features in the train set that are more generalizable. When combining transfer learning with AT at the source domain, the biases are modulated with the transferred features that are also filtered by AT, and the generalization of the classifier further increases. These results confirm that adversarial training improves transfer learning, even in the context of the challenging filling level classification task.

Overall, whenever the performance of ST is low, all transfer learning strategies lead to a significant improvement. On the contrary, whenever ST performs well, the contribution of transfer learning is insignificant, and sometimes it even decreases the final performance. Furthermore, applying AT on the \emph{target} domain, either alone or combined with transfer learning, may even be harmful for the classifier. 

For $\text{S}_1$, the accuracy of ST on the beer cup (middle) is already very high -- due to the similar shape of the small transparent cup in the training set -- and the other training strategies do not further improve it. The accuracy on the cocktail glass (right) is similar for all strategies,  with AT$\rightarrow$FT performing slightly better. 
As for the champagne flute (left), the performance of ST and AT is quite low ($\sim$46\%), which might be caused by the unique shape of the flute with respect to the shapes in the training set. However, the accuracy significantly improves with transfer learning, and AT$\rightarrow$FT outperforms all the other strategies by $\sim$30 percentage points (pp). 

For $\text{S}_2$, the accuracy of all strategies on the champagne flute (left) is similar to the one achieved on $\text{S}_1$. The accuracy on the cocktail glass (right) is much lower for most strategies ($\sim$10pp less compared to the performance on $\text{S}_1$), except AT$\rightarrow$FT, which drops only by 3pp and again outperforms the rest of the strategies. The drop of the other strategies could be caused by the lack of a container with a stem in the training set. Finally, the performance on the wine glass (middle) is similar for most strategies, with AT$\rightarrow$FT being again slightly better than the rest. Compared to the cocktail glass, the higher accuracy of all strategies on the wine glass could be caused by the similarity of its shape above the stem with the other transparent cups in the training set, despite the fact that no container with a stem is presented in the training set.

For $\text{S}_3$, the accuracy of ST on the beer cup (right) is high and the other training strategies do not improve it. Instead, the accuracy of ST on the green glass (middle) is lower and reaches an accuracy of 66\%. Although ST$\rightarrow$FT does not improve the accuracy, AT$\rightarrow$FT significantly increases it (almost 10pp). The red cup (left) obtains the most interesting improvement compared to the 0.005\% accuracy of ST: all transfer learning techniques achieve an accuracy above 90\%, with AT$\rightarrow$FT achieving 99.5\% classification accuracy. By inspecting the predictions of ST and AT, the classifier assigned the label full (filling level: $90\%$) almost 99\% of the times. In fact, predicting the $\text{\emph{unknown}}$ class is conceptually different from estimating the filling level, and it is more related to classifying non-transparent containers. In this sense, the features learned for transparent objects that are full with rice or pasta might be correlated with the features of the red cup.

\subsection{Discussion}
In this section, we found out that using adversarial training on the source dataset (ImageNet) followed by transfer learning on the target dataset (C-CCM), permits to consistently improve generalization to unseen containers. Our analysis demonstrates the possibilities of exploiting adversarial robustness for tasks that extend beyond classical image classification settings. 

However, adversarial robustness is not necessarily the best way to improve the robustness to certain distribution shifts, such as the one introduced by common corruptions of the data. This is gonna be the main focus of the next chapter, where we will design an efficient data augmentation scheme for conferring robustness to such shifts. In fact, this augmentation scheme can also be applied in the context of filling level estimation. As we will see, transfer learning from adversarially trained models -- and transfer learning in general -- is not necessary, since one can properly tune our data augmentation scheme of the next section in order to achieve similar or even better generalization performance.

\section{Conclusions}
\label{sec:hmt-conclusions}

In this chapter, we proposed a new geometric framework that permits to relate the features of a dataset with the distance to the decision boundary along specific directions. Through a series of carefully designed experiments, both in synthetic and real image datasets, we explained how the inductive bias of the learning algorithm shapes the decision boundaries of neural networks, by creating boundaries that are invariant to non-discriminative directions.

Furthermore, we demonstrated that the decision boundaries are very sensitive to the position of the training samples, and that very small changes along specific directions can cause large and sudden changes in these directions. In fact, adversarial training exploits this sensitivity of the decision boundaries, as well as their inductive bias towards invariance to non-discriminative features, in order to build more robust classifiers. Interestingly, modifying the position of just a \emph{minimal} number of training samples during adversarial training is enough to locally introduce excessive invariance.

Finally, we studied the implications that the invariance properties of robust models have in the downstream application of classifying the filling level within containers. In particular, we showed that, during transfer learning, using robust models in the source domain permits to consistently improve generalization to unseen containers in the target domain that come from a shifted distribution.

We believe that our new framework can be used in future research to investigate the connections between training features and the macroscopic geometry of deep models. This can serve as a tool to obtain new insights on the intriguing properties of deep networks, as we demonstrated for their catastrophic forgetting. On the practical side, there are some important applications that could benefit from our findings. In terms of robustness, identifying the small subspace of discriminative features of a network can lead to faster black box-attacks by restricting the search space of the perturbations. In fact our analysis answers why using low-frequency perturbations improves the query efficiency in recent attacks~\cite{Tsuzuku2019StructuralSens, Sharma2019EffectivenessLowFrequency, Rahmati2020GeoDA}. Simultaneously, the dependency of boundaries to just a few training samples can be exploited to design faster adversarial training schemes, and is a clear avenue for future research in active learning. Finally, having a better understanding about the mechanisms that lead to excessive invariance~\cite{TramerFundamentalTradeoffs} after adversarial training could help boost the standard accuracy of robust models.

\cleardoublepage
\chapter{Robustness to non-adversarial distribution shifts}
\label{ch:prime}

\begin{raggedleft}
    \textit{``Your favourite virtue... Simplicity!''} \\
    --- Karl Marx \\
\end{raggedleft}
\vspace*{2cm}

\section{Introduction}
\label{sec:prime-intro}

In the previous chapters, we mainly exploited adversarial proxies for evaluating the robustness of image classifiers, and for developing methodologies that enabled us to understand, analyze and further explore multiple properties of deep networks. Furthermore, we also saw that, in some tasks, the invariances that adversarially trained classifiers obtain can boost the robustness of deep networks to specific distribution shifts of the data. In this chapter, we steer our focus from the adversarial to the more general setting, and investigate the robustness of deep networks to common distortions of the images.
\blfootnote{Part of this chapter appears in}
\blfootnote{``PRIME: A few primitives can boost robustness to Common Corruptions''. In \textit{European Conference on Computer Vision (ECCV)}, 2022~\cite{Modas2021PRIME}.}
\blfootnote{``Data augmentation with mixtures of max-entropy transformations for filling-level classification''. In \textit{European Signal Processing Conference (EUSIPCO)}, 2022~\cite{Modas2022PrimeFilling}.}

In general, deep image classifiers do not work well in the presence of various types of distribution shifts~\cite{Dodge2017StudyAndComparisonOfHuman, Geirhos2018GeneralisationInHumans, Taori2020MeasuringRobustnessToNatural}. Most notably, their performance can severely drop when the input images are affected by common distortions that are not contained in the training data, such as digital artefacts, low contrast, or blurs~\cite{Hendrycks2019Benchmarking, Mintun2021Cbar}.

``Common corruptions'' is an umbrella term used to describe the set of distortions that can happen to natural images during their acquisition, storage, and processing lifetime, which can be very diverse.  Nevertheless, while the space of possible perturbations is huge, the term ``common corruptions'' is generally used to refer to typical image transformations that, while degrading the quality of the images, still preserve their semantic information.

Building classifiers that are robust to common corruptions is far from trivial. A naive solution is to include data with all sorts of corruptions during training, but the sheer scale of all possible types of typical perturbations that might affect an image is simply too large. Moreover, the problem is per se ill-defined since there exists no formal description of all possible common corruptions.

Due to the luck of such formal description, the research community has recently favoured increasing the ``diversity'' of the training data via data augmentation schemes~\cite{Cubuk2019Autoaugment,Hendrycks2020AugMix,Hendrycks2021TheManyFaces}. Intuitively, the hope is that showing very diverse augmentations of an image to a network would increase the chance that the latter becomes invariant to some common corruptions. Still, covering the full space of common corruptions is hard. Hence, current literature has mostly resorted to increasing the diversity of augmentations by designing intricate data augmentation pipelines, e.g., introducing DNNs for generating varied augmentations~\cite{Hendrycks2021TheManyFaces,Calian2022DefendingImageCorruptions}, or coalescing multiple techniques~\cite{Wang2021AugMax}, and thus achieve good performance on different benchmarks. This strategy, though, leaves a big range of unintuitive design choices, making it hard to pinpoint which elements of these methods meaningfully contribute to the overall robustness. Meanwhile, the high complexity of recent methods \cite{Wang2021AugMax,Calian2022DefendingImageCorruptions} makes them impractical for large-scale tasks. Whereas, some methods are tailored to particular datasets and might not be general enough. Nonetheless, the problem of building robust classifiers is far from completely solved, and the gap between robust and standard accuracy is still large.

In this chapter, we take a step back and provide a systematic way for designing a simple, yet effective data augmentation scheme. By focusing on first principles, we formulate a new model for semantically-preserving corruptions, and build on basic concepts to characterize the notions of transformation strength and diversity using a few transformation primitives. Relying on this model, we propose \emph{PRIME}, a data augmentation scheme that draws transformations from a max-entropy distribution to efficiently sample from a large space of possible distortions. The performance of PRIME, alone, already tops the current baselines on different common corruption datasets, whilst it can also be combined with other methods to further boost their performance. Moreover, the simplicity and flexibility of PRIME allows to easily understand how each of its components contributes to improving robustness. Finally, we demonstrate that PRIME provides a ready-to-use recipe for the problem of filling level classification.

The rest of the chapter is organized as follows: In \cref{sec:prime-towards-robustness-natural-corruptions} we demonstrate that simple pre-processing techniques might not be enough for inducing invariance to all possible corruptions, and we formulate a new mathematical model for semantically-preserving corruptions. Then, in \cref{sec:prime-prime-data-augmentations} we provide a data augmentation scheme that utilizes the proposed model of visual distortions, and evaluate its performance on multiple corruption benchmarks. Furthermore, in \cref{sec:prime-robustness-insights} we use our method to investigate different aspects behind robustness to common corruptions, while, in \cref{sec:corsmal-prime} we tune PRIME to confer robustness to distributions shifts in filling level classification. Finally, in \cref{sec:prime-discussion} we provide a discussion on open challenges and potential extensions of our method.

\section{Towards robustness to common corruptions}
\label{sec:prime-towards-robustness-natural-corruptions}

\subsection{Invariance by removing features}
\label{subsec:prime-invariance-pre-processing}


Data augmentation is the most common technique for improving the robustness of classifiers to common distortions of the images. However, designing efficient data augmentation schemes can be challenging, while at the same time the methods can be computationally intense. To this end, it is reasonable to explore if the ideas introduced in \cref{ch:hold-me} can also be beneficial. That is, apply some simple pre-processing operations for inducing specific invariances that can improve the robustness to common corruptions, instead of generating augmentations at each training iteration. 

In order to evaluate this approach, we will use the widely adopted \emph{Common Corruptions}~\cite{Hendrycks2019Benchmarking} benchmark. This benchmark consists of $15$ natural image distortions that can naturally occur during acquisition, processing or storing of an image, each applied with $5$ different severity levels. At a higher level, these corruptions can be grouped into four categories, namely ``noise'', ``blur'', ``weather'' and ``digital''. 

In previous works~\cite{Yin2019FourierPerspective,kireev2021} it was observed that some of these common corruptions affect the high frequency part of the image frequency spectrum (i.e., high-frequency distortions). Hence, following the exact same procedure as in \cref{subsec:hmt-invariance-and-elasticity}, we can train the classifier on a low-pass filtered version of the data, such that the network will be invariant to any high-frequency changes introduced by the corruptions.
Furthermore, color is another important image space that can be affected by the aforementioned common corruptions. In this sense, we might also want to introduce to the network some invariance to color changes, similarly to what can be achieved by projecting the images onto the low-frequency part of the spectrum to achieve invariance to the high-frequency one. Hence, we can remove the color information from the data by training the classifier on grayscale images. Note, however, that for retaining the architecture of the network such that it can still be able to process standard RGB images at test time, we repeat the grayscale training images along 3 channels. 

In order to evaluate the aforementioned operations we conduct a small experiment. Specifically, we train a ResNet-18 for $30$ epochs on CIFAR-10 and its pre-processed versions, using SGD and a cross-entropy loss function. For comparison, we also train a network on CIFAR-10 using AugMix~\cite{Hendrycks2020AugMix}, a widely-used data augmentation method that achieves very good robustness on the common corruptions benchmark. For simplicity, we do not deploy the Jensen-Shannon Divergence (JSD) consistency loss -- as introduced in the original AugMix paper -- when training with AugMix. Finally, we evaluate the performance of the resulting networks by measuring their accuracy on the validation set of CIFAR-10 (C-10) and its corruption counterpart, CIFAR-10-C (C-10-C).

\begin{table}[t]
    \centering
    \small
    \aboverulesep=0ex
    \belowrulesep=0ex
    \begin{tabular}{lccc|cc}
        \cmidrule[\heavyrulewidth]{2-6}
        & Standard & LP & Gray & AugMix & AugMix\texttt{+}LP \\
        \midrule
        Gauss. noise & 46.1 & \textbf{57.6} & 45.6 & 69.6 & \underline{70.9} \\
        Shot noise & 58.2 & \textbf{65.9} & 56.4 & 78.3 & \underline{79.1} \\
        Impulse noise & 54.5 & 56.2 & \textbf{57.4} & 77.5 & \underline{78.5} \\
        \midrule
        Defocus blur & 81.5 & \textbf{88.8} & 71.7 & 91.8 & \underline{91.5} \\
        Glass blur & 51.0 & \underline{\textbf{85.3}} & 35.9 & 64.0 & 65.4 \\
        Motion blur & 75.0 & \textbf{84.8} & 62.9 & \underline{88.1} & 88.0 \\
        Zoom blur & 75.3 & \textbf{87.6} & 62.0 & \underline{89.9} & 89.6 \\
        \midrule
        Snow & 80.6 & \underline{\textbf{85.7}} & 73.0 & 84.5 & 85.0 \\
        Frost & 75.5 & \underline{\textbf{86.9}} & 66.5 & 83.8 & 84.3\\
        Fog & \textbf{88.0} & 81.2 & 82.3 & \underline{89.9} & 89.7 \\
        Brightness & \textbf{92.9} & 89.0 & 91.0 & 92.7 & \underline{93.0} \\
        \midrule
        Contrast & 76.1 & 68.8 & \textbf{78.4} & 83.6 & \underline{84.5} \\
        Elastic & 81.7 & \textbf{86.5} & 73.1 & \underline{86.9} & 86.5 \\
        Pixelate & 73.9 & \underline{\textbf{90.1}} & 63.3 & 79.5 & 81.2 \\
        JPEG & 77.4 & \textbf{83.1} & 73.0 & 82.4 & \underline{83.2} \\
        \midrule
        Clean & \underline{\textbf{94.4}} & 91.1 & 92.4 & 94.1 & 94.1 \\
        Avg. Corruption & 72.5 & \textbf{79.8} & 66.2 & 82.8 & \underline{83.4}\\
        \bottomrule
    \end{tabular}
    \caption{Clean and corruption accuracy of ResNet-18 on CIFAR-10 using pre-processing methods, AugMix, or combining AugMIx with low-pass filtering (AugMix\texttt{+}LP). Bold: maximum value among Standard, LP, and Gray. Underline: maximum value per row.}
\label{tab:prime-results-pre-processing}
\end{table}

The results are presented in \cref{tab:prime-results-pre-processing}. Focusing on the pre-processing methods, the first thing to notice is that the low-pass (LP) filtering has a very strong effect on the robustness of the classifier, mainly improving the accuracy on blurs and digital corruptions (e.g., Pixelate). However, as expected from \cref{subsec:hmt-invariance-and-elasticity}, there is a drop of around $3\%$ in the clean accuracy of the model. It is quite interesting that such a simple operation of low-complexity can boost the robustness significantly, which verifies the observation that redundant features can hurt robustness to distribution shifts~\cite{OrtizJimenezRedundantFeatures}. Nevertheless, this might be just a CIFAR-10 artefact: recall from \cref{sec:hmt-discriminative-features-of-real} that for other datasets like ImageNet, the classifiers are already invariant to high-frequency features, and hence a low-pass filtering might not have an effect similar to the one observed here. On the other hand, training on grayscale images seems to hurt the robustness of the classifier, except for the case of impulse noise and contrast. Hence, deploying such pre-processing method to achieve robustness to color changes is not beneficial.

From the results with data augmentation, it is evident that AugMix achieves good results, without really hurting the accuracy on the clean images. Interestingly, though, in many corruption types low-pass filtering achieves better robustness than AugMix. Hence, and in order to investigate any complementary gains between the two methods, we trained the network by combining AugMix with low-pass filtering as well. Such combination can further increase the overall robustness, achieving better results than simply applying AugMix or LP alone. Nevertheless, it seems that in some cases (i.e., glass blur) the influence of AugMix dominates, and it constrains some beneficial properties of LP.

Overall, from our analysis we can highlight two main insights. First, pre-processing can sometimes be beneficial, but some others can even be hurtful. Beyond the techniques we deployed, there are many other ways to pre-process the data and possibly confer robustness to some corruptions. However, the problem of common corruptions is ill-posed, and it is difficult to introduce pre-processing methods that can cover every possible corruption that might occur. Second, using data augmentation seems to be a sensible direction. Yet, the AugMix\texttt{+}LP experiment hints that existing augmentation practices, such as AugMix, might not properly cover the space of corruptions and can be further improved. Hence, it is important to design a more general data augmentation framework that effectively increases the coverage over the space of possible distortions.

\subsection{General model of visual corruptions}
\label{subsec:prime-model-of-visual-corruptions}

Motivated by the ``semantically-preserving'' nature of common corruptions, we define a new model of typical distortions. Specifically, we leverage the long tradition of image processing in developing techniques to manipulate images while retaining their semantics and construct a principled framework to characterize a large space of visual corruptions.

Let $\bm x:[0,1]^2\to[0,1]^3$ be a continuous image\footnote{We define our model of common corruptions in the continuous domain for simplicity. However, as is common in image processing, in practice we will work with discrete images on a regular grid.} mapping pixel coordinates $\bm r=(r_1, r_2)$ to RGB values. We define our model of common corruptions as the action on $\bm x$ of the following additive subgroup of the near-ring of transformations~\cite{near-ring}
\begin{equation}
    \mathcal{T}_{\bm x}=\left\{\sum_{i=1}^n \lambda_i\; g^i_1\circ \dots\circ g^i_m(\bm x): \:g^i_j\in\{\omega, \tau, \gamma\}, \lambda_i \in \R\right\},
\label{eq:prime-cc_model}
\end{equation}
where $\omega, \tau$ and $\gamma$ are random primitive transformations which distort $\bm x$ along the spectral ($\omega$), spatial ($\tau$), and color ($\gamma$) domains. As we will see, defining each of these primitives in a principled and coherent fashion will be enough to construct a set of perturbations which covers most types of visual corruptions.

To guarantee as much diversity as possible in our model, we follow the principle of maximum entropy to define our distributions of transformations~\cite{cover_info}. Note that using a set of augmentations that guarantees maximum entropy comes naturally when trying to optimize the sample complexity derived from certain information-theoretic generalization bounds, both in the clean~\cite{xuInfoBound} and corrupted settings~\cite{OODInfoBound}. Specifically, the principle of maximum entropy postulates favoring those distributions that are as unbiased as possible given the set of constraints that define a family of distributions. In our case, these constraints are given in the form of an expected strength $\sigma^2$, some boundary conditions, e.g., the displacement field must be zero at the borders of an image, and finally the desired smoothness level $K$. The principle of smoothness helps formalize the notion of physical plausibility, as most naturally occurring processes are smooth.

Formally, let $\mathcal{I}$ denote the space of all images, and let $f:\mathcal{I}\to \mathcal{I}$ be a random image transformation distributed according to the law $\mu$. Further, let us define a set of constraints $\mathcal{C}\subseteq \mathcal{F}$, which restricts the domain of applicability of $f$, i.e., $f\in\mathcal{C}$, and where $\mathcal{F}$ denotes the space of functions $\mathcal{I}\to \mathcal{I}$. The principle of maximum entropy postulates using the distribution $\mu$ which has maximum entropy given the constraints:
\begin{align}
    \underset{\mu}{\text{maximize}}\quad &  H(\mu)=-\int_{\mathcal{F}}\, \mathrm{d}\mu(f) \log(\mu(f)) \label{eq:prime-max_entropy}\\
    \text{subject to}\quad & f\in\mathcal C\quad \forall f\in\operatorname{supp}(\mu), \nonumber
\end{align}
where $H(\mu)$ represents the entropy of the distribution $\mu$~\cite{cover_info}. In its general form, solving \cref{eq:prime-max_entropy} for any set of constraints $\mathcal{C}$ is intractable. However, as we show in \cref{sec:app-ch6-max-entropy}, for the distributions of each of our family of transformations we can derive an analytical expression in closed form, by leveraging results from statistical physics~\cite{beale}.

In what follows, we describe the analytical solutions to \cref{eq:prime-max_entropy} for each of our basic primitives. In general, these distributions are governed by two parameters: $K$ to control smoothness, and $\sigma^2$ to control strength. These transformations fall back to identity mappings when $\sigma^2=0$, independently of $K$. 

\textbf{Spectral domain}\quad 
We parameterize the distribution of random spectral transformations using random filters $\bm\omega(\bm r)$, such that the output of the transformation follows
\begin{equation}
    \omega(\bm x)(\bm r)=\left(\bm x * \left(\bm\delta+\bm \omega'\right)\right)(\bm r),
\label{eq:prime-spectral_domain}
\end{equation}
where, $*$ is the convolution operator,  $\bm\delta(\bm r)$ represents a Dirac delta, i.e., identity filter, and $\bm\omega'(\bm r)$ is implemented in the discrete grid as a finite impulse response (FIR) filter of size $K_\omega\times K_\omega$ with i.i.d random entries distributed according to $\mathcal{N}(0, \sigma^2_\omega)$. Here, $\sigma^2_\omega$ governs the transformation strength, while larger $K_\omega$ yields filters of higher spectral resolution. The bias $\bm\delta(\bm r)$ makes the output close to the original image.

\textbf{Spatial domain}\quad 
We define our distribution of random spatial transformations, which apply random perturbations over the coordinates of an image, using the following model
\begin{equation}
    \tau(\bm x)(\bm r) = \bm x(\bm r + \bm\tau'(\bm r)).
\end{equation}
This model has been recently proposed by Petrini \etal~\cite{Petrini2021Diffeo} to define a distribution of random smooth diffeomorphisms in order to study the stability of neural networks to small spatial transformations. To guarantee smoothness but preserve maximum entropy, Petrini \etal propose to parameterize the vector field $\bm\tau'$ as
\begin{equation}
    \bm\tau'(\bm r)=\sum_{i^2+j^2\leq K^2_\tau}\beta_{i,j}\sin(\pi i \bm r_1)\sin(\pi j \bm r_2), \label{eq:prime-spatial_transform}
\end{equation}
where $\beta_{i,j}\sim\mathcal{N}(0,\sfrac{\sigma^2_\tau}{(i^2+j^2)})$. This choice of values guarantees that the resulting mapping is smooth according to the cut frequency $K_\tau$, while $\sigma^2_\tau$ determines its strength.

\paragraph{Color domain} We follow a similar approach to define the distribution of random color transformations. That is, we build random mappings $\gamma$ between color spaces such that
\begin{equation}
    \gamma(\bm x)(\bm r) = \bm x(\bm r) + \sum_{n=0}^{K_\gamma} \bm\beta_n \odot\sin\left(\pi n \,\bm x(\bm r)\right),
\label{eq:prime-color_transform}
\end{equation}
where $\bm{\beta}_n\sim\mathcal{N}(0, \sigma^2_\gamma\bm I_3)$, with $\odot$ denoting elementwise multiplication. Again, $K_\gamma$ controls the smoothness of the transformations and $\sigma^2_\gamma$ their strength. Note however that, compared to~\cref{eq:prime-spatial_transform}, the coefficients in~\cref{eq:prime-color_transform} are not weighted by the inverse of the frequency, and have constant variance. In practice, we observe that reducing the variance of the coefficients for higher frequencies creates color mappings that are too smooth and almost imperceptible, so we decided to drop this dependency in our model.

Finally, we note that our model is very flexible with respect to its core primitives. In particular, it can be easily extended to include other distributions of maximum entropy transformations that suit an objective task. For example, one might add the distribution of maximum entropy additive perturbations given by $\eta(\bm x)(\bm r)= \bm x(\bm r)+\bm\eta'(\bm r)$, where $\bm \eta'(\bm r)\sim\mathcal{N}(0, \sigma^2_\eta)$. Nonetheless, since most benchmarks of visual corruptions disallow the use of additive perturbations during training~\cite{Hendrycks2019Benchmarking}, our model does not include an additive perturbation category.

Overall, based on the results in \cref{subsec:prime-performance-common-corruptions,subsec:prime-importance-of-mixing}, our model is flexible and covers a large part of the semantic-preserving distortions. It also allows to easily control the strength and style of the transformations with just a few parameters. Moreover, changing the transformation strength enables to control the trade-off between corruption robustness and standard accuracy, as shown in \cref{subsec:prime-tradeoff}. In what follows, we use our model to design an efficient augmentation scheme to build classifiers robust to common corruptions.

\section{PRIME data augmentations}
\label{sec:prime-prime-data-augmentations}

\AlgoDontDisplayBlockMarkers
\RestyleAlgo{ruled}
\SetAlgoNoLine
\LinesNumbered
\begin{algorithm}[t]
 	\footnotesize
    \algnewcommand{\LeftComment}[1]{\Statex \(\triangleright\) #1}
 	\KwIn{Image $\bm x$, primitives $\mathcal{G}=\{\operatorname{Id},\omega, \tau\, \gamma\}$, where $\operatorname{Id}$ is the identity operator} 
 	\KwOut{Augmented image $\tilde{\bm x}$}
 	\BlankLine
    $\tilde{\bm x}_0 \gets \bm x$\\
	\For{$i\in \{1,\dots,n\}$}
	{
	    $\tilde{\bm x}_i \gets \bm x$\\
        \For{$j\in\{1,\dots,m\}$}
        {
            $g\sim\mathcal{U}(\mathcal{G})$ \Comment{Strength $\sigma\sim\mathcal{U}(\sigma_{\text{min}}, \sigma_{\text{max}})$} \\
            $\tilde{\bm x}_i \gets g(\tilde{\bm x}_i)$
        }
    }
    $\bm \lambda\sim\operatorname{Dir}(\bm 1)$ \Comment{Random Dirichlet convex coefficients}\\
    $\tilde{\bm x}\gets \sum_{i=0}^n \lambda_i \tilde{\bm x}_i$ \\
\caption{PRIME}
\label{alg:prime-prime}
\end{algorithm}

\subsection{Instantiating the general model of visual corruptions}
\label{subsec:prime-instantiating-the-general-model}

We now introduce PRIME, a simple yet efficient augmentation scheme that uses our \textbf{PRI}mitives of \textbf{M}aximum \textbf{E}ntropy to confer robustness against common corruptions. The pseudo-code of PRIME is given in~\cref{alg:prime-prime}, which draws a random sample from~\cref{eq:prime-cc_model} using a convex combination of a composition of basic primitives. Below we describe the main implementation details of our algorithm.

\begin{figure}[t]
    \centering
    \includegraphics[width=0.8\columnwidth]{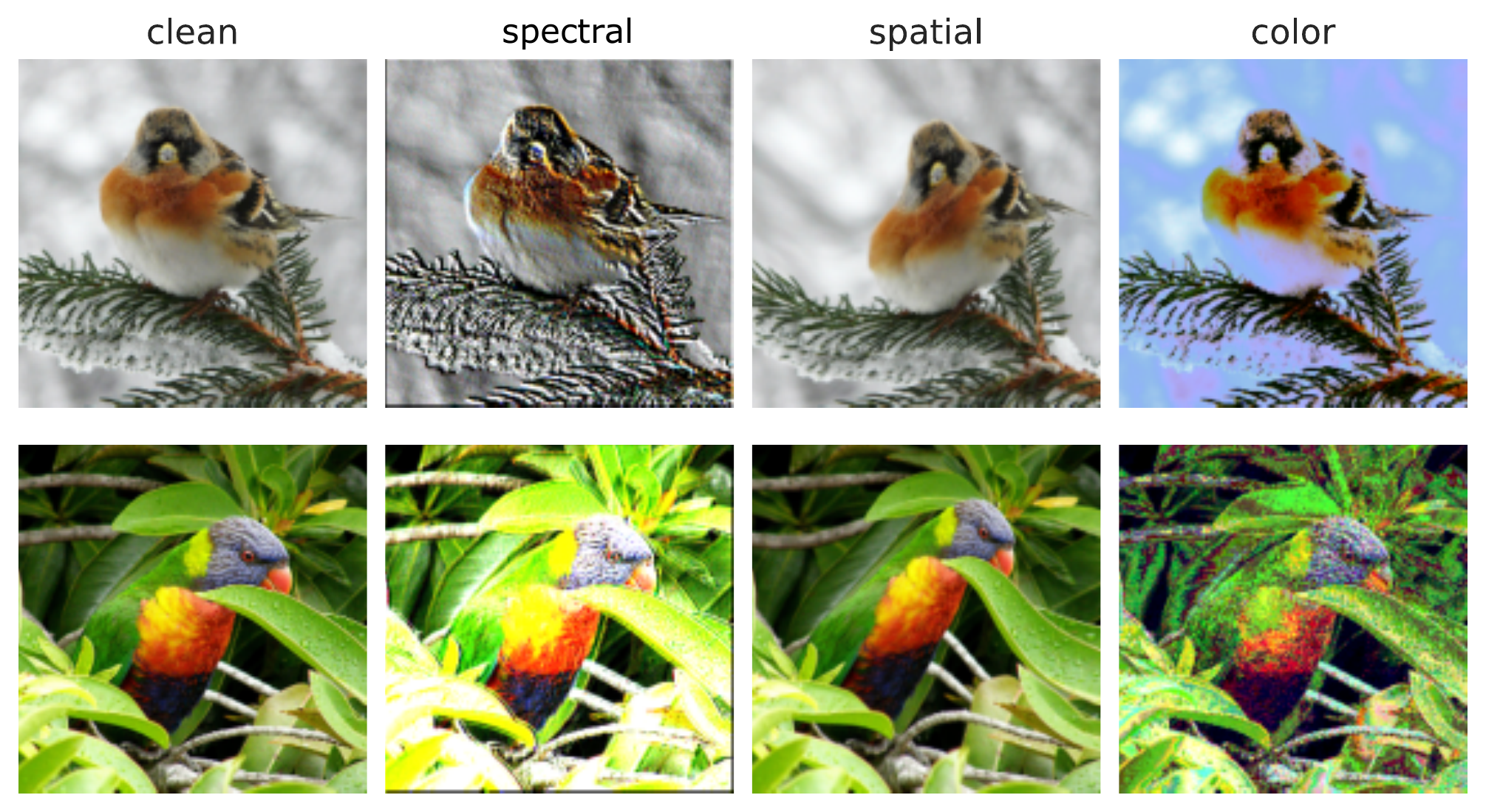}
    \caption{Example generated with the transformations of our common corruptions model. Despite the perceptibility of the distortion, the image semantics are preserved.}
    \label{fig:prime-visual_inspection}
\end{figure}

\textbf{Parameter selection}\quad
It is important to ensure that the semantic information of an image is preserved after it goes through PRIME. As measuring semantic preservation quantitatively is not simple, we subjectively select each primitive's parameters based on visual inspection, ensuring maximum permissible distortion while retaining the semantic content of the image\footnote{All the selected transformation parameters of PRIME are provided in \cref{sec:app-ch6-implementation_details}.}. To avoid relying on a specific strength for each transformation, PRIME stochastically generates augmentations of different strengths by sampling $\sigma$ from a uniform distribution, with different minimum and maximum values for each primitive. 

For the color primitive, we observed that fairly large values for $K_\gamma$ (in the order of $500$) are important for covering a large space of visual distortions. Unfortunately, implementing such a transformation can be memory inefficient. To avoid this issue, PRIME uses a slight modification of \cref{eq:prime-color_transform} and combines a fixed number $\Delta$ of consecutive frequencies randomly chosen in the range $[0, K_\gamma]$. Finally, as some of our transformations can push the images outside of their color range, we always clip the output of each transformation so that it lies on $[0,1]^3$.

\begin{figure}[t]
    \centering
    \includegraphics[width=0.85\columnwidth]{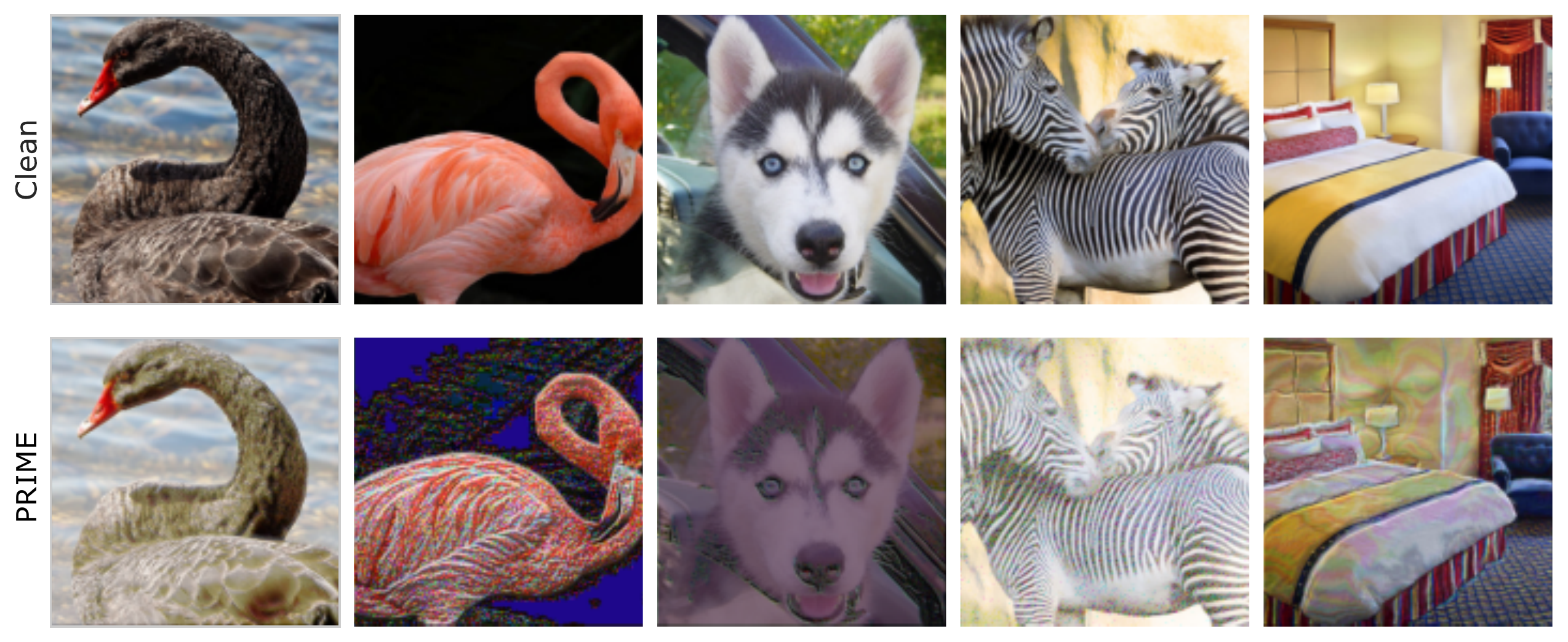}
    \caption{Images generated with PRIME, a simple method that uses primitive families of max-entropy transformations in different visual domains to create diverse data augmentations.}
    \label{fig:prime-prime_example}
\end{figure}

\textbf{Mixing transformations}\quad
The concept of mixing has been a recurring theme in the augmentation literature~\cite{Zhang2018Mixup, Yun2019CutMix, Hendrycks2020AugMix, Wang2021AugMax} and PRIME follows the same trend. In particular, \cref{alg:prime-prime} uses a convex combination of $n$ basic augmentations consisting of the composition of $m$ of our primitive transformations (see \cref{fig:prime-prime_example} for examples generated with PRIME). In our experiments, we fix the total number of generated transformed images (width) to be $n=3$. As for the composition of the transformations (depth), we follow a stochastic approach such that, on every iteration $i\in\{1,\dots,n\}$, only $\hat{m}\in[1,m]$ compositions are performed, with $m=3$. In fact, in \cref{alg:prime-prime} we do not explicitly select randomly a new $\hat{m}$ for every $i$ but we provide the identity operator $\operatorname{Id}$ instead. This guarantees that, in some cases, no transformation is performed.

In general, the convex mixing procedure (i) broadens the set of possible training augmentations, and (ii) ensures that the augmented image stay close to the original one. We later provide empirical results which underline the efficacy of mixing in~\cref{subsec:prime-importance-of-mixing}. Finally, note that, the basic skeleton of PRIME is similar to that of AugMix. However, as we will see next, incorporating our maximum entropy transformations leads to significant gains in common corruptions robustness over AugMix.

\subsection{Performance on common corruptions}
\label{subsec:prime-performance-common-corruptions}

\begin{table}[!t]
    \centering
    \small
    \begin{tabular}{clccc}
        \toprule
        \multirow{2}{*}{Dataset} & \multicolumn{1}{c}{\multirow{2}{*}{Method}} & Clean & \multicolumn{2}{c}{Common Corruption}\\
        & & Acc ($\uparrow$) & Acc ($\uparrow$) & mCE ($\downarrow$)\\
        \midrule
        \multirow{3}{*}{C-10} & Standard & 95.0 & 74.0 & 24.0\\
        & AugMix & 95.2 & 88.6 & 11.4\\
        & PRIME & 94.2 & \textbf{89.8} & \textbf{10.2}\\
        \midrule
        \multirow{3}{*}{C-100}& Standard & 76.7 & 51.9 & 48.1\\
        & AugMix & 78.2 & 64.9 & 35.1\\
        & PRIME & 78.4 & \textbf{68.2} & \textbf{31.8}\\
        \midrule
        \multirow{6}{*}{IN-100} & Standard & 88.0 & 49.7 & 100.0\\
        & AugMix & 88.7	& 60.7 & 79.1\\
        & DA & 86.3 & 67.7 & 68.1\\
        & PRIME & 85.9 & \textbf{71.6} & \textbf{61.0}\\
        \cmidrule{2-5}
        & DA\texttt{+}AugMix & 86.5 & 73.1 & 57.3\\
        & DA\texttt{+}PRIME & 84.9 & \textbf{74.9} & \textbf{54.6}\\
        \midrule
        \multirow{7}{*}{IN} & Standard$^*$ & 76.1 & 38.1 & 76.7\\
        & AugMix$^*$ & 77.5 & 48.3 & 65.3 \\
        & DA$^*$ & 76.7 & 52.6 & 60.4 \\
        & PRIME$^\dagger$ & 77.0 & \textbf{55.0} & \textbf{57.5}\\
        \cmidrule{2-5}
        & DA\texttt{+}AugMix & 75.8 & 58.1 & 53.6\\
        & DA\texttt{+}PRIME$^\dagger$ & 75.5 & \textbf{59.9} & \textbf{51.3}\\
        \bottomrule
    \end{tabular}
    \caption{Clean and corruption accuracy, and mean corruption error (mCE) for different methods with ResNet-18 on C-10, C-100, IN-100 and ResNet-50 on IN. mCE is the mean corruption error on common corruptions un-normalized for C-10 and C-100; normalized relative to standard model on IN-100 and IN. $^\dagger$ indicates that JSD consistency loss is not used. $^*$Models taken from~\protect\cite{robustbench2021}.}
\label{tab:prime-results-main}
\end{table}

We compare the classification performance of PRIME on the common corruption datasets~\cite{Hendrycks2019Benchmarking}, with that of two current approaches: AugMix~\cite{Hendrycks2020AugMix} and DeepAugment (DA)~\cite{Hendrycks2021TheManyFaces}. Regarding the training setup, we consider four datasets: CIFAR-10 (C-10), CIFAR-100 (C-100), ImageNet-100 (IN-100) and ImageNet (IN)~\cite{Deng2009ImageNet}. IN-100 is a $100$-class subset of IN obtained by selecting every $10$\textsuperscript{th} class in WordNet ID order. We train a ResNet-18~\cite{He2016ResNet} on C-10, C-100 and IN-100; and ResNet-50 on IN. Following AugMix, and for a complete comparison, we also integrate the Jensen-Shannon divergence (JSD)-based consistency loss with PRIME which compels the network to learn similar representations for differently augmented versions of the same input image. Note that all the models are trained for $100$ epochs.

Regarding the experimental details, all models are implemented in PyTorch~\cite{Paszke2019PyTorch} and are trained for $100$ epochs using a cyclic learning rate schedule with cosine annealing and a maximum learning rate of $0.2$ unless stated otherwise. For IN, we fine-tune a PyTorch regularly pretrained network with a maximum learning rate of $0.01$ following Hendrycks \etal~\cite{Hendrycks2021TheManyFaces}. We use SGD optimizer with momentum factor $0.9$ and Nesterov momentum. On C-10 \& C-100, we set the batch size to $128$ and use a weight decay of $0.0005$. On IN-100 and IN, the batch size is $256$ and weight decay is $0.0001$. We employ ResNet-18~\cite{He2016ResNet} on C-10, C-100 and IN-100; and use ResNet-50 for IN. The augmentation hyperparameters for AugMix and DeepAugment are the same as in their original implementations. 

We evaluate our models on the corrupted versions (C-10-C, C-100-C, IN-100-C, IN-C) of the aforementioned datasets, and the results are summarized in~\cref{tab:prime-results-main}. Amongst individual methods, PRIME yields superior results compared to those obtained by AugMix and DeepAugment alone and advances the baseline performance on the corrupted counterparts of the four datasets. As listed, PRIME pushes the corruption accuracy by $1.2\%$ and $3.3\%$ on C-10-C and C-100-C respectively over AugMix. 

On IN-100-C, a more complicated dataset, we observe significant improvements wherein PRIME outperforms AugMix by $10.9\%$. In fact, this increase in performance hints that our primitive transformations are actually able to cover a larger space of image corruptions, compared to the restricted set of AugMix. Interestingly, the random transformations in PRIME also lead to a $3.9\%$ boost in corruptions accuracy over DeepAugment despite the fact that DeepAugment leverages additional knowledge to augment the training data via its use of pre-trained architectures. Moreover, PRIME provides cumulative gains when combined with DeepAugment, reducing the mean corruption error (mCE) of prior art (DA\texttt{+}AugMix) by $2.7\%$ on IN-100-C.

Lastly, we evaluate the performance of PRIME on full IN-C, but we do not use JSD with PRIME to reduce the computational complexity. Yet, even without JSD, PRIME outperforms, in terms of corruption accuracy, both AugMix (with JSD) and DeepAugment by $6.7\%$ and $2.4\%$ respectively, while the mCE is reduced by $7.8\%$ and $2.9\%$. Also, when PRIME is combined with DeepAugment, it also surpasses the performance of DA\texttt{+}AugMix (with JSD), reaching a corruption accuracy of almost $60\%$ and an mCE of $51.3\%$.

Note here, that, not only PRIME achieves superior robustness, but it does so efficiently. Compared to standard training on IN-100, AugMix requires 1.20x time and PRIME requires 1.27x. In contrast, DA is tedious and we do not measure its runtime since it also requires the training of two large image-to-image networks for producing augmentations, and can only be applied offline.

Additionally, given the nuances amongst individual corruption types in common corruptions, we perform a fine-grained analysis with PRIME on IN-100-C to ensure that our method leads to general improvements against all corruption types. As evident from the comparison in \cref{tab:prime-results-in100}, PRIME alone, even without the JSD term, improves robustness over current techniques to almost every corruption type except blur. Further, incorporating the JSD term with PRIME attains the best results on all the corruption categories. Relative to the previous best by DeepAugment on IN-100-C, PRIME improves by $4.3\%$ on noises, $2\%$ on blurs, $3.4\%$ on weather changes and $5.8\%$ against digital distortions. This validates that PRIME helps against all common corruption types in IN-100-C, underlining the generality of our model of common corruptions.

\begin{table}[!t]
    \centering
    \small
    \begin{tabular}{lcccccc}
        \toprule
        {Method} & {IN-100-C} & {Noise} & {Blur} & {Weather} & {Digital} & {IN-100}\\
        \midrule
        AugMix$^\dagger$ & 55.2 & 38.9 & 56.8 & 57.0 & 64.2 & 88.0\\
        AugMix & 60.7 & 44.8	& 63.1 & 60.7 & 70.3 & \textbf{88.7} \\
        DA & 67.7 & 75.9 & 62.5 & 63.6 & 70.9 & 86.3\\
        PRIME$^\dagger$ & 68.8 & 78.8 & 58.3 & 66.0 & 74.8 & 87.1\\
        PRIME & \textbf{71.6} & \textbf{80.2} & \textbf{64.5} & \textbf{67.0} & \textbf{76.7} & 85.9\\
        \bottomrule
    \end{tabular}
    \caption{Classification accuracy ($\uparrow$) of various methods on the different corruption types contained in IN-100-C. $^\dagger$ indicates that JSD consistency loss is not used. Network used: ResNet-18.}
\label{tab:prime-results-in100}
\end{table}

\subsection{Unsupervised domain adaptation}
\label{subsec:prime-unsupervised-domain-adaptation}

Recently, robustness to common corruptions has also been of significant interest in the field of unsupervised domain adaptation~\cite{bnadapt_so2021, bnadapt_bethge2021}. The main difference is that, in domain adaptation, one exploits the limited access to test-time corrupted samples to adjust certain network parameters. Hence, it would be interesting to investigate the utility of PRIME under the setting of domain adaption.

To that end, we combine our method with the adaption trick by Schneider \etal~\cite{bnadapt_bethge2021}. Specifically, we adjust the batch normalization (BN) statistics of our models using a few corrupted samples. Suppose $z_s \in \{\mu_s$, $\sigma_s\}$ are the BN mean and variance estimated from the training data, and $z_t \in \{\mu_t$, $\sigma_t\}$ are the corresponding statistics computed from $n$ unlabelled, corrupted test samples, then we re-estimate the BN statistics as follows:
\begin{equation}
    \hat{z} = \frac{N}{N+n}z_s + \frac{n}{N+n}z_t
\end{equation}
We consider three adaptation scenarios: single sample ($n=1, N=16$), partial ($n=8, N=16$) and full ($n=400, N=0$) adaptation. Here, we do not perform parameter tuning for $N$. As shown in \cref{tab:prime-bn-adapt}, simply correcting BN statistics using as little as $8$ corrupted samples pushes the corruption accuracy of PRIME from $71.6\%$ to $75.3\%$. In general, PRIME yields cumulative gains in combination with adaptation and has the best IN-100-C accuracy.

\begin{table}[th]
    \centering
    \small
    \begin{tabular}{lccccc}
        \toprule
        \multicolumn{1}{c}{\multirow{3}{*}{Method}} & \multicolumn{4}{c}{IN-100-C accuracy ($\uparrow$)} & IN-100 ($\uparrow$)\\
        & w/o & single & partial & full & single\\
        & adapt & adapt & adapt & adapt & adapt\\
        \midrule
        Standard & 49.7 & 53.8 & 62.0 & 63.9 & 88.1\\
        AugMix & 60.7 & 65.5 & 71.3 & 73.0 & \textbf{88.3}\\
        DA & 67.7 & 70.2 & 72.7 & 74.6 & 86.3\\
        PRIME & \textbf{71.6} & \textbf{73.5} & \textbf{75.3} & \textbf{76.6} & 85.7\\
        \bottomrule
    \end{tabular}
    \caption{Performance when in concert with domain adaptation on IN-100. Partial adaptation uses $8$ samples; full adaptation uses $400$ corrupted samples. Network used: ResNet-18.}
    \label{tab:prime-bn-adapt}
\end{table}

\section{Robustness insights}
\label{sec:prime-robustness-insights}

In this section, we exploit the simplicity of PRIME to investigate different aspects behind robustness to common corruptions. We first analyze how each transformation domain contributes to the overall robustness of the network. Then, we empirically locate and justify the benefits of mixing the transformations of each domain. Moreover, we demonstrate the existence of a robustness-accuracy trade-off, and, finally, we comment on the low-complexity benefits of PRIME in different data augmentation settings. 

\subsection{Contribution of transformations}
\label{subsec:prime-contribution-transformations}

\begin{table}[!b]
    \centering
    \small
    \begin{tabular}{lcccccc}
        \toprule
        {Transform} & {IN-100-C} & {Noise} & {Blur} & {Weather} & {Digital} & {IN-100}\\
        \midrule
        None & 49.7 & 27.3 & 48.6 & 54.8 & 62.6 & \textbf{88.0}\\
        $\omega$ & 64.1 & 60.7 & 55.4 & 66.6 & 72.9 & 87.3\\
        $\tau$ & 53.8 & 30.1 & 56.2 & 57.6 & 65.4 & 87.0\\
        $\gamma$ & 59.9 & 67.4 & 52.6 & 54.4 & 67.1 & 86.9\\
        $\omega$\texttt{+}$\tau$ & 64.5 & 58.5 & 57.3 & \textbf{66.8} & 73.9 & 87.7\\
        $\omega$\texttt{+}$\gamma$ & 67.5 & 77.2 & 55.7 & 65.3 & 74.2 & 87.1\\
        $\tau$\texttt{+}$\gamma$ & 63.3 & 74.7 & 57.4 & 56.2 & 67.8 & 86.2\\
        $\omega$\texttt{+}$\tau$\texttt{+}$\gamma$ & \textbf{68.8} & \textbf{78.8} & \textbf{58.3} & 66.0 & \textbf{74.8} & 87.1\\
        \bottomrule
    \end{tabular}
    \caption{Impact of the different primitives of max-entropy ($\omega$: spectral, $\gamma$: color, $\tau$: spatial) from PRIME on common corruption accuracy ($\uparrow$). All the transformations are essential for the performance of PRIME. All models are trained without JSD loss. 
    Network used: ResNet-18.}
    \label{tab:prime-orthogonality-in100}
\end{table}

We want to understand how the transformations in each domain of~\cref{eq:prime-cc_model} contribute to the overall robustness. To that end, we conduct an ablation study on IN-100-C by training a ResNet-18 with the max-entropy transformations of PRIME individually or in combination. As shown in~\cref{tab:prime-orthogonality-in100}, spectral transformations mainly help against blur, weather and digital corruptions. Spatial operations also improve on blurs, but on elastic transforms as well (digital). On the contrary, color transformations excel on noises and certain high frequency digital distortions, \eg pixelate and JPEG artefacts, and have a minor effect on weather changes. Besides, incrementally combining these transformations lead to cumulative gains \eg spatial\texttt{+}color help on both noises and blurs. Yet, for obtaining the best results, the combination of all transformations is required. This means that each transformation increases the coverage over the space of possible distortions and the increase in robustness comes from their cumulative contribution.

\subsection{The importance of mixing}
\label{subsec:prime-importance-of-mixing}

\begin{figure}[t]
    \centering
    \begin{subfigure}[t]{0.495\columnwidth}
        \centering
        \small
        \includegraphics[width=\linewidth]{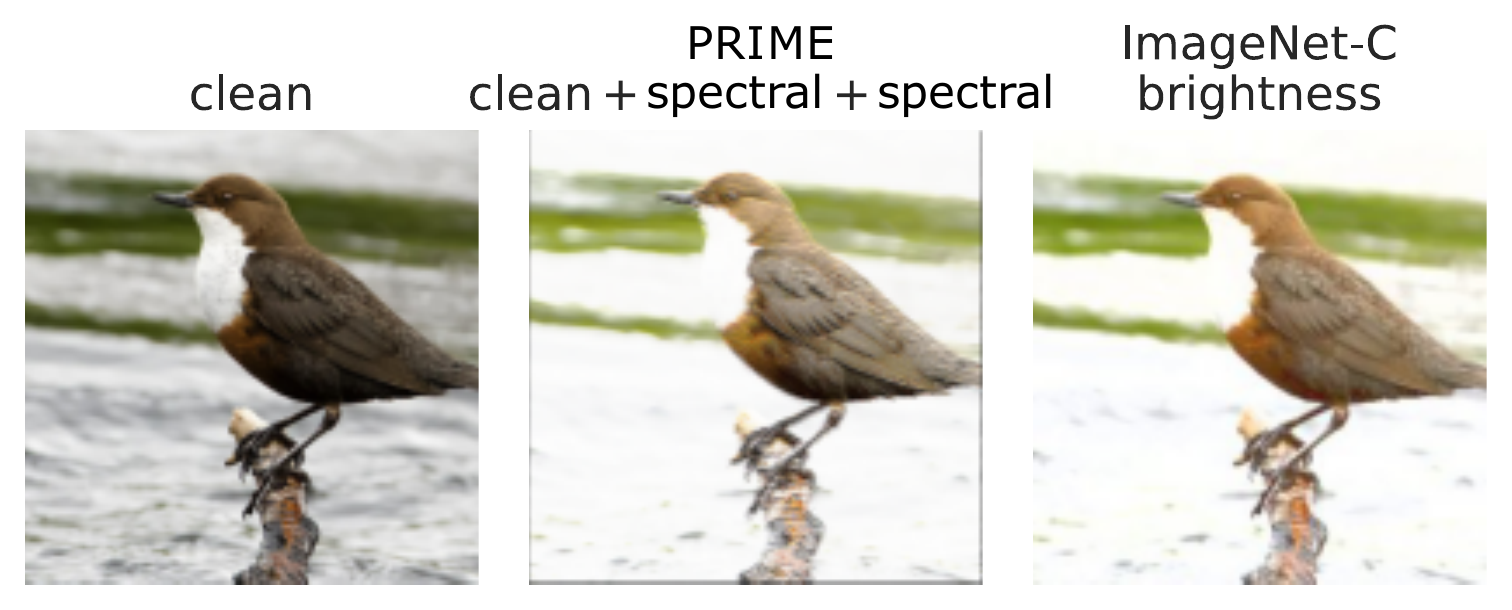}
        \captionsetup{justification=centering}
        \caption{$\texttt{clean+}\texttt{spectral}$ \\$\texttt{+spectral}\approx\texttt{brightness}$}
        \label{fig:prime-mixing-examples:a}
    \end{subfigure}\hfill
    \begin{subfigure}[t]{0.495\columnwidth}
        \centering
        \small
        \includegraphics[width=\linewidth]{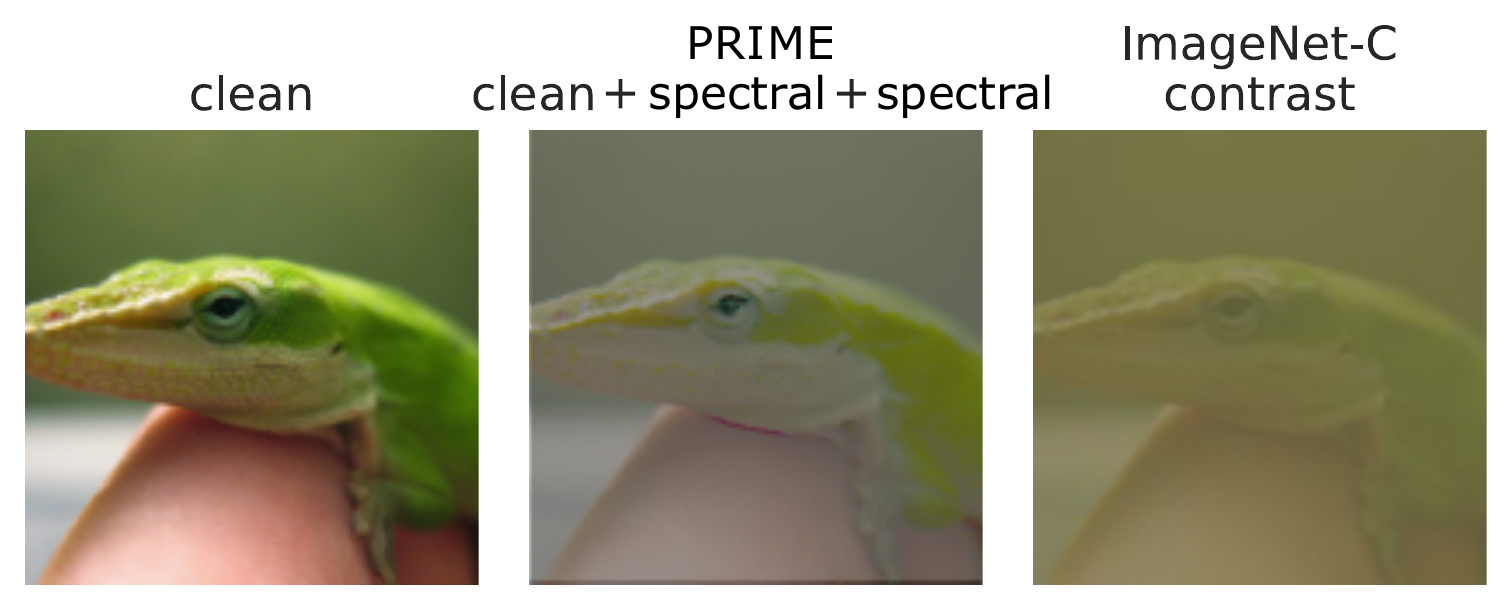}
        \captionsetup{justification=centering}
        \caption{$\texttt{clean+}\texttt{spectral}$ \\$\texttt{+spectral}\approx\texttt{contrast}$}
        \label{fig:prime-mixing-examples:b} 
    \end{subfigure}\hfill
    \begin{subfigure}[t]{0.495\columnwidth}
        \centering
        \includegraphics[width=\linewidth]{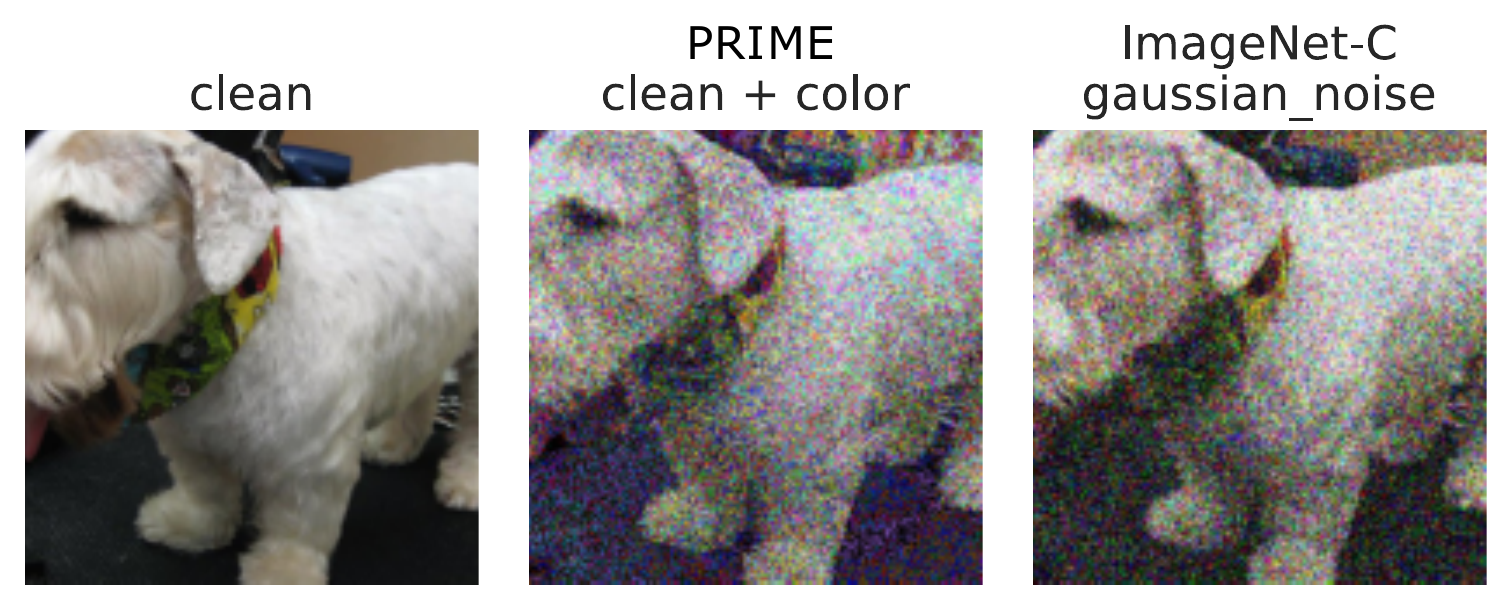}
        \small
        \captionsetup{justification=centering}
        \caption{$\texttt{clean+}\texttt{color}$
        \\$\approx\texttt{gaussian\_noise}$}
        \label{fig:prime-mixing-examples:d}
    \end{subfigure}\hfill
    \begin{subfigure}[t]{0.495\columnwidth}
        \small
        \centering
        \includegraphics[width=\linewidth]{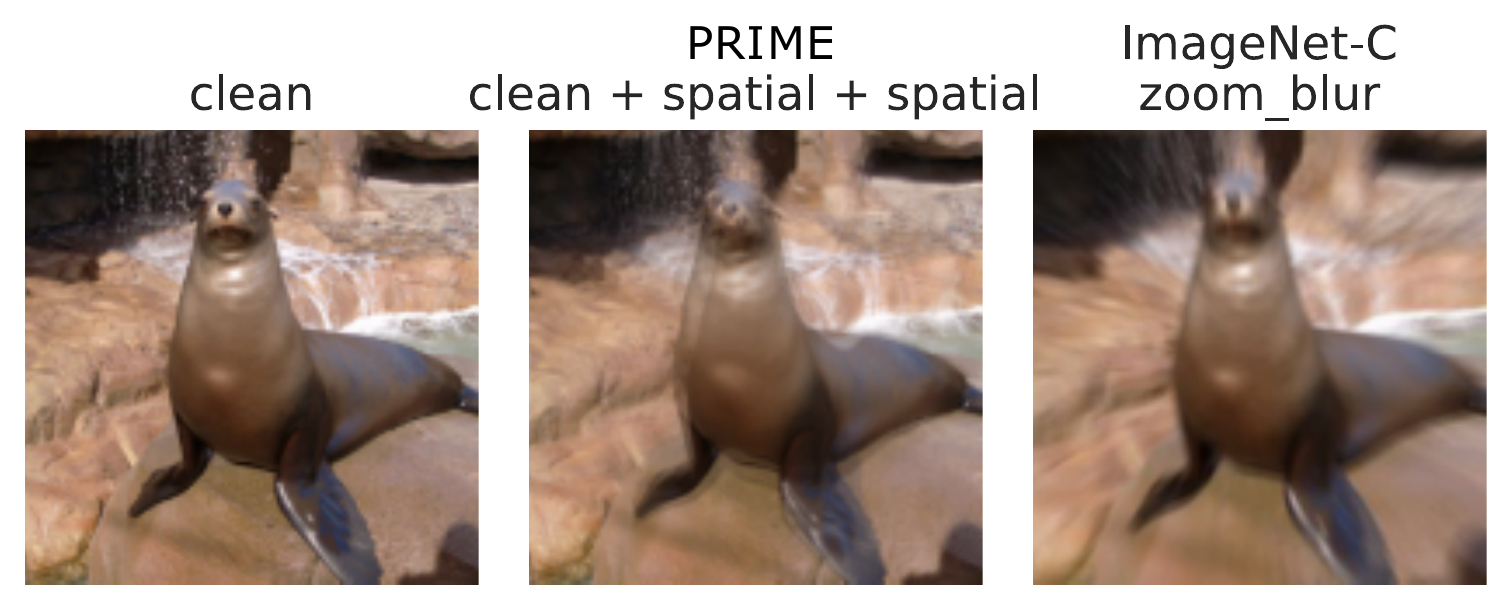}
        \captionsetup{justification=centering}
        \caption{$\texttt{clean+}\texttt{spatial+}\texttt{spatial}$
        \\$\approx\texttt{zoom\_blur}$}
        \label{fig:prime-mixing-examples:e}
    \end{subfigure}\hfill
    \caption{The mixing procedure creates distorted images that look visually similar to the test-time corruptions. In each example, we show the clean image, the PRIME image and the corresponding common corruption that resembles the image produced by mixing. We also report the mixing combination used for recreating the corruption. Additional examples can be found in App.~D of \cite{Modas2021PRIME}.}
    \label{fig:prime-mixing-examples}
\end{figure}

In most data augmentation methods, besides the importance of the transformations themselves, mixing has been claimed as an essential module for increasing diversity in the training process~\cite{Zhang2018Mixup, Yun2019CutMix, Hendrycks2020AugMix, Wang2021AugMax}. In our attempt to provide insights on the role of mixing in the context of common corruptions, we found out that it is capable of constructing augmented images that look perceptually similar to their corrupted counterparts. In fact, the improvements on specific corruption types observed in~\cref{tab:prime-orthogonality-in100} can be largely attributed to mixing. As exemplified in~\cref{fig:prime-mixing-examples:a,fig:prime-mixing-examples:b}, careful combinations of spectral transformations with the clean image introduce brightness and contrast-like artefacts that look similar to the corresponding corruptions in IN-C. Also, combining spatial transformations creates blur-like artefacts that look identical to zoom blur in IN-C (\cref{fig:prime-mixing-examples:e}). Finally, notice in~\cref{fig:prime-mixing-examples:d} how mixing color transformations helps fabricate corruptions of the ``noise'' category. This means that the max-entropy color model of PRIME enables robustness to different types of noise without explicitly adding any during training. This might explain the significant improvement over the ``noise'' category in~\cref{tab:prime-results-in100}.

Note that one of the main goals of data augmentation is to achieve maximum coverage of the space of possible distortions using a limited transformation budget, i.e., within a few training epochs. The principle of max-entropy guarantees this within each primitive, but the effect of mixing on the overall space is harder to quantify. In this regard, we can use the distance in the embedding space $\phi$ of a SimCLRv2~\cite{simclr} model, as a proxy for visual similarity~\cite{lpips,moayeri_sample_efficient_iccv}. We are interested in measuring how mixing the base transformations changes the likelihood that an augmentation scheme generates some sample during training that is visually similar to some of the common corruptions. To that end, we randomly select $N=1000$ training images $\{\bm x_n\}_{n=1}^N$ from IN, along with their $C=75$ ($15$ corruptions of $5$ severity levels) associated common corruptions $\{\hat{\bm x}_n^c\}_{c=1}^C$, and generate for each of the clean images another $T=100$ transformed samples $\{\tilde{\bm x}^t_n\}_{t=1}^T$ using each augmentation scheme. Moreover, for each corruption $\hat{\bm x}_n^c$ we find its closest neighbor $\tilde{\bm x}_n^t$ from the set of generated samples using the cosine distance in the embedding space\footnote{Examples of nearest neighbors can be found in App.~E of \cite{Modas2021PRIME}.}. Our overall measure of fitness is
\begin{equation}
    \cfrac{1}{NC} \sum_{n=1}^N\sum_{c=1}^C\min_{t}\left\{1-\left(\cfrac{\phi(\hat{\bm x}^c_n)^\top\phi(\tilde{\bm x}^t_n)}{\|\phi(\hat{\bm x}^c_n)\|_2~\|\phi(\tilde{\bm x}^t_n)\|_2}\right)\right\}.
\end{equation}


\Cref{tab:prime-simclr_distances} shows the values of this measure applied to AugMix and PRIME, with and without mixing. For reference, we also report the values of the clean (no transform) images $\{\bm x_n\}_{n=1}^N$. Clearly, mixing helps reducing the distance between the common corruptions and the augmented samples from both methods. We also observe that PRIME, even with only $100$ augmentations per image -- in the order of number of training epochs --  can generate samples which are twice as close to the common corruptions as AugMix. In fact, the feature similarity between training augmentations and test corruptions was also studied in~\cite{Mintun2021Cbar}, with an attempt to justify the good performance of AugMix on C-10. Yet, we see that the fundamental transformations of AugMix are not enough to span a broad space warranting high perceptual similarity to IN-C. The significant difference in terms of perceptual similarity in~\cref{tab:prime-simclr_distances} between AugMix and PRIME may explain the superior performance of PRIME on IN-100-C and IN-C (cf.~\cref{tab:prime-results-main}).

\begin{table}[t]
    \centering
    \small
    \aboverulesep=0ex
    \belowrulesep=0ex
    \begin{tabular}{lcc}
        \MyToprule{1-3}
        \rule{0pt}{1.1EM}
        \multirow{2}{*}{Method} & \multicolumn{2}{c}{Min. cosine distance \footnotesize{($\times10^{-3}$)}} \\
        \MyMidrule{2-3}
        & Avg. ($\downarrow$) & Median  ($\downarrow$) \\
        \midrule
        None (clean) & 25.38 & 6.44 \\
        \midrule
        AugMix (w/o mix) & 20.57 & 3.56 \\
        PRIME (w/o mix) & \textbf{10.61} & \textbf{1.88} \\
        \midrule
        AugMix & 17.48 & 2.61 \\
        PRIME & \textbf{~~7.71} & \textbf{1.61} \\
        \bottomrule
    \end{tabular}
    \caption{Minimum cosine distances in the ResNet-50 SimCLRv2 embedding space between $100$ augmented samples from $1000$ ImageNet images, and their corresponding common corruptions.}
    \label{tab:prime-simclr_distances}
\end{table}

\subsection{Robustness vs Accuracy trade-off}
\label{subsec:prime-tradeoff}

An important phenomenon observed in the literature of adversarial robustness is the so-called robustness-accuracy trade-off~\cite{Fawzi2018AnalysisOfClassifiers,Tsipras2019RobustnessOddsAccuracy,RaghunathanUnderstanding}, where technically adversarial training~\cite{Madry2018TowardsDeepLearning} with smaller perturbations (typically smaller $\varepsilon$) results in models with higher standard but lower adversarial accuracy, and vice versa. In this sense, we want to understand if the strength of the image transformations introduced through data augmentation can also cause such phenomenon in the context of robustness to common corruptions. As described in~\cref{subsec:prime-model-of-visual-corruptions}, each of the transformations of PRIME has a strength parameter $\sigma$, which can be seen as the analogue of $\varepsilon$ in adversarial robustness. Hence, we can easily reduce or increase the strength of the transformations by setting $\hat{\sigma} = \alpha\sigma$, where $\alpha\in\R^+$. Then, by training a network for different values of $\alpha$ we can monitor its accuracy on the clean and the corrupted datasets.

We train a ResNet-18 on C-10 and IN-100 using the setup of~\cref{subsec:prime-performance-common-corruptions}. For reducing complexity, we do not use the JSD loss and we train for $30$ epochs. This could cause some performance drop compared to the results of~\cref{tab:prime-results-main}, but we expect the overall trends in terms of accuracy and robustness to be preserved. Regarding the scaling of the parameters' strength, for C-10 we set $\alpha\in[10^{-3}, 10^2]$ and sample $100$ values spaced evenly on a log-scale, while for IN-100 we set $\alpha\in[10^{-2}, 10^2]$ and we sample $20$ values.

\begin{figure}[t]
\centering
    \begin{subfigure}[b]{0.47\linewidth}
    \centering
    \includegraphics[width=\linewidth]{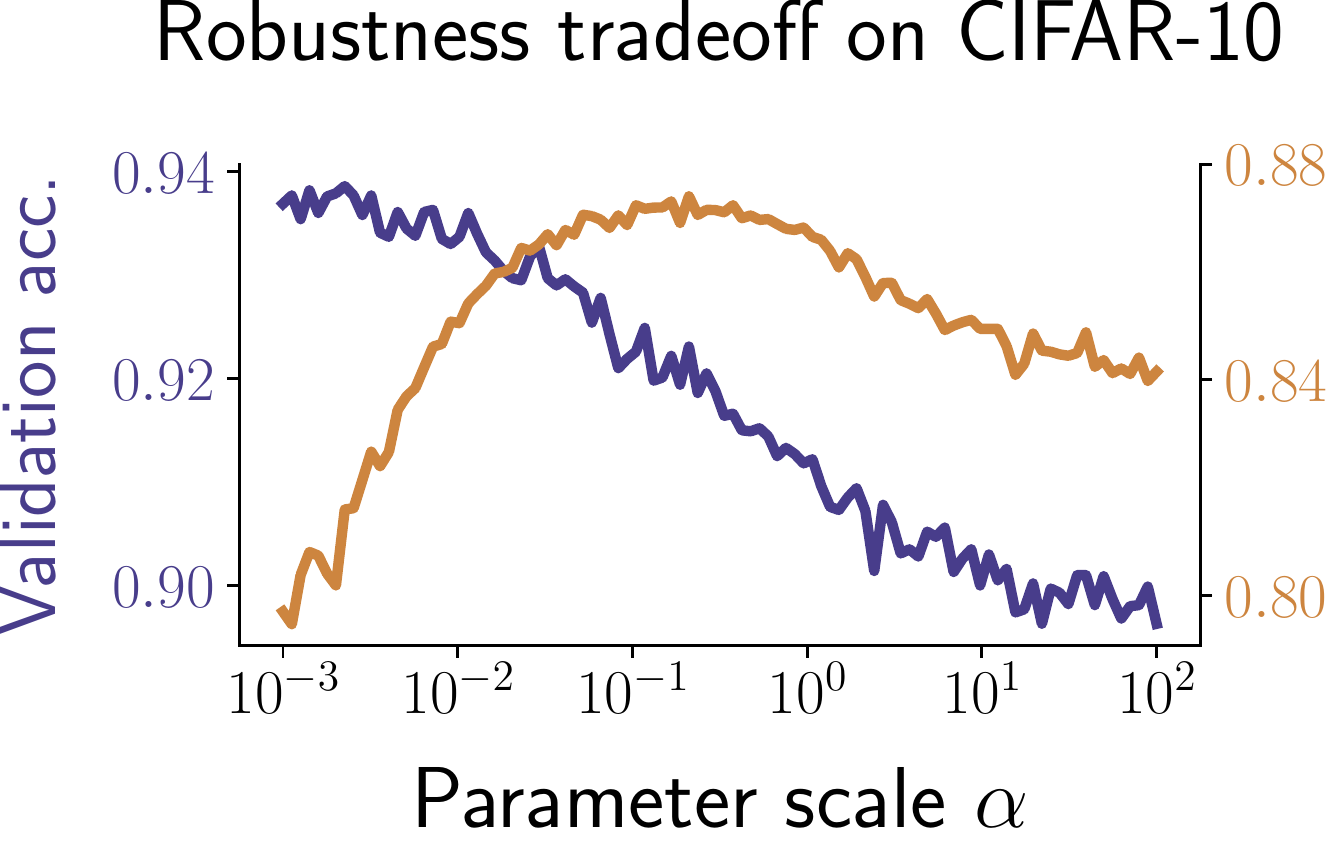} 
    \end{subfigure}\hfill%
    \begin{subfigure}[b]{0.462\linewidth}
    \centering
    \includegraphics[width=\linewidth]{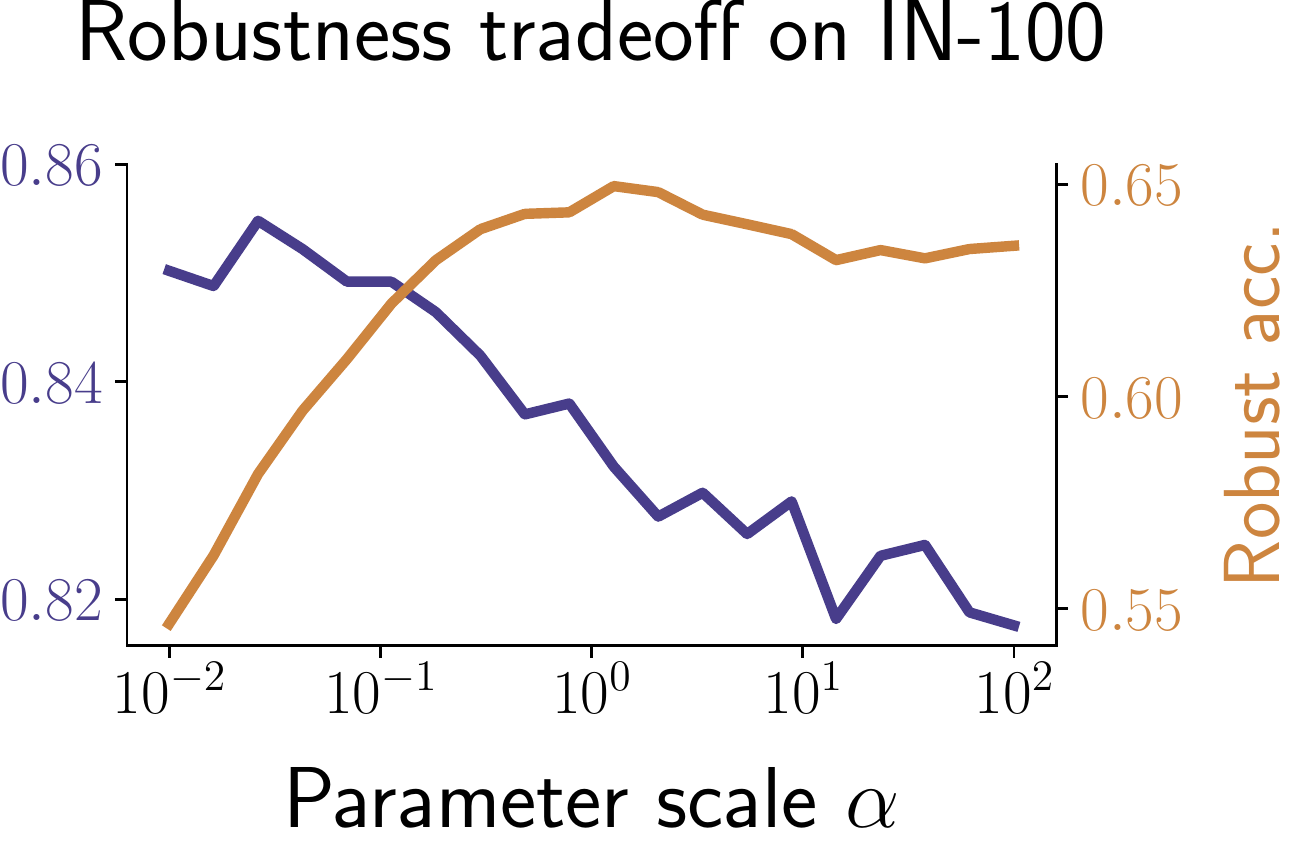}
    \end{subfigure} 
    \caption{Robustness vs. accuracy of a ResNet-18 (w/o JSD) on CIFAR-10 (left) and ImageNet-100 (right), when trained multiple times with PRIME. For each training instance, the transformation strength is scaled by $\alpha$. Note the different scale in axes.}
    \label{fig:prime-rob_acc}
\end{figure}

The results are presented in~\cref{fig:prime-rob_acc}. For both C-10 and IN-100, there is a sweet spot for the scale around $\alpha=0.2$ and $\alpha=1$ respectively, where the accuracy on common corruptions reaches its maximum. For $\alpha$ smaller than these values, we observe a clear trade-off between validation and robust accuracy. While the robustness to common corruptions increases, the validation accuracy decays. However, for $\alpha$ greater than the sweet-spot values, we observe that the trade-off ceases to exist since both the validation and robust accuracy present similar behaviour (slight decay). In fact, these observations indicate that robust and validation accuracies are not always positively correlated, and that one might have to slightly sacrifice validation accuracy in order to achieve robustness.

\subsection{Sample complexity}
\label{subsec:prime-sample-complexity}

Finally, we investigate the necessity of performing augmentation during training (on-line augmentation), compared to statically augmenting the dataset before training (off-line augmentation). On the one hand, on-line augmentation is useful when the dataset is huge and storing augmented versions requires a lot of memory. Besides, there are cases where offline augmentation is not feasible as it relies on pre-trained or generative models which are unavailable in certain scenarios, \eg DeepAugment~\cite{Hendrycks2021TheManyFaces} or AdA~\cite{Calian2022DefendingImageCorruptions} cannot be applied on C-100. On the other hand, off-line augmentation may be necessary to avoid the computational cost of generating augmentations during training.

To this end, for each of the C-10 and IN-100 training sets, we augment them off-line with $k=1,2,\dots,10$ i.i.d. PRIME transformed versions. Afterwards, for different values of $k$, we train a ResNet-18 on the corresponding augmented dataset and report the accuracy on the validation set and the common corruptions. For the training setup, we follow the settings of~\cref{subsec:prime-performance-common-corruptions}, but without JSD loss. Also, since we increase the size of the training set by $(k+1)$, we also divide the number of training epochs by the same factor, in order to keep the same overall number of gradient updates.

\begin{figure}[t]
\centering
    \begin{subfigure}[b]{0.40\columnwidth}
    \centering
    \includegraphics[width=\columnwidth]{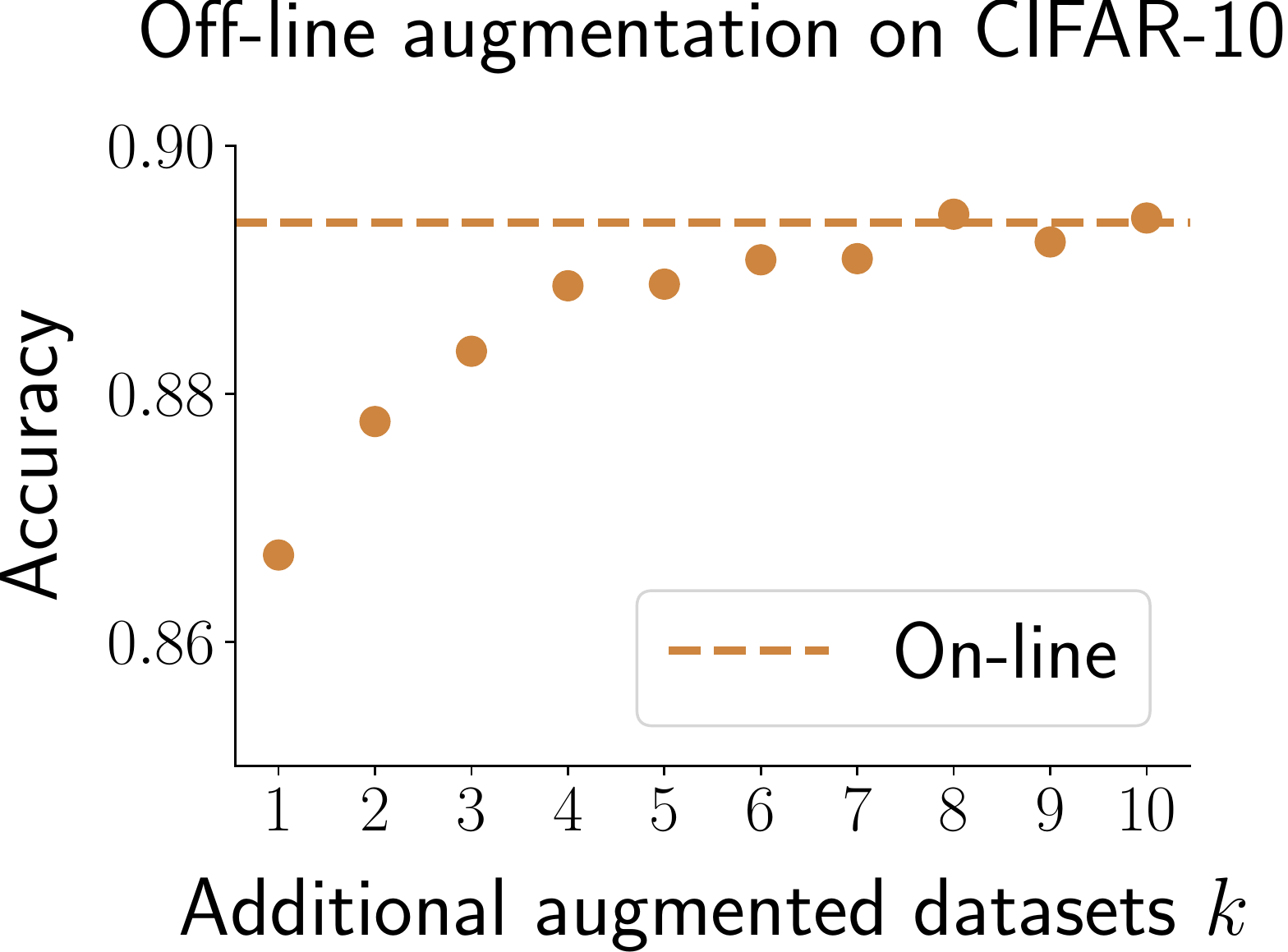}
    \label{fig:prime-acc_vs_aug_cifar} 
    \end{subfigure}\!
    \begin{subfigure}[b]{0.385\columnwidth}
    \centering
    \includegraphics[width=\columnwidth]{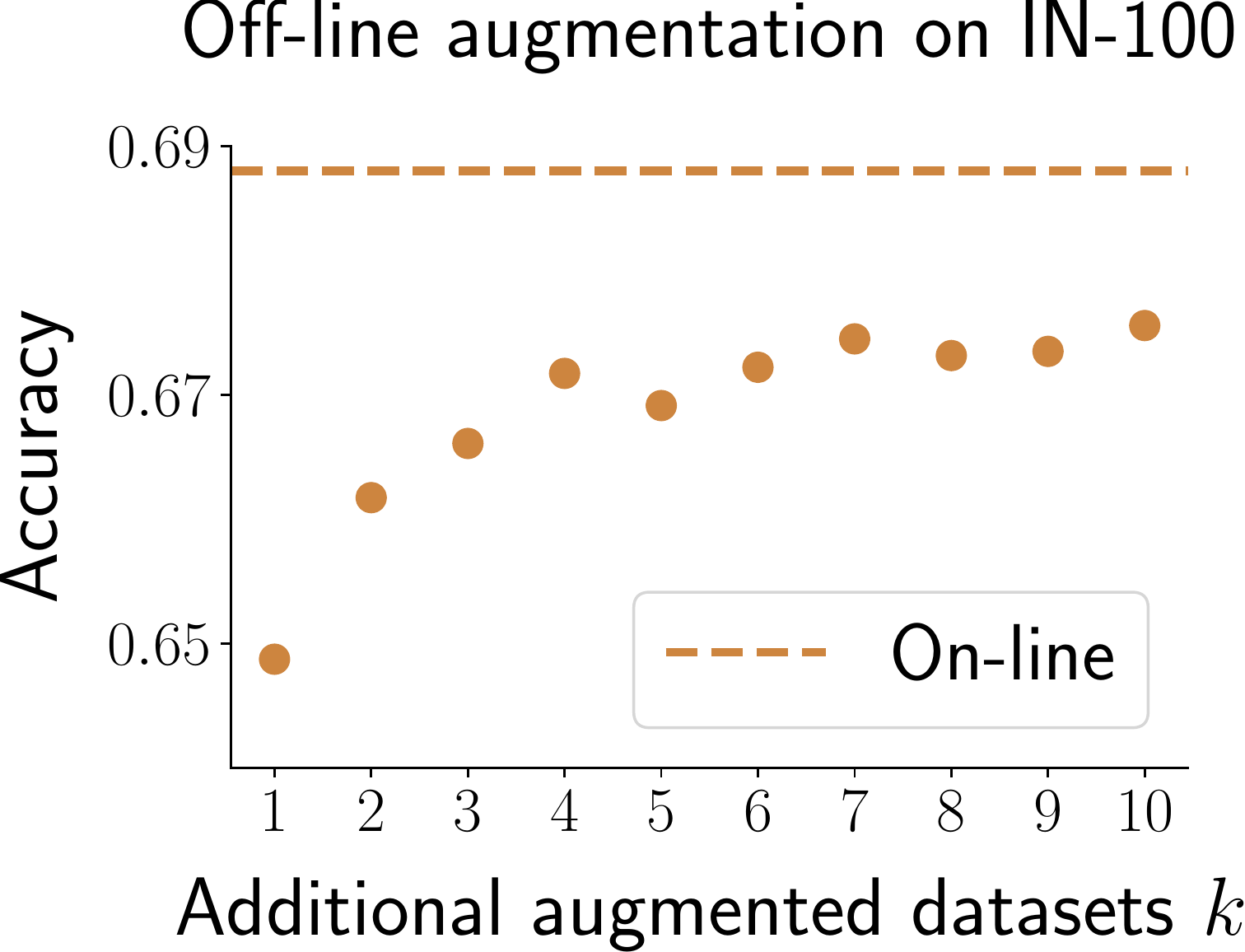}
    \label{fig:prime-acc_vs_aug_imagenet} 
    \end{subfigure} 
    \caption{Accuracy of ResNet-18 (w/o JSD) on CIFAR-10 (left) and ImageNet-100 (right) when augmenting the training sets with additional PRIME counterparts off-line. Dashed lines denote the accuracy when training under the same setup, but generating the augmentations during training (on-line). Validation accuracy is omitted because it is rather constant: around $93.4\%$ for CIFAR-10 and around $87\%$ for ImageNet-100.}
    \label{fig:prime-acc_vs_aug}
\end{figure}

The performance on common corruptions is presented in~\cref{fig:prime-acc_vs_aug}. The first thing to notice is that, even for $k=1$, the obtained robustness to common corruptions is already quite good. In fact, for IN-100 the accuracy ($65\%$) is already better than the best achievable result of on-line AugMix  ($60.7\%$ with JSD loss cf.~\cref{tab:prime-results-in100}). Regarding C-10 we observe that for $k=4$ the actual difference with respect to the on-line augmentation is almost negligible ($88.8\%$ vs. $89.3\%$), especially considering the overhead of transforming the data at every epoch. Technically, this means that augmenting C-10 with $4$ PRIME counterparts is enough for achieving good robustness to common corruptions. 

Finally, we also see in~\cref{fig:prime-acc_vs_aug} that the corruption accuracy on IN-100 presents a very slow improvement after $k=4$. Comparing the accuracy at this point ($67.2\%$) to the one obtained with on-line augmentation and without JSD ($68.8\%$ cf.~\cref{tab:prime-results-in100}) we observe a gap of $1.6\%$. Hence, given the cost of on-line augmentation on such large scale datasets, simply augmenting the training with $4$ extra PRIME samples presents a good compromise for achieving competitive robustness. Nevertheless, the increase of $1.6\%$ introduced by on-line augmentation is rather significant, hinting that generating transformed samples during training might be necessary for maximizing performance. In this regard, the lower computational complexity of PRIME allows it to easily achieve this $+1.6\%$ through on-line augmentation, since it only requires $1.27\times$ additional training time compared to standard training, and only $1.06\times$ compared to AugMix, but with much better performance. This can be a significant advantage with respect to complex methods, like DeepAugment, that cannot be even applied on-line (require heavy pretraining).

\section{Improving filling level classification with PRIME}
\label{sec:corsmal-prime}

So far we demonstrated that data augmentation with PRIME improves the robustness of classifiers to common corruptions of the data. In this section, we will see that PRIME can also be efficiently applied to off-the-self tasks, and improve robustness to other types of distribution shifts. In particular, we will focus again on the problem of estimating the filling level of a container introduced in \cref{sec:hmt-icip}, and we will show that PRIME can be easily tuned to effectively replace current transfer learning approaches, since it can significantly improve the generalization of the classifier.

The main limitation of transfer learning is that it requires the overhead of pre-training large models on very big datasets, while this overhead can further explode if these models are trained adversarially~\cite{salman2020adversarially,utrera2020adversariallytrained}. At the same time, there is actually no guarantee that the transferred features are relevant for the target task, while the exact reason why transfer learning is expected to work is rather obscure. An alternative for increasing the variability of the training data is to perform data augmentation. However, generating additional samples that resemble the previously unseen properties of the test-time data can be quite challenging, since imposing such properties (i.e., changing shape) might require more complex and sophisticated operations, i.e, composition of transformations or mixing strategies~\cite{Hendrycks2020AugMix}

In what follows, we will show that data augmentation using PRIME can improve significantly improve the filling level classification. We will demonstrate that PRIME can be easily tuned to generate augmentations that are tailored for learning classifiers that generalize on test-time images with containers of previously unseen shape, color, and spectral content. Using our scheme, we prevent the underlying classifier from overfitting to undesired features, achieving a filling level classification accuracy that is on-par or better than the one obtained with transfer learning. Yet, through a constructive approach and without the need of pre-training on large datasets. Finally, we also show that the performance of the classifier may further increase when our data augmentation scheme is used in concert with transfer learning itself.

\subsection{Distribution shifts and PRIME augmentations}
\label{subsec:corsmal-prime-overfitting-and-distribution-shifts}

Recall from \cref{sec:hmt-icip} that in an attempt to empirically understand the dataset features that the classifier relies on, we introduced C-CCM, a dataset of cups and glasses that can be empty or filled with water, pasta or rice. The containers can vary a lot in terms of shape and transparency, while they can be captured under different illuminations, backgrounds and occlusions. We further defined three different training and validation splits, where a distribution shift is introduced in the validation set, such that some validation containers always have some property that does not exist in the training set (i.e., a unique shape, a stem, or color).

By measuring the network performance, we observed that the accuracy on the ``shifted'' containers was systematically lower than the one on containers that share similarities with those in the training set (overfitting). This is probably due to some biases related to the shape, color, and spectral content of the images, and that, with transfer learning, the target model becomes more invariant to features related to these properties. When the source model is also adversarially trained, then its stronger invariance to unnecessary features~\cite{OrtizJimenezRedundantFeatures,OrtizJimenez2020HoldMeTight} enables the target model to avoid irrelevant correlations and to rather identify features related to the actual task (i.e., filling level rather than shape).

To avoid both the computational cost of training robust models and the obscurity of transfer learning, it is worth investigating if the variability of the training set can be increased through data augmentation. When performing data augmentation, the choice of the basic transformations used to compose the augmentations is very important. They must be general and diverse enough to cover the desired changes to be induced. In the context of filling level classification, and for the overfitting phenomena discussed in \cref{subsec:corsmal-prime-overfitting-and-distribution-shifts}, one should seek for image transformations that are sufficient to, i.e., change the shape of the container or its color/frequency content.

To this end, PRIME can generate diverse augmentations using a set of primitive max-entropy transformations on the spatial ($\tau$), color ($\gamma$) and spectral ($\omega$) domain. For filling level classification, we expect these transformations to relate to the changes we want to introduce during training: container shape through $\tau$, container color through $\gamma$ , and  illumination and texture through $\omega$.

\begin{figure}[t!]
    \centering
    \includegraphics[width=0.6\columnwidth]{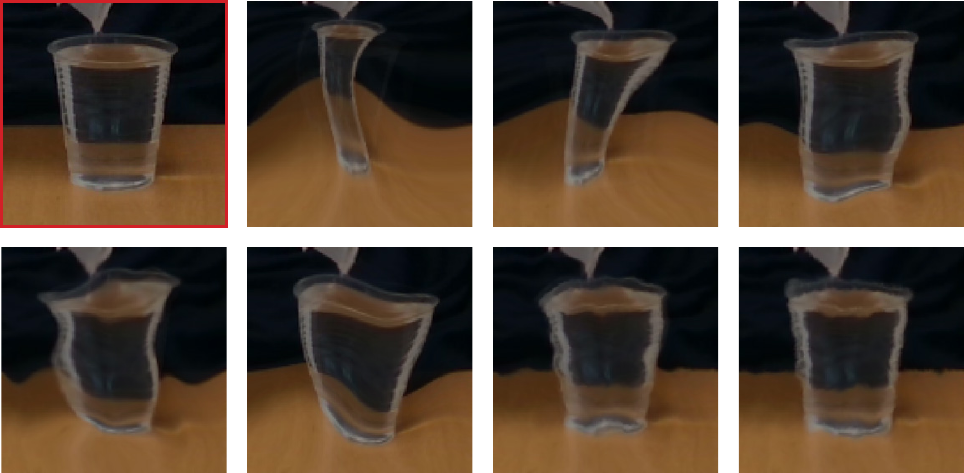}
    \caption{Sample transformed images  using  diffeomorphisms with varying smoothness level  $K_\tau$ (top row, from left to right: original image, transformed image with $K_\tau=2, 5, 10$; bottom row: $K_\tau=20, 40, 100, 300$).}
    \label{fig:corsmal-prime-diffeo_smoothnes}
\end{figure}

Recall from \cref{sec:prime-prime-data-augmentations} that each RIME transformation has two control parameters: $K$ for the smoothness and $\sigma^2$ for the strength. PRIME synthesizes a transformed image $\tilde{\bm x}$ through a convex combination of $n$ basic augmentations (width) consisting of the composition of $m$ of its max-entropy transformations (depth). We make a small modification on the last step of \cref{alg:prime-prime}, such that the final image $\hat{\bm x}$ is synthesized as a linear combination (mixing) of the original image $\bm x$ and the transformed image $\tilde{\bm x} = \sum_{i=1}^n \lambda_i \tilde{\bm x}_i$, with the coefficients of the linear combination drawn from a $\mathrm{Beta}$ distribution
\begin{equation}
    \hat{\bm x} = (1-p)\;\bm x + p\;\tilde{\bm x} \quad \text{with} \quad p\sim\mathrm{Beta}(\alpha,\beta).
\label{eq:corsmal-prime-mixing}
\end{equation}
Note that, for the shape parameters $\alpha,\beta$ of the $\mathrm{Beta}$ distribution, when $\alpha>\beta$ more importance is given to the pixels of the transformed image $\tilde{\bm x}$, while when $\alpha<\beta$ more importance is given to the pixels of the original image $\bm x$.

In the next section, we will investigate if transfer learning can be replaced by a more controlled data augmentation strategy tailored for filling level classification. We will explore how to tune PRIME parameters for tackling the dataset-specific distribution shifts and improve the performance of the network on estimating the filling level.

\subsection{PRIME transformation parameters}
\label{subsec:corsmal-prime-transformation-parameters}

\begin{figure}[t!]
    \centering
    \includegraphics[width=0.6\columnwidth]{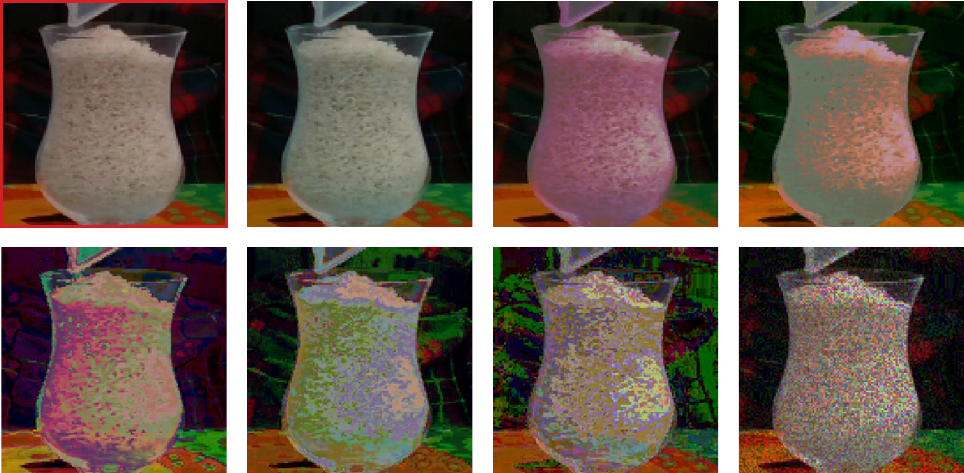}
    \caption{Sample transformed images  using color jittering with varying smoothness level  $K_\gamma$ (top row, from left to right: original image, transformed image with $K_\gamma=2, 5, 10$; bottom row: $K_\gamma=20, 40, 100, 300$).}
    \label{fig:corsmal-prime-color_smoothnes}
\end{figure}
\begin{figure}[t!]
    \centering
    \includegraphics[width=0.6\columnwidth]{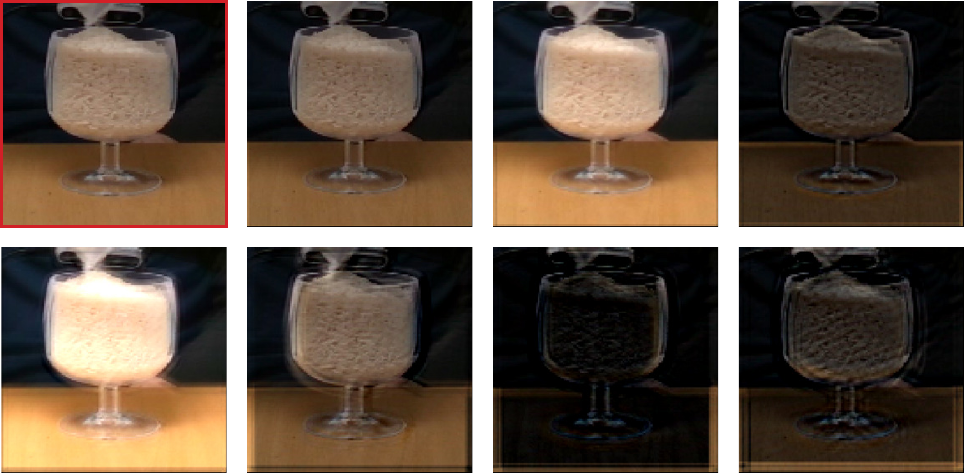}
    \caption{Sample transformed images  using spectral filtering with varying kernel size (smoothness level)  $K_\omega \times K_\omega$ (top row, from left to right: original image, transformed image with $K_\omega=3, 5, 7$; bottom row: $K_\omega=9, 11, 13, 15$).}
    \label{fig:corsmal-prime-spectral_smoothnes}
\end{figure}

We must first identify the proper parameter values, given the dataset-specific shifts that arise for each validation split. We focus first on the splits S$_1$ and S$_2$ of the C-CCM dataset, where the shifts are mostly related to the shape of the containers. Hence, the most relevant transformation in PRIME is the one in the spatial domain. Intuitively, we would like to enforce smooth, yet strong, diffeomorphisms that are able to alter the shape of the whole container so it becomes as narrow as a champagne flute, or just a part of it so it resembles the stem of a cocktail glass (see \cref{fig:corsmal-prime-diffeo_smoothnes}). Recall, that, for a fixed value of smoothness $K_\tau$ the authors in~\cite{Petrini2021Diffeo} propose to randomly sample the strength $\sigma_\tau^2$ from a specific interval, such that the resulting diffeomorphism remains bijective. In practice, for smaller values of $K_\tau$ (smoother), larger values of $\sigma_\tau^2$ are allowed to be sampled. Hence, we decided to set $K_\tau=10$ and let $\sigma_\tau^2$ to be properly sampled during training. In practice we observed that $K\tau\in[10,20]$ still leads to good results.

For the split S$_3$, since we mostly deal with shifts related to the color and frequency content of the containers (i.e., red and green glass, which can lead to different textures and reflections), we focus on the color (see \cref{fig:corsmal-prime-color_smoothnes}) and spectral (see \cref{fig:corsmal-prime-spectral_smoothnes}) transforms in PRIME. For the smoothness parameter $K$, we keep the values proposed in \cref{subsec:prime-instantiating-the-general-model}: $K_\gamma=500$ and $K_\omega=3$ for the color and the spectral domain respectively. As for the parameter strength, since very strong changes could potentially destroy the information in the images, we decide to only slightly manipulate the color of the pixels and the frequency information of the images, and hence we set $\sigma_\gamma^2=0.001$ and $\sigma_\omega^2=0.01$ respectively.

\subsection{Mixing parameters in PRIME}
\label{subsec:corsmal-prime-mixing-parameters}

Regarding the mixing strategy, the width $n$ specifies the number of transformed instances to be used in the convex combination for synthesizing the transformed image $\tilde{\bm x}$. It is reasonable to assume that we must set $n>1$ in order to increase the diversity of the generated transformed instances, and hence we decide to use the default value ($n=3$). 

\begin{table}[t]
    \centering
    \small
    \aboverulesep=0ex
    \belowrulesep=0ex
    \begin{tabular}{cccc}
        \rule{0pt}{1.1EM}
        $m$ & $\text{S}_1$ &  $\text{S}_2$ & $\text{S}_3$\\
        \midrule
        $1$ & 82.69 & \textbf{73.42} & 67.91 \\
        \midrule
        $2$ & 83.16 & 70.95 & 65.90 \\
        \midrule
        $3$ & \textbf{84.93} & 68.92 & \textbf{75.03}  \\
        \bottomrule
    \end{tabular}
\caption{Validation accuracy of a ResNet-18 on the three C-CCM dataset splits ($\text{S}_1, \text{S}_2, \text{S}_3$), when the composition depth $m$ of PRIME increases. Note that the transformation width is fixed to $n=3$.}
\label{tab:corsmal-prime-depth_analysis}
\end{table}

The depth $m$ specifies how many transformations will be sequentially applied on an image. In general, it is not always possible to determine what will be the exact outcome of such composition, and its impact on the overall performance. Hence, we let the mixing coefficient $p$ of \cref{eq:corsmal-prime-mixing} to be uniformly sampled ($\alpha=\beta=1$) and perform a sensitivity analysis on the values of $m$ (see \cref{sec:corsmal-prime-experimental-validation} for training details). The performance of a ResNet-18 on each dataset split is shown in \cref{tab:corsmal-prime-depth_analysis}. We observe that for S$_1$ and S$_3$ increasing the depth significantly improves the performance. For S$_2$ though we observe the opposite effect, indicating that applying multiple transformations on the image degrades some important information.

\begin{figure}[t]
\centering
\includegraphics[width=0.75\columnwidth]{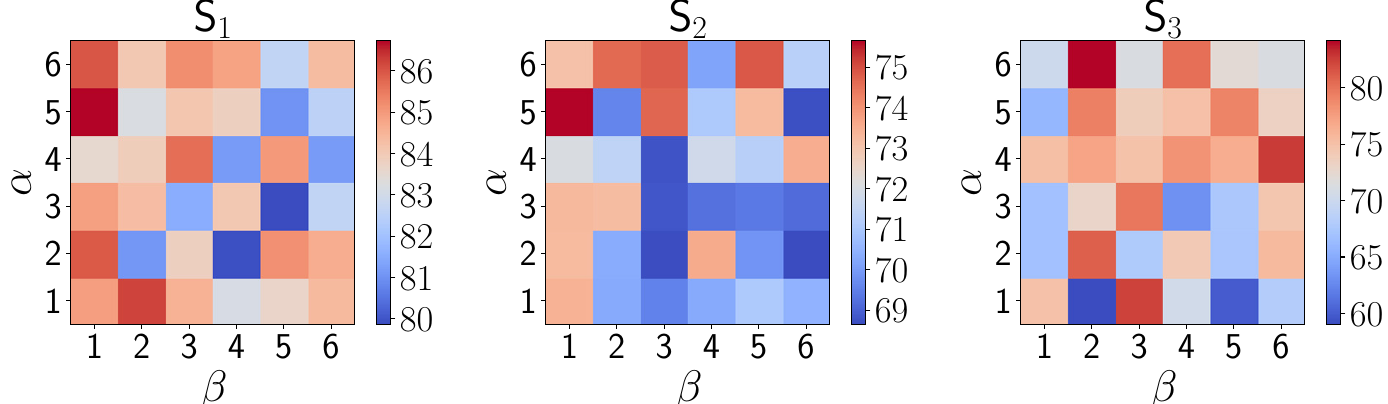}
\caption{Effect of the $\mathrm{Beta}(\alpha, \beta)$ distribution on the validation accuracy of a ResNet-18, trained with PRIME, on each split of C-CCM. Note that, during the mixing step of PRIME, $\alpha>\beta$ imposes more importance to the pixels of the transformed image, while $\alpha<\beta$ to the pixels of the original image.}
\label{fig:corsmal-prime-sensitivity_beta}
\end{figure}

Finally, for the best performing values of $m$ in \cref{tab:corsmal-prime-depth_analysis}, we explore the effect of the mixing coefficient $p$ in equation \cref{eq:corsmal-prime-mixing}. Specifically, we focus on the $\mathrm{Beta}$ distribution parameters, which control the relative importance of the pixels of $\bm x$ or $\tilde{\bm x}$. Intuitively, since the classifier overfits to the training data we expect that more importance on $\hat{\bm x}$ is necessary. To that end, we perform a sensitivity analysis on $\alpha$ and $\beta$, by measuring the performance of the network on their different combinations. Based on the results in \cref{fig:corsmal-prime-sensitivity_beta}, our initial intuition was right: on every dataset split, the highest validation accuracy is achieved when more importance is given to the pixels of the transformed image. In particular, on S$_1$ and S$_2$ the best performance ($86.73\%$ and $75.66\%$ respectively) is achieved for $\mathrm{Beta}(5, 1)$, while on S$_3$ ($84.21\%$) it is achieved for $\mathrm{Beta}(6, 2)$.

\subsection{Experimental validation}
\label{sec:corsmal-prime-experimental-validation}


We conduct experiments on C-CCM~\cite{Modas2021Improving} using a ResNet-18~\cite{He2016ResNet}. From the C-CCM pre-trained models provided by~\cite{Modas2021Improving}, we evaluate the ones trained with ST (baseline), ST$\rightarrow$FT, and AT$\rightarrow$FT, with the latter currently being the best one for classifying C-CCM. Furthermore, we train a model directly on C-CCM using data augmentations generated with PRIME (DA$_\text{PRIME}$), and, finally, we also explore the combination of fine-tuning an adversarially trained model~\cite{salman2020adversarially} with DA$_\text{PRIME}$. We denote this strategy as AT$\rightarrow$DA$_\text{PRIME}$. We evaluate and compare the different methods on the different splits of C-CCM. Whenever PRIME is used, we use the parameters specified in \cref{subsec:corsmal-prime-transformation-parameters,subsec:corsmal-prime-mixing-parameters}. All model definitions and training procedures are implemented in PyTorch~\cite{Paszke2019PyTorch}.

For DA$_\text{PRIME}$ and AT$\rightarrow$DA$_\text{PRIME}$ strategies we train or fine-tune the classifier for $50$ epochs, using a  cross-entropy loss and stochastic gradient descent. The maximum learning rate for updating the weights is set to $0.05$ and $0.005$ when performing DA$_\text{PRIME}$ and AT$\rightarrow$DA$_\text{PRIME}$ respectively. The learning rate decays linearly during training. Note that the models we evaluate are the ones that achieve the highest validation accuracy (early-stopping), while for dealing with class imbalances, the training images in a batch are randomly sampled with probabilities that are inversely proportional to the number of images of each class.

\subsubsection{Classification results}
\label{subsec:corsmal-prime-results}

We now evaluate and compare the different methods on the different splits of C-CCM. For the case of AT$\rightarrow$DA$_\text{PRIME}$, since there are multiple source models adversarially trained with perturbations of different strength $\epsilon$, we decided to choose those that lead to the highest validation accuracy. Hence, for S$_1$ we select a network trained with $\epsilon=0.05$, while for S$_2$ and S$_3$ a network trained with $\epsilon=0.5$. Recall that, for the AT$\rightarrow$FT models used in~\cite{Modas2021Improving} the selected values of $\epsilon$ were $0.05$, $1$ and $0.5$ for each dataset split respectively.

\textbf{Overall performance}
\Cref{fig:corsmal-prime-shapeanalysis} shows the classification performance of different strategies on the three configurations, $\text{S}_1$,  $\text{S}_2$ and  $\text{S}_3$. The results indicate that pre-training might not be necessary: properly tuning data augmentation to compensate for the dataset-specific distribution shifts, improves performance with a lower computational cost than using transfer learning. DA$_\text{PRIME}$ requires only $1.2\times$ additional training time compared to ST on C-CCM, which is many orders of magnitude lower than (adversarially) training a model on ImageNet for using transfer learning. When training time is not an issue, transfer learning with AT at the source domain combined with DA$_\text{PRIME}$ (AT$\rightarrow$DA$_\text{PRIME}$) generally improves performance. Note that when the performance of ST is low, all strategies lead to significant improvements; whereas  when ST performs well,  AT$\rightarrow$FT has an  insignificant contribution or decreases the final performance. 

\definecolor{ts1}{RGB}{ 0  0    0}    
\definecolor{ts2}{RGB}{229, 194, 36}  
\definecolor{ts3}{RGB}{238,113,27}  
\definecolor{ts4}{RGB}{19,219,30}  
\definecolor{ts5}{RGB}{0,58,236}  

\begin{figure*}[!t]
\includegraphics[width=\textwidth]{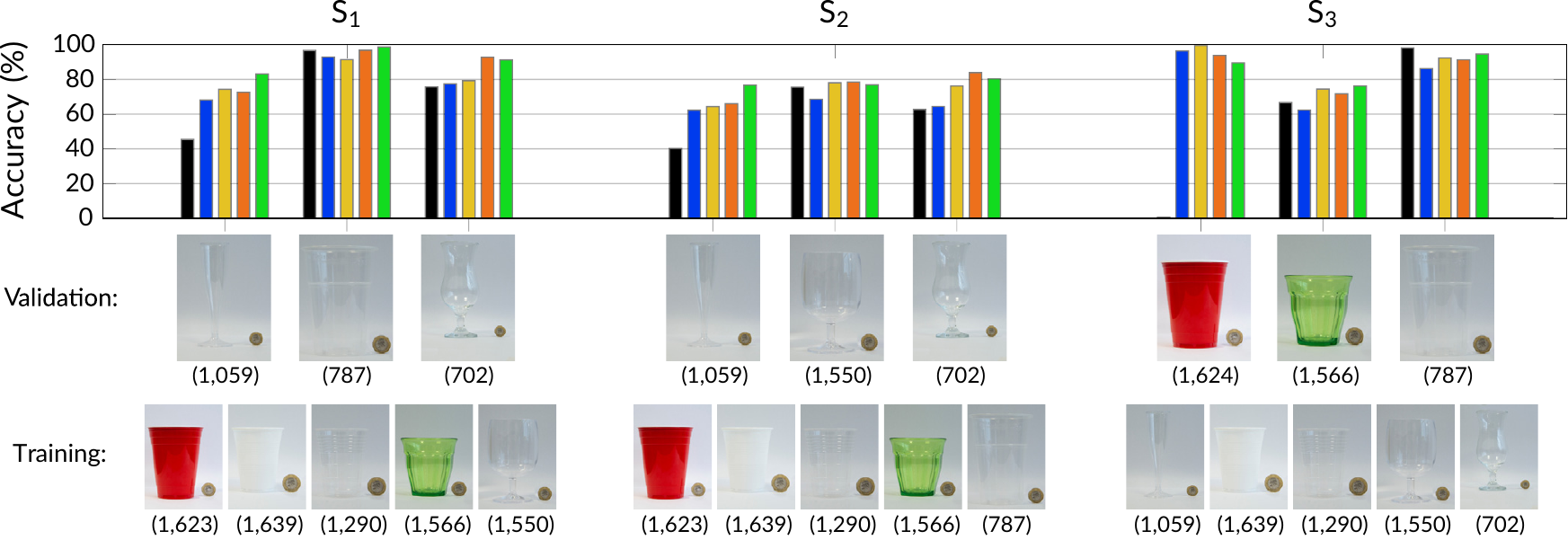}
\caption{Per-container filling level classification accuracy (top) on the three dataset splits (bottom) of C-CCM. Parentheses denote the number of images for each type of container. Legend:
\protect\raisebox{2pt}{\protect\tikz \protect\draw[ts1,line width=2] (0,0) -- (0.3,0);}~ST,
\protect\raisebox{2pt}{\protect\tikz \protect\draw[ts5,line width=2] (0,0) -- (0.3,0);}~ST$\rightarrow$FT.
\protect\raisebox{2pt}{\protect\tikz \protect\draw[ts2,line width=2] (0,0) -- (0.3,0);}~AT$\rightarrow$FT,
\protect\raisebox{2pt}{\protect\tikz \protect\draw[ts3,line width=2] (0,0) -- (0.3,0);}~DA$_\text{PRIME}$,
\protect\raisebox{2pt}{\protect\tikz \protect\draw[ts4,line width=2] (0,0) -- (0.3,0);}~AT$\rightarrow$DA$_\text{PRIME}$.}
\label{fig:corsmal-prime-shapeanalysis}
\vspace*{-2em}
\end{figure*}

\textbf{Detailed analysis for each split}
For  $\text{S}_1$, the low performance of ST on the champagne flute (left) is improved by both AT$\rightarrow$FT and DA$_\text{PRIME}$, and even more so by  AT$\rightarrow$DA$_\text{PRIME}$, suggesting that diffeomorphisms compensate for the unique narrow shape of the flute. The accuracy of ST on the beer cup (middle) is high, due to the  shape similarity  of the small transparent cup in the training set.  AT$\rightarrow$FT causes a small accuracy drop, whereas DA$_\text{PRIME}$ retains the performance and AT$\rightarrow$DA$_\text{PRIME}$ improves it. The accuracy of ST on the cocktail glass (right) is  slightly improved with AT$\rightarrow$FT and considerably improved by DA$_\text{PRIME}$ and AT$\rightarrow$DA$_\text{PRIME}$. Although there is another container with a stem in the training set (wine glass), it seems that the diffeomorphisms  better compensate for the different shape above the stem of the cocktail glass.

For  $\text{S}_2$, the accuracy of all strategies on the champagne flute (left) and the cocktail glass (right) is somehow similar in trend  to that on $\text{S}_1$. Note that there are no containers with a stem in the training set. Yet, the performance on the wine glass (middle) is similar for most strategies, which might be due to the similarity of its shape above the stem with the other transparent cups in the training set.

For  $\text{S}_3$, there is no colored container in the training set. ST is unable to generalize for the red cup (left), unlike AT$\rightarrow$FT, DA$_\text{PRIME}$ and AT$\rightarrow$DA$_\text{PRIME}$. Still, the accuracy with data augmentation is not on the same level as with AT$\rightarrow$FT, which sets this specific container case as an example of the benefits of adversarially pre-training the network on a large and diverse source dataset.
As for the green glass (middle) AT$\rightarrow$FT increases on ST, similarly to  AT$\rightarrow$DA$_\text{PRIME}$.
Finally, the accuracy of ST on the beer cup (right) is high and the other strategies cannot reach that level, with AT$\rightarrow$DA$_\text{PRIME}$ featuring the lowest performance drop.

\section{Discussion}
\label{sec:prime-discussion}

\textbf{Coverage over the space of corruptions}\quad
In many parts of this chapter we implied that, for conferring robustness to common corruptions, a good augmentation method should generate augmentations that cover a large space of possible corruptions. In general, formally identifying the space of semantic-preserving corruptions, and providing solid guarantees, is an utmost challenge. However, as with other problems in computer vision, we can rely on empirical proxies to gauge the coverage of an augmentation method. Specifically for PRIME, (i) its superior performance on multiple benchmarks (\cref{tab:prime-results-main}) and (ii) the quantitative study of \cref{subsec:prime-importance-of-mixing} along with the SimCLR embedding distances of \cref{tab:prime-simclr_distances} suggest that PRIME achieves a broader coverage of the space of common corruptions than other methods. Additionally, the max-entropy principle formally guarantees a good coverage over the space of each of the three primitive transformations. Note here, that similar transformations to our three primitives are commonly used to model many types of image corruptions (e.g., color jitters, or lens artifacts).

\begin{figure}[!ht]
    \begin{center}
    \includegraphics[width=0.8\linewidth]{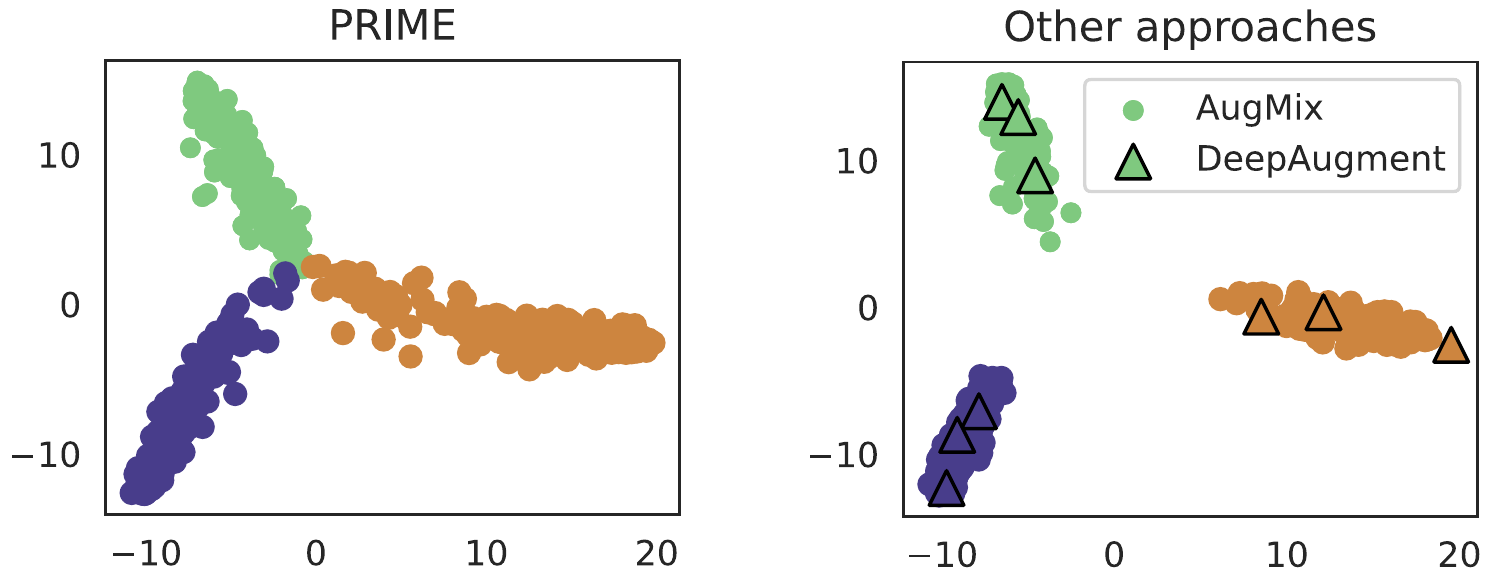}
    \end{center}
    \caption{Projections of augmentations generated by different methods on the embedding space of a ResNet-50.}
    \label{fig:prime-diversity-pca-feat-space}
\end{figure}

The SimCLR embedding distances of \cref{tab:prime-simclr_distances} can be seen as a measure of ``fitness'' for the common corruption benchmark; that is, how similar/representative are the generated augmentations with respect to the actual corrupted images in the benchmark. Another measure of interest would be to investigate how diverse are the generated augmentations; that is, how large is the variance of the augmentations as measured in a given space (i.e., embedding space of a network). Hence, to qualitatively compare the diversity the augmentations of PRIME with respect to other methods, we can follow the procedure in~\cite{Wang2021AugMax}. We randomly select 3 images from ImageNet, each one belonging to a different class. For each image, we generate 100 transformed instances using AugMix and PRIME, while with DeepAugment we can only use the original images and the 2 transformed instances that are pre-generated with the EDSR and CAE image-to-image networks that DeepAugment uses. Then, we pass the transformed instances of each method through a ResNet-50 pre-trained on ImageNet and extract the features of its embedding space. On the features extracted for each method, we perform PCA and then visualize the projection of the features onto the first two principal components. We visualize the projected augmented space in \cref{fig:prime-diversity-pca-feat-space}, which suggests that PRIME, not only generates augmentations that fit better the benchmark of common corruptions (cf. SimCLR embedding distances), but that are also more diverse than AugMix and DeepAugment.

Furthermore, for understanding the potential of PRIME in covering a large space of the common corruptions, one can measure the capability of PRIME to create augmented images that are very similar to those in the benchmark. Since the transformations of PRIME are all differentiable, apart from the SimCLR embedding space distances, one can try to optimize the parameters of the transformations to minimize $\|\bm x^t-\bm{x}^c\|_2$; that is, the distance between a PRIME-transformed image $\bm x^t$ and the corresponding image from the corruption benchmark $\bm{x}^c$. Alternatively, one can focus on the mixing parameters and follow the same approach as in~\cite{Wang2021AugMax}: to find the worst-case (adversarial) mixing coefficients $\bm\lambda$ such that the mixed augmented images reach a part of the space that causes the most significant change on the classifier.  

\textbf{Extensions} The general model of common corruptions of \cref{eq:prime-cc_model} can be extended with more transformations, as long as they follow the same principles as the ones we introduced (i.e., sampled from a max-entropy distribution). We already discussed the use of additive random noise $\eta$, which we do not include in our experiments due to its overlap with corruptions that exist in the evaluation benchmarks. Another possibility is to consider the transformations in the spatial domain. We chose to use diffeomorphisms, but one can also introduce random affine or projective transformations instead. Finally, our model can also easily accept other types of modifications, such as occlusions, or transformations based on task-specific priors.

\section{Conclusions}
\label{sec:prime-conclusions}

In this chapter, we took a systematic approach to formulate a universal model that captures a wide variety of semantic-preserving image transformations. In particular, we defined a set of primitive max-entropy transformations in the spatial, color and spectral image domain, which guarantee that we provide the network with the maximum additional information possible about the domain of the transformation. Relying on this model, we proposed a novel data augmentation scheme called \textit{PRIME}, which instantiates our model of corruptions, to confer robustness against common corruptions. 

From a practical perspective, our method is principled yet efficient and can be conveniently incorporated into existing training procedures. Moreover, it yields a strong baseline on existing corruption benchmarks outperforming current standalone methods. Additionally, our thorough ablations demonstrate that diversity among basic augmentations (primitives) -- which AugMix and other approaches lack -- is essential, and that mixing plays a crucial role in the success of both prior methods and PRIME.
In general, while complicated methods like DeepAugment perform well, it is difficult to understand, ablate and apply these online. Instead, we show that a simple model-based stance with a few guiding principles can be used to build an efficient augmentation scheme that can be easily understood, ablated and tuned.

We believe that our insights and PRIME pave the way for building robust models in real-life scenarios. For instance, we demonstrated that PRIME already provides a ready-to-use recipe for conferring robustness in the problem of filling level classification. We believe that our method can be easily adapted to other for data-scarce domains such as medical imaging.
\chapter{Conclusion}
\label{ch:conclusion}

\section{Summary}
\label{sec:conclusion-summary}

In this thesis, we provided novel algorithmic tools, a large a set of new insights and experimental evidences, as well as constructive solutions for understanding and improving the robustness and invariance properties of image classifiers. Our tools allowed us to efficiently evaluate the robustness of deep networks to non-standard adversarial perturbation regimes, as well as to establish strong connections between the data features, the geometric properties of deep classifiers, and their inductive biases. Furthermore, our insights enabled us to further design principled methods for analyzing and improving the robustness of deep classifiers to different types of distribution shifts in multiple tasks.

We first studied methods for measuring the robustness of deep classifiers to sparse additive adversarial perturbations. For avoiding the NP-hardness of minimizing the $\ell_0$ norm in computing sparse perturbations, we focused on finding an efficient relaxation method. To this end, we exploited the low mean curvature of the decision boundaries in the vicinity of the data samples and designed an iterative method that we coin SparseFool. At each iteration SparseFool performs a linear approximation of the decision boundary and solves the simpler $\ell_1$ box-constrained problem. Our method computes very sparse perturbations, is by orders of magnitude faster than existing methods, and can easily scale to high-dimensional datasets. By visually inspecting the generated adversarial examples, we observed that SparseFool altered features that are shared among different images, and that, in many cases, the perturbations resembled image features that are correlated with the fooling class. Such observation suggested that adversarial perturbations might not necessarily be a ``hole'' in the system, but they might actually reflect some strong connection/correlation between the features of the dataset and the features that the networks use for taking their decisions. 

Then, we studied this connection in-depth, and we proposed a new geometric framework that permits to relate the features of a dataset with the distance to the decision boundary along specific directions. Through a series of carefully designed experiments, both in synthetic and real image datasets, we explained how the inductive bias of the learning algorithm shapes the decision boundaries of neural networks by creating boundaries that are invariant to non-discriminative directions. Furthermore, we demonstrated that the decision boundaries are very sensitive to the position of the training samples, and that small changes along specific directions can cause large and sudden changes in orthogonal ones. In fact, adversarial training exploits this sensitivity, as well as the inductive bias towards invariance, in order to shape the boundaries and build robust classifiers. Interestingly, modifying the position of just a minimal number of training samples during adversarial training is enough to locally introduce excessive invariance. In general, our framework can be used to identify parts of the input space that are important for the classifier, to understand intriguing properties of deep networks, such as their catastrophic forgetting, and to design stronger defenses and black-box attacks that exploit small discriminative subspaces. Furthermore, we studied the implications that the invariance properties of robust classifiers have in the downstream, data-scarce application of classifying the filling level within containers. In particular, we showed that, during transfer learning, using robust models in the source domain permits to consistently improve generalization to unseen containers in the target domain that come from a shifted distribution, e.g., containers of unseen shape or color.


However, adversarial robustness is not necessarily the best way to improve the robustness of classifiers to certain distribution shifts, such as common corruptions of the data. Hence, we took a systematic approach to understand the notion of common corruptions and formulated a universal model that captures a wide variety of semantic preserving, common image transformations. In particular, we defined a set of primitive max-entropy transformations in the spatial, color and spectral image domains, in order to guarantee that, with each new augmentation, we provide the network with the maximum additional information possible about the domain of the transformation. Relying on this model, we proposed a novel data augmentation scheme called \textit{PRIME}, which instantiates our semantic-preserving transformations, to confer robustness against common corruptions. From a practical perspective, our method is principled yet efficient and can be conveniently incorporated into existing training procedures. Moreover, it yields a strong baseline on existing corruption benchmarks and outperforms current similar  methods. Additionally, our thorough ablations demonstrate that diversity among basic augmentations (primitives) -- which AugMix and other approaches lack -- is essential, and that mixing plays a crucial role in the success of both prior methods and PRIME. Our findings also underlined the benefits of using PRIME in on-line augmentation settings, due to its scalability and efficiency. Finally, we demonstrated that PRIME already provides a ready-to-use recipe for conferring robustness in the data-scarce problem of filling level classification. PRIME can be easily tuned to generate samples, which are tailored for building classifiers that generalize on images with containers of unknown shape, color, and spectral content.

\section{Future directions}
\label{sec:conclusion-future}

First, we mostly studied the geometric and invariance characteristics of the decision boundaries through the lens of the data features. However, we did not investigate other factors that are responsible for shaping the decision boundaries, such as the architecture. From our experiments we can observe that, when the frequency representation of the data is flipped, the margin distribution is not an exact mirroring of the original one, and that the margin along low-frequency directions cannot reach excessive levels. In fact, this triggered some recent research, which demonstrated that the architecture itself has a very strong inductive bias towards specific input directions, which are typically aligned with low-frequency directions. Such directions are coined Neural Anisotropy Directions (NADS)~\cite{OrtizJimenez2020NADs}, and can shed new light onto the different types of inductive biases that deep networks have, such as the one towards simple solutions~\cite{OrtizJimenez2021Underspecification}. Developing methodologies for revealing, understanding, and controlling the inductive bias is very important, since it can have strong implications in the generalization and robustness properties of deep networks~\cite{Ortiz2021WhatCanLinearized,Yuce2022StructuredDictionary}.

On the practical side, the invariance insights of our geometric analysis have already been exploited in~\cite{Rade2021HelperBased} to control the excessive margin caused by adversarial training, and eventually balance the robustness/accuracy tradeoff. Furthermore, in our experiments we used the DCT basis. However, such basis is independent of the data and the training process. Incorporating other task- or data-related priors to measure the geometric properties or manipulate the data is quite important. This could have significant application in fairness-sensitive tasks for identifying (or eliminating) the undesired bias from the model. In addition, our framework could also be extended beyond image classification, for example towards natural language processing tasks. 

Then, we have seen that formally identifying the space of common corruptions is very difficult and, probably, ill-posed. Currently, one can only follow qualitative or empirical proxies to measure the efficacy of a method, i.e., through accuracy or through the features of the embedding space (i.e., measure similarity or visualize projections on principal feature directions). Nevertheless, it is important to analytically characterize this space, even partially, since it will enable the development of more fundamental methods that provide a good coverage over that space. Additionally, the same principles for identifying the space of corruptions can be followed for characterizing other types of manipulations or distribution shifts~\cite{Hendrycks2021TheManyFaces,Koh2021WILDS}. 

In this regard, and from a practical perspective, it is worth investigating how the primitives of PRIME can be extended for distributions shifts beyond common corruptions, but also to downstream tasks and applications. For instance, it is interesting to explore the applicability of PRIME on semantic segmentation and object detection, or even speech recognition, but also in safety-critical applications, i.e., autonomous driving, or data-scarce domains, i.e., medical imaging. 

\addtocontents{toc}{\vspace{\normalbaselineskip}}
\cleardoublepage
\bookmarksetup{startatroot}

\appendix


\chapter{Appendix of Chapter 4}
\label{ch:appendix-ch4}
\section{Margin distribution of a linear classifier}
\label{sec:appendix-ch4-theoretical-margin}

In this section we demonstrate that even for linear classifiers trained on $\mathcal{T}_1(\epsilon, \rho, N)$ the distribution of margins along non-discriminative directions will never be infinite. We demonstrate this effect in practice by repeating the experiment of \cref{subsec:hmt-synthetic-example-discriminative-features}, where instead of an MLP we use a simple logistic regression (see \cref{tab:hmt-invariance_synthetic_logreg}). Clearly, although the values along $\operatorname{span}\{\bm{u}_1\}^\perp$ are quite large, they are still finite. This demonstrates that due to the finiteness of the training set and its high-dimensionality the influence of the non-discriminative directions in the final solution is significant.

\begin{table}[ht]
\begin{center}
\begin{scriptsize}
\begin{sc}
\begin{tabular}{lcccc}
\toprule
& $\bm{u}_1$ & $\operatorname{span}\{\bm{u}_1\}^\perp$ & $\mathcal{S}_{\text{orth}}$ & $\mathcal{S}_{\text{rand}}$ \\
\midrule
5-perc. & $2.39$ & $36.7$ & $184.95$  & $11.57$ \\
Median & $2.49$ & $38.3$ & $192.98$  & $12.08$ \\
95-perc. & $2.60$ & $39.9$ & $201.16$  & $12.59$\\
\bottomrule
\end{tabular}
\end{sc}
\end{scriptsize}
\end{center}
\caption{Margin statistics of a logistic regressor trained on $\mathcal{T}_1(\epsilon=5, \sigma=1)$ along different directions ($N=10,000$, $M=1,000$, $S=3$).}
\label{tab:hmt-invariance_synthetic_logreg}
\end{table}

\section{Training parameters}
\label{sec:appendix-ch4-train_params}

\Cref{tab:appendix-ch4-train_performance} shows the performance and training parameters of the different networks used in the paper. Note that the hyperparameters of these networks were not optimized in any form during this work. Instead they were selected from a set of best practices from the DAWNBench submissions that have been empirically shown to give a good trade-off in terms of convergence speed and performance. In this sense, especially for the non-standard datasets (e.g., ``flipped'' datasets), the final performance might not be the best reflection of the highest achievable performance of a given architecture. In fact, since the goal of our experiments is not to achieve the most robust models on such non-standard datasets, but rather investigate how the previously observed trends are represented in these new classifiers, no further hyperparameter tuning was applied. All the experiments with synthetic data were trained in the same way, namely using SGD with a linearly decaying learning rate (max lr. 0.1), no explicit regularization, and trained for 500 epochs.

\begin{table}[ht]
\begin{center}
\begin{sc}
\scriptsize
\begin{tabular}{lcccccc}
\toprule
Dataset & Network & \makecell[c]{Test \\ Acc.} & Epochs & \makecell[c]{LR \\ Schedule} & max. LR & Batch \\
\midrule
\multirow{2}{*}{MNIST}
& LeNet & $99.35\%$ & \multirow{2}{*}{$30$} & \multirow{2}{*}{Triang.} & \multirow{2}{*}{$0.21$} & \multirow{2}{*}{$128$} \\ 
& ResNet-18 & $99.53\%$ & & & & \\
\midrule
\multirow{2}{*}{\makecell[l]{MNIST \\ Flipped}}
& LeNet & $99.34\%$ & \multirow{2}{*}{$30$} & \multirow{2}{*}{Triang.} & \multirow{2}{*}{$0.21$} & \multirow{2}{*}{$128$} \\ 
& ResNet-18 & $99.52\%$ & & & & \\
\midrule
\multirow{3}{*}{CIFAR-10}
& VGG-19 & $89.39\%$ & \multirow{3}{*}{$50$} & \multirow{3}{*}{Triang.} & \multirow{3}{*}{$0.21$} & \multirow{3}{*}{$128$} \\
& ResNet-18 & $90.05\%$ & & & & \\
& DenseNet-121 & $93.03\%$ & & & & \\
\midrule
\multirow{3}{*}{\makecell[l]{CIFAR-10 \\ Low Pass}}
& VGG-19 & $84.81\%$ & \multirow{3}{*}{$50$} & \multirow{3}{*}{Triang.} & \multirow{3}{*}{$0.21$} & \multirow{3}{*}{$128$} \\
& ResNet-18 & $84.77\%$ & & & & \\
& DenseNet-121 & $88.51\%$ & & & & \\
\midrule
\multirow{3}{*}{\makecell[l]{CIFAR-10 \\ Flipped}}
& VGG-19 & $87.42\%$ & \multirow{3}{*}{$50$} & \multirow{3}{*}{Triang.} & \multirow{3}{*}{$0.21$} & \multirow{3}{*}{$128$} \\
& ResNet-18 & $88.67\%$ & & &  & \\
& DenseNet-121 & $91.19\%$ & & & & \\
\midrule
\multirow{3}{*}{ImageNet}
& VGG-16 & $71.59\%$ & \multirow{3}{*}{--} & \multirow{3}{*}{--} & \multirow{3}{*}{--} & \multirow{3}{*}{--} \\
& ResNet-50& $76.15\%$ & & & & \\
& DenseNet-121 & $74.65\%$ & & & & \\
\midrule
\makecell[l]{ImageNet \\ Flipped} & ResNet-50 & $68.12\%$ & $90(68)$ & \makecell[c]{Piecewise \\ Constant} & $0.1$ & $256$ \\
\bottomrule
\end{tabular}
\end{sc}
\caption{Performance and training parameters of multiple networks trained on different datasets. All networks have been trained using SGD with momentum $0.9$ and a weight decay of $5\times10^{-4}$. For ImageNet, we use the pretrained models from PyTorch. For ``flipped'' ImageNet, the weight decay was set to $10^{-4}$, while for computational reasons the training was executed until the $68^\text{th}$ epoch.}
\label{tab:appendix-ch4-train_performance}
\end{center}
\end{table}

\clearpage
\newpage

\section{Cross-dataset performance}
\label{sec:appendix-ch4-cross_dataset}

We now show the performance of different networks trained with different variants of the standard computer vision datasets and tested on the rest.

\begin{table}[!ht]
\begin{center}
\begin{sc}
\scriptsize
\begin{tabular}{lccccc}
\toprule
& & MNIST & MNIST Flipped & MNIST High Pass \\
\midrule
\multirow{2}{*}{MNIST}
& LeNet & $99.35\%$ & $18.73\%$ & $44.09\%$ \\ 
& ResNet-18 & $99.53\%$ & $11.88\%$ & $15.73\%$ \\ 
\midrule
\multirow{2}{*}{\makecell[l]{MNIST \\ Flipped}}
& LeNet & $10.52\%$ & $99.34\%$ & $9.87\%$ \\ 
& ResNet-18 & $16.59\%$ & $99.52\%$ & $11.23\%$ \\ 
\midrule
\multirow{2}{*}{\makecell[l]{MNIST \\ High Pass}}
& LeNet & $96.35\%$ & $42.36\%$ & $98.65\%$ \\
& ResNet-18 & $88.38\%$ & $21.48\%$ & $98.71\%$ \\
\bottomrule
\end{tabular}
\end{sc}
\caption{Networks trained on a specific version of MNIST, but evaluated on different variations of it. Rows denote the dataset that a network is trained on, and columns the dataset they are evaluated on. Values on the diagonal correspond to the same variation.}
\label{tab:appendix-ch4-cross_mnist}
\end{center}
\end{table}

\begin{table}[!ht]
\begin{center}
\begin{sc}
\scriptsize
\begin{tabular}{lccccc}
\toprule
& & CIFAR-10 & CIFAR-10 Flipped & CIFAR-10 Low Pass \\
\midrule
\multirow{3}{*}{CIFAR-10}
& VGG-19 & $89.39\%$ & $10.63\%$ & $61.4\%$ \\ 
& ResNet-18 & $90.05\%$ & $10\%$ & $46.99\%$ \\ 
& DenseNet-121 & $93.03\%$ & $10.3\%$ & $27.45\%$ \\ 
\midrule
\multirow{3}{*}{\makecell[l]{CIFAR-10 \\ Flipped}}
& VGG-19 & $10.77\%$ & $87.42\%$ & $10.79\%$ \\ 
& ResNet-18 & $9.91\%$ & $88.67\%$ & $9.97\%$ \\ 
& DenseNet-121 & $9.98\%$ & $91.19\%$ & $10\%$ \\ 
\midrule
\multirow{3}{*}{\makecell[l]{CIFAR-10 \\ Low Pass}}
& VGG-19 & $85.16\%$ & $10.52\%$ & $84.81\%$ \\ 
& ResNet-18 & $85.47\%$ & $10.45\%$ & $84.77\%$ \\ 
& DenseNet-121 & $89.67\%$ & $10.45\%$ & $88.51\%$ \\ 
\bottomrule
\end{tabular}
\end{sc}
\caption{Networks trained on a specific version of CIFAR-10, but evaluated on different variations of it. Rows denote the dataset that a network is trained on, and columns the dataset they are evaluated on. Values on the diagonal correspond to the same variation.}
\label{tab:appendix-ch4-cross_cifar}
\end{center}
\end{table}

\begin{table}[!ht]
\begin{center}
\begin{sc}
\scriptsize
\begin{tabular}{lcccc}
\toprule
& & ImageNet & ImageNet Flipped \\
\midrule
ImageNet
& VGG-16 & $71.59\%$ & $0.106\%$ \\
& ResNet-50 & $76.15\%$ & $0.292\%$ \\
& DenseNet-121 & $74.65\%$ & $0.22\%$ \\
\midrule
\makecell[l]{ImageNet \\ Flipped}
& ResNet-50 & $0.184\%$ & $68.12\%$ \\ 
\bottomrule
\end{tabular}
\end{sc}
\caption{Networks trained on a specific version of ImageNet, but evaluated on different variations of it. Rows denote the dataset that a network is trained on, and columns the dataset they are evaluated on. Values on the diagonal correspond to the same variation.}
\label{tab:appendix-ch4-cross_imagenet}
\end{center}
\end{table}

\clearpage
\newpage

\section{Adversarial training parameters}
\label{sec:appendix-ch4-adv_train_params}

\Cref{tab:appendix-ch4-adv_train_performance} shows the performance and adversarial training parameters of the different networks used in the paper. Note that the hyperparameters of these networks were not optimized in any form during this work. Instead they were selected from a set of best practices from the DAWNBench submissions that have been empirically shown to give a good trade-off in terms of convergence speed and performance. Again, as stated in \cref{sec:appendix-ch4-train_params}, especially for the non-standard datasets (e.g., ``flipped'' datasets), the final performance might not be the best reflection of the highest achievable performance or robustness of a given architecture, since no further hyperparameter tuning was applied.

\begin{table}[!ht]
\begin{center}
\begin{sc}
\scriptsize
\begin{tabular}{lccccccc}
\toprule
Dataset & Network & \makecell[c]{Standard \\ Test Acc.} & \makecell[c]{Adv. \\ Test Acc.} & Epochs & \makecell[c]{$\ell_2$ ball \\ radius} & Steps\\
\midrule
\multirow{2}{*}{MNIST}
& LeNet & $98.32\%$ & $76.01\%$ & \multirow{2}{*}{$25$} & \multirow{2}{*}{$2$} & \multirow{2}{*}{$7$}\\ 
& ResNet-18 & $98.89\%$ & $80.26\%$ & & &\\ 
\midrule
\multirow{2}{*}{\makecell[l]{MNIST \\ Flipped}}
& LeNet & $98.29\%$ & $74.68\%$ & \multirow{2}{*}{$25$} & \multirow{2}{*}{$2$} & \multirow{2}{*}{$7$}\\ 
& ResNet-18 & $98.75\%$ & $81.97\%$ & & &\\ 
\midrule
\multirow{3}{*}{CIFAR-10}
& VGG-19 & $73.76\%$ & $50.15\%$ & \multirow{3}{*}{$50$} & \multirow{3}{*}{$1$} & \multirow{3}{*}{$7$}\\
& ResNet-18 & $82.20\%$ & $52.38\%$ & & & \\ 
& DenseNet-121 & $82.90\%$ & $54.86\%$ & & & \\ 
\midrule
\multirow{3}{*}{\makecell[l]{CIFAR-10 \\ Flipped}}
& VGG-19 & $71.39\%$ & $35.64\%$ & \multirow{3}{*}{$50$} & \multirow{3}{*}{$1$} & \multirow{3}{*}{$7$}\\
& ResNet-18 & $73.64\%$ & $37.24\%$ & & &\\ 
& DenseNet-121 & $78.32\%$ & $42.32\%$ & & & \\ 
\midrule
ImageNet & ResNet-50 & $57.90\%$ & $35.16$ & -- & $3$ & $20$\\
\bottomrule
\end{tabular}
\end{sc}
\caption{Performance and attack parameters of multiple networks adversarially trained using $\ell_2$-PGD. The training parameters are similar to the ones of \cref{tab:appendix-ch4-train_performance}. For ImageNet we use the adversarially trained ResNet-50 provided by~\cite{robustness}.}
\label{tab:appendix-ch4-adv_train_performance}
\end{center}
\end{table}

\section{Spectral decomposition on frequency ``flipped'' data}
\label{sec:appendix-ch4-spectral-decomp-flipped}

\begin{figure}[ht]
\begin{center}
\begin{subfigure}[b]{0.48\textwidth}
\includegraphics[width=\linewidth]{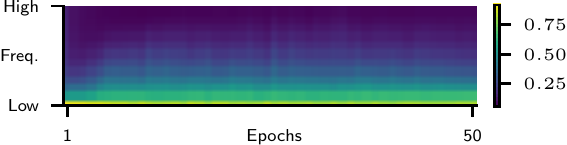}
\caption{Adversarial training.}
\end{subfigure}\hfill
\begin{subfigure}[b]{0.48\textwidth}
\includegraphics[width=\linewidth]{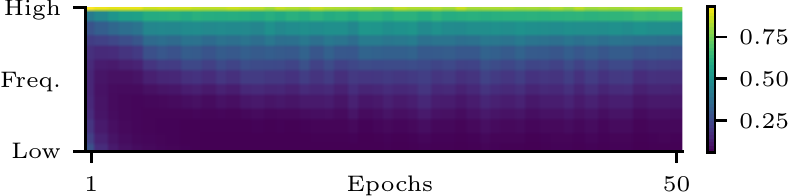}
\caption{Frequency ``flipped'' adv. training.}
\end{subfigure}\hfill
\caption{Energy decomposition in subspaces of the DCT diagonal of adversarial perturbations used during adversarial training ($\ell_2$ PGD with $\epsilon=1$) on 1,000 (a) CIFAR-10 and (b) frequency ``flipped'' CIFAR-10 training samples per epoch for a DenseNet-121. The plot shows 95-percentile of energy.}
\label{fig:appendix-ch4-spectral_standard_flip}
\end{center}
\end{figure}

Following the results presented in \cref{subsec:hmt-connections-to-adversarial-training}, we now show in \cref{fig:appendix-ch4-spectral_standard_flip} the spectral decomposition of the adversarial perturbations crafted during adversarial training for the frequency ``flipped'' CIFAR-10 dataset on a DenseNet-121 network. In contrast to the spectral decomposition of the perturbations on CIFAR-10 (left), the energy of the frequency ``flipped'' CIFAR-10 perturbations (right) remains concentrated in the high part of the spectrum during the whole training process, and has hardly any presence in the low frequencies. In other words, the frequency content of the $\ell_2$-PGD adversarial perturbations also ``flips''.

\clearpage
\newpage

\chapter{Appendix of Chapter 6}
\label{ch:appendix-ch6}

\section{Maximum entropy transformations}
\label{sec:app-ch6-max-entropy}

To guarantee as much diversity as possible in our model of common corruptions, we follow the principle of maximum entropy to define our distributions of transformations~\cite{cover_info}. Note that using a set of augmentations that guarantees maximum entropy comes naturally when trying to optimize the sample complexity derived from certain information theoretic generalization bounds, both in the clean~\cite{xuInfoBound} and corrupted setting~\cite{OODInfoBound}. Specifically, the principle of maximum entropy postulates favoring those distributions that are as unbiased as possible given the set of constraints that defines a family of distributions. In our case, these constraints are given in the form of an expected strength, i.e., $\sigma^2$, desired smoothness, i.e., $K$, and/or some boundary conditions, \eg, the displacement field must be zero at the borders of an image.  

Let us make this formal. In particular, let $\mathcal{I}$ denote the space of all images $\bm x:\R^2\to\R^3$, and let $f:\mathcal{I}\to \mathcal{I}$ denote a random image transformation distributed according to the law $\mu$. Further, let us define a set of constraints $\mathcal{C}\subseteq \mathcal{F}$, which restrict the domain of applicability of $f$, i.e., $f\in\mathcal{C}$, and where $\mathcal{F}$ denotes the space of functions $\mathcal{I}\to \mathcal{I}$. The principle of maximum entropy postulates using the distribution $\mu$ which has maximum entropy given the constraints:
\begin{align}
    \underset{\mu}{\text{maximize}}\quad &  H(\mu)=\int_{\mathcal{F}}\, \mathrm{d}\mu(f) \log(\mu(f)) \label{eq:app-ch6-max_entropy}\\
    \text{subject to}\quad & f\in\mathcal C\quad \forall f\sim\mu, \nonumber
\end{align}
where $H(\mu)$ represents the entropy of the distribution $\mu$~\cite{cover_info}. In its general form, solving \cref{eq:app-ch6-max_entropy} for any set of constraints $\mathcal{C}$ is intractable. However, leveraging results from statistical physics, we will see that for our domains of interest, \cref{eq:app-ch6-max_entropy} has a simple solution. In what follows we derive those distributions for each of our family of transformations.

\subsection{Spectral domain}

As we introduced in \cref{subsec:prime-model-of-visual-corruptions}, we propose to parameterize our family of spectral transformations using an FIR filter of size $K_\omega\times K_\omega$. That is, we are interested in finding a maximum entropy distribution over the space of spectral transformations with a finite spatial support. 

Nevertheless, on top of this smoothness constraint we are also interested in controlling the strength of the transformations.  We define the strength of a distribution of random spectral transformations applied to an image $\bm x$, as the expected $L_2$ norm of the difference between the clean and transformed images, i.e.,
\begin{equation}
   \mathbb{E}_{\omega} \|\bm x-\omega(\bm x)\|^2_2=\mathbb{E}_{\bm\omega'}\|\bm \omega' * \bm x\|_2^2,
\end{equation}
which using Young's convolution inequality is bounded as
\begin{equation}
   \mathbb{E}_{\bm\omega'}\|\bm \omega' * \bm x\|_2^2\leq \|\bm x\|_1^2 \;\mathbb{E}_{\bm \omega'}\|\bm\omega'\|_2^2.
\end{equation}
Indeed, we can see that the strength of a distribution of random smooth spectral transformations is governed by the expected norm of its filter. In the discrete domain, this can be simply computed as
\begin{equation}
   \mathbb{E}_{\bm \omega'}\|\bm\omega'\|_2^2=\sum_{i=1}^{K_\omega}\sum_{j=1}^{K_\omega}\mathbb{E}_{\bm\omega'}\bm {\omega'}^2_{i,j}.\label{eq:app-ch6-norm_omega}
\end{equation}

Considering this, we should then look for a maximum entropy distribution whose samples satisfy
\begin{equation}
    \mathcal{C}=\left\{\bm\omega'\in\R^{K_\omega\times K_\omega}\wedge \mathbb{E}_{\bm\omega'}\|\bm\omega'\|_2^2=K_\omega^2 \sigma^2_\omega \,|\, \omega\sim \mu_\omega\right\}.\label{eq:app-ch6-spectral_constraint}
\end{equation}

Now, note that this set is defined by an equality constraint involving a sum of $K_\omega^2$ quadratic random variables. In this sense, we know that the Equipartition Theorem~\cite{beale} applies and can be used to identify the distribution of maximum entropy. That is, the solution of \cref{eq:app-ch6-max_entropy} in the case that $\mathcal{C}$ is given by \cref{eq:app-ch6-spectral_constraint}, is equal to the distribution of FIR filters whose coefficients are i.i.d. with law $\mathcal{N}(0, \sigma^2_\omega)$.

\subsection{Spatial domain}
The distribution of diffeomorphisms of maximum entropy with a fixed norm was derived by Petrini \etal in \cite{Petrini2021Diffeo}. The derivation is similar to the spectral domain, but with the additional constraint that the diffeomorphisms produce a null displacement at the borders of the image.

\subsection{Color domain}
We can follow a very similar route to derive the distribution of maximum entropy among all color transformations, where, specifically, we constraint the transformations to yield $\gamma(0)=0$ and $\gamma(1)=1$ on every channel independently. Doing so, the derivation of the maximum entropy distribution can follow the same steps as in \cite{Petrini2021Diffeo}.

\section{PRIME implementation details}
\label{sec:app-ch6-implementation_details}

In this section, we provide additional details regarding the implementation of PRIME described in \cref{sec:prime-prime-data-augmentations}. Since the parameters of the transformations are empirically selected, we first provide more visual examples for different values of smoothness $K$ and strength $\sigma$. Then, we give the exact values of the parameters we use in our experiments supported by additional visual examples.

\subsection{Additional transformed examples}

\begin{figure}[!b]
    \centering
    \includegraphics[width=0.7\columnwidth]{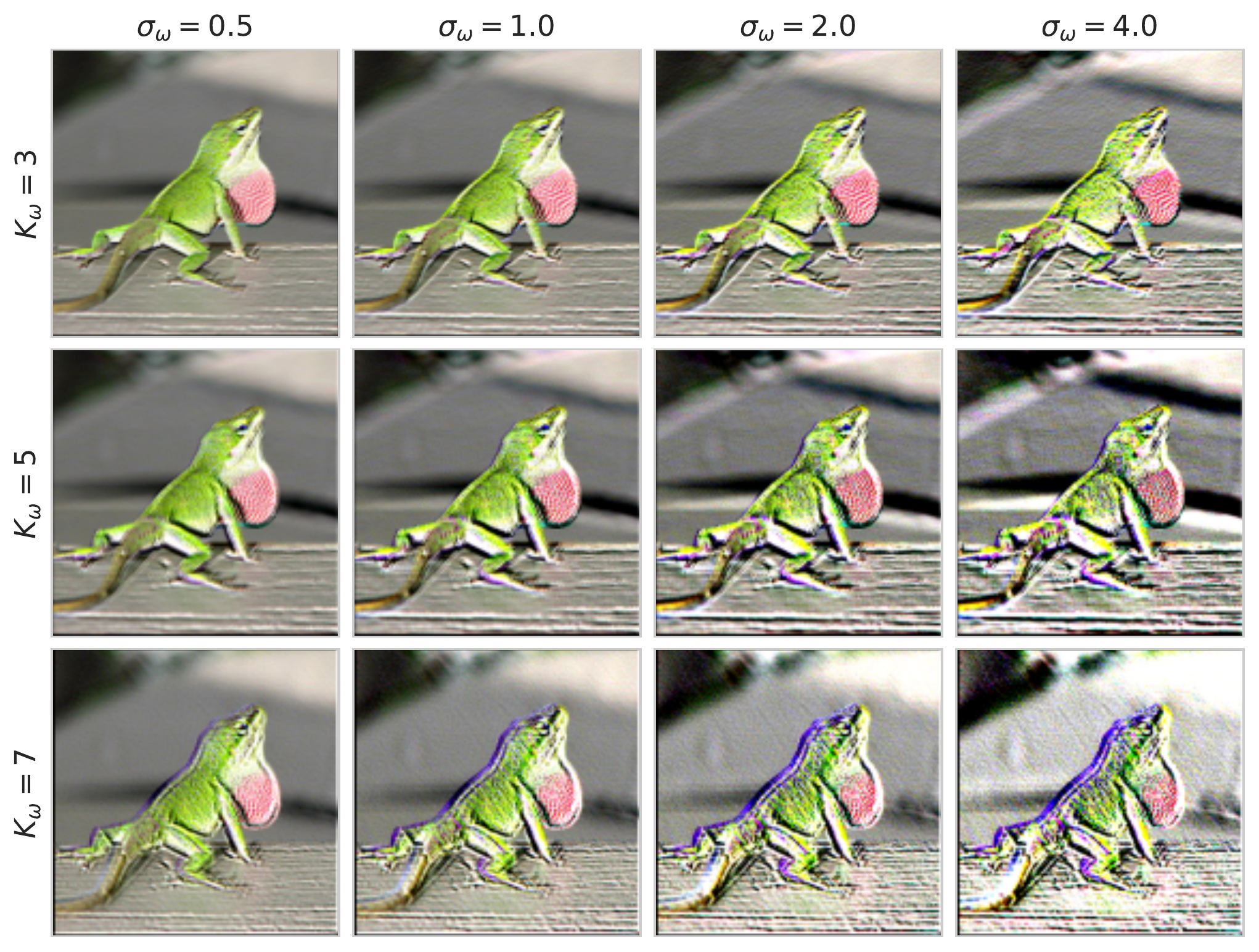}
    \caption{Example images (IN-100) generated with spectral transformations from our common corruptions model. In each row, we enlarge the transformation strength $\sigma_{\omega}$ from left to right. From top to bottom, we increase the spectral resolution of the filter $K_\omega$.}
    \label{fig:app-ch6-ex_spectral}
\end{figure}

\begin{figure}[!t]
    \centering
    \includegraphics[width=0.7\columnwidth]{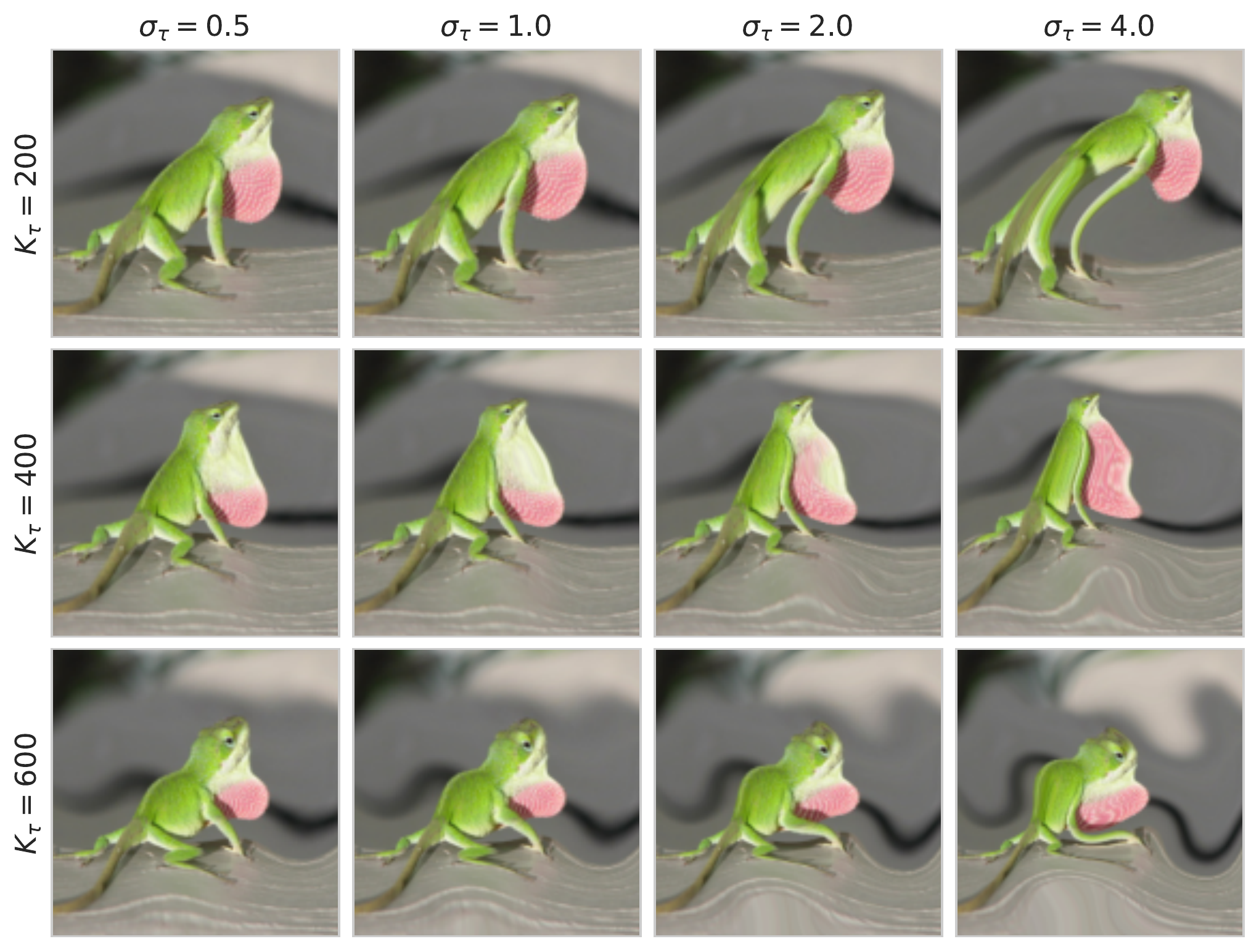}
    \caption{Example images (IN-100) generated with spatial transformations from our common corruptions model. In each row, we enlarge the transformation strength $\sigma_{\tau}$ from left to right. From top to bottom, we increase the cut frequency $K_\tau$.}
    \label{fig:app-ch6-ex_spatial}
\end{figure}

\begin{figure}[!t]
    \centering
    \includegraphics[width=0.7\columnwidth]{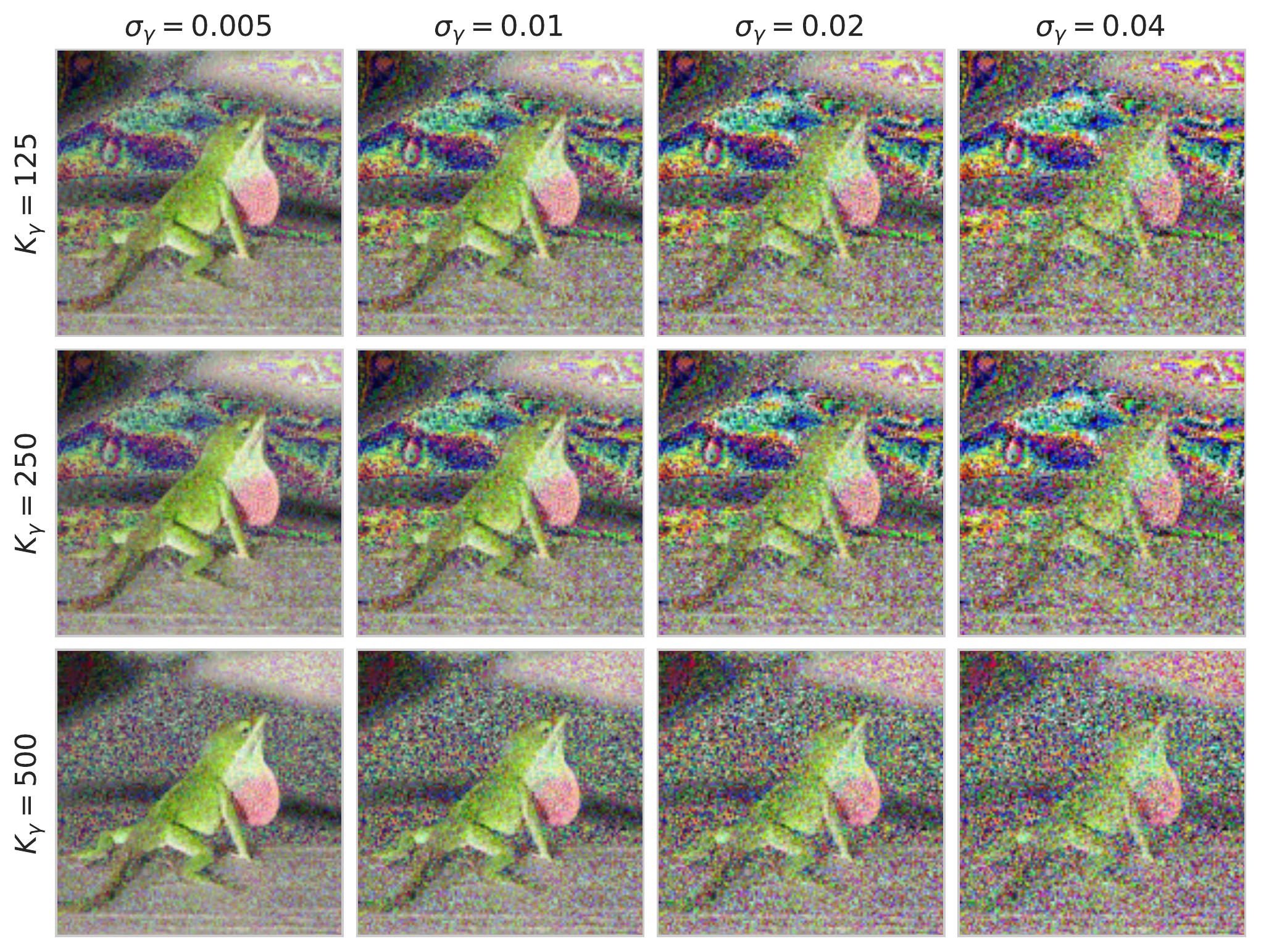}
    \caption{Example images (IN-100) generated with color transformations from our common corruptions model. In each row, we enlarge the transformation strength $\sigma_{\gamma}$ from left to right. From top to bottom, we increase $K_\gamma$.}
    \vspace{-0.8em}
\label{fig:app-ch6-ex_color}
\end{figure}

Here we provide additional visual examples for each of the primitives of PRIME illustrating the effect of the following two factors: (i) smoothness controlled by parameter $K$, and (ii) strength of the transformation $\sigma$ on the resulting transformed images created by the primitives. \cref{fig:app-ch6-ex_spectral,,fig:app-ch6-ex_spatial,,fig:app-ch6-ex_color} demonstrate the resulting spectrum of images created by applying spectral, spatial and color transformations while varying the parameters $K$ and $\sigma$. Notice how increasing the strength $\sigma$ of each transformation drifts the augmented image farther away from its clean counterpart, yet produces plausible images when appropriately controlled.

\clearpage
\newpage
\subsection{Transformation parameters}

\textbf{Spectral transform}\quad
Regarding the spectral transform of \cref{eq:prime-spectral_domain} we found out that, for the FIR filter $\bm \omega'$, a size of $K_\omega=3$ results into semantically preserving images for CIFAR-10/100 and ImageNet. For the latter, one can stretch the filter size to $5\times5$ or even $7\times7$, but then slight changes on the strength, $\sigma_\omega$, might destroy the image semantics. Eventually, given $K_\omega=3$, we observed that $\sigma_\omega=4$ is good enough for CIFAR-10/100 and ImageNet. 

\textbf{Spatial transform}\quad
Concerning the spatial transform of \cref{eq:prime-spatial_transform}, for the cut-off parameter $K_\tau$ we followed the value regimes proposed by Petrini \etal~\cite{Petrini2021Diffeo} and set $K_\tau=100$ for CIFAR-10/100; $K_\tau=500$ for ImageNet. Furthermore, for a given $K_\tau$, Petrini \etal also compute the appropriate bounds for the transformation strength, $\sigma^2_{\tau_\text{min}} \leq \sigma^2_\tau \leq \sigma^2_{\tau_\text{max}}$, such that the resulting diffeomorphism remains bijective and the pixel displacement does not destroy the image. In fact, in their original implementation, which can be fount at \url{https://github.com/pcsl-epfl/diffeomorphism}, Petrini \etal directly sample $\sigma_\tau\sim U(\sigma_{\tau_\text{min}},\sigma_{\tau_\text{max}})$ instead of explicitly setting the strength. In our implementation, we also follow the same approach. 

\textbf{Color transform}\quad
Regarding the color transform of \cref{eq:prime-color_transform} we found out that for CIFAR-10/100 a cut-off value of $K_\gamma=10$ and a strength of $\sigma_\gamma=0.01$ result into semantically preserving images for CIFAR-10/100; while for ImageNet, the corresponding values are $K_\gamma=500$ and $\sigma_\gamma=0.05$. As for the bandwidth (consecutive frequencies) $\Delta$ we observed that a value of $\Delta=20$ was memory sufficient for ImageNet, but for CIFAR-10/100 we can even afford all the frequencies to be used, e.g., $\Delta=K_\gamma$. 

Finally, as mentioned in \cref{sec:prime-prime-data-augmentations}, we randomly sample the strength of the transformations $\sigma$ from a uniform distribution of given minimum and maximum values. Regarding the maximum, we always set it to be the one we selected through visual inspection, while the minimum is set to $0$.

\section{Performance per corruption}

\begin{table*}[b]
\resizebox{\linewidth}{!}
{%
    \centering
    \aboverulesep=0ex
    \belowrulesep=0ex
    \begin{tabular}{clcc|ccc|cccc|cccc|cccc}
        \toprule
        \multirow{2}{*}{Dataset} & \multicolumn{1}{c}{\multirow{2}{*}{Method}} & \multicolumn{1}{c}{\multirow{2}{*}{Clean}} & \multicolumn{1}{c|}{\multirow{2}{*}{CC}} & \multicolumn{3}{c|}{Noise} & \multicolumn{4}{c|}{Blur} & \multicolumn{4}{c|}{Weather} & \multicolumn{4}{c}{Digital} \\
        & & & & Gauss. & Shot & Impulse & Defoc. & Glass & Motion & Zoom & Snow & Frost & Fog & Bright. & Contr. & Elastic & Pixel. & JPEG\\
        \midrule
        \multirow{3}{*}{C-10} & Std. & 95.0 & 74.0 & 45.1 & 58.7 & 54.9 & 83.2 & 53.3 & 76.9 & 79.1 & 83.1 & 79.3 & 89.0 & 93.6 & 76.3 & 83.9 & 75.1 & 77.9\\
        & AugMix & 95.2 & 88.6 & 79.3 & 84.8 & 85.8 & 94.1 & 78.9 & 92.4 & 93.4 & 89.7 & 89.0 & 91.9 & 94.3 & 90.5 & 90.5 & 87.6 & 87.5\\
        & PRIME & 94.2 & 89.8 & 86.9 & 88.1 & 88.6 & 92.6 & 85.3 & 90.8 & 92.2 & 89.3 & 90.5 & 89.8 & 93.7 & 92.4 & 90.1 & 88.1 & 88.8\\
        \midrule
        \multirow{3}{*}{C-100}& Std. & 76.7 & 51.9 & 25.3 & 33.7 & 26.6 & 60.8 & 47.1 & 55.5 & 57.6 & 60.8 & 56.2 & 62.5 & 72.2 & 53.2 & 63.4 & 50.1 & 52.7\\
        & AugMix & 78.2 & 64.9 & 46.7 & 55.1 & 60.6 & 76.2 & 47.3 & 72.6 & 74.3 & 67.4 & 64.4 & 69.9 & 75.5 & 67.4 & 69.6 & 64.9 & 61.8\\
        & PRIME & 78.4 & 68.2 & 59.0 & 62.1 & 68.1 & 74.0 & 58.3 & 70.5 & 72.3 & 68.9 & 68.5 & 69.8 & 76.8 & 74.4 & 70.1 & 65.5 & 64.4\\
        \bottomrule
    \end{tabular}
}
    \caption{Per-corruption accuracy of different methods on C-10 and C-100 (ResNet-18).}
    \label{tab:app-ch6-results-per-corruption-cifar}
\end{table*}

\begin{table*}[tp]
\resizebox{\linewidth}{!}
{%
    \centering
    \aboverulesep=0ex
    \belowrulesep=0ex
    \begin{tabular}{clcc|ccc|cccc|cccc|cccc}
        \toprule
        \multirow{2}{*}{Dataset} & \multicolumn{1}{c}{\multirow{2}{*}{Method}} & \multicolumn{1}{c}{\multirow{2}{*}{Clean}} & \multicolumn{1}{c|}{\multirow{2}{*}{CC}} & \multicolumn{3}{c|}{Noise} & \multicolumn{4}{c|}{Blur} & \multicolumn{4}{c|}{Weather} & \multicolumn{4}{c}{Digital} \\
        & & & & Gauss. & Shot & Impulse & Defoc. & Glass & Motion & Zoom & Snow & Frost & Fog & Bright. & Contr. & Elastic & Pixel. & JPEG\\
        \midrule
        \multirow{6}{*}{IN-100} & Standard & 88.0 & 49.7 & 30.9 & 29.0 & 22.0 & 45.6 & 44.6 & 50.4 & 53.9 & 43.8 & 46.2 & 50.5 & 78.6 & 42.9 & 68.8 & 68.0 & 70.6 \\
        & AugMix & 88.7 & 60.7 & 45.2 & 45.8 & 43.4 & 58.7 & 53.3 & 69.5 & 71.0 & 49.1 & 52.7 & 60.2 & 80.7 & 59.6 & 73.3 & 73.6 & 74.7\\
        & DA & 86.3 & 67.7 & 76.3 & 75.6 & 75.7 & 64.2 & 61.7 & 61.3 & 62.7 & 54.4 & 62.8 & 55.7 & 81.6 & 49.7 & 69.9 & 83.3 & 80.6\\
        & PRIME & 85.9 & 71.6 & 80.6 & 80.0 & 80.1 & 57.2 & 66.3 & 66.2 & 68.2 & 61.5 & 68.2 & 57.2 & 81.2 & 68.3 & 73.7 & 82.9 & 81.9\\
        \cmidrule{2-19}
        & DA\texttt{+}AugMix & 86.5 & 73.1 & 75.2 & 75.8 & 74.9 & 74.1 & 68.5 & 76.0 & 72.1 & 59.9 & 66.8 & 61.4 & 82.1 & 72.4 & 73.1 & 83.8 & 81.1\\
        & DA\texttt{+}PRIME & 84.9 & 74.9 & 81.1 & 80.9 & 81.2 & 70.5 & 74.2 & 72.0 & 71.5 & 66.3 & 73.6 & 56.6 & 81.9 & 72.8 & 74.8 & 83.4 & 82.3\\
        \midrule
        \multirow{7}{*}{IN} & Standard$^*$ & 76.1 & 39.2 & 29.3 & 27.0 & 23.8 & 38.8 & 26.8 & 38.7 & 36.2 & 32.5 & 38.1 & 45.4 & 68.0 & 39.0 & 45.3 & 44.8 & 53.4  \\
        & AugMix$^*$ & 77.5 & 48.3 & 40.6 & 41.1 & 37.7 & 47.7 & 34.9 & 53.5 & 49.0 & 39.9 & 43.8 & 47.1 & 69.5 & 51.1 & 52.0 & 57.0 & 60.3 \\
        & DA$^*$ & 76.7 & 52.6 & 56.6 & 54.9 & 56.3 & 51.7 & 40.1 & 48.7 & 39.5 & 44.2 & 50.3 & 52.1 & 71.1 & 48.3 & 50.9 & 65.5 & 59.3 \\
        & PRIME$^\dagger$ & 77.0 & 55.0 & 61.9 & 60.6 & 60.9 & 47.6 & 39.0 & 48.4 & 46.0 & 47.4 & 50.8 & 54.1 & 71.7 & 58.2 & 56.3 & 59.5 & 62.2\\
        \cmidrule{2-19}
        & DA\texttt{+}AugMix & 75.8 & 58.1 & 59.4 & 59.6 & 59.1 & 59.0 & 46.8 & 61.1 & 51.5 & 49.4 & 53.3 & 55.9 & 70.8 & 58.7 & 54.3 & 68.8 & 63.3 \\
        & DA\texttt{+}PRIME$^\dagger$ & 75.5 & 59.9 & 67.4 & 67.2 & 66.8 & 56.2 & 47.5 & 54.3 & 47.3 & 52.8 & 56.4 & 56.3 & 71.7 & 62.3 & 57.3 & 70.3 & 65.1 \\
        \bottomrule
    \end{tabular}
}
    \caption{Per-corruption accuracy of different methods on IN-100 (ResNet-18) and IN (ResNet-50). $^\dagger$ indicates that JSD consistency loss is not used. $^*$Models taken from \texttt{RobustBench}~\cite{robustbench2021}.}
    \label{tab:app-ch6-results-per-corruption-imagenet}
\end{table*}

Beyond the average corruption accuracy that we report in \cref{tab:prime-results-main}, we also provide here the performance of each method on the individual corruptions. The results on CIFAR-10/100 and ImageNet/ImageNet-100 are shown on \cref{tab:app-ch6-results-per-corruption-cifar,tab:app-ch6-results-per-corruption-imagenet} respectively. Compared to AugMix on CIFAR-10/100, the improvements from PRIME are mostly observed against Gaussian noise ($+7.6\%/12.3\%$), shot noise ($+3.3\%/7.0\%$), glass blur ($+6.4\%/11.0\%$) and JPEG compression ($+1.3\%/2.6\%$). These results show that PRIME can really push the performance against certain corruptions in CIFAR-10/100-C despite the fact that AugMix is already good on these datasets. However, AugMix turns out to be slightly better than PRIME against impulse noise, defocus blur and motion blur modifications; all of which have been shown to be resembled by AugMix created images. With ImageNet-100, PRIME enhances the diversity of augmented images, and leads to general improvements against all corruptions except certain blurs.
On ImageNet, we observe that, in comparison to DeepAugment, the supremacy of PRIME is reflected on almost every corruption type, except some blurs and pixelate corruptions where DeepAugment is slightly better. When PRIME is used in conjunction with DeepAugment, compared to AugMix combined with DeepAugment, our method seems to lack behind only on blurs, while on the rest of the corruptions achieves higher robustness.

\section{Performance per severity level}

We also want to investigate the robustness of each method on different severity levels of the corruptions. The results for CIFAR-10/100 and ImageNet/ImageNet-100 are presented in \cref{tab:app-ch6-results-per-severity-cifar,tab:app-ch6-results-per-severity-imagenet} respectively. With CIFAR-10/100, PRIME predominantly helps against corruptions with maximal severity and yields $+3.9\%$ and $+7.1\%$ gains on CIFAR-10 and CIFAR-100 respectively. Besides on ImageNet-100, PRIME again excels at corruptions with moderate to higher severity. This observations also holds when PRIME is employed in concert with DeepAugment. With ImageNet too this trend continues, and we observe that, compared to DeepAugment, PRIME improves significantly on corruptions of larger severity ($+3.4\%$ and $+5.5\%$ on severity levels 4 and 5 respectively). Also, this behaviour is consistent even when PRIME is combined with DeepAugment and is compared to DeepAugment\texttt{+}AugMix, where we see that again on levels 4 and 5 there is a significant improvement of $+2.1\%$ and $+3.7\%$ respectively.

\clearpage
\newpage

\begin{table*}[!th]
\centering
\resizebox{0.6\linewidth}{!}
{%
    \aboverulesep=0ex
    \belowrulesep=0ex
    \begin{tabular}{clcc|ccccc}
        \toprule
        \multirow{2}{*}{Dataset} & \multicolumn{1}{c}{\multirow{2}{*}{Method}} & \multicolumn{1}{c}{\multirow{2}{*}{Clean}} & \multicolumn{1}{c|}{\multirow{2}{*}{CC Avg.}} & \multicolumn{5}{c}{Severity} \\
        & & & & 1 & 2 & 3 & 4 & 5\\
        \midrule
        \multirow{3}{*}{C-10} & Standard & 95.0 & 74.0 & 87.4 & 81.7 & 75.7 & 68.3 & 56.7\\
        & AugMix & 95.2 & 88.6 & 93.1 & 91.8 & 89.9 & 86.7 & 81.7\\
        & PRIME & 94.2 & 89.8 & 92.8 & 91.6 & 90.4 & 88.6 & 85.6\\
        \midrule
        \multirow{3}{*}{C-100}& Standard & 76.7 & 51.9 & 66.7 & 59.4 & 52.8 & 45.0 & 35.4\\
        & AugMix & 78.2 & 64.9 & 73.3 & 70.0 & 66.6 & 61.3 & 53.4\\
        & PRIME & 78.4 & 68.2 & 74.0 & 71.6 & 69.2 & 65.6 & 60.5\\
        \bottomrule
    \end{tabular}
}
    \caption{Average accuracy for each corruption severity level of different methods on C-10 and C-100 (ResNet-18).}
    
    \label{tab:app-ch6-results-per-severity-cifar}
\end{table*}

\begin{table*}[!th]
\centering
\resizebox{0.65\linewidth}{!}
{%
    \aboverulesep=0ex
    \belowrulesep=0ex
    \small
    \begin{tabular}{clcc|ccccc}
        \toprule
        \multirow{2}{*}{Dataset} & \multicolumn{1}{c}{\multirow{2}{*}{Method}} & \multicolumn{1}{c}{\multirow{2}{*}{Clean}} & \multicolumn{1}{c|}{\multirow{2}{*}{CC Avg.}} & \multicolumn{5}{c}{Severity} \\
        & & & & 1 & 2 & 3 & 4 & 5\\
        \midrule
        \multirow{6}{*}{IN-100} & Standard & 88.0 & 49.7 & 73.5 & 61.0 & 49.8 & 37.2 & 27.0\\
        & AugMix & 88.7 & 60.7 & 80.4 & 71.8 & 63.8 & 50.3 & 37.2\\
        & DA & 86.3 & 67.7 & 81.2 & 75.4 & 69.9 & 61.2 & 50.8\\
        & PRIME & 85.9 & 71.6 & 81.7 & 77.5 & 73.4 & 66.9 & 58.4\\
        \cmidrule{2-9}
        & DA\texttt{+}AugMix & 86.5 & 73.1 & 82.7 & 78.0 & 75.5 & 69.6 & 59.9\\
        & DA\texttt{+}PRIME & 84.9 & 74.9 & 82.0 & 78.7 & 76.4 & 71.8 & 65.5\\
        \midrule
        \multirow{7}{*}{IN} & Standard$^*$ & 76.1 & 39.2 & 60.6 & 49.8 & 39.8 & 27.7 & 18.0 \\
        & AugMix$^*$ & 77.5 & 48.3 & 66.7 & 58.3 & 51.1 & 39.1 & 26.5 \\
        & DA$^*$ & 76.7 & 52.6 & 69.0 & 61.7 & 55.4 & 44.9 & 32.1 \\
        & PRIME$^\dagger$ & 77.0 & 55.0 & 68.9 & 63.1 & 56.9 & 48.3 & 37.6 \\
        \cmidrule{2-9}
        & DA\texttt{+}AugMix & 75.8 & 58.1 & 70.3 & 64.5 & 60.5 & 53.0 & 42.2 \\
        & DA\texttt{+}PRIME$^\dagger$ & 75.5 & 59.9 & 70.8 & 66.3 & 61.6 & 55.1 & 45.9 \\
        \bottomrule
    \end{tabular}
}
    \caption{Average accuracy for each corruption severity level of different methods on IN-100 (ResNet-18) and IN (ResNet-50). $^\dagger$ indicates that JSD consistency loss is not used. $^*$Models taken from \texttt{RobustBench}~\cite{robustbench2021}.}
    
    \label{tab:app-ch6-results-per-severity-imagenet}
\end{table*}

\clearpage

\backmatter

\cleardoublepage
\phantomsection
\addcontentsline{toc}{chapter}{Bibliography}

\bibliographystyle{IEEEtran}
\bibliography{strings.bib,refs.bib}

\end{document}